\documentclass[10pt,twocolumn,letterpaper]{article}

\pdfoutput=1

\usepackage{cvpr}
\usepackage{times}
\usepackage{epsfig}
\usepackage{graphicx}
\usepackage{amsmath}
\usepackage{amssymb}
\usepackage{booktabs}
\usepackage{multirow}
\usepackage{subfigure}
\usepackage{lib}
\usepackage[font=small,labelfont=bf,skip=5pt]{caption}

\usepackage[pagebackref=true,breaklinks=true,letterpaper=true,colorlinks,bookmarks=false]{hyperref}

\cvprfinalcopy

\renewcommand{\cf}{\textit{cf.}\xspace}

\newcommand{\pool}[1]{$\textrm{pool}_{#1}$\xspace}

\newcommand{\fc}[1]{$\textrm{fc}_{#1}$\xspace}
\newcommand{\OtwoP}{O$_2$P\xspace}
\newcommand{\val}{val\xspace}
\newcommand{\vala}{val$_1$\xspace}
\newcommand{\valb}{val$_2$\xspace}
\newcommand{\train}{train\xspace}
\newcommand{\test}{test\xspace}
\newcommand{\trainvalN}{val$_1+$train$_N$\xspace}
\newcommand{\trainvala}{val$_1+$train$_{.5\textrm{k}}$\xspace}
\newcommand{\trainvalb}{val$_1+$train$_{1\textrm{k}}$\xspace}
\newcommand{\trainvalab}{val$+$train$_{1\textrm{k}}$\xspace}

\makeatletter
\renewcommand{\paragraph}{%
  \@startsection{paragraph}{4}%
  {\z@}{1.5ex \@plus 1ex \@minus .2ex}{-0.5em}%
  {\normalfont\normalsize\bfseries}%
}
\makeatother

\begin{document}

\abovedisplayskip=12pt plus 3pt minus 9pt
\abovedisplayshortskip=0pt plus 3pt
\belowdisplayskip=12pt plus 3pt minus 9pt
\belowdisplayshortskip=7pt plus 3pt minus 4pt

\title{Rich feature hierarchies for accurate object detection and semantic segmentation\\{\large Tech report (v5)}}

\author{Ross Girshick~~~Jeff Donahue~~~Trevor Darrell~~~Jitendra Malik\\
UC Berkeley\\
{\tt\small \{rbg,jdonahue,trevor,malik\}@eecs.berkeley.edu}
}

\maketitle

\begin{abstract}
Object detection performance, as measured on the canonical PASCAL VOC dataset, has plateaued in the last few years.
The best-performing methods are complex ensemble systems that typically combine multiple low-level image features with high-level context.
In this paper, we propose a simple and scalable detection algorithm that improves mean average precision (mAP) by more than 30\% relative to the previous best result on VOC 2012---achieving a mAP of 53.3\%.
Our approach combines two key insights: (1) one can apply high-capacity convolutional neural networks (CNNs) to bottom-up region proposals in order to localize and segment objects and (2) when labeled training data is scarce, supervised pre-training for an auxiliary task, followed by domain-specific fine-tuning, yields a significant performance boost.
Since we combine region proposals with CNNs, we call our method \emph{R-CNN:} Regions with CNN features.
We also compare R-CNN to OverFeat, a recently proposed sliding-window detector based on a similar CNN architecture.
We find that R-CNN outperforms OverFeat by a large margin on the 200-class ILSVRC2013 detection dataset.
Source code for the complete system is available at 
\url{http://www.cs.berkeley.edu/~rbg/rcnn}.
\vspace{-1em}
\end{abstract}


\section{Introduction}

Features matter.
The last decade of progress on various visual recognition tasks has been based considerably on the use of SIFT \cite{SIFT} and HOG \cite{Dalal05}.
But if we look at performance on the canonical visual recognition task, PASCAL VOC object detection \cite{PASCAL-IJCV}, it is generally acknowledged that progress has been slow during 2010-2012, with small gains obtained by building ensemble systems and employing minor variants of successful methods.

SIFT and HOG are blockwise orientation histograms, a representation we could associate roughly with complex cells in V1, the first cortical area in the primate visual pathway.
But we also know that recognition occurs several stages downstream, which suggests that there might be hierarchical, multi-stage processes for computing features that are even more informative for visual recognition.

\begin{figure}[t!]
\centering
\includegraphics[width=\linewidth]{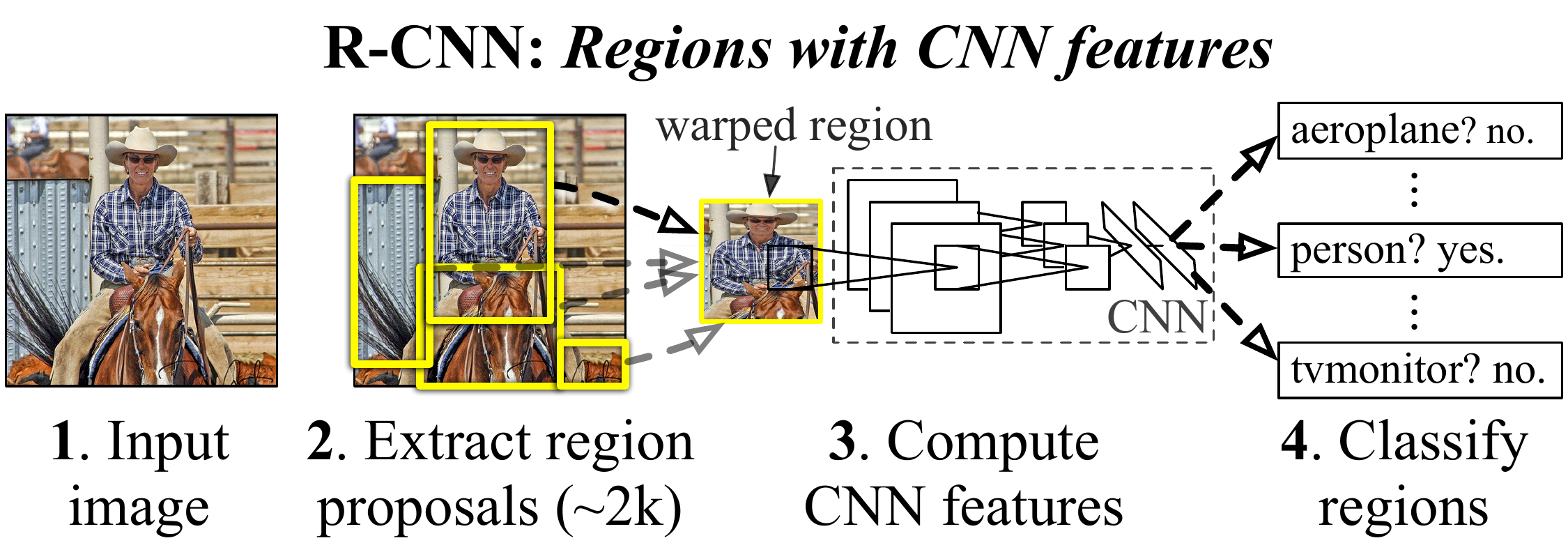}
\caption{\textbf{Object detection system overview.}
Our system (1) takes an input image, (2) extracts around 2000 bottom-up region proposals, (3) computes features for each proposal using a large convolutional neural network (CNN), and then (4) classifies each region using class-specific linear SVMs.
R-CNN achieves a mean average precision (mAP) of \textbf{53.7\% on PASCAL~VOC~2010}. 
For comparison, \cite{UijlingsIJCV2013} reports 35.1\% mAP using the same region proposals, but with a spatial pyramid and bag-of-visual-words approach.
The popular deformable part models perform at 33.4\%.
On the 200-class \textbf{ILSVRC2013 detection dataset, R-CNN's mAP is 31.4\%}, a large improvement over OverFeat \cite{overfeat}, which had the previous best result at 24.3\%.
}
\figlabel{splash}
\vspace{-1em}
\end{figure}

Fukushima's ``neocognitron'' \cite{fukushima1980neocognitron}, a biologically-inspired hierarchical and shift-invariant model for pattern recognition, was an early attempt at just such a process.
The neocognitron, however, lacked a supervised training algorithm.
Building on Rumelhart \etal \cite{rumelhart86}, LeCun \etal \cite{lecun-89e} showed that stochastic gradient descent via backpropagation was effective for training convolutional neural networks (CNNs), a class of models that extend the neocognitron.

CNNs saw heavy use in the 1990s (\eg, \cite{lecun-98}), but then fell out of fashion with the rise of support vector machines.
In 2012, Krizhevsky \etal \cite{krizhevsky2012imagenet} rekindled interest
in CNNs by showing substantially higher image classification accuracy on the ImageNet Large Scale Visual Recognition Challenge (ILSVRC) \cite{ILSVRC12,imagenet_cvpr09}.
Their success resulted from training a large CNN on 1.2 million labeled images, together with a few twists on LeCun's CNN (\eg, $\max(x,0)$ rectifying non-linearities and ``dropout'' regularization).

The significance of the ImageNet result was vigorously debated during the ILSVRC 2012 workshop.
The central issue can be distilled to the following: To what
extent do the CNN classification results on ImageNet generalize
to object detection results on the PASCAL VOC Challenge?

We answer this question by bridging the gap between image classification and object detection.
This paper is the first to show that a CNN can lead to dramatically higher object detection performance on PASCAL VOC as compared to systems based on simpler HOG-like features.
To achieve this result, we focused on two problems: localizing objects with a deep network and training a high-capacity model with only a small quantity of annotated detection data.

Unlike image classification, detection requires localizing (likely many) objects within an image.
One approach frames localization as a regression problem.
However, work from Szegedy \etal \cite{szegedy2013deep}, concurrent with our own, 
indicates that this strategy may not fare well in practice (they report a mAP of 30.5\% on VOC 2007 compared to the 58.5\% achieved by our method).
An alternative is to build a sliding-window detector.
CNNs have been used in this way for at least two decades, typically on constrained object categories, such as faces \cite{rowley1998neural,lecun94} and pedestrians \cite{sermanetCVPR13}.
In order to maintain high spatial resolution, these CNNs typically only have two convolutional and pooling layers.
We also considered adopting a sliding-window approach.
However, units high up in our network, which has five convolutional layers, have very large receptive fields ($195 \times 195$ pixels) and strides ($32 \times 32$ pixels) in the input image, which makes precise localization within the sliding-window paradigm an open technical challenge.

Instead, we solve the CNN localization problem by operating within the ``recognition using regions'' paradigm \cite{gu2009recognition}, which has been successful for both object detection \cite{UijlingsIJCV2013} and semantic segmentation \cite{carreira2012cpmc}.
At test time, our method generates around 2000 category-independent region proposals for the input image, extracts a fixed-length feature vector from each proposal using a CNN, and then classifies each region with category-specific linear SVMs.
We use a simple technique (affine image warping) to compute a fixed-size CNN input from each region proposal, regardless of the region's shape.
\figref{splash} presents an overview of our method and highlights some of our results.
Since our system combines region proposals with CNNs, we dub the method R-CNN: Regions with CNN features.

In this updated version of this paper, we provide a head-to-head comparison of R-CNN and the recently proposed OverFeat \cite{overfeat} detection system by running R-CNN on the 200-class ILSVRC2013 detection dataset.
OverFeat uses a sliding-window CNN for detection and until now was the best performing method on ILSVRC2013 detection.
We show that R-CNN significantly outperforms OverFeat, with a mAP of 31.4\% versus 24.3\%.

A second challenge faced in detection is that labeled data is scarce and the amount currently available is insufficient for training a large CNN.
The conventional solution to this problem is to use \emph{unsupervised} pre-training, followed by supervised fine-tuning (\eg, \cite{sermanetCVPR13}).
The second principle contribution of this paper is to show that \emph{supervised} pre-training on a large auxiliary dataset (ILSVRC), followed by domain-specific fine-tuning on a small dataset (PASCAL), is an effective paradigm for learning high-capacity CNNs when data is scarce.
In our experiments, fine-tuning for detection improves mAP performance by 8 percentage points.
After fine-tuning, our system achieves a mAP of 54\% on VOC 2010 compared to 33\% for the highly-tuned, HOG-based deformable part model (DPM) \cite{lsvm-pami,release5}.
We also point readers to contemporaneous work by Donahue \etal \cite{decafICML}, who show that Krizhevsky's CNN can be used (without fine-tuning) as a blackbox feature extractor, yielding excellent performance on several recognition tasks including scene classification, fine-grained sub-categorization, and domain adaptation.

Our system is also quite efficient.
The only class-specific computations are a reasonably small matrix-vector product and greedy non-maximum suppression.
This computational property follows from features that are shared across all categories and that are also two orders of magnitude lower-dimensional than previously used region features (\cf \cite{UijlingsIJCV2013}).

Understanding the failure modes of our approach is also critical for improving it, and so we report results from the detection analysis tool of Hoiem \etal
\cite{hoiem2012diagnosing}.
As an immediate consequence of this analysis, we demonstrate that a simple bounding-box regression method significantly reduces mislocalizations, which are the dominant error mode.

Before developing technical details, we note that because R-CNN operates on regions it is natural to extend it to the task of semantic segmentation.
With minor modifications, we also achieve competitive results on the PASCAL VOC segmentation task, with an average segmentation accuracy of 47.9\% on the VOC 2011 test set.

\section{Object detection with R-CNN}
\seclabel{detection}

Our object detection system consists of three modules.
The first generates category-independent region proposals. 
These proposals define the set of candidate detections available to our detector.
The second module is a large convolutional neural network that extracts a fixed-length feature vector from each region. 
The third module is a set of class-specific linear SVMs.
In this section, we present our design decisions for each module, describe their test-time usage, detail how their parameters are learned, and show detection results on PASCAL VOC 2010-12 and on ILSVRC2013.

\subsection{Module design}

\paragraph{Region proposals.}
A variety of recent papers offer methods for generating category-independent 
region proposals. Examples include: objectness \cite{objectness-pami},
selective search \cite{UijlingsIJCV2013}, category-independent
object proposals \cite{endres2010category}, constrained parametric min-cuts (CPMC)
\cite{carreira2012cpmc}, multi-scale combinatorial grouping \cite{mcg2014}, and Cire{\c{s}}an \etal \cite{cirecsan2013mitosis}, who detect mitotic cells by applying a CNN to regularly-spaced square crops, which are a special case of region proposals.
While R-CNN is agnostic to the particular region proposal method, we use selective search to enable a controlled comparison with prior detection work (\eg, \cite{UijlingsIJCV2013,regionlets}).

\paragraph{Feature extraction.}
We extract a 4096-dimensional feature vector from each region proposal using the Caffe \cite{Jia13caffe} implementation of the CNN described by Krizhevsky \etal
\cite{krizhevsky2012imagenet}.
Features are computed by forward propagating a mean-subtracted $227 \times 227$ 
RGB image through five convolutional layers and two fully connected layers.
We refer readers to \cite{Jia13caffe,krizhevsky2012imagenet} for more network architecture details.

In order to compute features for a region proposal, we must first 
convert the image data in that region into a form that is compatible
with the CNN (its architecture requires inputs of a fixed
$227 \times 227$ pixel size). 
Of the many possible transformations of our arbitrary-shaped regions, we opt for the simplest. 
Regardless of the size or aspect ratio of the candidate region, we warp all pixels in a tight bounding box around it to the required size.
Prior to warping, we dilate the tight bounding box so that at the warped size there are exactly $p$ pixels of warped image context around the original box (we use $p = 16$).
\figref{warped-samples} shows a random sampling of warped training regions.
Alternatives to warping are discussed in \asecref{altwarp}.

\begin{figure}[t!]
\def \sz {-1.15em}
\centering
\includegraphics[height=0.8in]{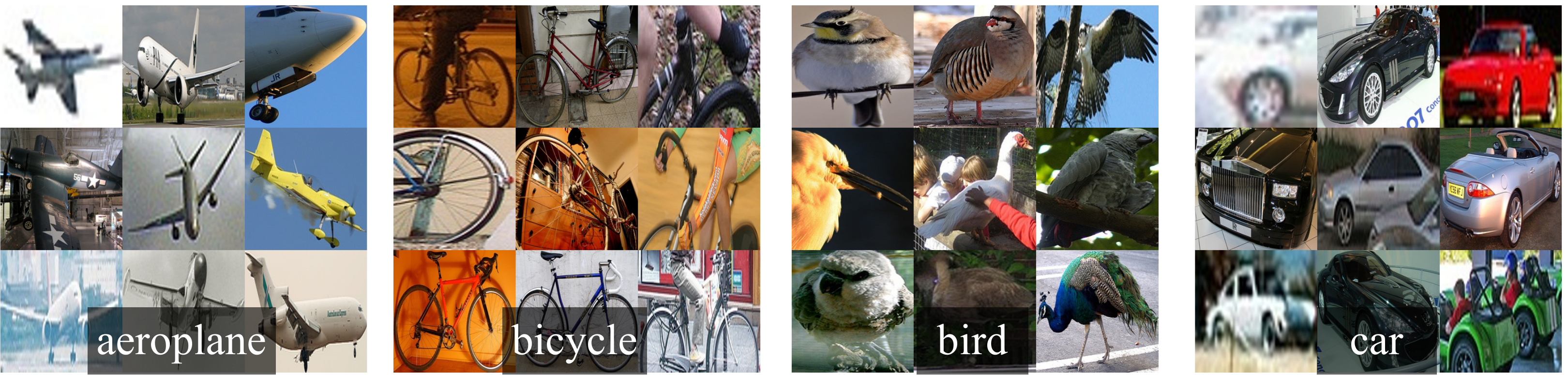}
\vspace{-1em}
\caption{\textbf{Warped training samples} from VOC 2007 train.}
\figlabel{warped-samples}
\vspace{-1em}
\end{figure}

\subsection{Test-time detection}
\seclabel{runtime}

At test time, we run selective search on the test image to extract around 2000 region proposals (we use selective search's ``fast mode'' in all experiments).
We warp each proposal and forward propagate it through the CNN in order to compute features. 
Then, for each class, we score each extracted feature vector using the SVM trained for that class.  
Given all scored regions in an image, we apply a greedy non-maximum suppression (for each class independently) that rejects a region if it has an intersection-over-union (IoU) overlap with a higher scoring selected region larger than a learned threshold.

\paragraph{Run-time analysis.}
Two properties make detection efficient. First,
all CNN parameters are shared across all categories.  Second, the
feature vectors computed by the CNN are low-dimensional when compared
to other common approaches, such as spatial pyramids with bag-of-visual-word
encodings. 
The features used in the UVA detection system \cite{UijlingsIJCV2013}, for example, are two orders of magnitude larger than ours (360k \vs 4k-dimensional).

The result of such sharing is that the time spent computing region
proposals and features (13s/image on a GPU or
53s/image on a CPU) is amortized over all classes. The only
class-specific computations are dot products between features and
SVM weights and non-maximum suppression. 
In practice, all dot products for an image are batched
into a single matrix-matrix product. The feature matrix is typically
$2000 \times 4096$ and the SVM weight matrix is $4096 \times
N$, where $N$ is the number of classes. 

This analysis shows that R-CNN can
scale to thousands of object classes without resorting to approximate
techniques, such as hashing. 
Even if there were 100k classes, the resulting
matrix multiplication takes only 10 seconds on a modern multi-core CPU.
This efficiency is not merely the result
of using region proposals and shared features. The UVA system, 
due to its high-dimensional features, would be two orders
of magnitude slower while requiring
134GB of memory just to store 100k linear predictors, compared
to just 1.5GB for our lower-dimensional features.

It is also interesting to contrast R-CNN with the recent work from Dean \etal
on scalable detection using DPMs and hashing \cite{dean2013fast}. They report a mAP of around 16\% on VOC 2007 at a run-time of 5 minutes per image
when introducing 10k distractor classes. With our approach, 10k
detectors can run in about a minute on a CPU, and because 
no approximations are made mAP would remain at 59\% (\secref{ablation}).

\subsection{Training}
\seclabel{training}

\paragraph{Supervised pre-training.}
We discriminatively pre-trained the CNN on a large auxiliary dataset (ILSVRC2012 classification) using \emph{image-level annotations} only (bounding-box labels are not available for this data).
Pre-training was performed using the open source Caffe CNN library \cite{Jia13caffe}.
In brief, our CNN nearly matches the performance of Krizhevsky \etal
\cite{krizhevsky2012imagenet}, obtaining a top-1 error rate
2.2 percentage points higher on the ILSVRC2012 classification validation
set. This discrepancy is due to simplifications in the training process.

\paragraph{Domain-specific fine-tuning.}
To adapt our CNN to the new task (detection) and the new domain (warped proposal windows), we continue stochastic gradient descent (SGD) training of the CNN parameters using only warped region proposals.
Aside from replacing the CNN's ImageNet-specific 1000-way classification layer with a randomly initialized ($N+1$)-way classification layer (where $N$ is the number of object classes, plus 1 for background), the CNN architecture is unchanged.
For VOC, $N = 20$ and for ILSVRC2013, $N = 200$.
We treat all region proposals with $\geq 0.5$ IoU overlap with a ground-truth
box as positives for that box's class and the rest as negatives. 
We start SGD at a learning rate of 0.001 (1/10th of the initial pre-training rate),
which allows fine-tuning to make progress while not clobbering the initialization.
In each SGD iteration, we uniformly sample 32 positive windows (over all classes) and 96 background windows to construct a mini-batch of size 128.
We bias the sampling towards positive windows because they are extremely rare compared to background.

\begin{table*}[t!]
\centering
\renewcommand{\arraystretch}{1.2}
\renewcommand{\tabcolsep}{1.2mm}
\resizebox{\linewidth}{!}{
\begin{tabular}{@{}l|r*{19}{c}|c@{}}
\textbf{VOC 2010 test}         & aero      & bike      & bird      & boat      & bottle     & bus        & car        & cat        & chair      & cow        & table      & dog        & horse      & mbike      & person     & plant      & sheep      & sofa       & train      & tv         & mAP       \\
\hline
DPM v5 \cite{release5}$^\dagger$   &  49.2  &  53.8  &  13.1  &  15.3  &  35.5  &  53.4  &  49.7  &  27.0  &  17.2  &  28.8  &  14.7  &  17.8  &  46.4  &  51.2  &  47.7  &  10.8  &  34.2  &  20.7  &  43.8  &  38.3  &  33.4 \\
UVA \cite{UijlingsIJCV2013}  &  56.2  &  42.4  &  15.3  &  12.6  &  21.8  &  49.3  &  36.8  &  46.1  &  12.9  &  32.1  &  30.0  &  36.5  &  43.5  &  52.9  &  32.9  &  15.3  &  41.1  &  31.8  &  47.0  &  44.8  &  35.1 \\
Regionlets \cite{regionlets}  &  65.0  &  48.9  &  25.9  &  24.6  &  24.5  &  56.1  &  54.5  &  51.2  &  17.0  &  28.9  &  30.2  &  35.8  &  40.2  &  55.7  &  43.5  &  14.3  &  43.9  &  32.6  &  54.0  &  45.9  &  39.7 \\
SegDPM \cite{fidler2013bottom}$^\dagger$  &  61.4  &  53.4  &  25.6  &  25.2  &  35.5  &  51.7  &  50.6  &  50.8  &  19.3  &  33.8  &  26.8  &  40.4  &  48.3  &  54.4  &  47.1  &  14.8  &  38.7  &  35.0  &  52.8  &  43.1  &  40.4 \\
\hline
R-CNN  &  67.1  &  64.1  &  46.7  &  32.0  &  30.5  &  56.4  &  57.2  &  65.9  &  27.0  &  47.3  &  40.9  &  66.6  &  57.8  &  65.9  &  53.6  &  26.7  &  56.5  &  38.1  &  52.8  &  50.2  &  50.2 \\
R-CNN BB  &  \bf{71.8}  &  \bf{65.8}  &  \bf{53.0}  &  \bf{36.8}  &  \bf{35.9}  &  \bf{59.7}  &  \bf{60.0}  &  \bf{69.9}  &  \bf{27.9}  &  \bf{50.6}  &  \bf{41.4}  &  \bf{70.0}  &  \bf{62.0}  &  \bf{69.0}  &  \bf{58.1}  &  \bf{29.5}  &  \bf{59.4}  &  \bf{39.3}  &  \bf{61.2}  &  \bf{52.4}  &  \bf{53.7} \\
\end{tabular}
}
\caption{\textbf{Detection average precision (\%) on VOC 2010 test.} R-CNN is most directly comparable to UVA and Regionlets since all methods use selective search region proposals. 
Bounding-box regression (BB) is described in \secref{bboxreg}.
At publication time, SegDPM was the top-performer on the PASCAL VOC leaderboard. $^\dagger$DPM and SegDPM use context rescoring not used by the other methods.}
\tablelabel{voc2010}
\end{table*}

\paragraph{Object category classifiers.} 
Consider training a binary classifier to detect cars.
It's clear that an image region tightly enclosing a car should be a positive example.
Similarly, it's clear that a background region, which has nothing to do with cars, should be a negative example.
Less clear is how to label a region that partially overlaps a car.
We resolve this issue with an IoU overlap threshold, below which regions are defined as negatives.
The overlap threshold, $0.3$,  was selected by a grid search over $\{0, 0.1, \ldots, 0.5\}$ on a validation set. 
We found that selecting this threshold carefully is important. 
Setting it to $0.5$, as in \cite{UijlingsIJCV2013}, decreased mAP by $5$ points. Similarly, setting it to $0$ decreased mAP by $4$ points.
Positive examples are defined simply to be the ground-truth bounding boxes for each class.

Once features are extracted and training labels are applied, we optimize one linear SVM per class.
Since the training data is too large to fit in memory, we adopt the standard hard negative mining method \cite{lsvm-pami,Sung94}. 
Hard negative mining converges quickly and in practice mAP stops increasing after only a single pass over all images.

In \asecref{posneg} we discuss why the positive and negative examples are defined differently in fine-tuning versus SVM training.
We also discuss the trade-offs involved in training detection SVMs rather than simply using the outputs from the final softmax layer of the fine-tuned CNN.


\subsection{Results on PASCAL VOC 2010-12}
Following the PASCAL VOC best practices \cite{PASCAL-IJCV}, we validated 
all design decisions and hyperparameters on the VOC 2007 dataset (\secref{ablation}).
For final results on the VOC 2010-12 datasets, we fine-tuned
the CNN on VOC 2012 train and optimized our detection SVMs on VOC 2012 trainval. 
We submitted test results to the evaluation server only once for each of the two major algorithm variants (with and without bounding-box regression).

\tableref{voc2010} shows complete results on VOC 2010. 
We compare our method against four strong baselines, including SegDPM \cite{fidler2013bottom}, which combines DPM detectors with the output of a semantic segmentation system \cite{o2p} and uses additional inter-detector context and image-classifier rescoring.
The most
germane comparison is to the UVA system from
Uijlings \etal \cite{UijlingsIJCV2013}, since our systems use the
same region proposal algorithm. To classify regions, their method builds a four-level
spatial pyramid and populates it with densely sampled SIFT,
Extended OpponentSIFT, and RGB-SIFT descriptors, each vector quantized with 4000-word codebooks. 
Classification is performed with a histogram intersection
kernel SVM. Compared to their multi-feature, non-linear kernel SVM
approach, we achieve a large improvement in mAP, from 35.1\% to
53.7\% mAP, while also being much faster (\secref{runtime}).
Our method achieves similar performance (53.3\% mAP) on VOC 2011/12 test.

\subsection{Results on ILSVRC2013 detection}
We ran R-CNN on the 200-class ILSVRC2013 detection dataset using the same system hyperparameters that we used for PASCAL VOC.
We followed the same protocol of submitting test results to the ILSVRC2013 evaluation server only twice, once with and once without bounding-box regression.

\figref{ilsvrc13} compares R-CNN to the entries in the ILSVRC 2013 competition and to the post-competition OverFeat result \cite{overfeat}.
R-CNN achieves a mAP of 31.4\%, which is significantly ahead of the second-best result of 24.3\% from OverFeat.
To give a sense of the AP distribution over classes, box plots are also presented and a table of per-class APs follows at the end of the paper in \tableref{classaps}.
Most of the competing submissions (OverFeat, NEC-MU, UvA-Euvision, Toronto A, and UIUC-IFP) used convolutional neural networks, indicating that there is significant nuance in how CNNs can be applied to object detection, leading to greatly varying outcomes.

In \secref{ilsvrc}, we give an overview of the ILSVRC2013 detection dataset and provide details about choices that we made when running R-CNN on it.

\begin{figure*}[t]
\begin{center}
\includegraphics[height=2.5in]{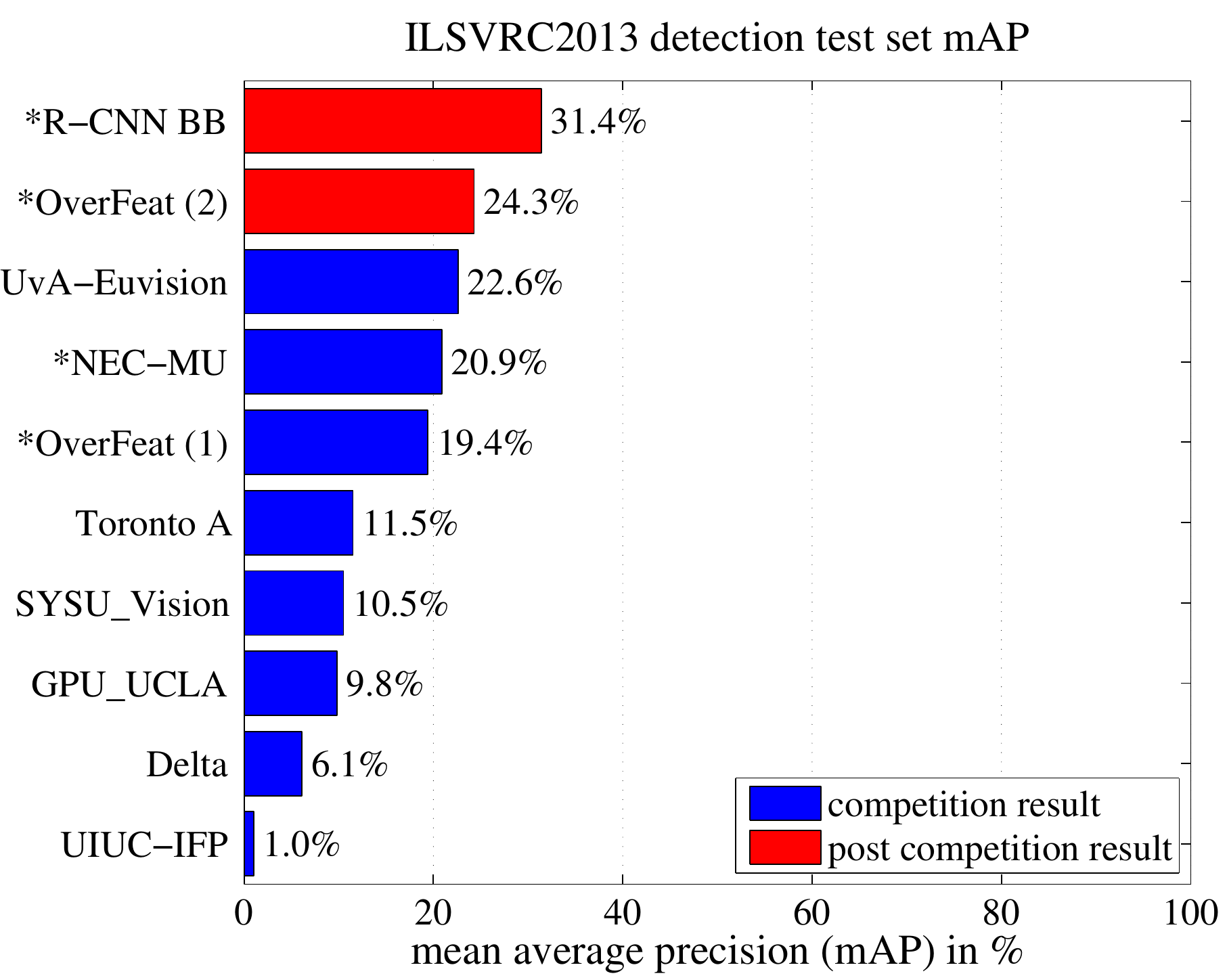}
\hspace{2em}
\includegraphics[height=2.5in]{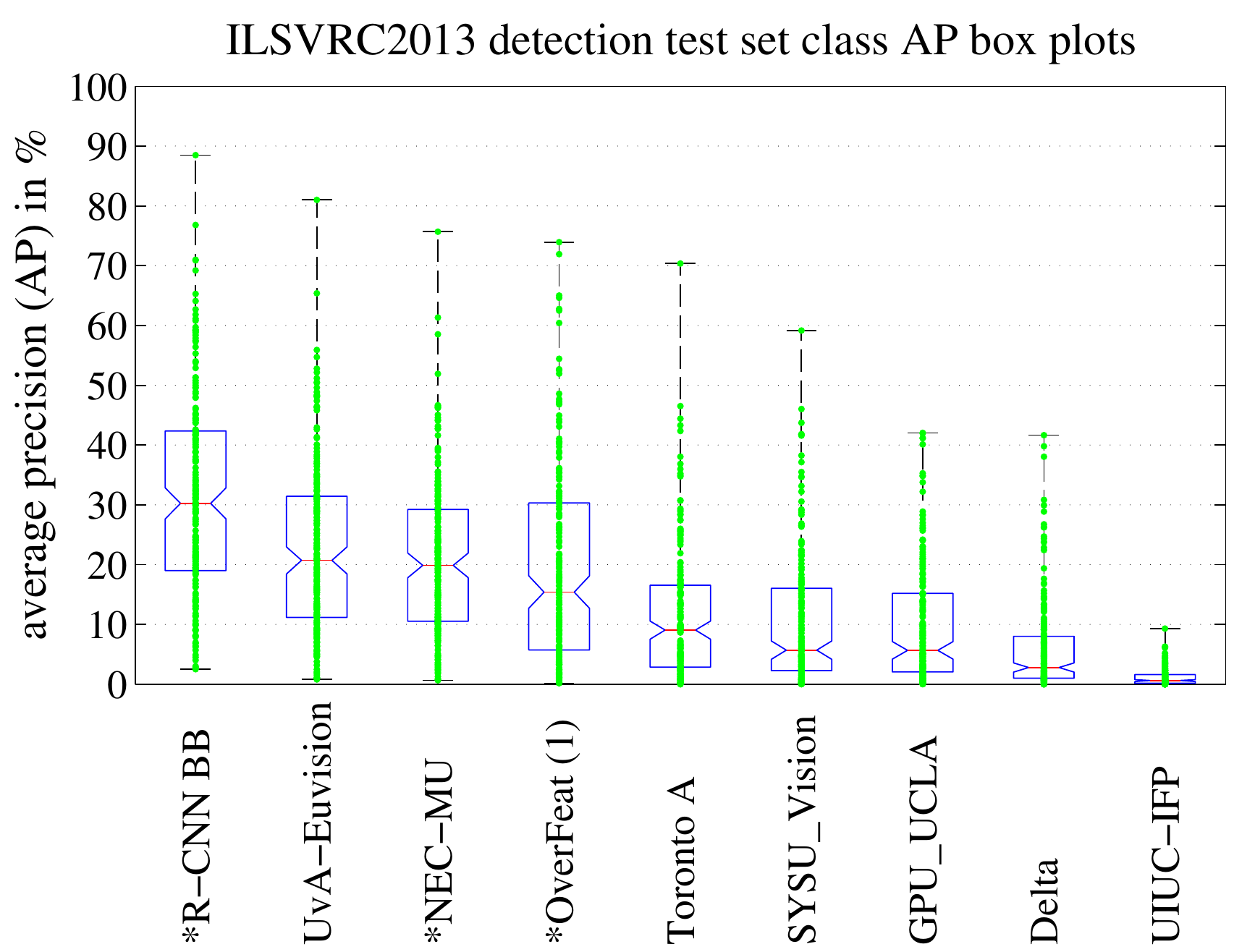}
\end{center}
\vspace{-0.5em}
\caption{\textbf{(Left) Mean average precision on the ILSVRC2013 detection test set.} 
Methods preceeded by * use outside training data (images and labels from the ILSVRC classification dataset in all cases). 
\textbf{(Right) Box plots for the 200 average precision values per method.}
A box plot for the post-competition OverFeat result is not shown because per-class APs are not yet available (per-class APs for R-CNN are in \tableref{classaps} and also included in the tech report source uploaded to arXiv.org; see \texttt{R-CNN-ILSVRC2013-APs.txt}).
The red line marks the median AP, the box bottom and top are the 25th and 75th percentiles.
The whiskers extend to the min and max AP of each method.
Each AP is plotted as a green dot over the whiskers (best viewed digitally with zoom).
}
\figlabel{ilsvrc13}
\end{figure*}

\section{Visualization, ablation, and modes of error}
\seclabel{experiments}

\subsection{Visualizing learned features}
\seclabel{vis}

\begin{figure*}[t!]
\centering
\includegraphics[width=\linewidth,clip=true,trim=0 0 0.2em 0]{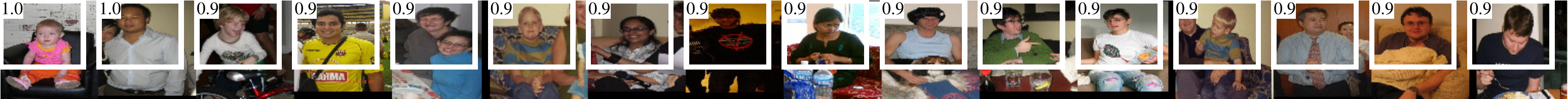}\\
\includegraphics[width=\linewidth,clip=true,trim=0 0 0.2em 0]{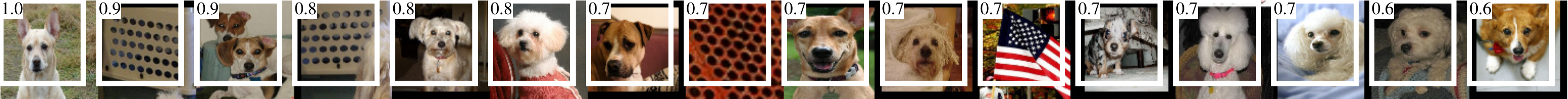}\\
\includegraphics[width=\linewidth,clip=true,trim=0 0 0.2em 0]{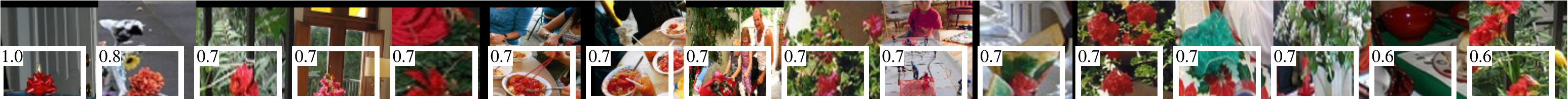}\\
\includegraphics[width=\linewidth,clip=true,trim=0 0 0.2em 0]{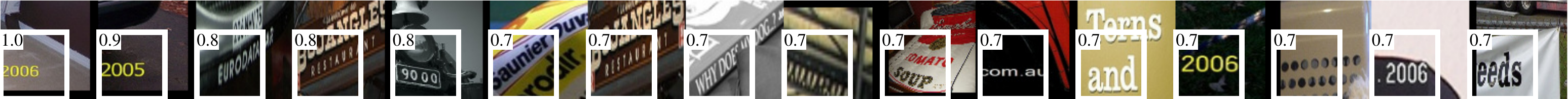}\\
\includegraphics[width=\linewidth,clip=true,trim=0 0 0.2em 0]{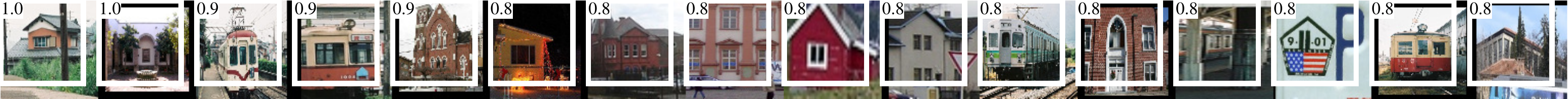}\\
\includegraphics[width=\linewidth,clip=true,trim=0 0 0.2em 0]{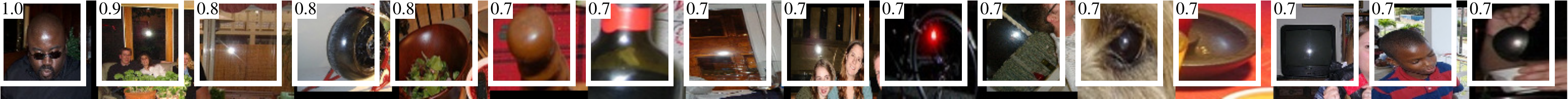}
\caption{\textbf{Top regions for six \pool{5} units.} 
Receptive fields and activation values are drawn in white. 
Some units are aligned to concepts, such as people (row 1) or text (4).
Other units capture texture and material properties, such as dot arrays (2) and specular reflections (6).
}
\figlabel{vislittle}
\vspace{-0.5em}
\end{figure*}
\begin{table*}[t!]
\centering
\renewcommand{\arraystretch}{1.2}
\renewcommand{\tabcolsep}{1.2mm}
\resizebox{\linewidth}{!}{
\begin{tabular}{@{}l|r*{19}{c}|c@{}}
\textbf{VOC 2007 test}        & aero      & bike      & bird      & boat      & bottle     & bus        & car        & cat        & chair      & cow        & table      & dog        & horse      & mbike      & person     & plant      & sheep      & sofa       & train      & tv         & mAP       \\
\hline
R-CNN \pool{5}  &  51.8  &  60.2  &  36.4  &  27.8  &  23.2  &  52.8  &  60.6  &  49.2  &  18.3  &  47.8  &  44.3  &  40.8  &  56.6  &  58.7  &  42.4  &  23.4  &  46.1  &  36.7  &  51.3  &  55.7  &  44.2 \\
R-CNN \fc{6}  &  59.3  &  61.8  &  43.1  &  34.0  &  25.1  &  53.1  &  60.6  &  52.8  &  21.7  &  47.8  &  42.7  &  47.8  &  52.5  &  58.5  &  44.6  &  25.6  &  48.3  &  34.0  &  53.1  &  58.0  &  46.2 \\
R-CNN \fc{7}  &  57.6  &  57.9  &  38.5  &  31.8  &  23.7  &  51.2  &  58.9  &  51.4  &  20.0  &  50.5  &  40.9  &  46.0  &  51.6  &  55.9  &  43.3  &  23.3  &  48.1  &  35.3  &  51.0  &  57.4  &  44.7 \\
\hline
R-CNN FT \pool{5}  &  58.2  &  63.3  &  37.9  &  27.6  &  26.1  &  54.1  &  66.9  &  51.4  &  26.7  &  55.5  &  43.4  &  43.1  &  57.7  &  59.0  &  45.8  &  28.1  &  50.8  &  40.6  &  53.1  &  56.4  &  47.3 \\
R-CNN FT \fc{6}  &  63.5  &  66.0  &  47.9  &  37.7  &  29.9  &  62.5  &  70.2  &  60.2  &  32.0  &  57.9  &  47.0  &  53.5  &  60.1  &  64.2  &  52.2  &  31.3  &  55.0  &  50.0  &  57.7  &  63.0  &  53.1 \\
R-CNN FT \fc{7}  &  64.2  &  69.7  &  50.0  &  41.9  &  32.0  &  62.6  &  71.0  &  60.7  &  32.7  &  58.5  &  46.5  &  56.1  &  60.6  &  66.8  &  54.2  &  31.5  &  52.8  &  48.9  &  57.9  &  64.7  &  54.2 \\
\hline
R-CNN FT \fc{7} BB  &  \bf{68.1}  &  \bf{72.8}  &  \bf{56.8}  &  \bf{43.0}  &  \bf{36.8}  &  \bf{66.3}  &  \bf{74.2}  &  \bf{67.6}  &  \bf{34.4}  &  \bf{63.5}  &  \bf{54.5}  &  \bf{61.2}  &  \bf{69.1}  &  \bf{68.6}  &  \bf{58.7}  &  \bf{33.4}  &  \bf{62.9}  &  \bf{51.1}  &  \bf{62.5}  &  \bf{64.8}  &  \bf{58.5} \\
\hline
\hline
DPM v5 \cite{release5}  &  33.2  &  60.3  &  10.2  &  16.1  &  27.3  &  54.3  &  58.2  &  23.0  &  20.0  &  24.1  &  26.7  &  12.7  &  58.1  &  48.2  &  43.2  &  12.0  &  21.1  &  36.1  &  46.0  &  43.5  &  33.7 \\
DPM ST \cite{lim2013sketch}  &  23.8  &  58.2  &  10.5  &  \phz8.5  &  27.1  &  50.4  &  52.0  &  \phz7.3  &  19.2  &  22.8  &  18.1  &  \phz8.0  &  55.9  &  44.8  &  32.4  &  13.3  &  15.9  &  22.8  &  46.2  &  44.9  &  29.1 \\
DPM HSC \cite{HSC}  &  32.2  &  58.3  &  11.5  &  16.3  &  30.6  &  49.9  &  54.8  &  23.5  &  21.5  &  27.7  &  34.0  &  13.7  &  58.1  &  51.6  &  39.9  &  12.4  &  23.5  &  34.4  &  47.4  &  45.2  &  34.3 \\
\end{tabular}
}
\caption{\textbf{Detection average precision (\%) on VOC 2007 test.}
Rows 1-3 show R-CNN performance without fine-tuning.
Rows 4-6 show results for the CNN pre-trained on ILSVRC 2012
and then fine-tuned (FT) on VOC 2007 trainval.
Row 7 includes a simple bounding-box regression (BB) stage that reduces localization errors (\secref{bboxreg}).
Rows 8-10 present DPM methods as a strong baseline.  
The first uses only HOG, while the next two
use different feature learning approaches to augment or replace HOG. 
}
\tablelabel{voc2007}
\end{table*}

\begin{table*}[t!]
\centering
\renewcommand{\arraystretch}{1.2}
\renewcommand{\tabcolsep}{1.2mm}
\resizebox{\linewidth}{!}{
\begin{tabular}{@{}l|r*{19}{c}|c@{}}
\textbf{VOC 2007 test}        & aero      & bike      & bird      & boat      & bottle     & bus        & car        & cat        & chair      & cow        & table      & dog        & horse      & mbike      & person     & plant      & sheep      & sofa       & train      & tv         & mAP       \\
\hline
R-CNN T-Net&  64.2  &  69.7  &  50.0  &  41.9  &  32.0  &  62.6  &  71.0  &  60.7  &  32.7  &  58.5  &  46.5  &  56.1  &  60.6  &  66.8  &  54.2  &  31.5  &  52.8  &  48.9  &  57.9  &  64.7  &  54.2 \\
R-CNN T-Net BB &  68.1  &  72.8  &  56.8  &  43.0  &  36.8  &  66.3  &  74.2  &  67.6  &  34.4  &  63.5  &  54.5  &  61.2  &  69.1  &  68.6  &  58.7  &  33.4  &  62.9  &  51.1  &  62.5  &  64.8  &  58.5 \\
\hline
R-CNN O-Net & 71.6 & 73.5 & 58.1 & 42.2 & 39.4 & 70.7 & 76.0 & 74.5 & 38.7 & 71.0 & 56.9 & 74.5 & 67.9 & 69.6 & 59.3 & \bf{35.7} & 62.1 & 64.0 & 66.5 & \bf{71.2} & 62.2 \\
R-CNN O-Net BB & \bf{73.4} & \bf{77.0} & \bf{63.4} & \bf{45.4} & \bf{44.6} & \bf{75.1} & \bf{78.1} & \bf{79.8} & \bf{40.5} & \bf{73.7} & \bf{62.2} & \bf{79.4} & \bf{78.1} & \bf{73.1} & \bf{64.2} & 35.6 & \bf{66.8} & \bf{67.2} & \bf{70.4} & 71.1 & \bf{66.0} \\
\end{tabular}
}
\caption{\textbf{Detection average precision (\%) on VOC 2007 test for two different CNN architectures.}
The first two rows are results from \tableref{voc2007} using Krizhevsky \etal's architecture (T-Net).
Rows three and four use the recently proposed 16-layer architecture from Simonyan and Zisserman (O-Net) \cite{vggverydeep}.
}
\tablelabel{voc2007oxfordnet}
\vspace{-1em}
\end{table*}

First-layer filters can be visualized directly
and are easy to understand \cite{krizhevsky2012imagenet}. They capture
oriented edges and opponent colors. Understanding the subsequent layers
is more challenging. Zeiler and Fergus present a visually attractive deconvolutional
approach in \cite{zeiler2011adaptive}. We propose a simple (and complementary) non-parametric method that directly shows what the network learned. 

The idea is to single out a particular unit (feature) in the network and use it as if it were an object
detector in its own right. That is, we compute the unit's activations on a
large set of held-out region proposals (about 10 million), sort the proposals from
highest to lowest activation, perform non-maximum suppression, and then display the top-scoring regions. Our method lets the selected unit
``speak for itself'' by showing exactly which inputs it fires on. We avoid
averaging in order to see different visual modes and gain
insight into the invariances computed by the unit.

We visualize units from layer \pool{5}, which is the max-pooled output of
the network's fifth and final convolutional layer.  
The \pool{5} feature map is $6 \times 6 \times 256 = 9216$-dimensional. 
Ignoring boundary effects, each \pool{5} unit has a receptive field of $195
\times 195$ pixels in the original $227 \times 227$ pixel input.
A central \pool{5} unit has a nearly global view, while one near the edge has a smaller, clipped support.

Each row in \figref{vislittle} displays the top 16 activations for a \pool{5} unit from a CNN that we fine-tuned on VOC 2007 trainval.
Six of the 256 functionally unique units are visualized (\asecref{extravis} includes more).
These units were selected to show a representative sample of what the network learns.
In the second row, we see a unit that fires on dog faces and dot arrays.
The unit corresponding to the third row is a red blob detector.
There are also detectors for human faces and more abstract patterns such as text and triangular structures with windows.
The network appears to learn a representation that combines a small number of class-tuned features together with a distributed representation of shape, texture, color, and material properties.
The subsequent fully connected layer \fc{6} has the ability to model a large set of compositions of these rich features. 

\subsection{Ablation studies}
\seclabel{ablation}
\paragraph{Performance layer-by-layer, without fine-tuning.}
To understand which layers are critical for detection performance, we analyzed results on the VOC 2007 dataset for each of the CNN's last three layers.
Layer \pool{5} was briefly described in \secref{vis}.
The final two layers are summarized below.

Layer \fc{6} is fully connected to \pool{5}. To compute features,
it multiplies a $4096 \times 9216$ weight matrix by the \pool{5} feature map 
(reshaped as a 9216-dimensional vector) and then adds a vector of
biases. 
This intermediate vector is component-wise half-wave rectified ($x \leftarrow \max(0, x)$).

Layer \fc{7} is the final layer of the network. It is implemented
by multiplying the features computed by \fc{6} by a $4096 \times 4096$
weight matrix, and similarly adding a vector of biases and 
applying half-wave rectification.

We start by looking at results from the CNN \emph{without fine-tuning} on PASCAL, \ie all CNN parameters were pre-trained on ILSVRC 2012 only.
Analyzing performance layer-by-layer (\tableref{voc2007} rows 1-3) reveals that
features from \fc{7} generalize worse than 
features from \fc{6}. This means that 29\%, or about 16.8 million,
of the CNN's parameters can be removed without degrading mAP. 
More surprising is that removing \emph{both} \fc{7} and
\fc{6} produces quite good results even though \pool{5}
features are computed using \emph{only 6\%}
of the CNN's parameters. Much of the
CNN's representational power comes from its convolutional layers,
rather than from the much larger densely connected layers. This finding
suggests potential utility in computing a dense feature map,
in the sense of HOG, of an arbitrary-sized image by using only the convolutional layers
of the CNN. This representation would enable experimentation with sliding-window detectors,
including DPM, on top of \pool{5} features.


\paragraph{Performance layer-by-layer, with fine-tuning.}

We now look at results from our CNN after having fine-tuned its parameters on VOC 2007 trainval.
The improvement is striking (\tableref{voc2007} rows 4-6):
fine-tuning increases mAP by 8.0 percentage points to 54.2\%. 
The boost from fine-tuning is much larger for \fc{6} and \fc{7} than for \pool{5},
which suggests that the \pool{5} features learned from ImageNet are general and that most of the improvement is gained from learning domain-specific non-linear classifiers on top of them.

\paragraph{Comparison to recent feature learning methods.}
Relatively few feature learning methods have been tried on PASCAL VOC detection.
We look at two recent approaches that build on deformable part models. For reference, we also include results for the standard HOG-based DPM \cite{release5}.

The first DPM feature learning method, DPM ST \cite{lim2013sketch},
augments HOG features with histograms of ``sketch
token'' probabilities. Intuitively, a sketch token is
a tight distribution of contours passing through the center of
an image patch.
Sketch token probabilities are computed at
each pixel by a random forest that was trained to classify
$35 \times 35$ pixel patches into one of 150 sketch tokens or background.

The second method, DPM HSC \cite{HSC}, replaces HOG with histograms of sparse codes (HSC). To compute an HSC, sparse code
activations are solved for at each pixel using a learned dictionary
of 100 $7 \times 7$ pixel (grayscale) atoms. The resulting activations are 
rectified in three ways (full and both half-waves), spatially pooled,
unit $\ell_2$ normalized,
and then power transformed ($x \leftarrow \textrm{sign}(x)|x|^\alpha$).

All R-CNN variants strongly outperform the three DPM baselines (\tableref{voc2007} rows 8-10), including
the two that use feature learning. Compared to the latest version of
DPM, which uses only HOG features, our mAP is more than 20 percentage points higher: 54.2\% \vs 33.7\%---\emph{a 61\% relative improvement}.
The combination of HOG and sketch tokens 
yields 2.5 mAP points over HOG alone, while HSC improves over HOG
by 4 mAP points (when compared internally
to their private DPM baselines---both use non-public implementations of DPM
that underperform the open source version \cite{release5}). These methods achieve mAPs of 29.1\% and 34.3\%, respectively.

\subsection{Network architectures}
\seclabel{netarch}
Most results in this paper use the network architecture from Krizhevsky \etal \cite{krizhevsky2012imagenet}.
However, we have found that the choice of architecture has a large effect on R-CNN detection performance.
In \tableref{voc2007oxfordnet} we show results on VOC 2007 test using the 16-layer deep network recently proposed by Simonyan and Zisserman \cite{vggverydeep}.
This network was one of the top performers in the recent ILSVRC 2014 classification challenge.
The network has a homogeneous structure consisting of 13 layers of $3 \times 3$ convolution kernels,
with five max pooling layers interspersed, and topped with three fully-connected layers.
We refer to this network as ``O-Net'' for OxfordNet and the baseline as ``T-Net'' for TorontoNet.

To use O-Net in R-CNN, we downloaded the publicly available pre-trained network weights for the \texttt{VGG\_ILSVRC\_16\_layers} model from the Caffe Model Zoo.\footnote{\url{https://github.com/BVLC/caffe/wiki/Model-Zoo}}
We then fine-tuned the network using the same protocol as we used for T-Net.
The only difference was to use smaller minibatches (24 examples) as required in order to fit within GPU memory.
The results in \tableref{voc2007oxfordnet} show that R-CNN with O-Net substantially outperforms R-CNN with T-Net, increasing mAP from 58.5\% to 66.0\%.
However there is a considerable drawback in terms of compute time, with the forward pass of O-Net taking roughly 7 times longer than T-Net.

\subsection{Detection error analysis}
\seclabel{analysis}
We applied the excellent detection analysis tool from Hoiem \etal
\cite{hoiem2012diagnosing} in order to reveal our
method's error modes, understand how fine-tuning changes them, and
to see how our error types compare with DPM.  A full summary of the
analysis tool is beyond the scope of this paper and we encourage readers to consult \cite{hoiem2012diagnosing} to
understand some finer details (such as ``normalized AP''). Since the
analysis is best absorbed in the context of the associated plots, we
present the discussion within the captions of \figref{fp-type-dist}
and \figref{obj-char}.
\begin{figure}[h!]
\def \sz {1.2in}
\def \sp {-0.7em}
\centering
\includegraphics[height=\sz]{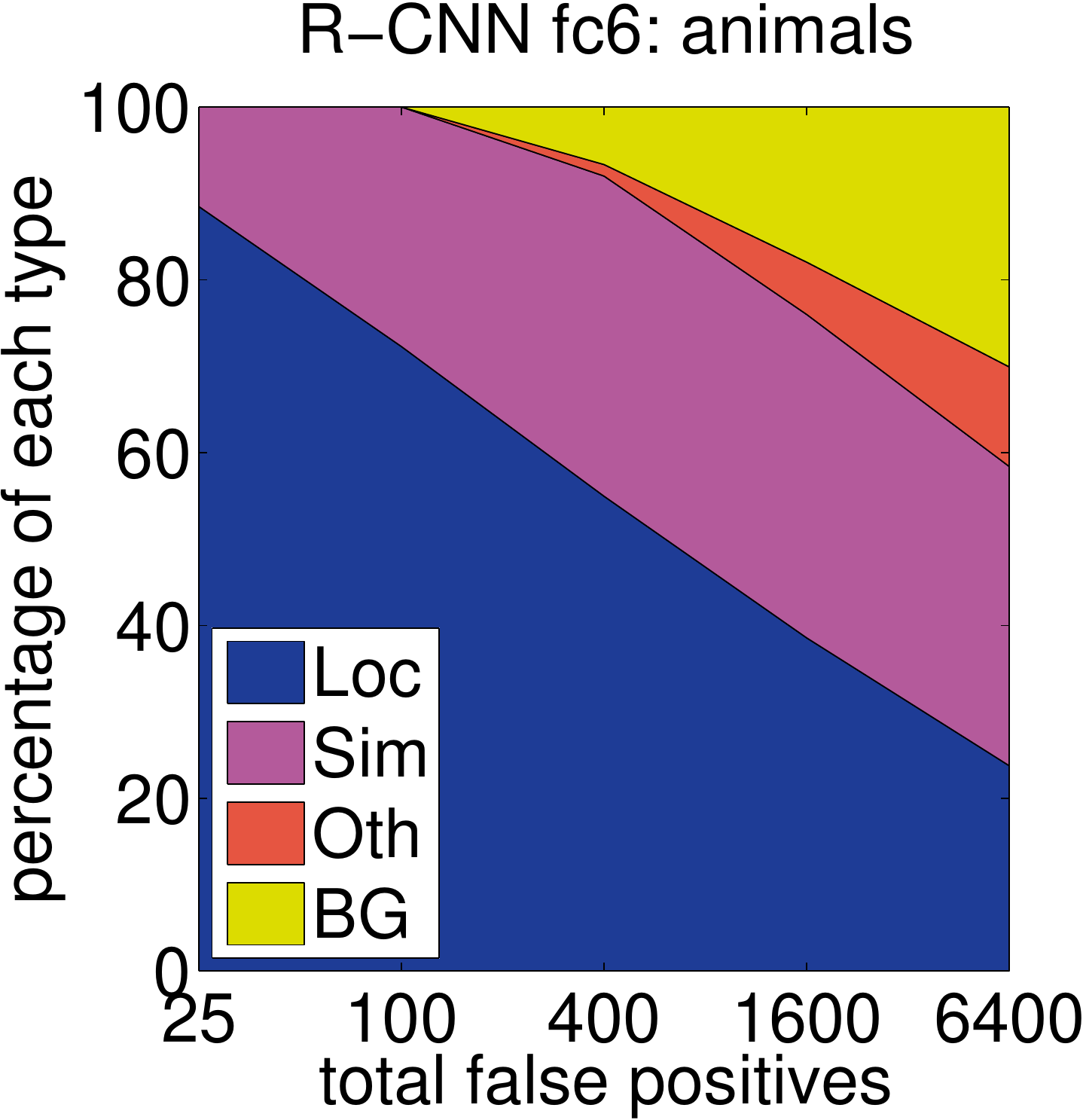}
\hspace{\sp}
\includegraphics[height=\sz,clip=true,trim=1.05cm 0 0 0]{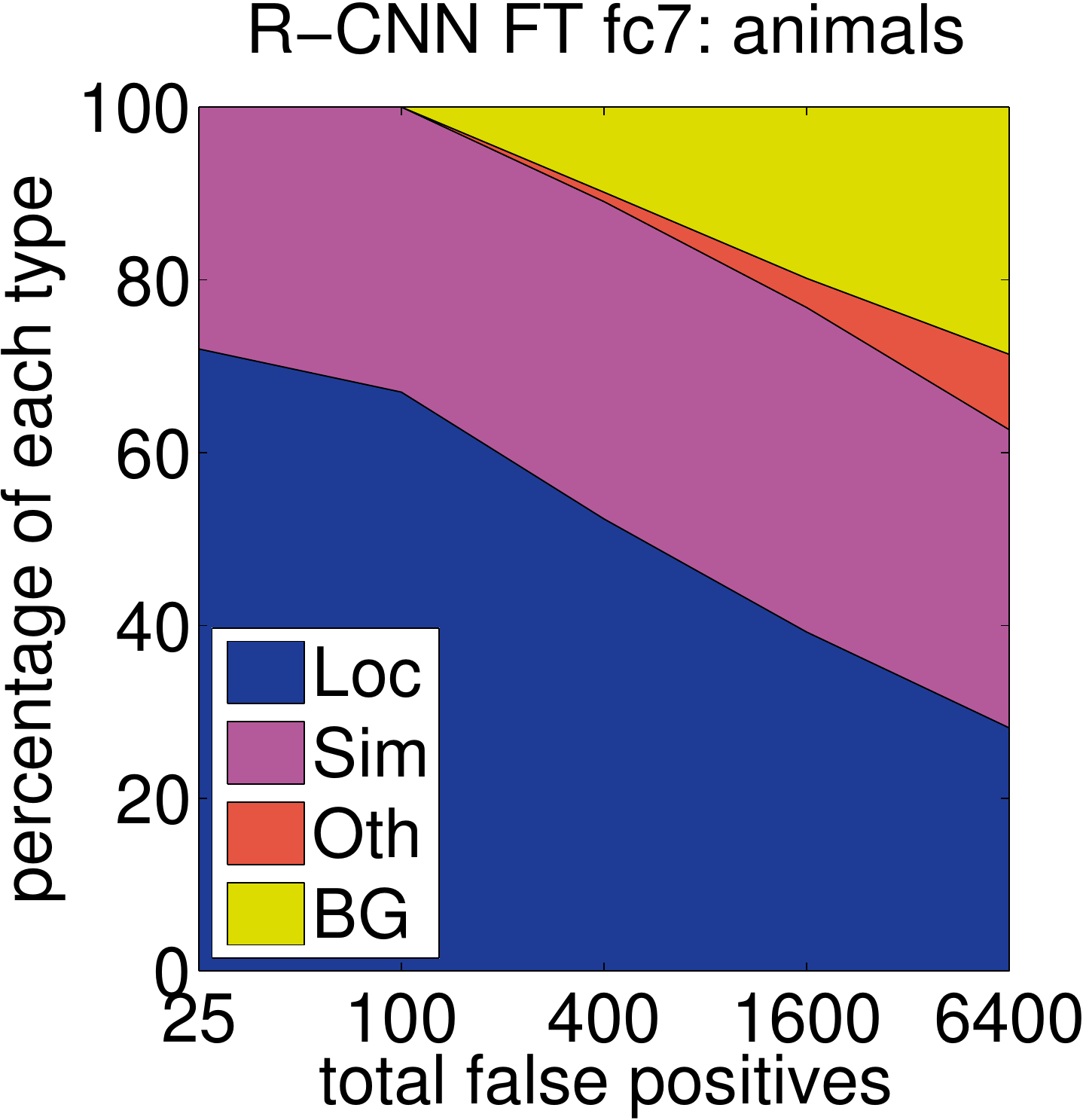}
\hspace{\sp}
\includegraphics[height=\sz,clip=true,trim=1.05cm 0 0 0]{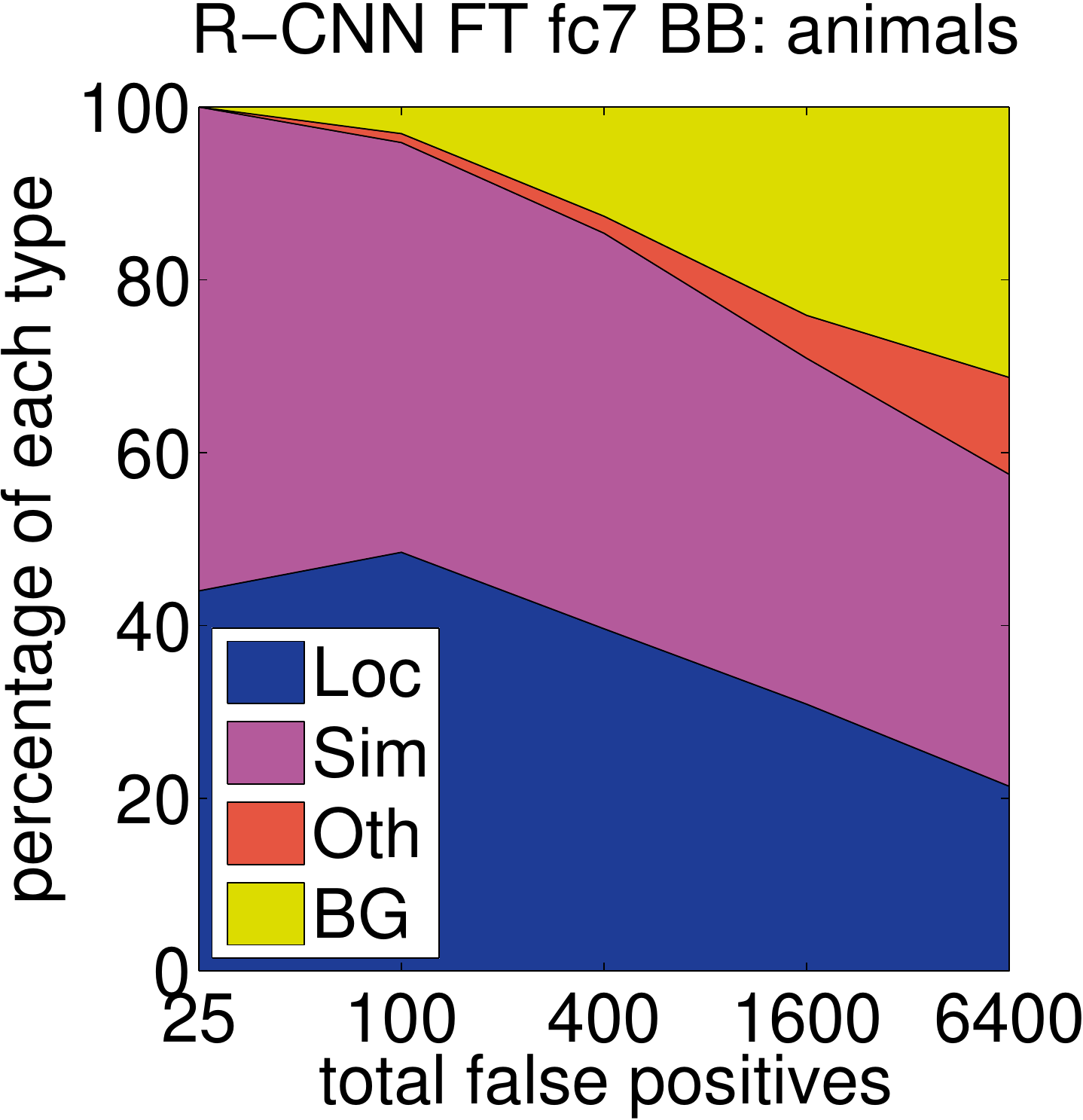}\\
\vspace{2pt}
\includegraphics[height=\sz]{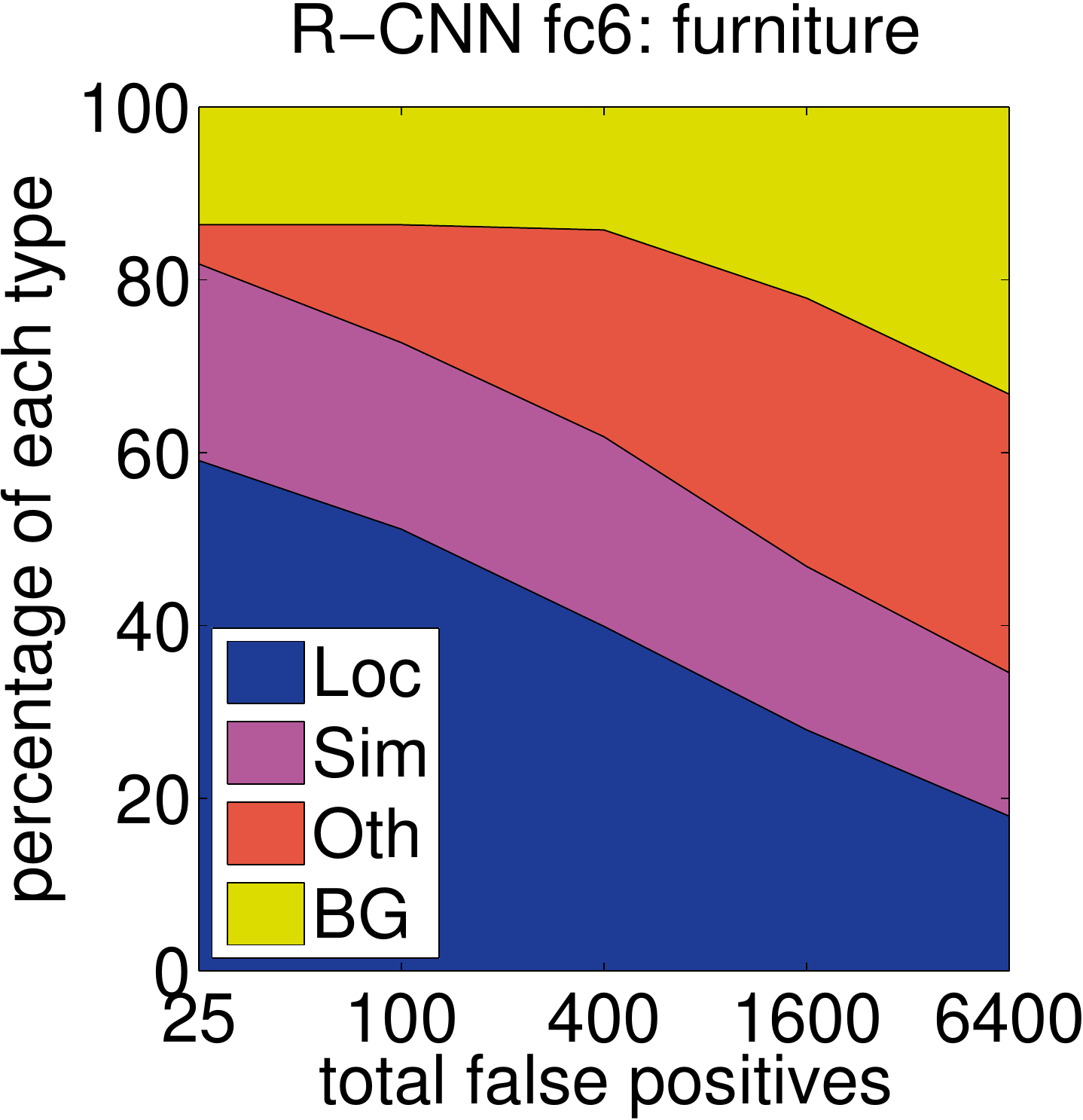}
\hspace{\sp}
\includegraphics[height=\sz,clip=true,trim=1.05cm 0 0 0]{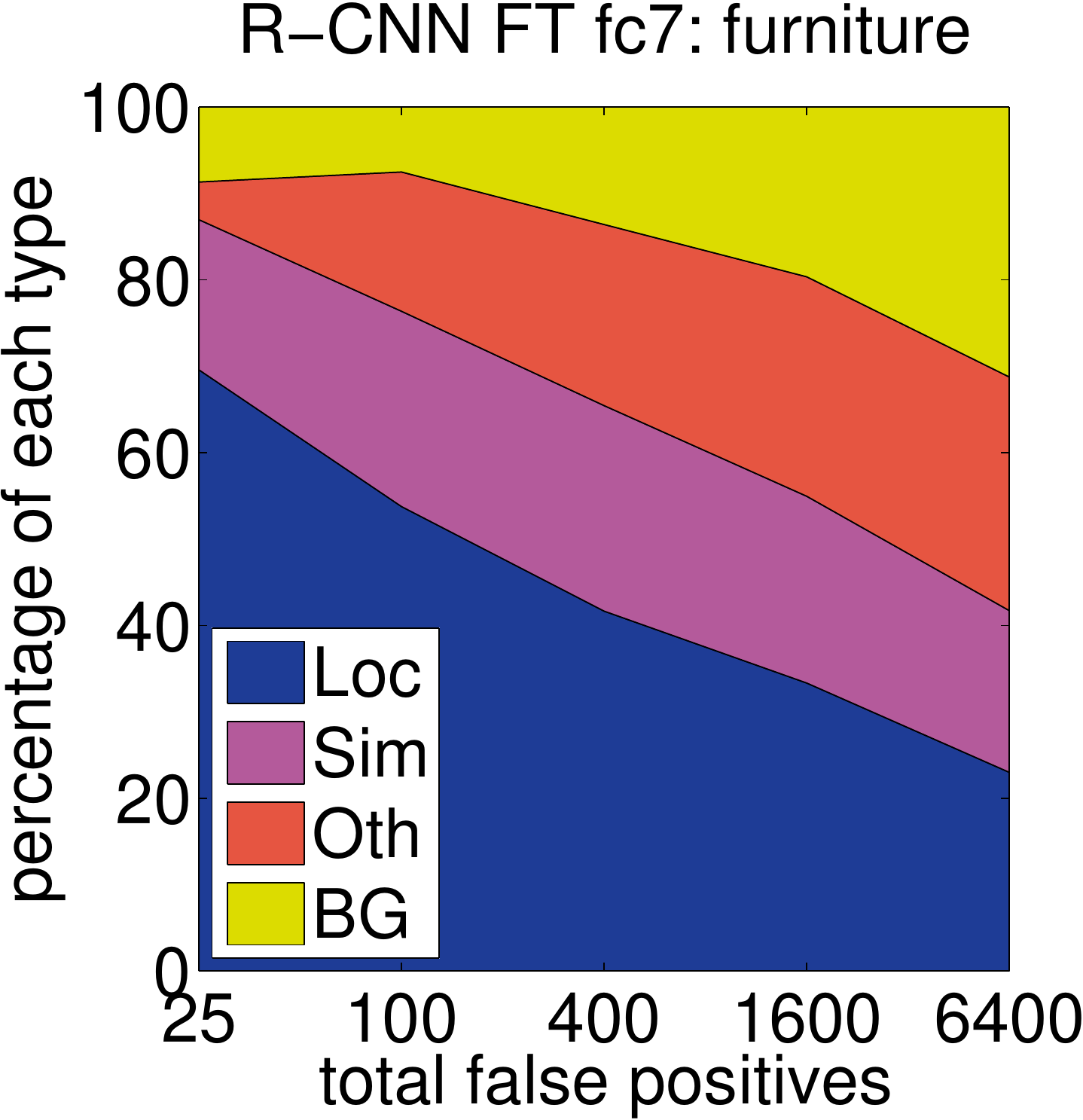}
\hspace{\sp}
\includegraphics[height=\sz,clip=true,trim=1.05cm 0 0 0]{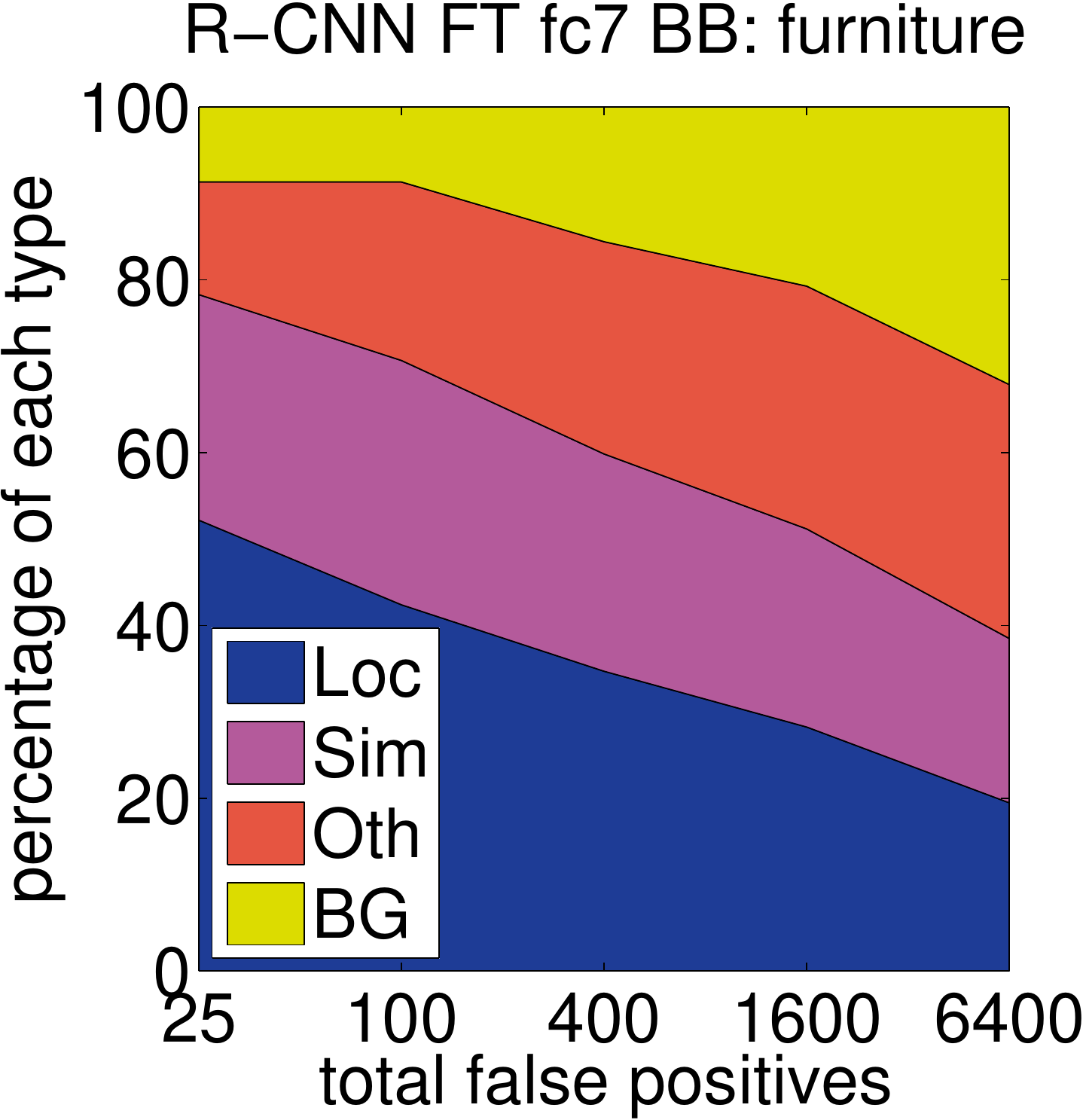}\\
\caption{\textbf{Distribution of top-ranked false positive (FP)
types.} Each plot shows the evolving distribution of FP types as
more FPs are considered in order of decreasing score. Each FP is
categorized into 1 of 4 types: Loc---poor localization (a detection with an IoU overlap
with the correct class between 0.1 and 0.5, or a duplicate); Sim---confusion with a similar category;
Oth---confusion with a dissimilar object category; BG---a FP that fired
on background.  Compared with DPM (see \cite{hoiem2012diagnosing}), significantly more of our errors
result from poor localization, rather than confusion with background
or other object classes, indicating that the CNN features are much
more discriminative than HOG. Loose localization likely results from our
use of bottom-up region proposals and the positional invariance
learned from pre-training the CNN for whole-image classification.
Column three shows how our simple bounding-box regression method fixes many localization errors.}
\figlabel{fp-type-dist}
\vspace{-1em}
\end{figure}

\begin{figure*}[t!]
\centering
\def \sz {1.275in}
\includegraphics[height=\sz]{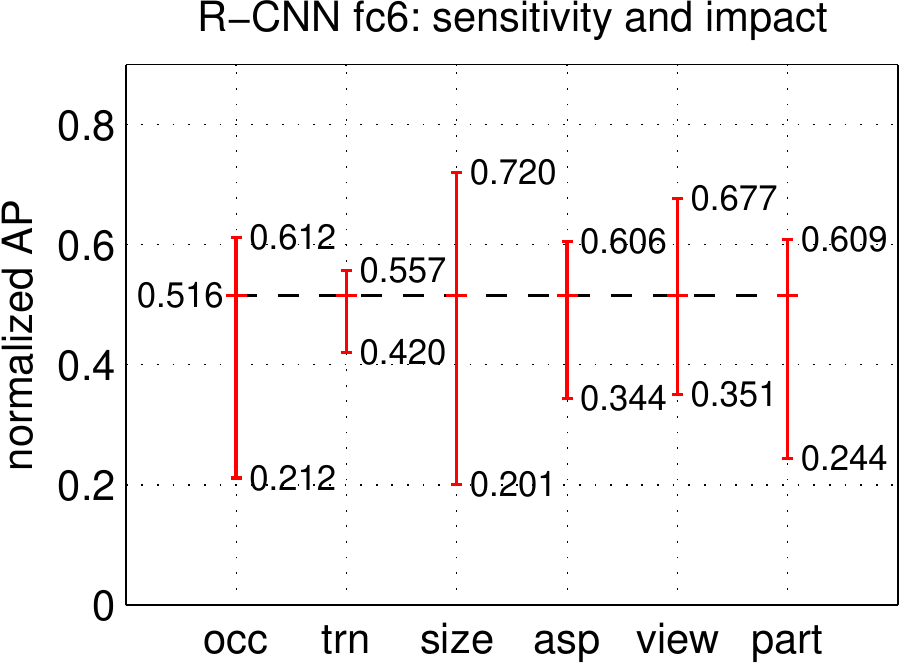}
\hspace{0em}
\includegraphics[height=\sz,clip=true,trim=0.5cm 0 0 0]{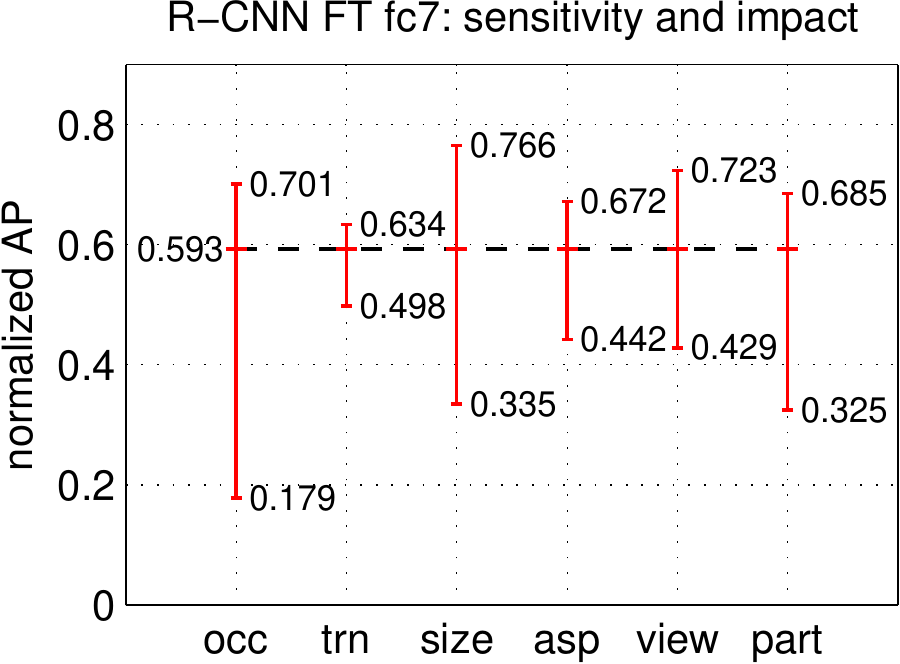}
\hspace{0em}
\includegraphics[height=\sz,clip=true,trim=0.5cm 0 0 0]{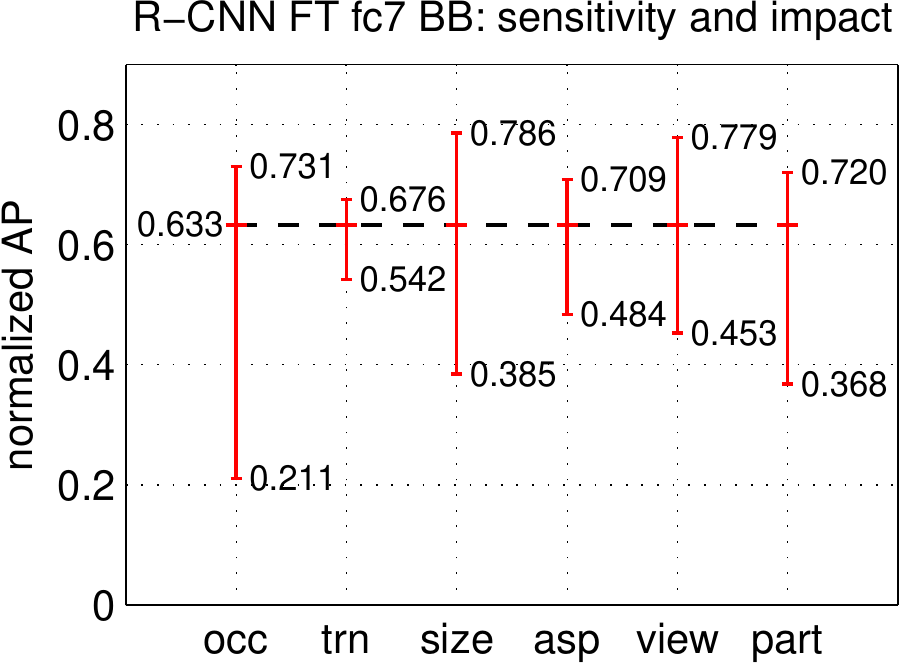}
\hspace{0em}
\includegraphics[height=\sz,clip=true,trim=0.5cm 0 0 0]{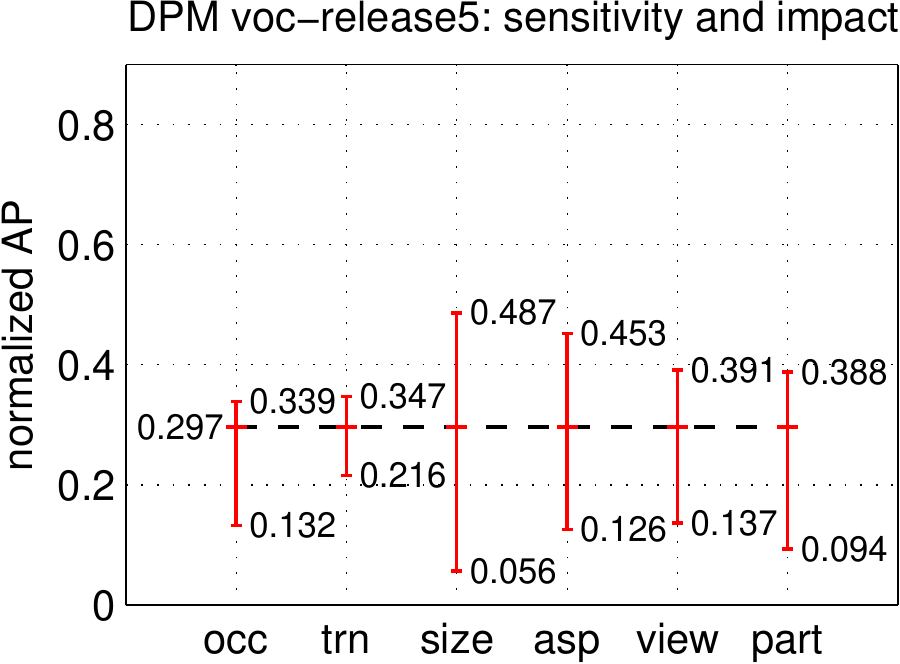}
\caption{\textbf{Sensitivity to object characteristics.} Each plot shows the mean (over classes)
normalized AP (see \cite{hoiem2012diagnosing}) for the highest and lowest performing subsets within six different
object characteristics (occlusion, truncation, bounding-box area, aspect ratio, viewpoint, part visibility).
We show plots for our method (R-CNN) with and without fine-tuning (FT) and bounding-box regression (BB) as well as for DPM voc-release5.
Overall, fine-tuning does not reduce sensitivity (the difference between max and min), but does substantially improve
both the highest and lowest performing subsets for nearly all characteristics. This indicates that fine-tuning does
more than simply improve the lowest performing subsets for aspect ratio and bounding-box area, as one might
conjecture based on how we warp network inputs. Instead, fine-tuning improves robustness
for all characteristics including occlusion, truncation, viewpoint, and part visibility.}
\figlabel{obj-char}
\vspace{-1em}
\end{figure*}

\subsection{Bounding-box regression}
\seclabel{bboxreg}
Based on the error analysis, we implemented a simple method to reduce localization errors.
Inspired by the bounding-box regression employed in DPM \cite{lsvm-pami}, we train a linear regression model to predict a new detection window given the \pool{5} features for a selective search region proposal.
Full details are given in \asecref{bboxreg}.
Results in \tableref{voc2010}, \tableref{voc2007}, and \figref{fp-type-dist} show that this simple approach fixes a large number of mislocalized detections, boosting mAP by 3 to 4 points.

\subsection{Qualitative results}
Qualitative detection results on ILSVRC2013 are presented in \figref{examples1} and \figref{examples2} at the end of the paper.
Each image was sampled randomly from the \valb set and all detections from all detectors with a precision greater than 0.5 are shown.
Note that these are not curated and give a realistic impression of the detectors in action.
More qualitative results are presented in \figref{cexamples1} and \figref{cexamples2}, but these have been curated.
We selected each image because it contained interesting, surprising, or amusing results.
Here, also, all detections at precision greater than 0.5 are shown.

\section{The ILSVRC2013 detection dataset}
\seclabel{ilsvrc}
In \secref{detection} we presented results on the ILSVRC2013 detection dataset.
This dataset is less homogeneous than PASCAL VOC, requiring choices about how to use it.
Since these decisions are non-trivial, we cover them in this section.

\subsection{Dataset overview}
The ILSVRC2013 detection dataset is split into three sets: \train (395,918), \val (20,121), and \test (40,152), where the number of images in each set is in parentheses.
The \val and \test splits are drawn from the same image distribution.
These images are scene-like and similar in complexity (number of objects, amount of clutter, pose variability, \etc) to PASCAL VOC images.
The \val and \test splits are exhaustively annotated, meaning that in each image all instances from all 200 classes are labeled with bounding boxes.
The \train set, in contrast, is drawn from the ILSVRC2013 \emph{classification} image distribution.
These images have more variable complexity with a skew towards images of a single centered object.
Unlike \val and \test, the \train images (due to their large number) are not exhaustively annotated.
In any given \train image, instances from the 200 classes may or may not be labeled.
In addition to these image sets, each class has an extra set of negative images.
Negative images are manually checked to validate that they do not contain any instances of their associated class.
The negative image sets were not used in this work.
More information on how ILSVRC was collected and annotated can be found in \cite{DengCHI14,HJL_AAAI12}.

The nature of these splits presents a number of choices for training R-CNN.
The \train images cannot be used for hard negative mining, because annotations are not exhaustive.
Where should negative examples come from?
Also, the \train images have different statistics than \val and \test.
Should the \train images be used at all, and if so, to what extent?
While we have not thoroughly evaluated a large number of choices, we present what seemed like the most obvious path based on previous experience.

Our general strategy is to rely heavily on the val set and use some of the \train images as an auxiliary source of positive examples.
To use \val for both training and validation, we split it into roughly equally sized ``\vala'' and ``\valb'' sets.
Since some classes have very few examples in \val (the smallest has only 31 and half have fewer than 110), it is important to produce an approximately class-balanced partition.
To do this, a large number of candidate splits were generated and the one with the smallest maximum relative class imbalance was selected.\footnote{Relative imbalance is measured as $|a - b| / (a+b)$ where $a$ and $b$ are class counts in each half of the split.}
Each candidate split was generated by clustering \val images using their class counts as features, followed by a randomized local search that may improve the split balance.
The particular split used here has a maximum relative imbalance of about 11\% and a median relative imbalance of 4\%.
The \vala/\valb split and code used to produce them will be publicly available to allow other researchers to compare their methods on the \val splits used in this report.

\subsection{Region proposals}
We followed the same region proposal approach that was used for detection on PASCAL.
Selective search \cite{UijlingsIJCV2013} was run in ``fast mode'' on each image in \vala, \valb, and \test (but not on images in \train).
One minor modification was required to deal with the fact that selective search is not scale invariant and so the number of regions produced depends on the image resolution.
ILSVRC image sizes range from very small to a few that are several mega-pixels, and so we resized each image to a fixed width (500 pixels) before running selective search.
On \val, selective search resulted in an average of 2403 region proposals per image with a 91.6\% recall of all ground-truth bounding boxes (at 0.5 IoU threshold).
This recall is notably lower than in PASCAL, where it is approximately 98\%, indicating significant room for improvement in the region proposal stage.

\subsection{Training data}
For training data, we formed a set of images and boxes that includes all selective search and ground-truth boxes from \vala together with up to $N$ ground-truth boxes per class from \train (if a class has fewer than $N$ ground-truth boxes in \train, then we take all of them).
We'll call this dataset of images and boxes \trainvalN.
In an ablation study, we show mAP on \valb for $N \in \{0, 500, 1000\}$ (\secref{ilsvrcablation}).

Training data is required for three procedures in R-CNN: (1) CNN fine-tuning, (2) detector SVM training, and (3) bounding-box regressor training.
CNN fine-tuning was run for 50k SGD iteration on \trainvalN using the exact same settings as were used for PASCAL.
Fine-tuning on a single NVIDIA Tesla K20 took 13 hours using Caffe.
For SVM training, all ground-truth boxes from \trainvalN were used as positive examples for their respective classes.
Hard negative mining was performed on a randomly selected subset of 5000 images from \vala.
An initial experiment indicated that mining negatives from all of \vala, versus a 5000 image subset (roughly half of it), resulted in only a 0.5 percentage point drop in mAP, while cutting SVM training time in half.
No negative examples were taken from \train because the annotations are not exhaustive.
The extra sets of verified negative images were not used.
The bounding-box regressors were trained on \vala.

\begin{table*}[t]
\centering
\renewcommand{\arraystretch}{1.2}
\renewcommand{\tabcolsep}{1.2mm}
\resizebox{\linewidth}{!}{
\begin{tabular}{@{}ccccccc|cc@{}}
  \textbf{test set}
& \valb 
& \valb 
& \valb 
& \valb 
& \valb 
& \valb 
& \test 
& \test \\
  \textbf{SVM training set}
& \vala
& \trainvala
& \trainvalb
& \trainvalb
& \trainvalb
& \trainvalb
& \trainvalab
& \trainvalab \\
  \textbf{CNN fine-tuning set}
& n/a 
& n/a
& n/a
& \vala
& \trainvalb
& \trainvalb 
& \trainvalb
& \trainvalb \\
  \textbf{bbox reg set}
& n/a 
& n/a
& n/a
& n/a
& n/a
& \vala 
& n/a
& \val \\
  \textbf{CNN feature layer}
& \fc{6}
& \fc{6}
& \fc{6}
& \fc{7}
& \fc{7}
& \fc{7}
& \fc{7}
& \fc{7} \\
\hline
  \textbf{mAP}
& 20.9
& 24.1
& 24.1
& 26.5
& 29.7
& \textbf{31.0}
& 30.2
& \textbf{31.4} \\
  \textbf{median AP}
& 17.7
& 21.0
& 21.4
& 24.8
& 29.2
& \textbf{29.6}
& 29.0
& \textbf{30.3} \\
\end{tabular}
}
\caption{\textbf{ILSVRC2013 ablation study} of data usage choices, fine-tuning, and bounding-box regression.}
\tablelabel{ablation}
\end{table*}

\subsection{Validation and evaluation}
Before submitting results to the evaluation server, we validated data usage choices and the effect of fine-tuning and bounding-box regression on the \valb set using the training data described above.
All system hyperparameters (\eg, SVM C hyperparameters, padding used in region warping, NMS thresholds, bounding-box regression hyperparameters) were fixed at the same values used for PASCAL.
Undoubtedly some of these hyperparameter choices are slightly suboptimal for ILSVRC, however the goal of this work was to produce a preliminary R-CNN result on ILSVRC without extensive dataset tuning.
After selecting the best choices on \valb, we submitted exactly two result files to the ILSVRC2013 evaluation server.
The first submission was without bounding-box regression and the second submission was with bounding-box regression.
For these submissions, we expanded the SVM and bounding-box regressor training sets to use \trainvalab and \val, respectively.
We used the CNN that was fine-tuned on \trainvalb to avoid re-running fine-tuning and feature computation.

\subsection{Ablation study}
\seclabel{ilsvrcablation}
\tableref{ablation} shows an ablation study of the effects of different amounts of training data, fine-tuning, and bounding-box regression.
A first observation is that mAP on \valb matches mAP on \test very closely.
This gives us confidence that mAP on \valb is a good indicator of test set performance.
The first result, 20.9\%, is what R-CNN achieves using a CNN pre-trained on the ILSVRC2012 classification dataset (no fine-tuning) and given access to the small amount of training data in \vala (recall that half of the classes in \vala have between 15 and 55 examples).
Expanding the training set to \trainvalN improves performance to 24.1\%, with essentially no difference between $N = 500$ and $N = 1000$.
Fine-tuning the CNN using examples from just \vala gives a modest improvement to 26.5\%, however there is likely significant overfitting due to the small number of positive training examples.
Expanding the fine-tuning set to \trainvalb, which adds up to 1000 positive examples per class from the train set, helps significantly, boosting mAP to 29.7\%.
Bounding-box regression improves results to 31.0\%, which is a smaller relative gain that what was observed in PASCAL.

\subsection{Relationship to OverFeat}
There is an interesting relationship between R-CNN and OverFeat:
OverFeat can be seen (roughly) as a special case of R-CNN.
If one were to replace selective search region proposals with a multi-scale pyramid of regular square regions and change the per-class bounding-box regressors to a single bounding-box regressor, then the systems would be very similar (modulo some potentially significant differences in how they are trained: CNN detection fine-tuning, using SVMs, etc.).
It is worth noting that OverFeat has a significant speed advantage over R-CNN: it is about 9x faster, based on a figure of 2 seconds per image quoted from \cite{overfeat}.
This speed comes from the fact that OverFeat's sliding windows (\ie, region proposals) are not warped at the image level and therefore computation can be easily shared between overlapping windows.
Sharing is implemented by running the entire network in a convolutional fashion over arbitrary-sized inputs.
Speeding up R-CNN should be possible in a variety of ways and remains as future work.

\section{Semantic segmentation}
\seclabel{segmentation}

Region classification is a standard technique for semantic segmentation, allowing us to easily apply R-CNN to the PASCAL VOC segmentation challenge.
To facilitate a direct comparison with the current leading semantic segmentation system (called \OtwoP for ``second-order pooling'') \cite{o2p}, we work within their open source framework.
\OtwoP uses CPMC to generate 150 region proposals per image and then predicts the quality of each region, for each class, using support vector regression (SVR).
The high performance of their approach is due to the quality of the CPMC regions and the powerful second-order pooling of multiple feature types (enriched variants of SIFT and LBP).
We also note that Farabet \etal \cite{farabet-pami-13} recently demonstrated good results on several dense scene labeling datasets (not including PASCAL) using a CNN as a multi-scale per-pixel classifier.

We follow \cite{arbelaez2012semantic,o2p} and extend the PASCAL segmentation training set to include the extra annotations made available by Hariharan \etal \cite{hariharan2012inverse}. 
Design decisions and hyperparameters were cross-validated on the VOC 2011 validation set. Final test results were evaluated only once.

\paragraph{CNN features for segmentation.} We evaluate three strategies for computing features on CPMC regions, all of which begin by warping the rectangular window around the region to $227 \times 227$.
The first strategy (\textit{full}) ignores the region's shape and computes CNN features directly on the warped window, exactly as we did for detection.
However, these features ignore the non-rectangular shape of the region.
Two regions might have very similar bounding boxes while having very little overlap.
Therefore, the second strategy (\textit{fg}) computes CNN features only on a region's foreground mask.
We replace the background with the mean input so that background regions are zero after mean subtraction.
The third strategy (\textit{full+fg}) simply concatenates the \textit{full} and \textit{fg} features; our experiments validate their complementarity.
\begin{table}[h!]
\centering
{\small
\begin{tabular}{@{}c|c|c|c|c|c|c}
& \multicolumn{2}{c|}{\textit{full} R-CNN}
& \multicolumn{2}{c|}{\textit{fg} R-CNN}
& \multicolumn{2}{c}{\textit{full+fg} R-CNN}
\\
\hline
\OtwoP \cite{o2p}
& \fc{6} & \fc{7} & \fc{6} & \fc{7} & \fc{6} & \fc{7} \\
\hline
  46.4   
& 43.0   
& 42.5   
& 43.7    
& 42.1    
& \textbf{47.9}   
& 45.8    
\end{tabular}
}
\caption{\textbf{Segmentation mean accuracy (\%) on VOC 2011 validation.}
Column 1 presents \OtwoP;
2-7 use our CNN pre-trained on ILSVRC 2012.
}
\tablelabel{voc2011val}
\vspace{-1em}
\end{table}
\paragraph{Results on VOC 2011.}
\tableref{voc2011val} shows a summary of our results on the VOC 2011 validation set compared with \OtwoP.
(See \asecref{segperclass} for complete per-category results.)
Within each feature computation strategy, layer \fc{6} always outperforms \fc{7} and the following discussion refers to the \fc{6} features.
The \textit{fg} strategy slightly outperforms \textit{full}, indicating that the masked region shape provides a stronger signal, matching our intuition.
However, \textit{full+fg} achieves an average accuracy of 47.9\%, our best result by a margin of 4.2\% (also modestly outperforming \OtwoP), indicating that the context provided by the \textit{full} features is highly informative even given the \textit{fg} features.
Notably, training the 20 SVRs on our \textit{full+fg} features takes an hour on a single core, compared to 10+ hours for training on \OtwoP features.

In \tableref{voc2011test} we present results on the VOC 2011 test set, comparing our best-performing method, \fc{6} (\textit{full+fg}), against two strong baselines.
Our method achieves the highest segmentation accuracy for 11 out of 21 categories, and the highest overall segmentation accuracy of 47.9\%, averaged across categories (but likely ties with the \OtwoP result under any reasonable margin of error).
Still better performance could likely be achieved by fine-tuning.

\begin{table*}[t!]
\centering
\renewcommand{\arraystretch}{1.2}
\renewcommand{\tabcolsep}{1.2mm}
\resizebox{\linewidth}{!}{
\begin{tabular}{@{}l|c|r*{19}{c}|c@{}}
\textbf{VOC 2011 test} & bg & aero      & bike      & bird      & boat      & bottle     & bus        & car        & cat        & chair      & cow        & table      & dog        & horse      & mbike      & person     & plant      & sheep      & sofa       & train      & tv         & mean \\
\hline
R\&P \cite{arbelaez2012semantic} &
83.4 &
46.8 & 18.9 & 36.6 & 31.2 & 42.7 & 57.3 & 47.4 & 44.1 & \phz8.1 & 39.4 &
\bf{36.1} & 36.3 & 49.5 & 48.3 & 50.7 & 26.3 & 47.2 & 22.1 & 42.0 & 43.2 &
40.8 \\
\OtwoP \cite{o2p} &
\bf{85.4} &
\bf{69.7} & 22.3 & 45.2 & \bf{44.4} & 46.9 & 66.7 & 57.8 & 56.2 & \bf{13.5} & \bf{46.1} &
32.3 & 41.2 & \bf{59.1} & 55.3 & 51.0 & \bf{36.2} & 50.4 & \bf{27.8} & 46.9 & \bf{44.6} &
47.6 \\
\hline
\textbf{ours} (\textit{full+fg} R-CNN \fc{6}) &
84.2 &
66.9 & \bf{23.7} & \bf{58.3} & 37.4 & \bf{55.4} & \bf{73.3} & \bf{58.7} & \bf{56.5} & \phz9.7 & 45.5 &
29.5 & \bf{49.3} & 40.1 & \bf{57.8} & \bf{53.9} & 33.8 & \bf{60.7} & 22.7 & \bf{47.1} & 41.3 &
\bf{47.9} \\
\end{tabular}
}
\caption{\textbf{Segmentation accuracy (\%) on VOC 2011 test.}
We compare against two strong baselines: the ``Regions and Parts'' (R\&P) method of~\cite{arbelaez2012semantic} and the second-order pooling (\OtwoP) method of~\cite{o2p}.
Without any fine-tuning, our CNN achieves top segmentation performance, outperforming R\&P and roughly matching \OtwoP.
}
\tablelabel{voc2011test}
\vspace{-1em}
\end{table*}

\section{Conclusion}

In recent years, object detection performance had stagnated.
The best performing systems were complex ensembles combining multiple low-level image features with high-level context from object detectors and scene classifiers.
This paper presents a simple and scalable object detection algorithm that gives a 30\% relative improvement over the best previous results on PASCAL VOC 2012.

We achieved this performance through two insights.
The first is to apply high-capacity convolutional neural networks to bottom-up region proposals in order to localize and segment objects.
The second is a paradigm for training large CNNs when labeled training data is scarce. We show that it is highly effective to pre-train the network---\emph{with supervision}---for a auxiliary task with abundant data (image classification) and then to fine-tune the network for the target task where data is scarce (detection).
We conjecture that the ``supervised pre-training/domain-specific fine-tuning'' paradigm will be highly effective for a variety of data-scarce vision problems.


We conclude by noting that it is significant that we achieved these results by using a combination of classical tools from computer vision \emph{and} deep learning (bottom-up region proposals and convolutional neural networks).
Rather than opposing lines of scientific inquiry, the two are natural and inevitable partners.

\paragraph{Acknowledgments.}
This research was supported in part by DARPA Mind's Eye and MSEE
programs, by NSF awards IIS-0905647, IIS-1134072, and IIS-1212798, MURI N000014-10-1-0933, and by support from Toyota.
The GPUs used in this research were generously donated by the NVIDIA Corporation.

\vspace{2em}
{\noindent \Large \bf Appendix}

\appendix
\section{Object proposal transformations}
\seclabel{altwarp}
The convolutional neural network used in this work requires a fixed-size input of $227 \times 227$ pixels.
For detection, we consider object proposals that are arbitrary image rectangles.
We evaluated two approaches for transforming object proposals into valid CNN inputs.

The first method (``tightest square with context'') encloses each object proposal inside the tightest square and then scales (isotropically) the image contained in that square to the CNN input size.
\figref{crops} column (B) shows this transformation.
A variant on this method (``tightest square without context'') excludes the image content that surrounds the original object proposal.
\figref{crops} column (C) shows this transformation.
The second method (``warp'') anisotropically scales each object proposal to the CNN input size.
\figref{crops} column (D) shows the warp transformation.

For each of these transformations, we also consider including additional image context around the original object proposal.
The amount of context padding ($p$) is defined as a border size around the original object proposal in the transformed input coordinate frame.
\figref{crops} shows $p = 0$ pixels in the top row of each example and $p = 16$ pixels in the bottom row.
In all methods, if the source rectangle extends beyond the image, the missing data is replaced with the image mean (which is then subtracted before inputing the image into the CNN).
A pilot set of experiments showed that warping with context padding ($p = 16$ pixels) outperformed the alternatives by a large margin (3-5 mAP points).
Obviously more alternatives are possible, including using replication instead of mean padding.
Exhaustive evaluation of these alternatives is left as future work.
\begin{figure}[t!]
\centering
\includegraphics[width=1\linewidth]{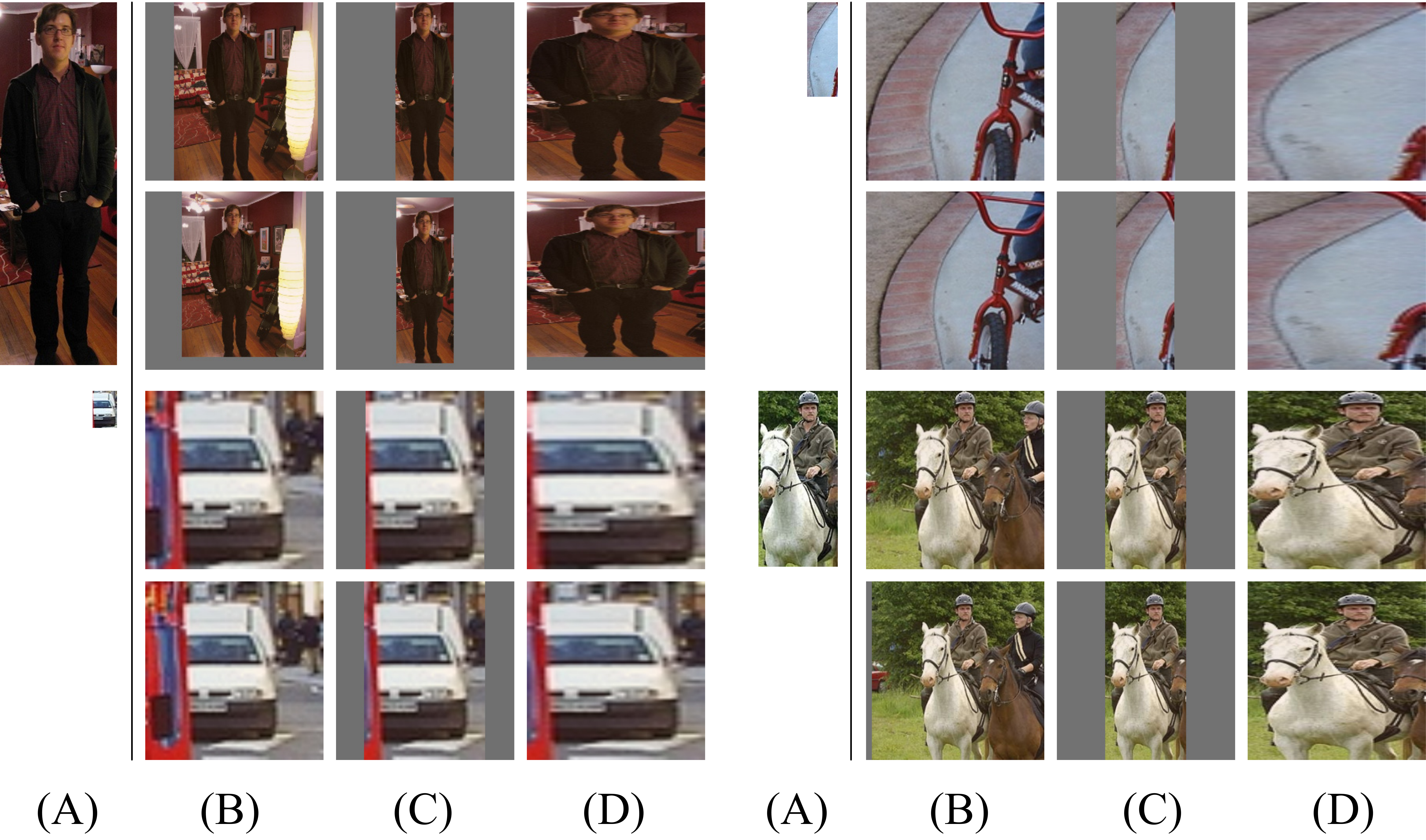}
\vspace{-1em}
\caption{\textbf{Different object proposal transformations.}
(A) the original object proposal at its actual scale relative to the transformed CNN inputs; (B) tightest square with context; (C) tightest square without context; (D) warp.
Within each column and example proposal, the top row corresponds to $p = 0$ pixels of context padding while the bottom row has $p = 16$ pixels of context padding.}
\figlabel{crops}
\end{figure}

\section{Positive \vs negative examples and softmax}
\seclabel{posneg}
Two design choices warrant further discussion.
The first is: Why are positive and negative examples defined differently for fine-tuning the CNN versus training the object detection SVMs?
To review the definitions briefly, for fine-tuning we map each object proposal to the ground-truth instance with which it has maximum IoU overlap (if any) and label it as a positive for the matched ground-truth class if the IoU is at least 0.5.
All other proposals are labeled ``background'' (\ie, negative examples for all classes).
For training SVMs, in contrast, we take only the ground-truth boxes as positive examples for their respective classes and label proposals with less than 0.3 IoU overlap with all instances of a class as a negative for that class.
Proposals that fall into the grey zone (more than 0.3 IoU overlap, but are not ground truth) are ignored.

Historically speaking, we arrived at these definitions because we started by training SVMs on features computed by the ImageNet pre-trained CNN, and so fine-tuning was not a consideration at that point in time.
In that setup, we found that our particular label definition for training SVMs was optimal within the set of options we evaluated (which included the setting we now use for fine-tuning).
When we started using fine-tuning, we initially used the same positive and negative example definition as we were using for SVM training.
However, we found that results were much worse than those obtained using our current definition of positives and negatives.

Our hypothesis is that this difference in how positives and negatives are defined is not fundamentally important and arises from the fact that fine-tuning data is limited.
Our current scheme introduces many ``jittered'' examples (those proposals with overlap between 0.5 and 1, but not ground truth), which expands the number of positive examples by approximately 30x.
We conjecture that this large set is needed when fine-tuning the \emph{entire} network to avoid overfitting.
However, we also note that using these jittered examples is likely suboptimal because the network is not being fine-tuned for precise localization.

This leads to the second issue: Why, after fine-tuning, train SVMs at all?
It would be cleaner to simply apply the last layer of the fine-tuned network, which is a 21-way softmax regression classifier, as the object detector.
We tried this and found that performance on VOC 2007 dropped from 54.2\% to 50.9\% mAP.
This performance drop likely arises from a combination of several factors including that the definition of positive examples used in fine-tuning does not emphasize precise localization and the softmax classifier was trained on randomly sampled negative examples rather than on the subset of ``hard negatives'' used for SVM training.

This result shows that it's possible to obtain close to the same level of performance without training SVMs after fine-tuning.
We conjecture that with some additional tweaks to fine-tuning the remaining performance gap may be closed.
If true, this would simplify and speed up R-CNN training with no loss in detection performance.

\section{Bounding-box regression}
\seclabel{bboxreg}
We use a simple bounding-box regression stage to improve localization performance.
After scoring each selective search proposal with a class-specific detection SVM, we predict a new bounding box for the detection using a class-specific bounding-box regressor.
This is similar in spirit to the bounding-box regression used in deformable part models \cite{lsvm-pami}.
The primary difference between the two approaches is that here we regress from features computed by the CNN, rather than from geometric features computed on the inferred DPM part locations.

The input to our training algorithm is a set of $N$ training pairs $\{(P^i, G^i)\}_{i = 1,\ldots,N}$, where $P^i = (P_x^i, P_y^i, P_w^i, P_h^i)$ specifies the pixel coordinates of the center of proposal $P^i$'s bounding box together with $P^i$'s  width and height in pixels.
Hence forth, we drop the superscript $i$ unless it is needed.
Each ground-truth bounding box $G$ is specified in the same way: $G = (G_x, G_y, G_w, G_h)$.
Our goal is to learn a transformation that maps a proposed box $P$ to a ground-truth box $G$.

We parameterize the transformation in terms of four functions $d_x(P)$, $d_y(P)$, $d_w(P)$, and $d_h(P)$.
The first two specify a scale-invariant translation of the center of $P$'s bounding box, while the second two specify log-space translations of the width and height of $P$'s bounding box.
After learning these functions, we can transform an input proposal $P$ into a predicted ground-truth box $\hat{G}$ by applying the transformation
\begin{align}
\hat{G}_x &= P_w d_x(P) + P_x \\
\hat{G}_y &= P_h d_y(P) + P_y \\
\hat{G}_w &= P_w \exp(d_w(P)) \\
\hat{G}_h &= P_h \exp(d_h(P)).
\end{align}

Each function $d_\star(P)$ (where $\star$ is one of $x,y,h,w$) is modeled as a linear function of the \pool{5} features of proposal $P$, denoted by $\bphi_5(P)$.
(The dependence of $\bphi_5(P)$ on the image data is implicitly assumed.) 
Thus we have $d_\star(P) = \bfw_\star^\trn \bphi_5(P)$, where $\bfw_\star$ is a vector of learnable model parameters.
We learn $\bfw_\star$ by optimizing the regularized least squares objective (ridge regression):
\begin{align}
\bfw_\star = \argmin_{\hat{\bfw}_\star} \sum_i^N (t^i_\star - \hat{\bfw}_\star^\trn \bphi_5(P^i))^2 + \lambda \norm{\hat{\bfw}_\star}^2.
\end{align}
The regression targets $t_\star$ for the training pair $(P, G)$ are defined as
\begin{align}
t_x &= (G_x - P_x) / P_w \\
t_y &= (G_y - P_y) / P_h \\
t_w &= \log(G_w / P_w) \\
t_h &= \log(G_h / P_h).
\end{align}
As a standard regularized least squares problem, this can be solved efficiently in closed form.

We found two subtle issues while implementing bounding-box regression.
The first is that regularization is important: we set $\lambda = 1000$ based on a validation set.
The second issue is that care must be taken when selecting which training pairs $(P,G)$ to use.
Intuitively, if $P$ is far from all ground-truth boxes, then the task of transforming $P$ to a ground-truth box $G$ does not make sense.
Using examples like $P$ would lead to a hopeless learning problem.
Therefore, we only learn from a proposal $P$ if it is \emph{nearby} at least one ground-truth box.
We implement ``nearness'' by assigning $P$ to the ground-truth box $G$ with which it has maximum IoU overlap (in case it overlaps more than one) if and only if the overlap is greater than a threshold (which we set to 0.6 using a validation set).
All unassigned proposals are discarded.
We do this once for each object class in order to learn a set of class-specific bounding-box regressors.

At test time, we score each proposal and predict its new detection window only once.
In principle, we could iterate this procedure (\ie, re-score the newly predicted bounding box, and then predict a new bounding box from it, and so on).
However, we found that iterating does not improve results.

\section{Additional feature visualizations}
\seclabel{extravis}
\figref{vis21} shows additional visualizations for 20 \pool{5} units.
For each unit, we show the 24 region proposals that maximally activate that unit out of the full set of approximately 10 million regions in all of VOC 2007 test.

We label each unit by its (y, x, channel) position in the $6 \times 6 \times 256$ dimensional \pool{5} feature map.
Within each channel, the CNN computes exactly the same function of the input region, with the (y, x) position changing only the receptive field.

\section{Per-category segmentation results}
\seclabel{segperclass}
In \tableref{voc2011valsupp} we show the per-category segmentation accuracy on VOC 2011 val for each of our six segmentation methods in addition to the \OtwoP method~\cite{o2p}. These results show which methods are strongest across each of the 20 PASCAL classes, plus the background class.

\begin{table*}[t!]
\centering
\renewcommand{\arraystretch}{1.2}
\renewcommand{\tabcolsep}{1.2mm}
\resizebox{\linewidth}{!}{
\begin{tabular}{@{}l|c|r*{19}{c}|c@{}}
\textbf{VOC 2011 val} & bg & aero      & bike      & bird      & boat      & bottle     & bus        & car        & cat        & chair      & cow        & table      & dog        & horse      & mbike      & person     & plant      & sheep      & sofa       & train      & tv         & mean \\
\hline
\OtwoP \cite{o2p} &
\textbf{84.0} &
\textbf{69.0} & 21.7 & 47.7 & 42.2 & 42.4 & \textbf{64.7} & \textbf{65.8} & 57.4 & \textbf{12.9} & 37.4 &
20.5 & 43.7 & 35.7 & 52.7 & 51.0 & \textbf{35.8} & \textbf{51.0} & 28.4 & 59.8 & 49.7 &
46.4 \\
\hline
\textit{full} R-CNN \fc{6} &
81.3 &
56.2 & 23.9 & 42.9 & 40.7 & 38.8 & 59.2 & 56.5 & 53.2 & 11.4 & 34.6 &
16.7 & 48.1 & 37.0 & 51.4 & 46.0 & 31.5 & 44.0 & 24.3 & 53.7 & 51.1 &
43.0 \\
\textit{full} R-CNN \fc{7} &
81.0 &
52.8 & \textbf{25.1} & 43.8 & 40.5 & 42.7 & 55.4 & 57.7 & 51.3 & \phz8.7 & 32.5 &
11.5 & 48.1 & 37.0 & 50.5 & 46.4 & 30.2 & 42.1 & 21.2 & 57.7 & \textbf{56.0} &
42.5 \\
\hline
\textit{fg} R-CNN \fc{6} &
81.4 &
54.1 & 21.1 & 40.6 & 38.7 & \textbf{53.6} & 59.9 & 57.2 & 52.5 & \phz9.1 & 36.5 &
\textbf{23.6} & 46.4 & 38.1 & 53.2 & 51.3 & 32.2 & 38.7 & \textbf{29.0} & 53.0 & 47.5 &
43.7 \\
\textit{fg} R-CNN \fc{7} &
80.9 &
50.1 & 20.0 & 40.2 & 34.1 & 40.9 & 59.7 & 59.8 & 52.7 & \phz7.3 & 32.1 &
14.3 & 48.8 & 42.9 & 54.0 & 48.6 & 28.9 & 42.6 & 24.9 & 52.2 & 48.8 &
42.1 \\
\hline
\textit{full+fg} R-CNN \fc{6} &
83.1 &
60.4 & 23.2 & 48.4 & \textbf{47.3} & 52.6 & 61.6 & 60.6 & \textbf{59.1} & 10.8 & \textbf{45.8} &
20.9 & \textbf{57.7} & 43.3 & \textbf{57.4} & \textbf{52.9} & 34.7 & 48.7 & 28.1 & 60.0 & 48.6 &
\textbf{47.9} \\
\textit{full+fg} R-CNN \fc{7} &
82.3 &
56.7 & 20.6 & \textbf{49.9} & 44.2 & 43.6 & 59.3 & 61.3 & 57.8 & \phz7.7 & 38.4 &
15.1 & 53.4 & \textbf{43.7} & 50.8 & 52.0 & 34.1 & 47.8 & 24.7 & \textbf{60.1} & 55.2 &
45.7 \\
\end{tabular}
}
\vspace{0.1em}
\caption{Per-category segmentation accuracy (\%) on the VOC 2011 validation set.
}
\tablelabel{voc2011valsupp}
\end{table*}

\section{Analysis of cross-dataset redundancy}
One concern when training on an auxiliary dataset is that there might be redundancy between it and the test set.
Even though the tasks of object detection and whole-image classification are substantially different, making such cross-set redundancy much less worrisome, we still conducted a thorough investigation that quantifies the extent to which PASCAL test images are contained within the ILSVRC 2012 training and validation sets.
Our findings may be useful to researchers who are interested in using ILSVRC 2012 as training data for the PASCAL image classification task.

We performed two checks for duplicate (and near-duplicate) images.
The first test is based on exact matches of flickr image IDs, which are included in the VOC 2007 test annotations (these IDs are intentionally kept secret for subsequent PASCAL test sets).
All PASCAL images, and about half of ILSVRC, were collected from flickr.com.
This check turned up 31 matches out of 4952 (0.63\%).

The second check uses GIST \cite{GIST} descriptor matching, which was shown in \cite{douze2009evaluation} to have excellent performance at near-duplicate image detection in large ($>$ 1 million) image collections.
Following \cite{douze2009evaluation}, we computed GIST descriptors on warped $32 \times 32$ pixel versions of all ILSVRC 2012 trainval and PASCAL 2007 test images.

Euclidean distance nearest-neighbor matching of GIST descriptors revealed 38 near-duplicate images (including all 31 found by flickr ID matching).
The matches tend to vary slightly in JPEG compression level and resolution, and to a lesser extent cropping.
These findings show that the overlap is small, less than 1\%.
For VOC 2012, because flickr IDs are not available, we used the GIST matching method only.
Based on GIST matches, 1.5\% of VOC 2012 test images are in ILSVRC 2012 trainval.
The slightly higher rate for VOC 2012 is likely due to the fact that the two datasets were collected closer together in time than VOC 2007 and ILSVRC 2012 were.

\begin{table*}[t]
\centering
\renewcommand{\arraystretch}{1.2}
\renewcommand{\tabcolsep}{1.2mm}
\resizebox{\linewidth}{!}{
\begin{tabular}{@{}lr|lr|lr|lr|lr@{}}
\textbf{class} & \textbf{AP} &
\textbf{class} & \textbf{AP} &
\textbf{class} & \textbf{AP} &
\textbf{class} & \textbf{AP} &
\textbf{class} & \textbf{AP} \\
\hline
accordion & 50.8 & centipede & 30.4 & hair spray & 13.8 & pencil box & 11.4 & snowplow & 69.2 \\
airplane & 50.0 & chain saw & 14.1 & hamburger & 34.2 & pencil sharpener & 9.0 & soap dispenser & 16.8 \\
ant & 31.8 & chair & 19.5 & hammer & 9.9 & perfume & 32.8 & soccer ball & 43.7 \\
antelope & 53.8 & chime & 24.6 & hamster & 46.0 & person & 41.7 & sofa & 16.3 \\
apple & 30.9 & cocktail shaker & 46.2 & harmonica & 12.6 & piano & 20.5 & spatula & 6.8 \\
armadillo & 54.0 & coffee maker & 21.5 & harp & 50.4 & pineapple & 22.6 & squirrel & 31.3 \\
artichoke & 45.0 & computer keyboard & 39.6 & hat with a wide brim & 40.5 & ping-pong ball & 21.0 & starfish & 45.1 \\
axe & 11.8 & computer mouse & 21.2 & head cabbage & 17.4 & pitcher & 19.2 & stethoscope & 18.3 \\
baby bed & 42.0 & corkscrew & 24.2 & helmet & 33.4 & pizza & 43.7 & stove & 8.1 \\
backpack & 2.8 & cream & 29.9 & hippopotamus & 38.0 & plastic bag & 6.4 & strainer & 9.9 \\
bagel & 37.5 & croquet ball & 30.0 & horizontal bar & 7.0 & plate rack & 15.2 & strawberry & 26.8 \\
balance beam & 32.6 & crutch & 23.7 & horse & 41.7 & pomegranate & 32.0 & stretcher & 13.2 \\
banana & 21.9 & cucumber & 22.8 & hotdog & 28.7 & popsicle & 21.2 & sunglasses & 18.8 \\
band aid & 17.4 & cup or mug & 34.0 & iPod & 59.2 & porcupine & 37.2 & swimming trunks & 9.1 \\
banjo & 55.3 & diaper & 10.1 & isopod & 19.5 & power drill & 7.9 & swine & 45.3 \\
baseball & 41.8 & digital clock & 18.5 & jellyfish & 23.7 & pretzel & 24.8 & syringe & 5.7 \\
basketball & 65.3 & dishwasher & 19.9 & koala bear & 44.3 & printer & 21.3 & table & 21.7 \\
bathing cap & 37.2 & dog & 76.8 & ladle & 3.0 & puck & 14.1 & tape player & 21.4 \\
beaker & 11.3 & domestic cat & 44.1 & ladybug & 58.4 & punching bag & 29.4 & tennis ball & 59.1 \\
bear & 62.7 & dragonfly & 27.8 & lamp & 9.1 & purse & 8.0 & tick & 42.6 \\
bee & 52.9 & drum & 19.9 & laptop & 35.4 & rabbit & 71.0 & tie & 24.6 \\
bell pepper & 38.8 & dumbbell & 14.1 & lemon & 33.3 & racket & 16.2 & tiger & 61.8 \\
bench & 12.7 & electric fan & 35.0 & lion & 51.3 & ray & 41.1 & toaster & 29.2 \\
bicycle & 41.1 & elephant & 56.4 & lipstick & 23.1 & red panda & 61.1 & traffic light & 24.7 \\
binder & 6.2 & face powder & 22.1 & lizard & 38.9 & refrigerator & 14.0 & train & 60.8 \\
bird & 70.9 & fig & 44.5 & lobster & 32.4 & remote control & 41.6 & trombone & 13.8 \\
bookshelf & 19.3 & filing cabinet & 20.6 & maillot & 31.0 & rubber eraser & 2.5 & trumpet & 14.4 \\
bow tie & 38.8 & flower pot & 20.2 & maraca & 30.1 & rugby ball & 34.5 & turtle & 59.1 \\
bow & 9.0 & flute & 4.9 & microphone & 4.0 & ruler & 11.5 & tv or monitor & 41.7 \\
bowl & 26.7 & fox & 59.3 & microwave & 40.1 & salt or pepper shaker & 24.6 & unicycle & 27.2 \\
brassiere & 31.2 & french horn & 24.2 & milk can & 33.3 & saxophone & 40.8 & vacuum & 19.5 \\
burrito & 25.7 & frog & 64.1 & miniskirt & 14.9 & scorpion & 57.3 & violin & 13.7 \\
bus & 57.5 & frying pan & 21.5 & monkey & 49.6 & screwdriver & 10.6 & volleyball & 59.7 \\
butterfly & 88.5 & giant panda & 42.5 & motorcycle & 42.2 & seal & 20.9 & waffle iron & 24.0 \\
camel & 37.6 & goldfish & 28.6 & mushroom & 31.8 & sheep & 48.9 & washer & 39.8 \\
can opener & 28.9 & golf ball & 51.3 & nail & 4.5 & ski & 9.0 & water bottle & 8.1 \\
car & 44.5 & golfcart & 47.9 & neck brace & 31.6 & skunk & 57.9 & watercraft & 40.9 \\
cart & 48.0 & guacamole & 32.3 & oboe & 27.5 & snail & 36.2 & whale & 48.6 \\
cattle & 32.3 & guitar & 33.1 & orange & 38.8 & snake & 33.8 & wine bottle & 31.2 \\
cello & 28.9 & hair dryer & 13.0 & otter & 22.2 & snowmobile & 58.8 & zebra & 49.6 \\
\end{tabular}
}
\caption{Per-class average precision (\%) on the ILSVRC2013 detection \test set.}
\tablelabel{classaps}
\end{table*}

\begin{figure*}[t]
\begin{center}
\def \sz {1in}
\includegraphics[height=\sz]{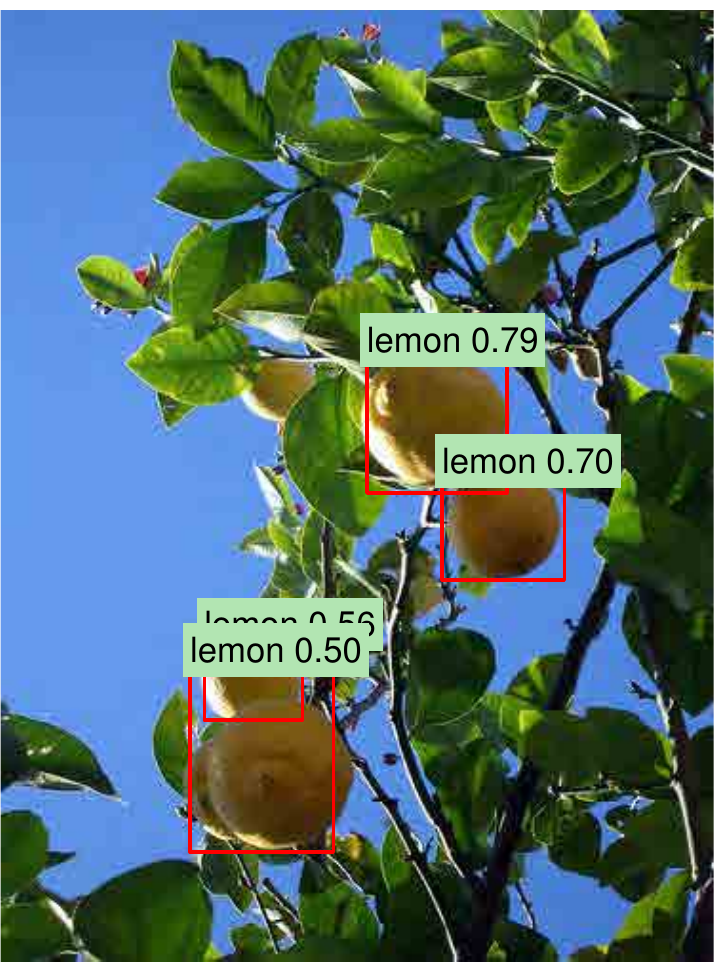}
\includegraphics[height=\sz]{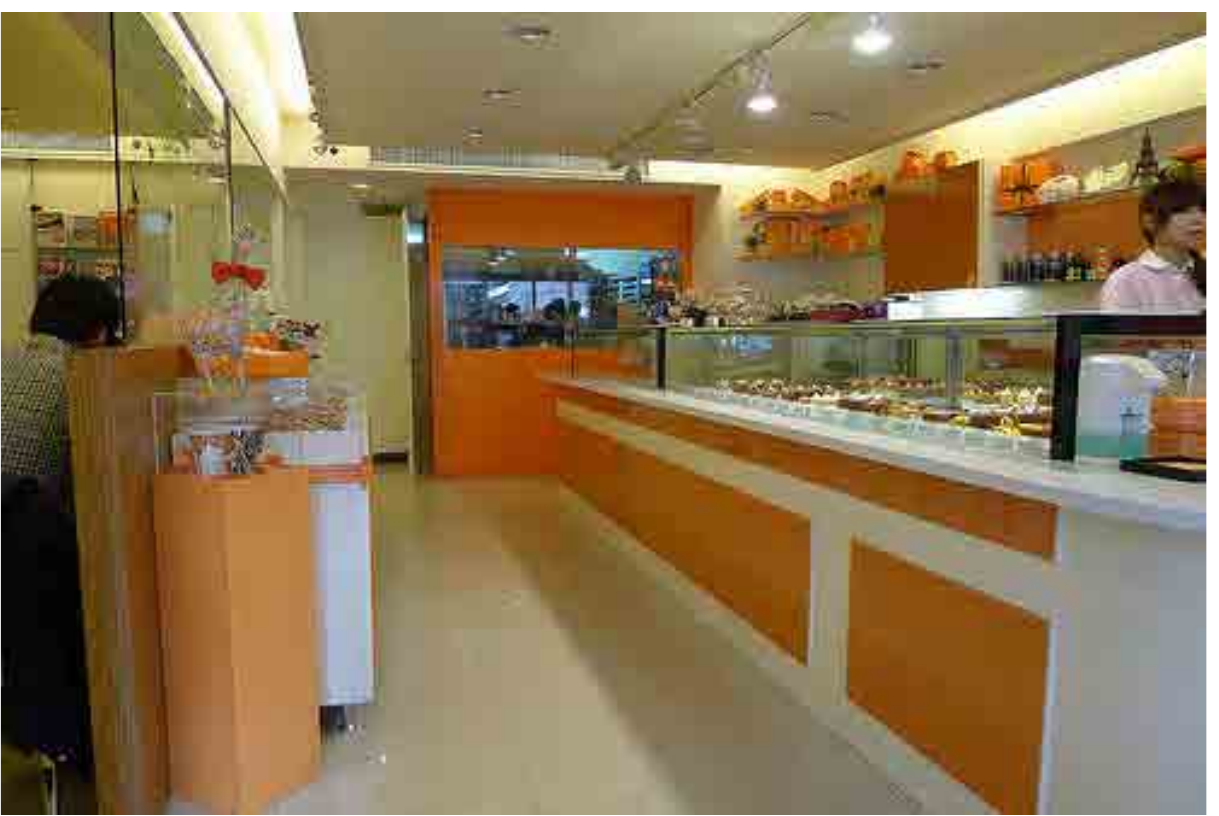}
\includegraphics[height=\sz]{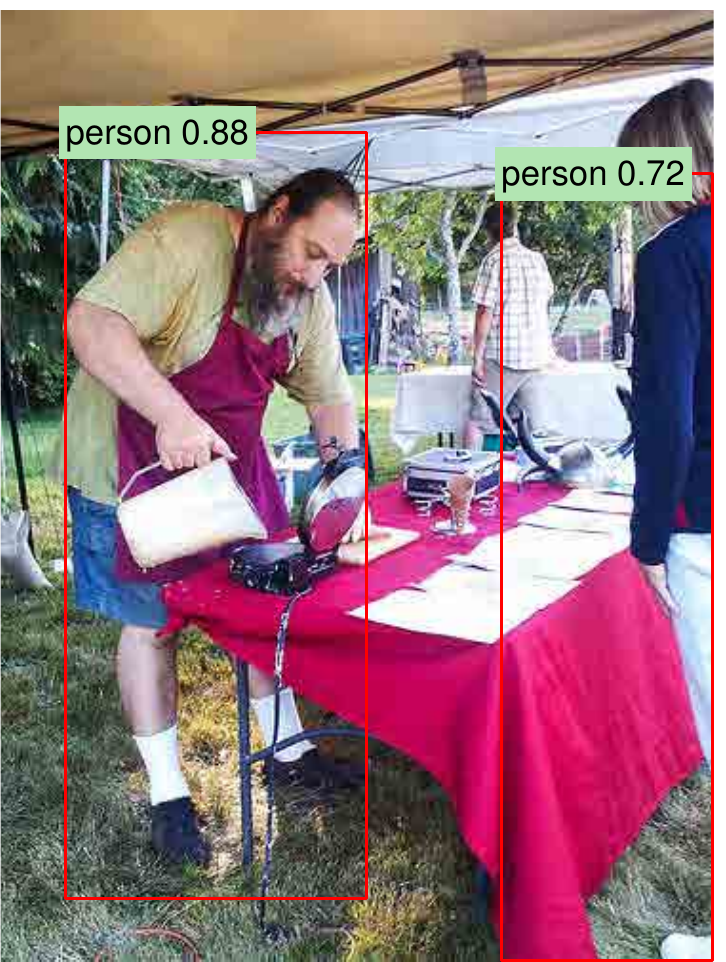}
\includegraphics[height=\sz]{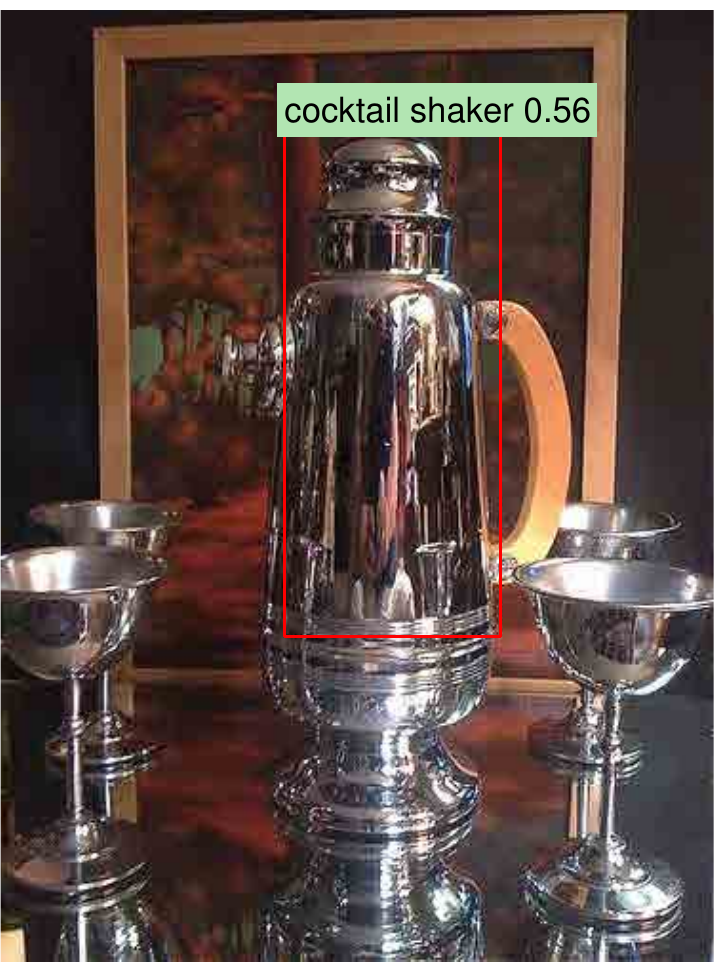}
\includegraphics[height=\sz]{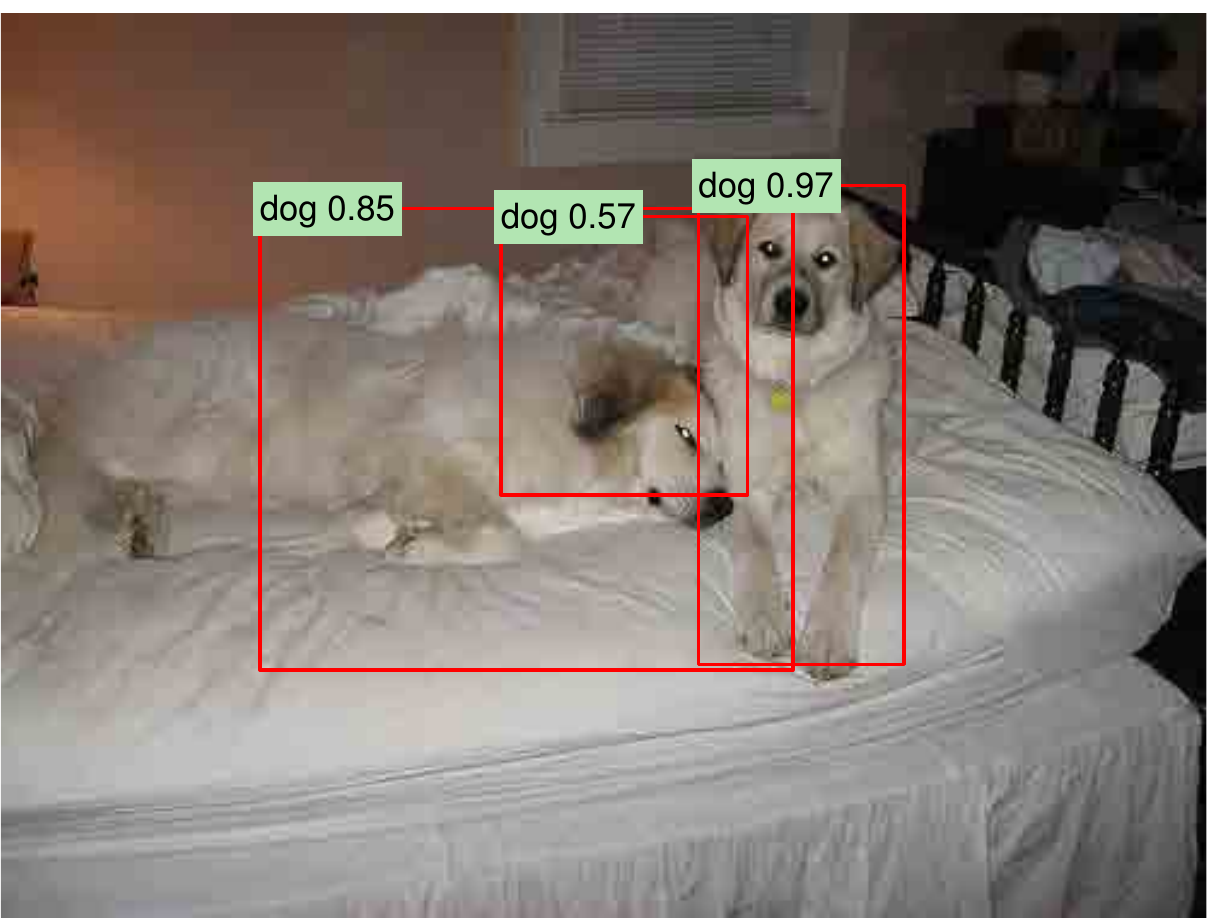}
\includegraphics[height=\sz]{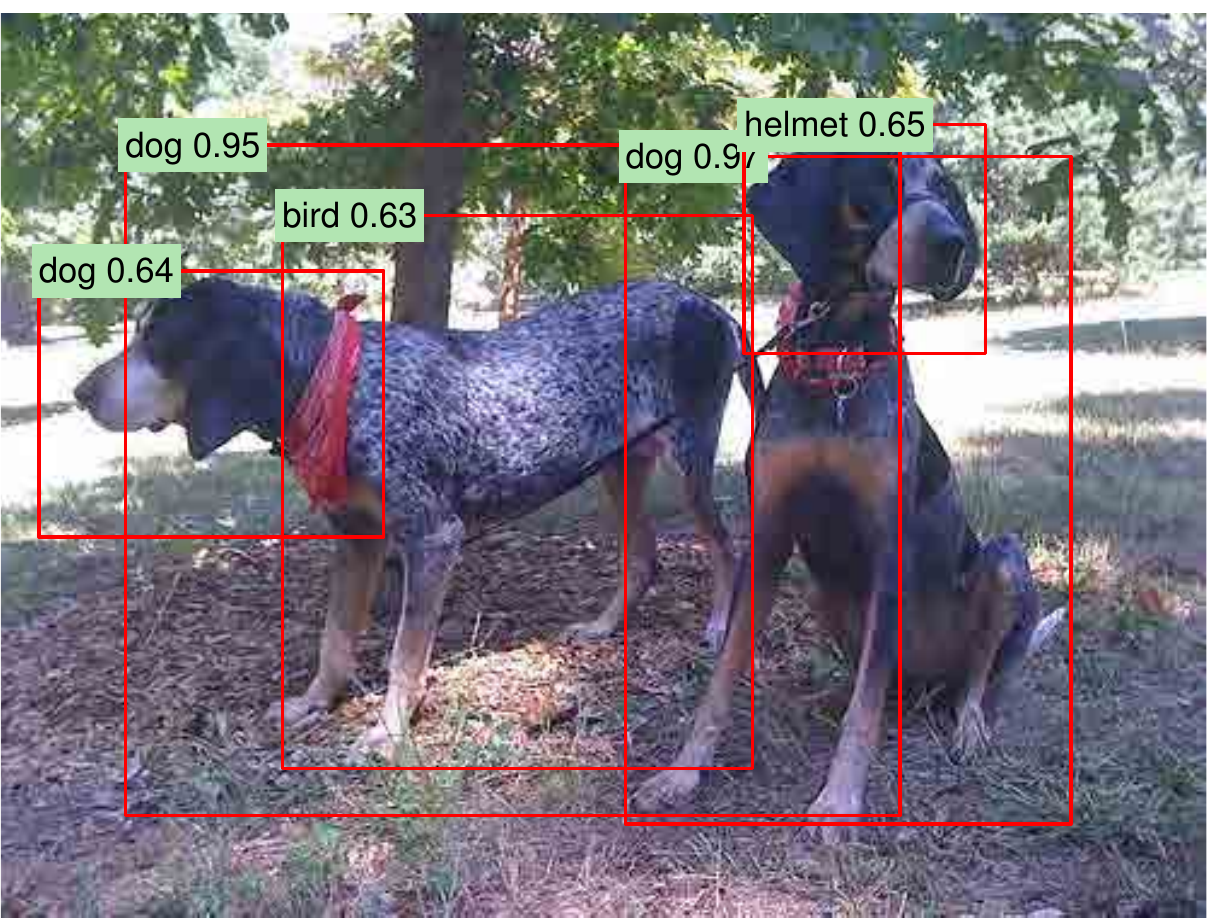}
\includegraphics[height=\sz]{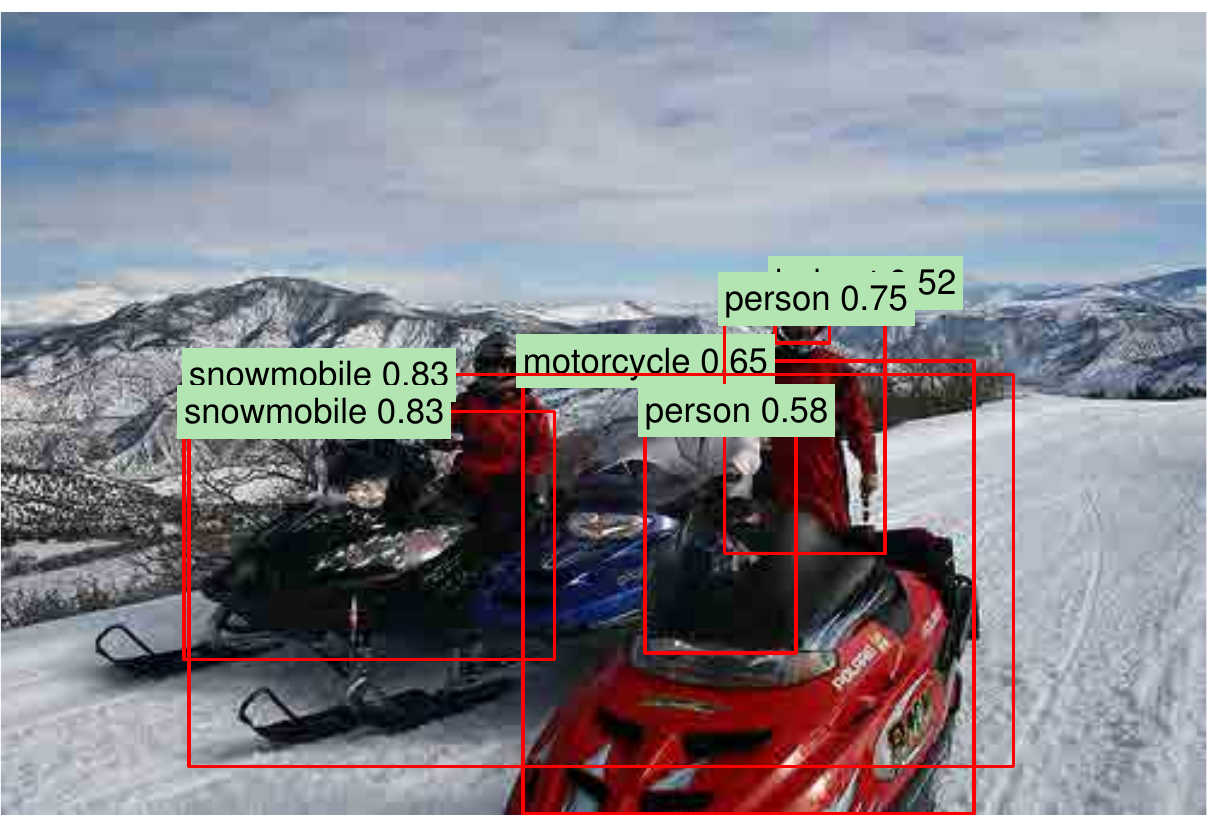}
\includegraphics[height=\sz]{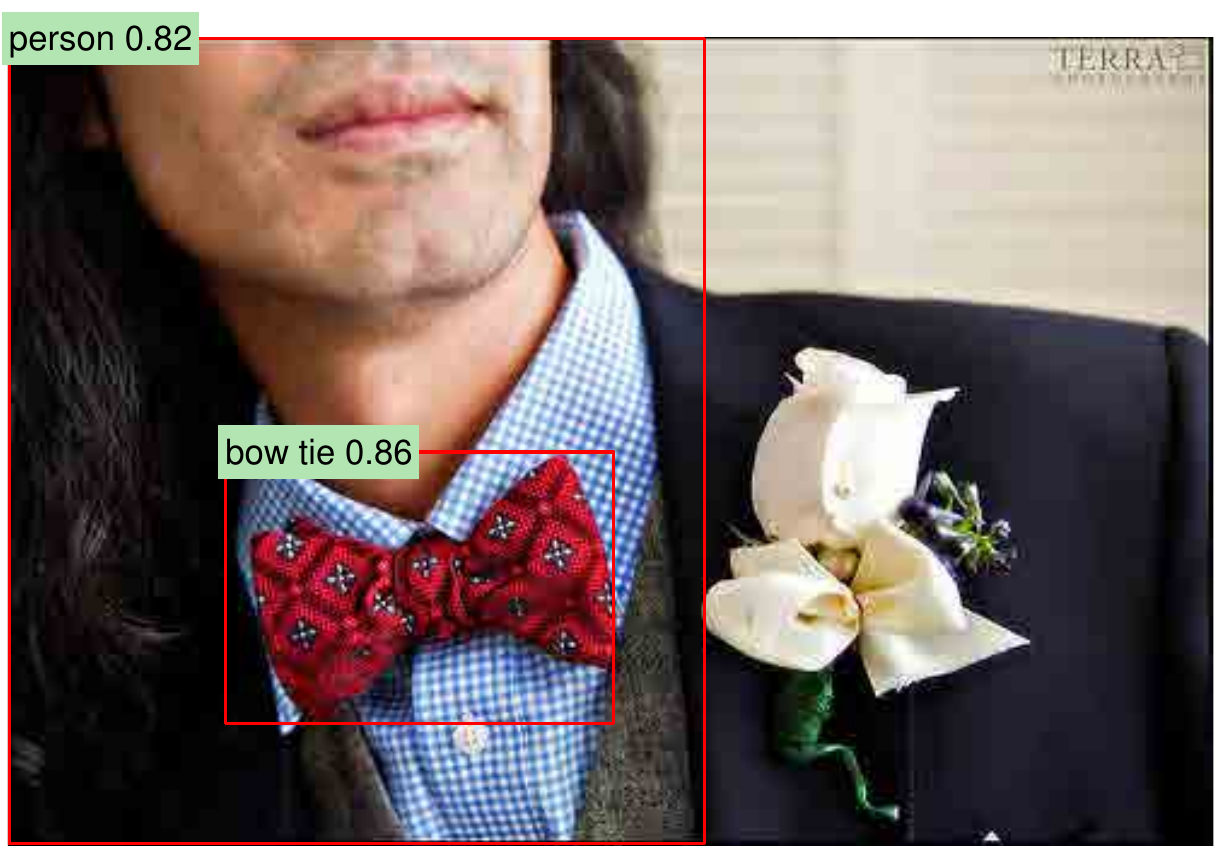}
\includegraphics[height=\sz]{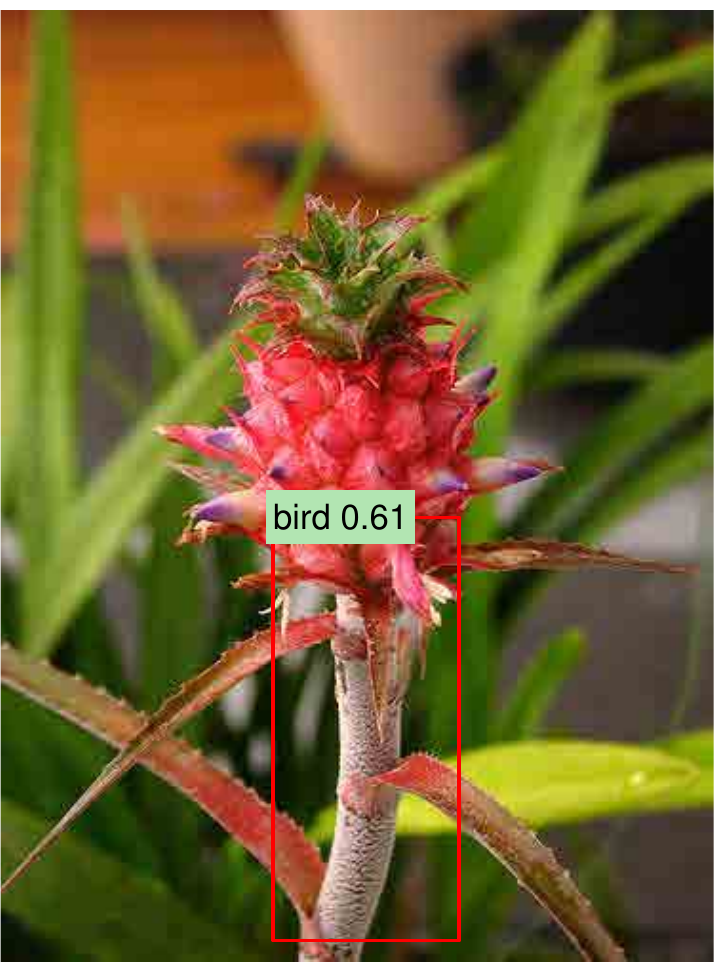}
\includegraphics[height=\sz]{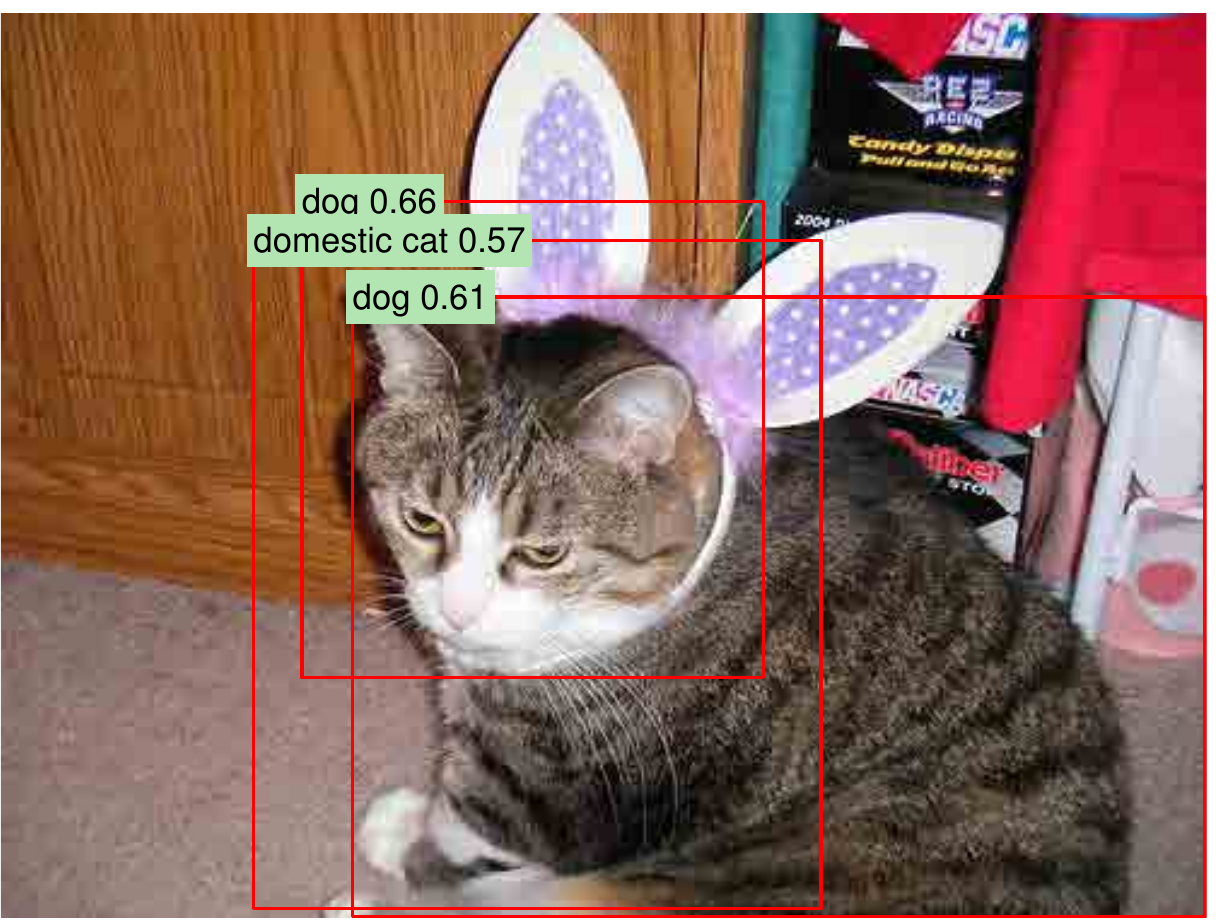}
\includegraphics[height=\sz]{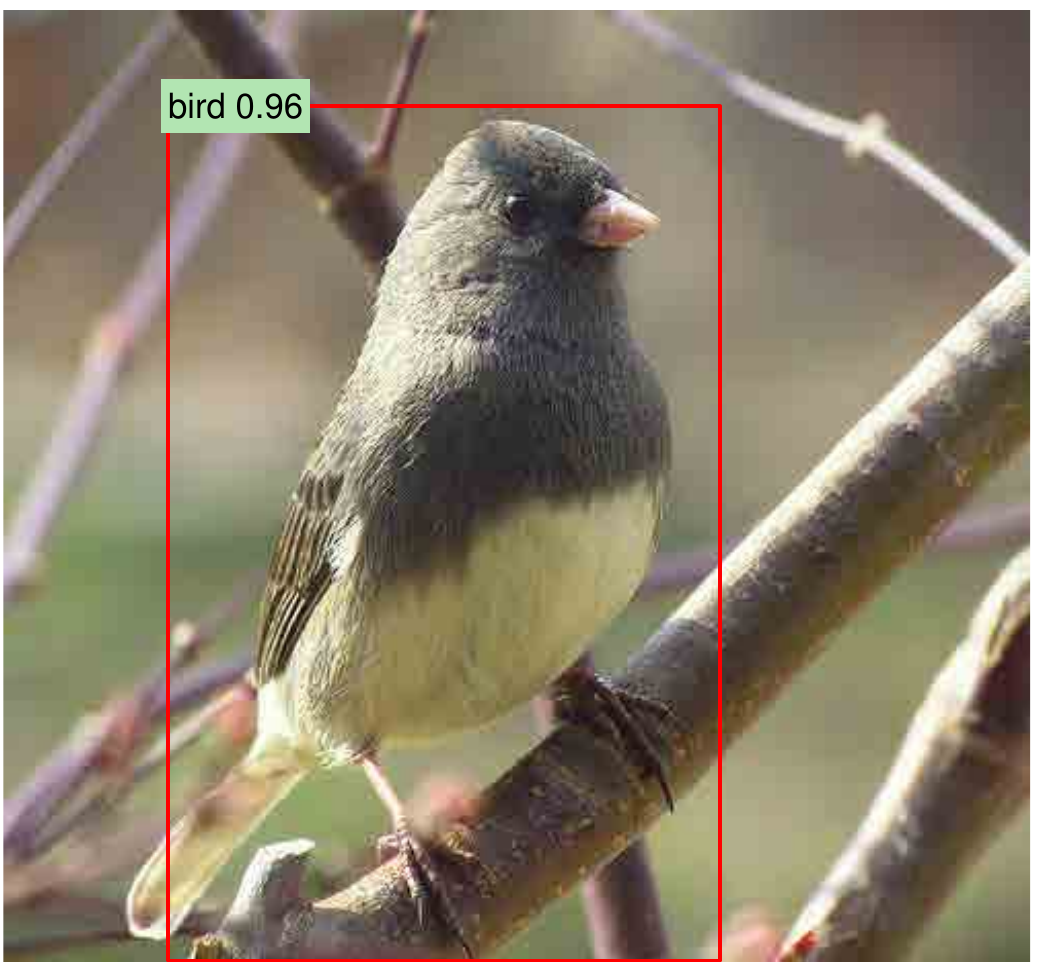}
\includegraphics[height=\sz]{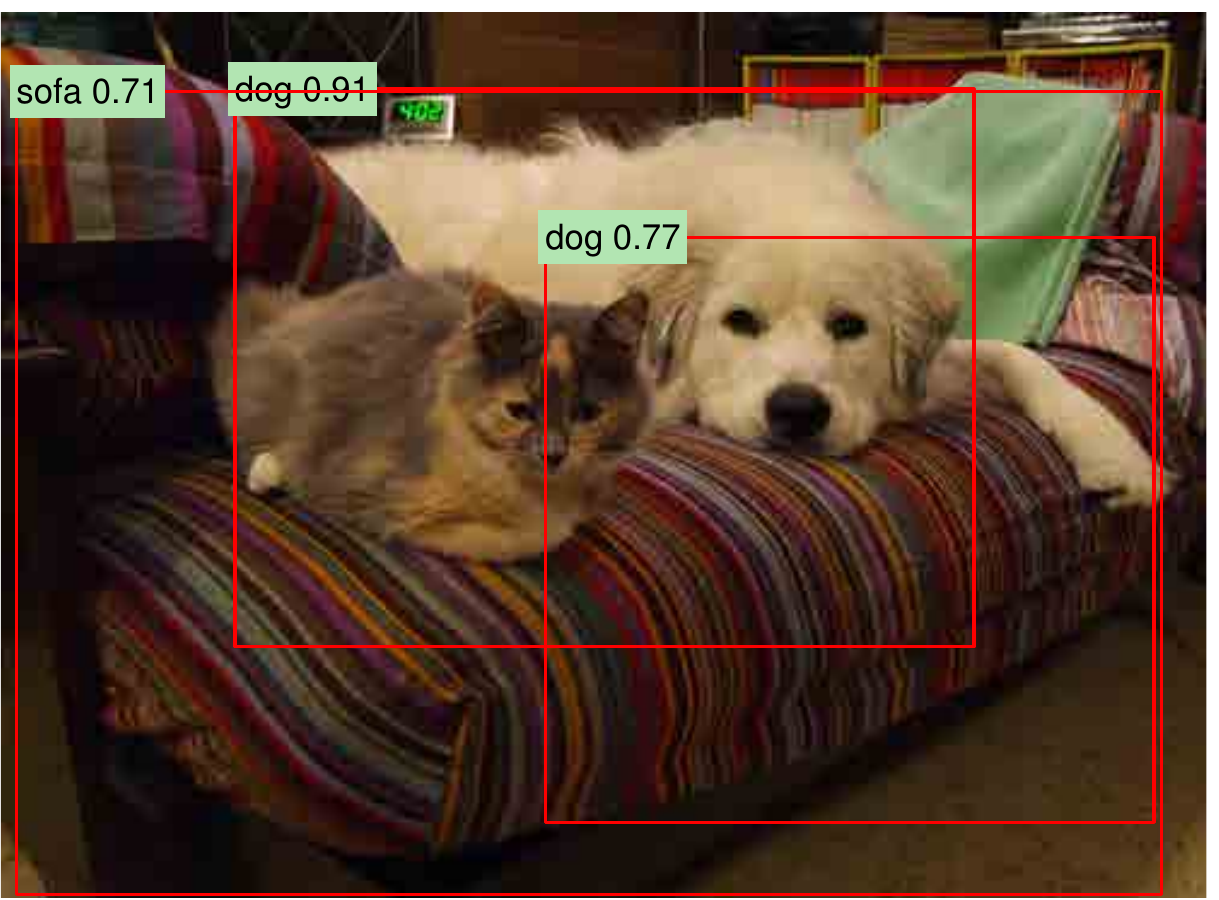}
\includegraphics[height=\sz]{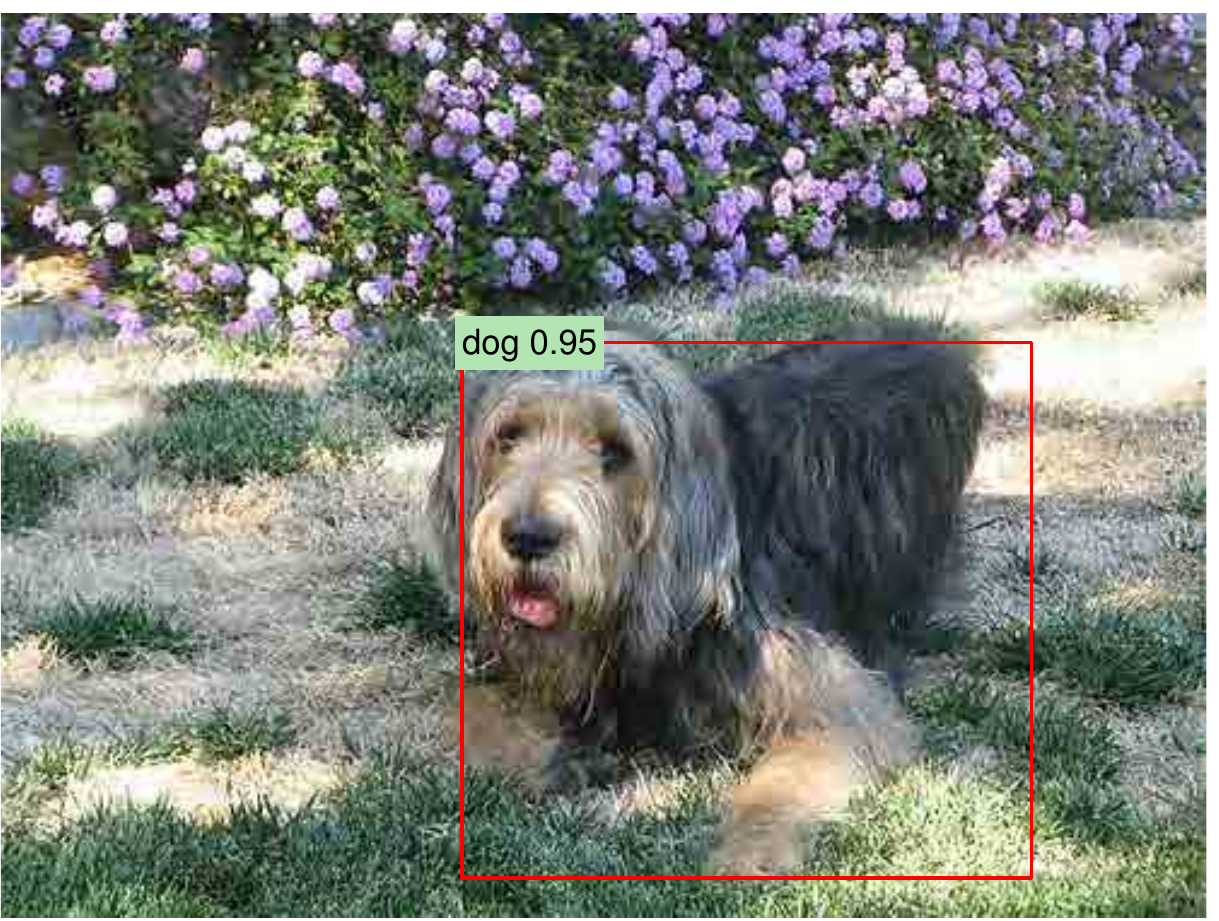}
\includegraphics[height=\sz]{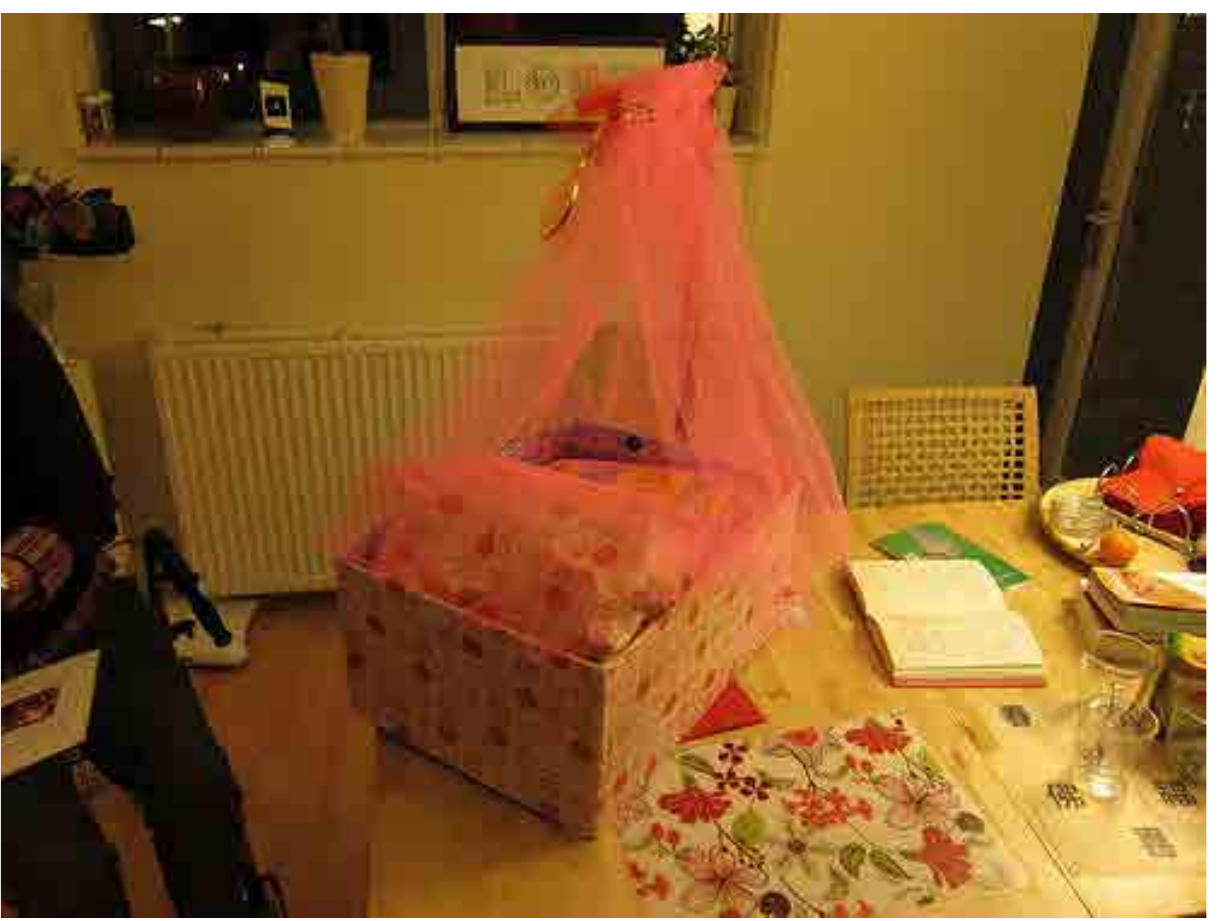}
\includegraphics[height=\sz]{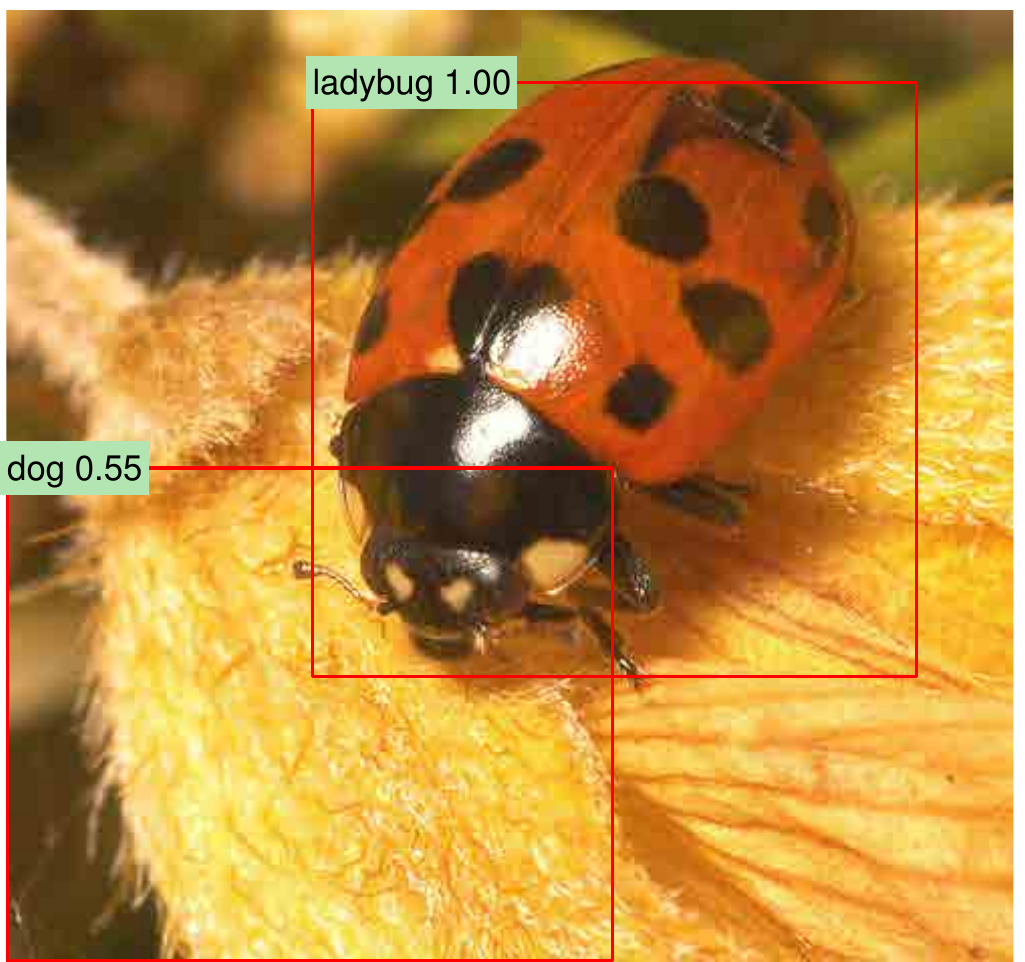}
\includegraphics[height=\sz]{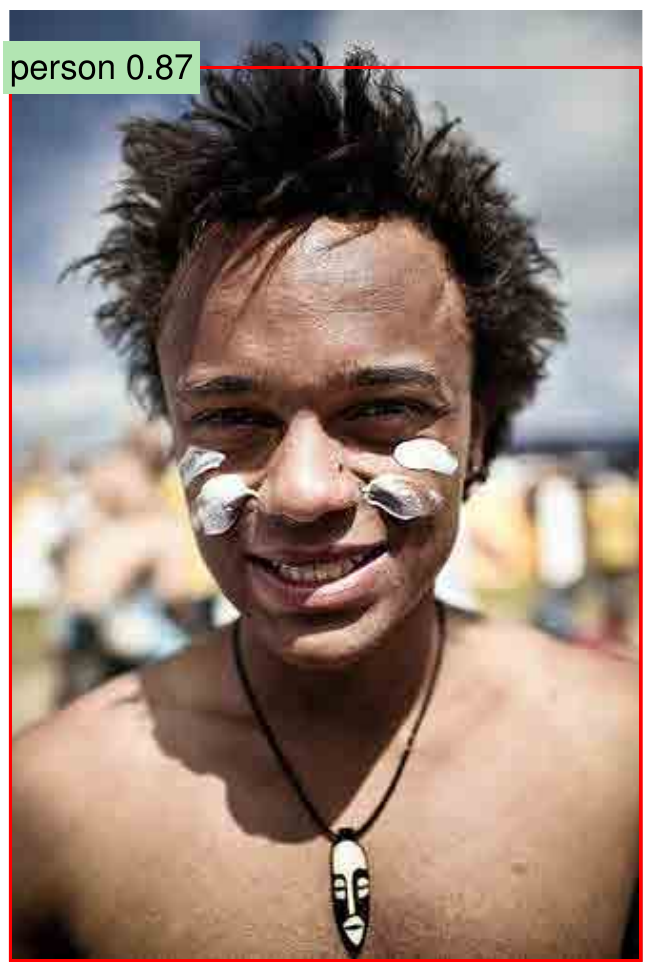}
\includegraphics[height=\sz]{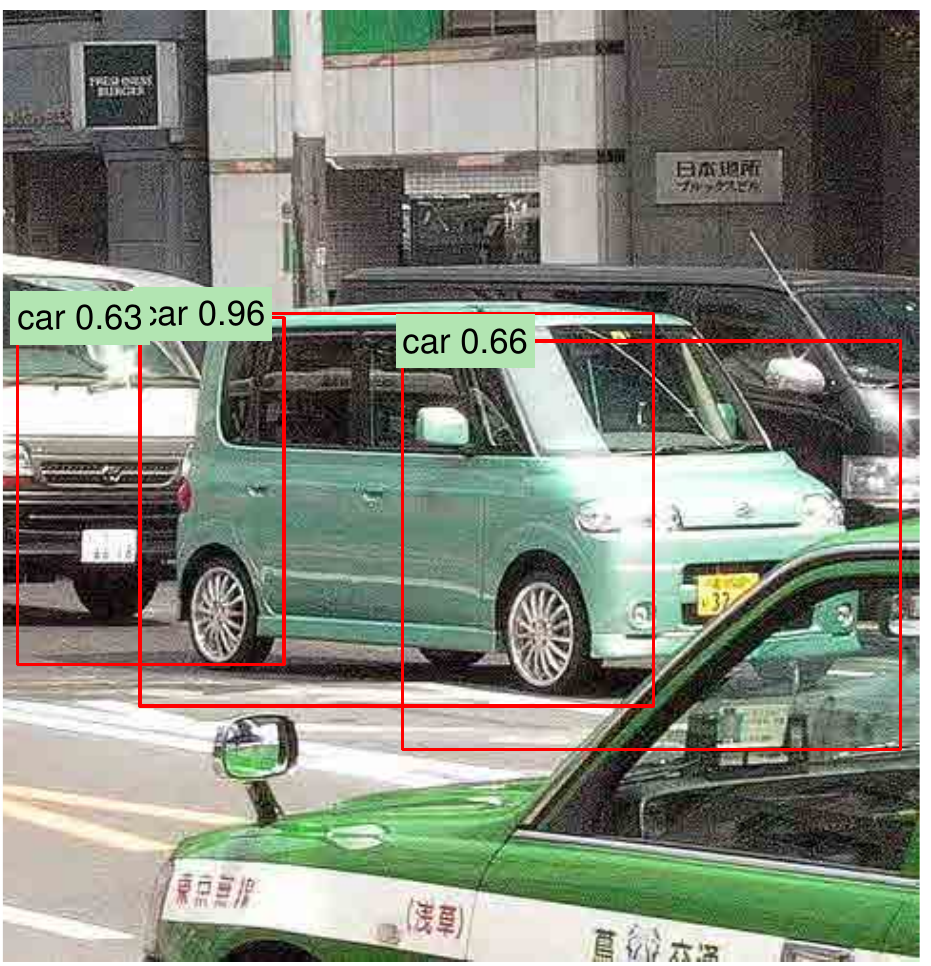}
\includegraphics[height=\sz]{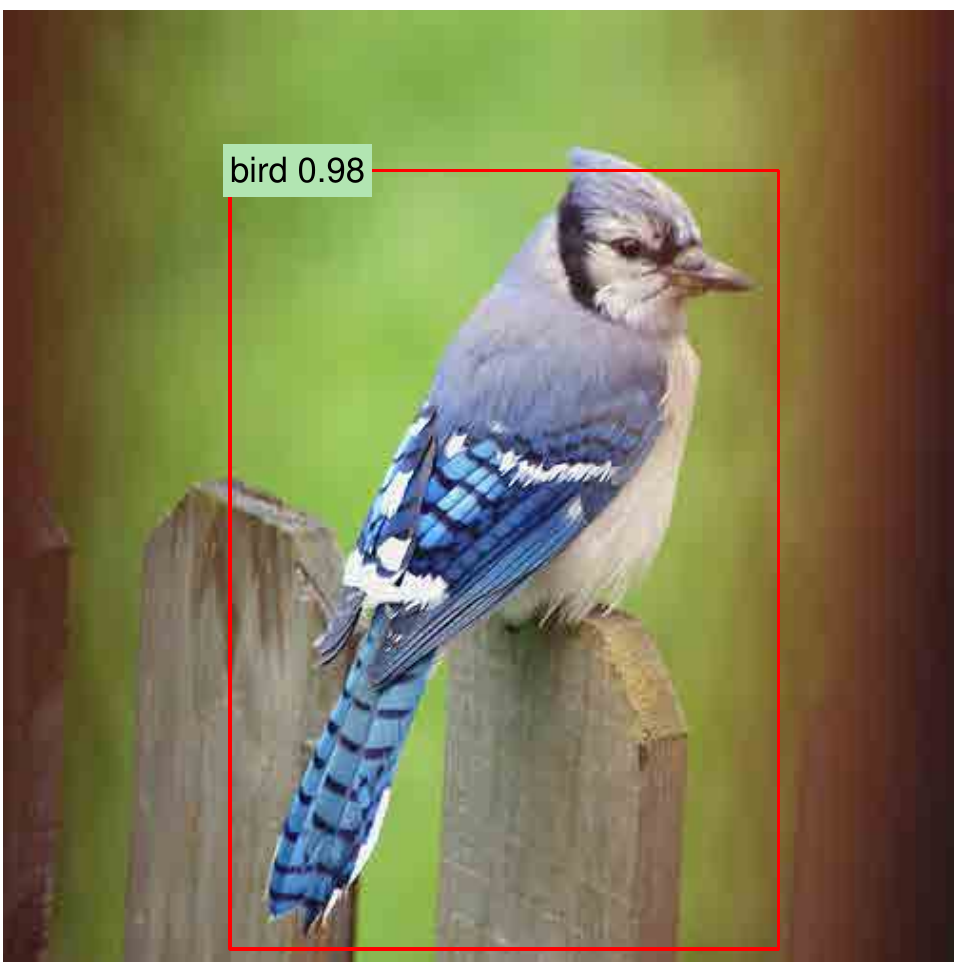}
\includegraphics[height=\sz]{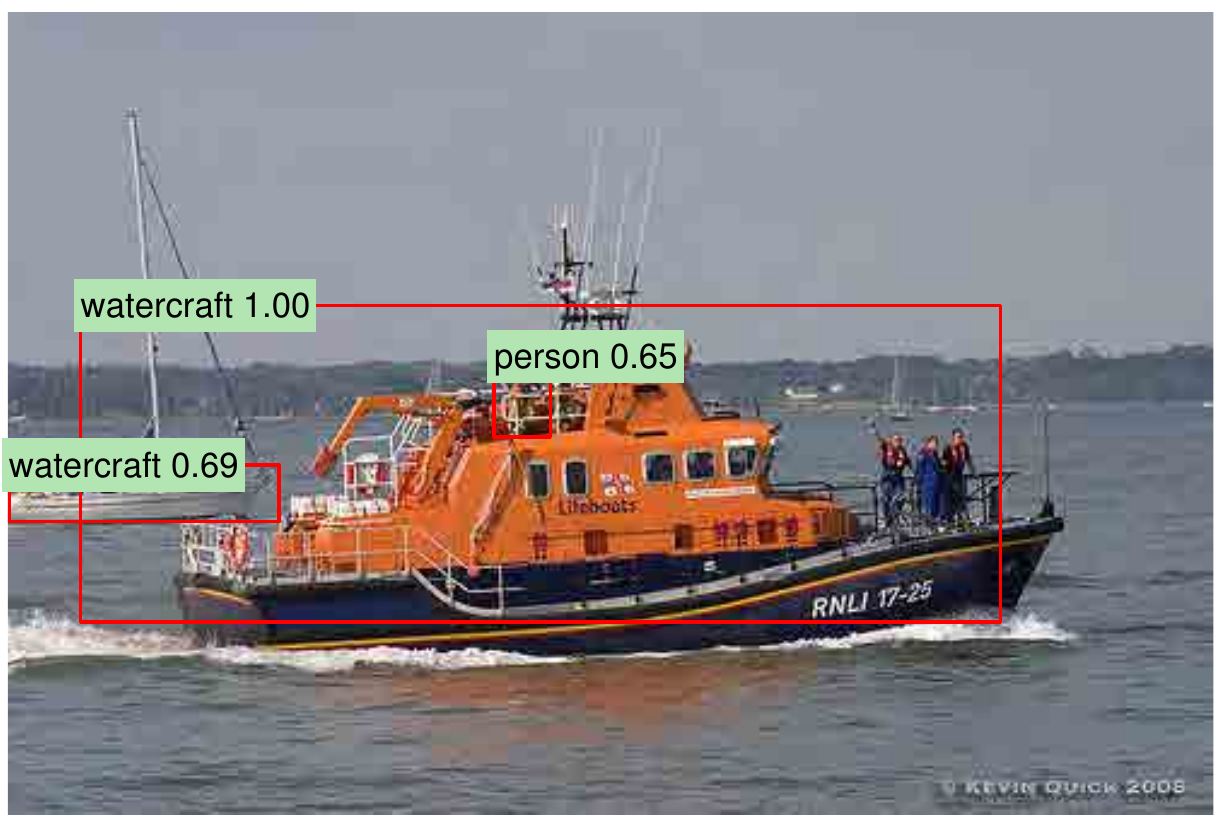}
\includegraphics[height=\sz]{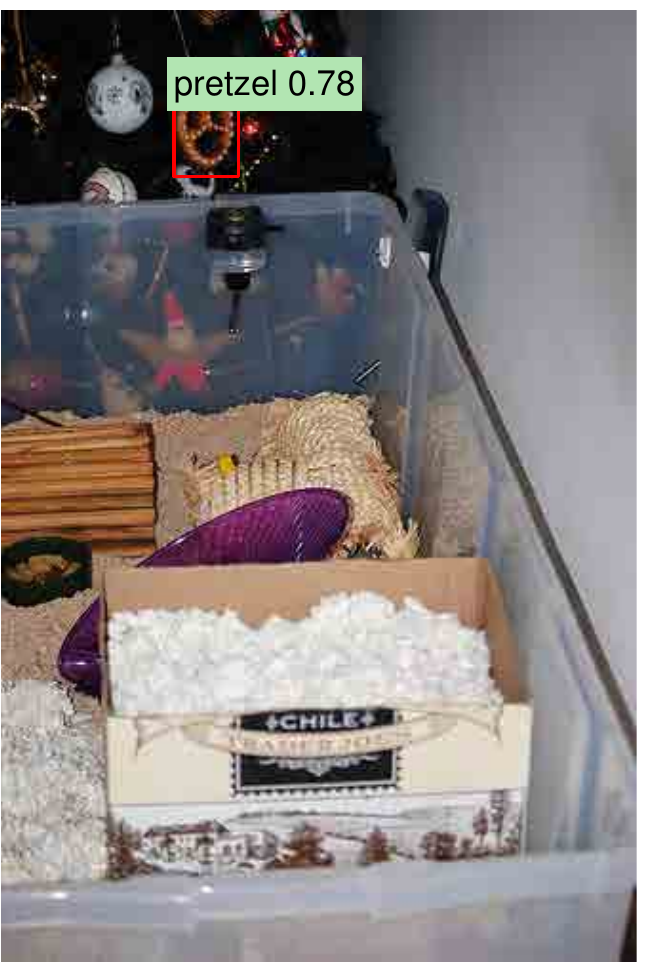}
\includegraphics[height=\sz]{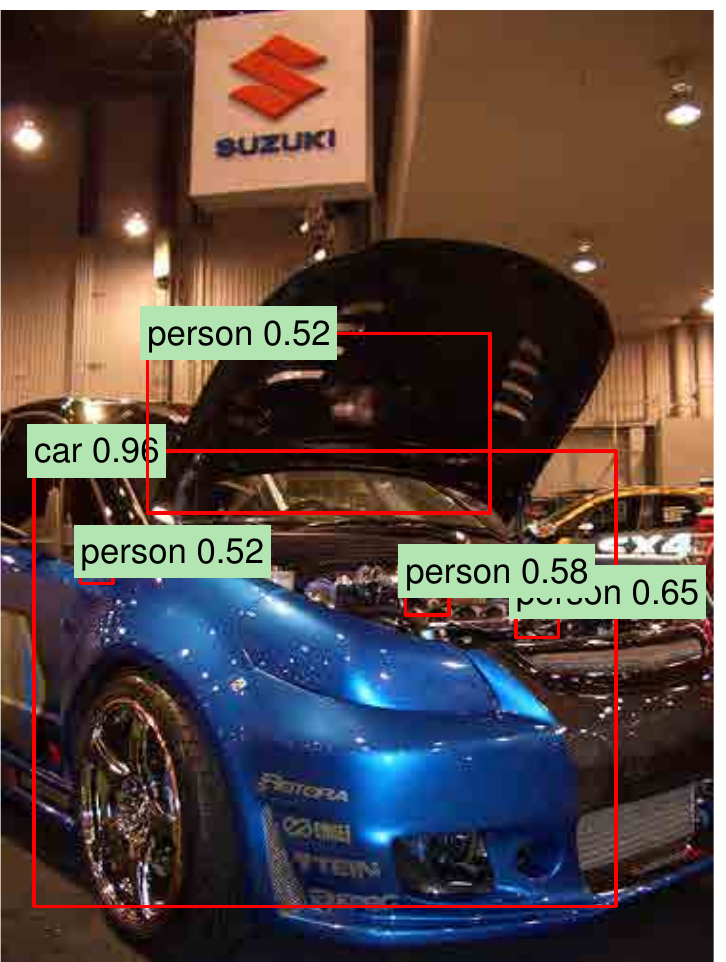}
\includegraphics[height=\sz]{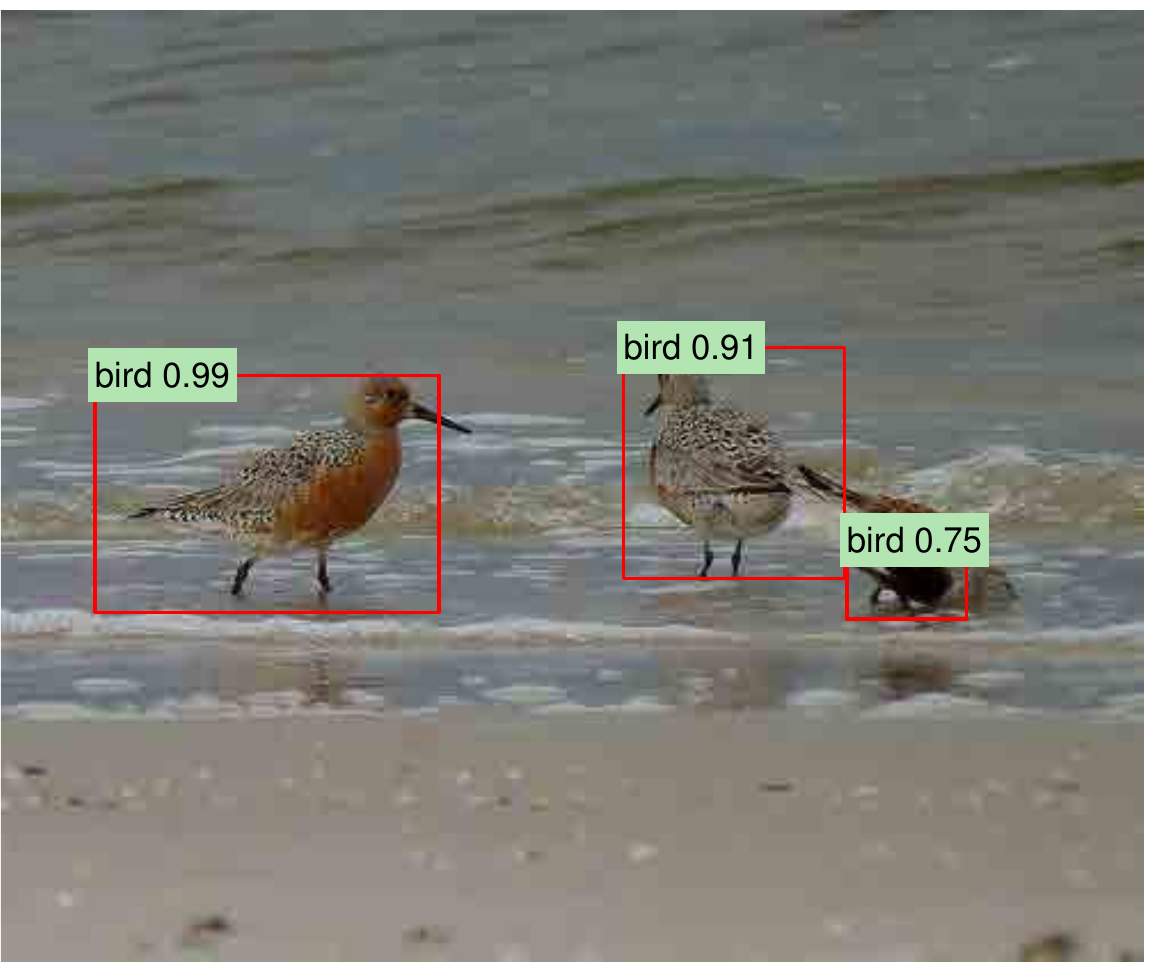}
\includegraphics[height=\sz]{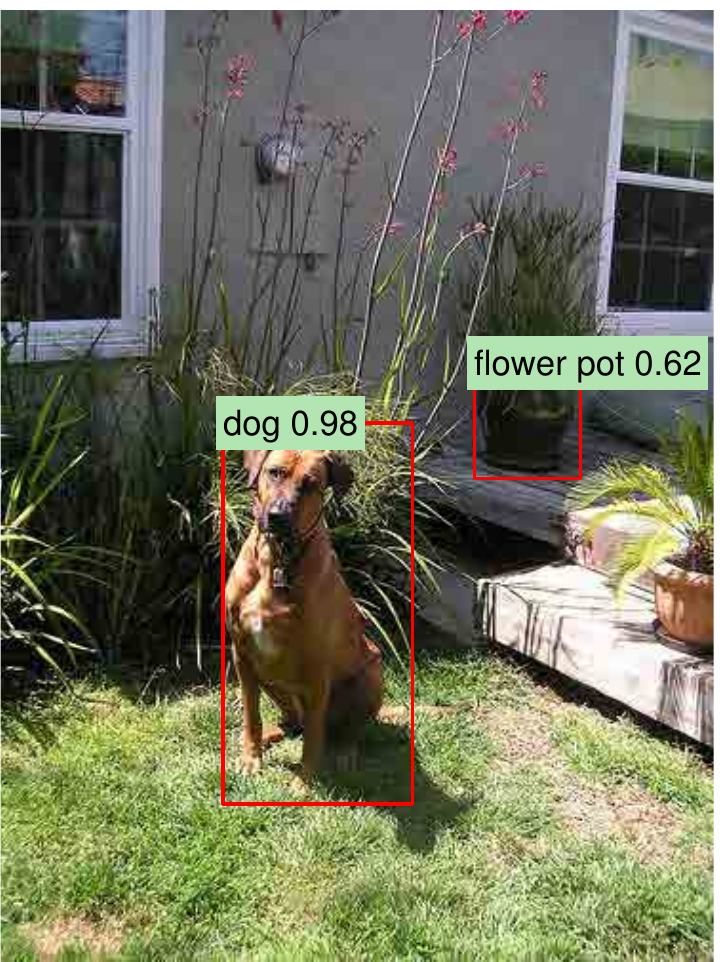}
\includegraphics[height=\sz]{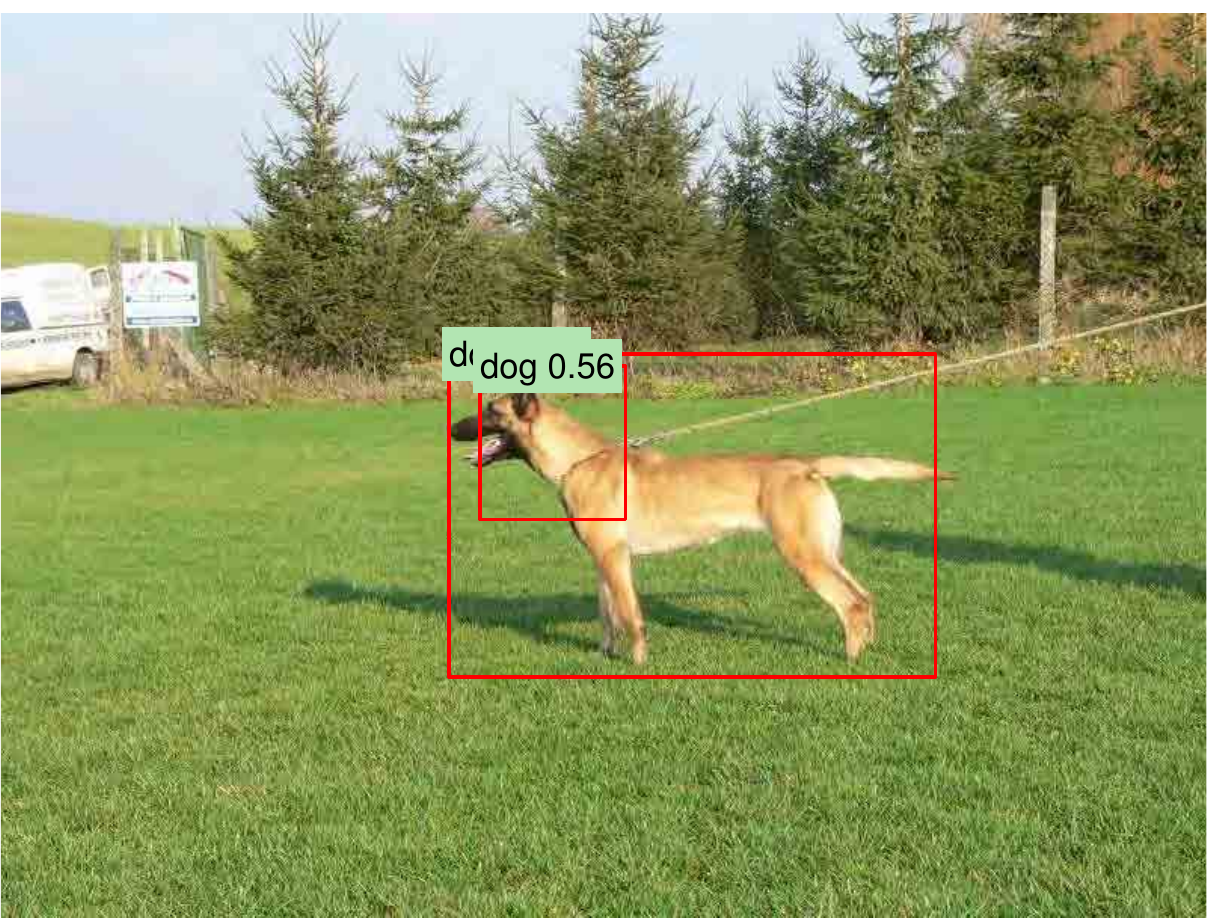}
\includegraphics[height=\sz]{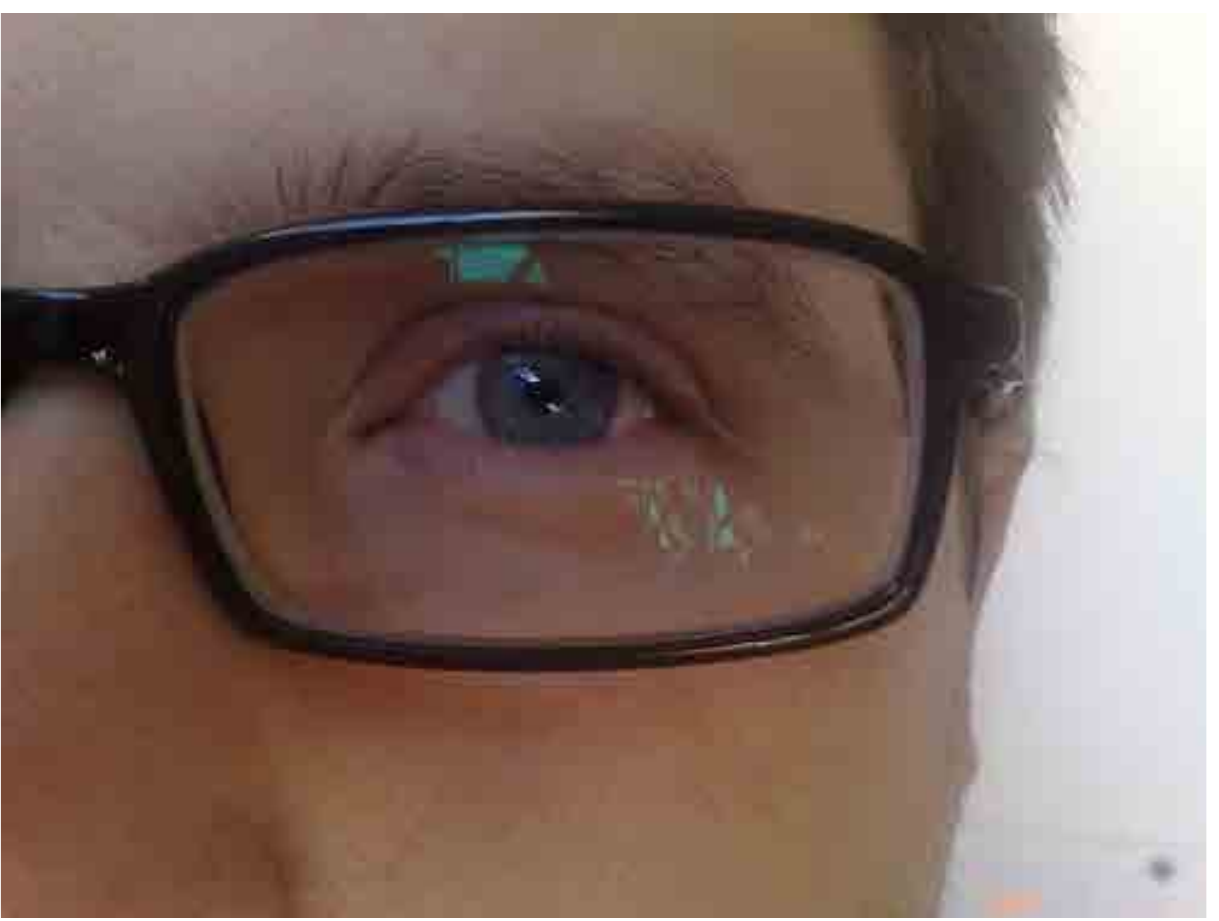}
\includegraphics[height=\sz]{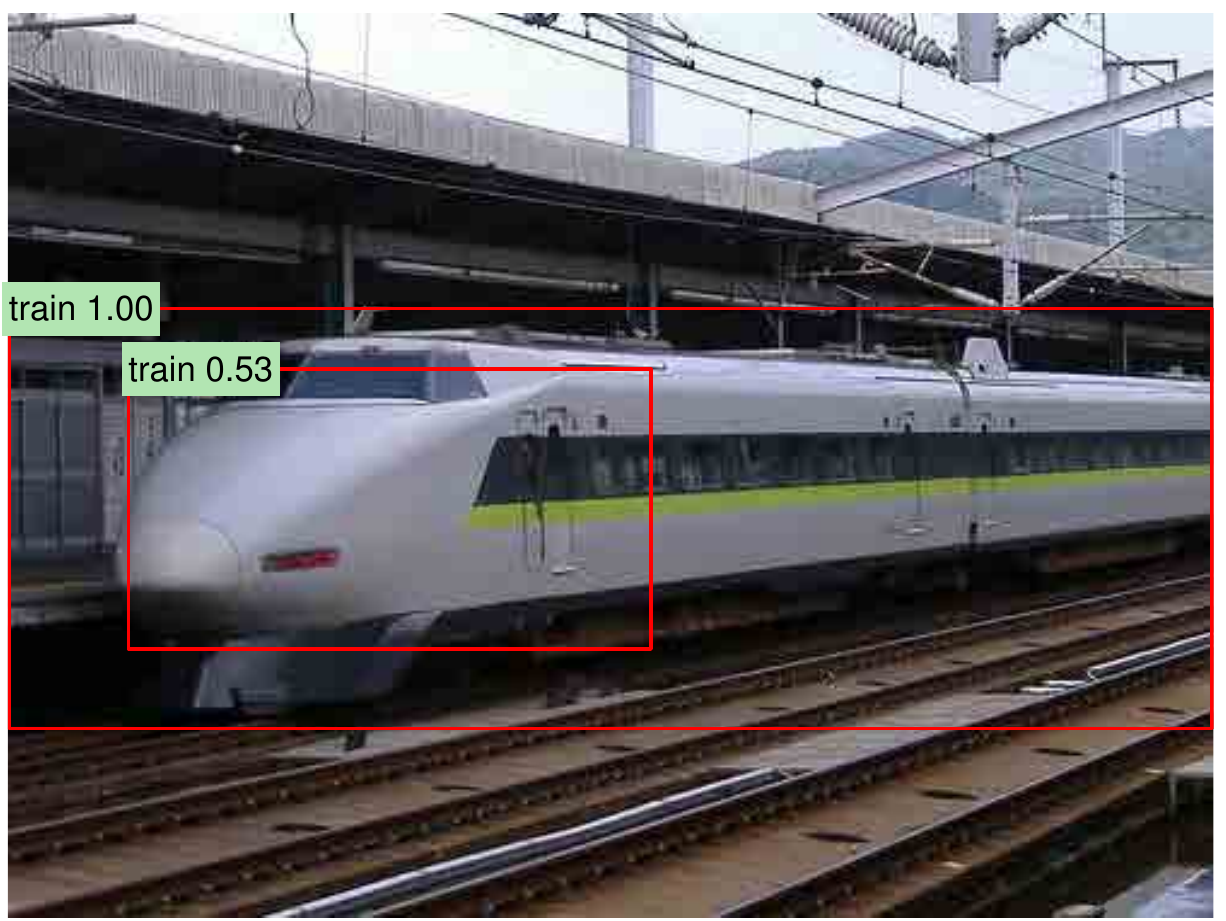}
\includegraphics[height=\sz]{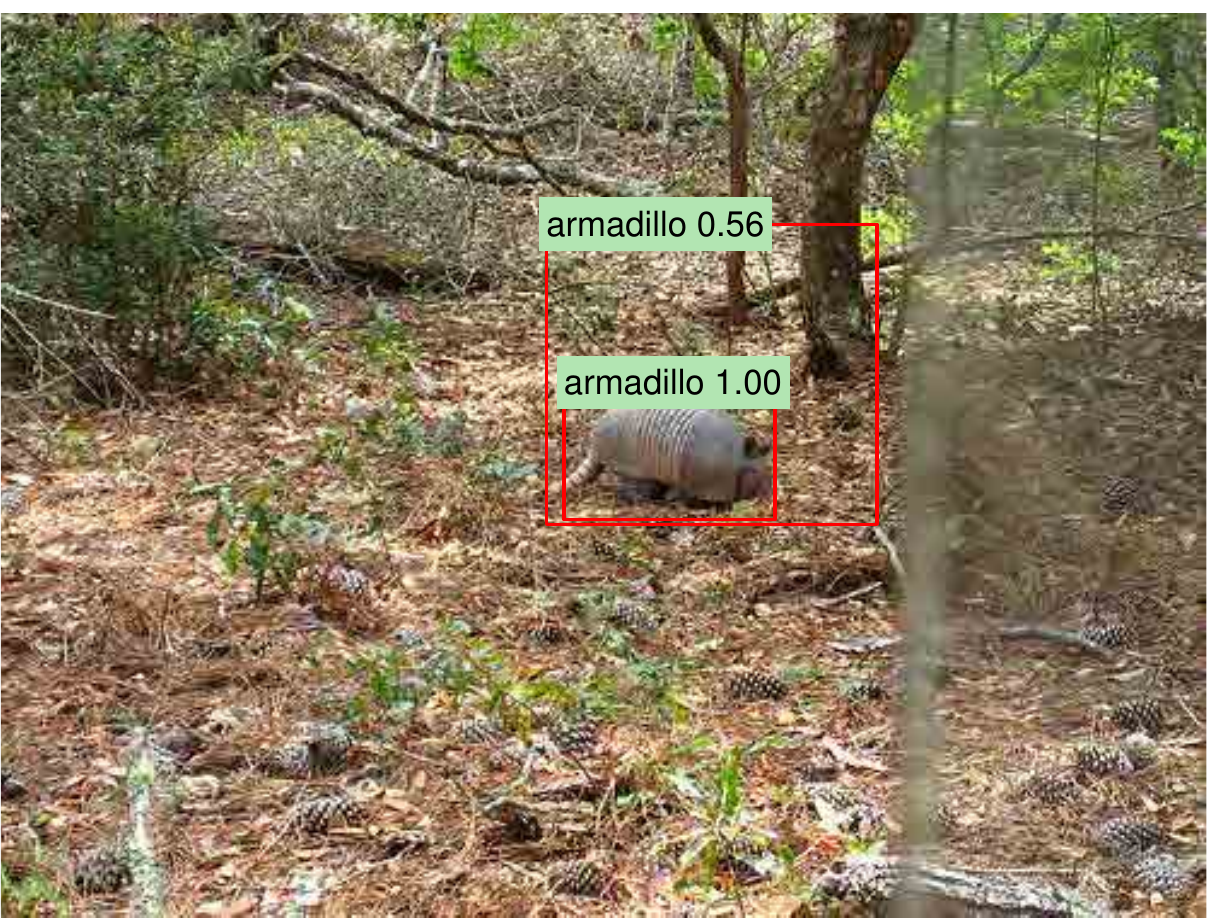}
\includegraphics[height=\sz]{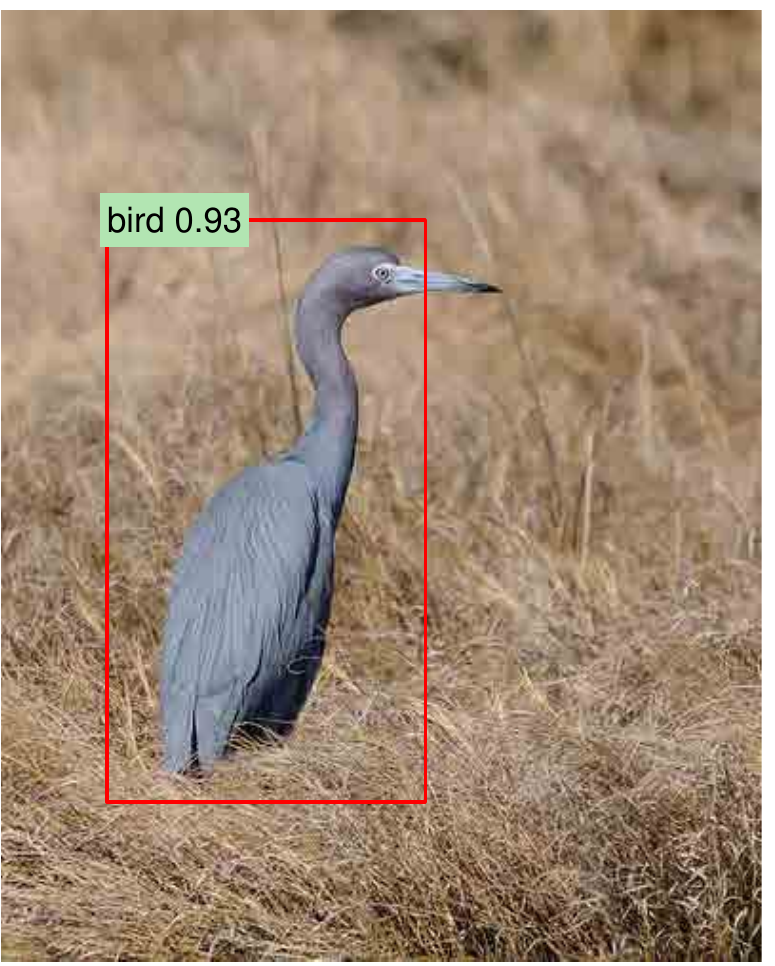}
\includegraphics[height=\sz]{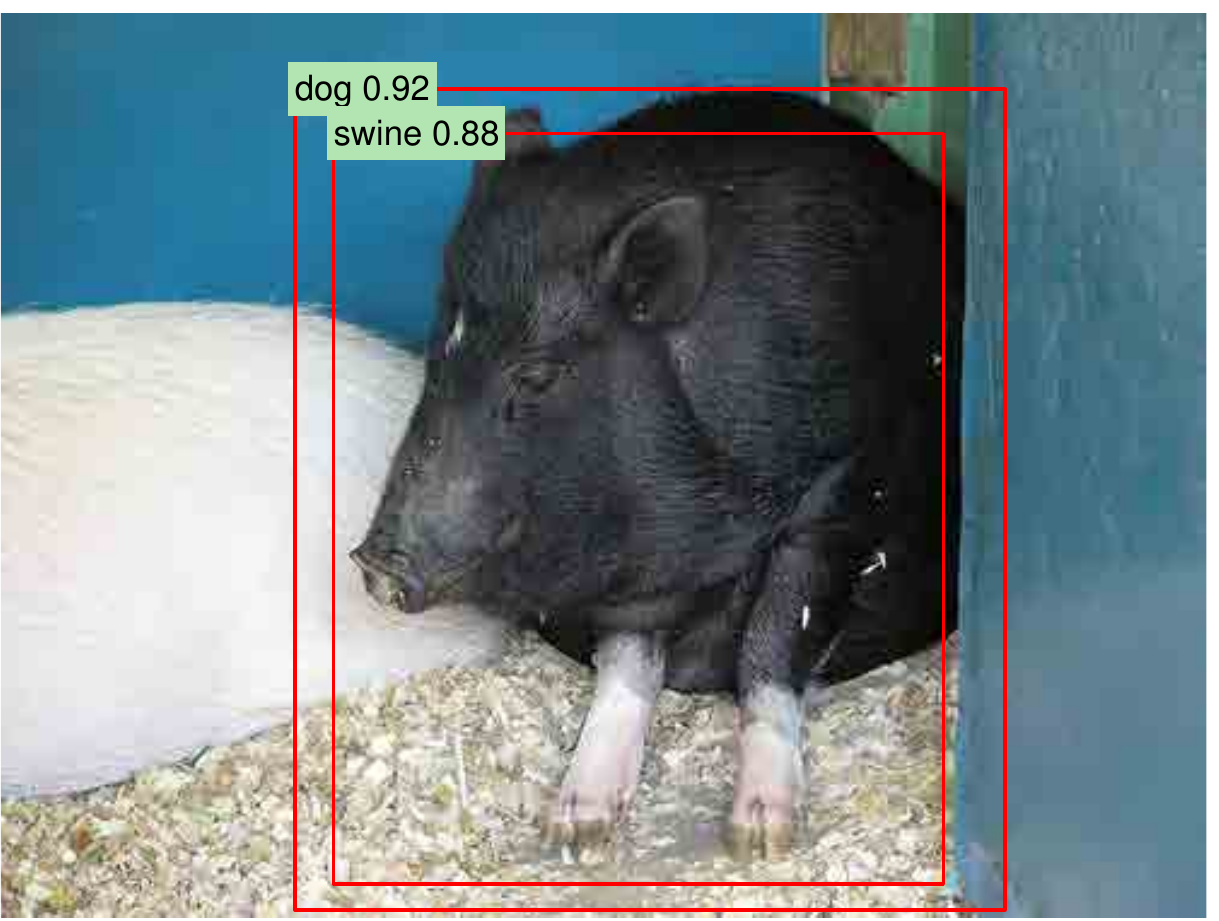}
\includegraphics[height=\sz]{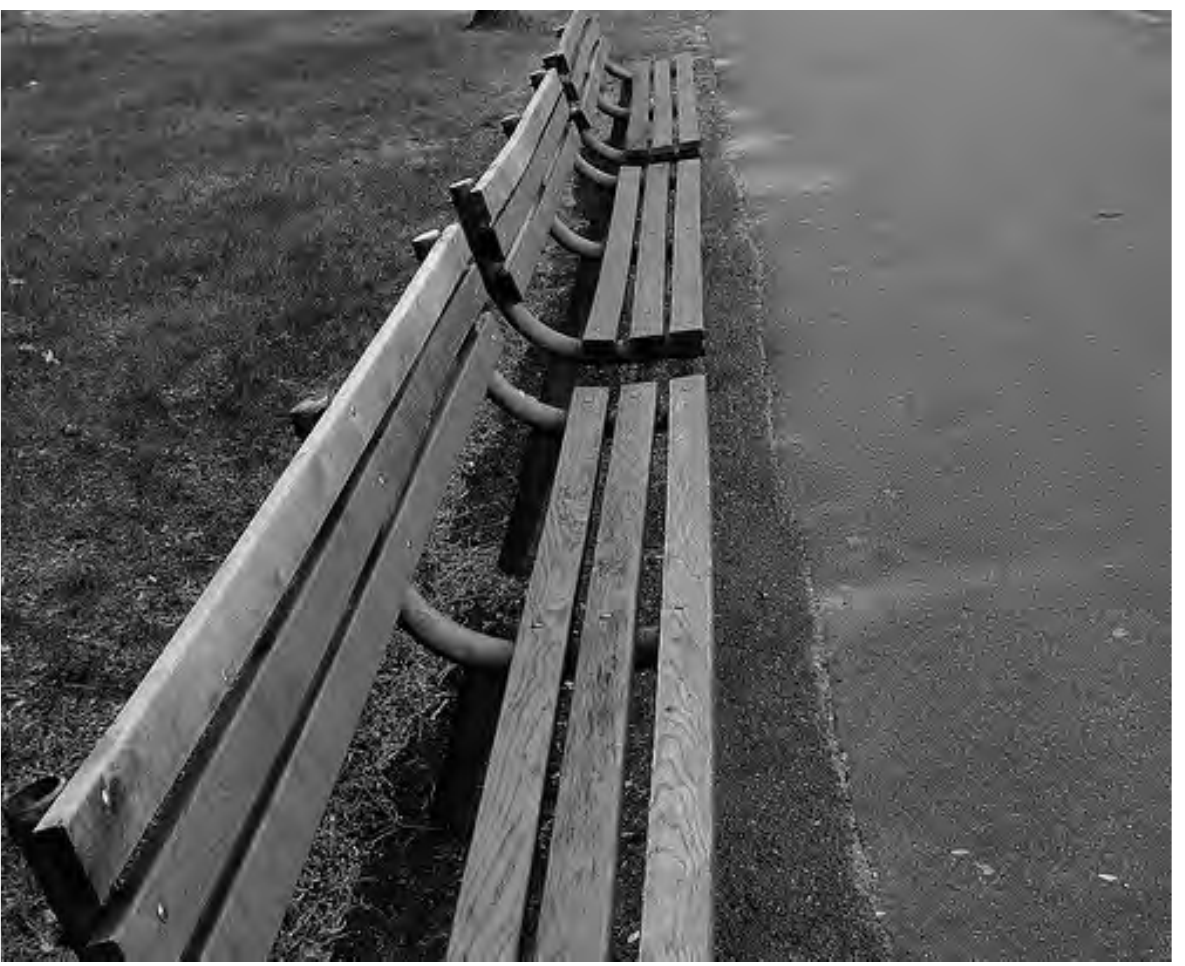}
\includegraphics[height=\sz]{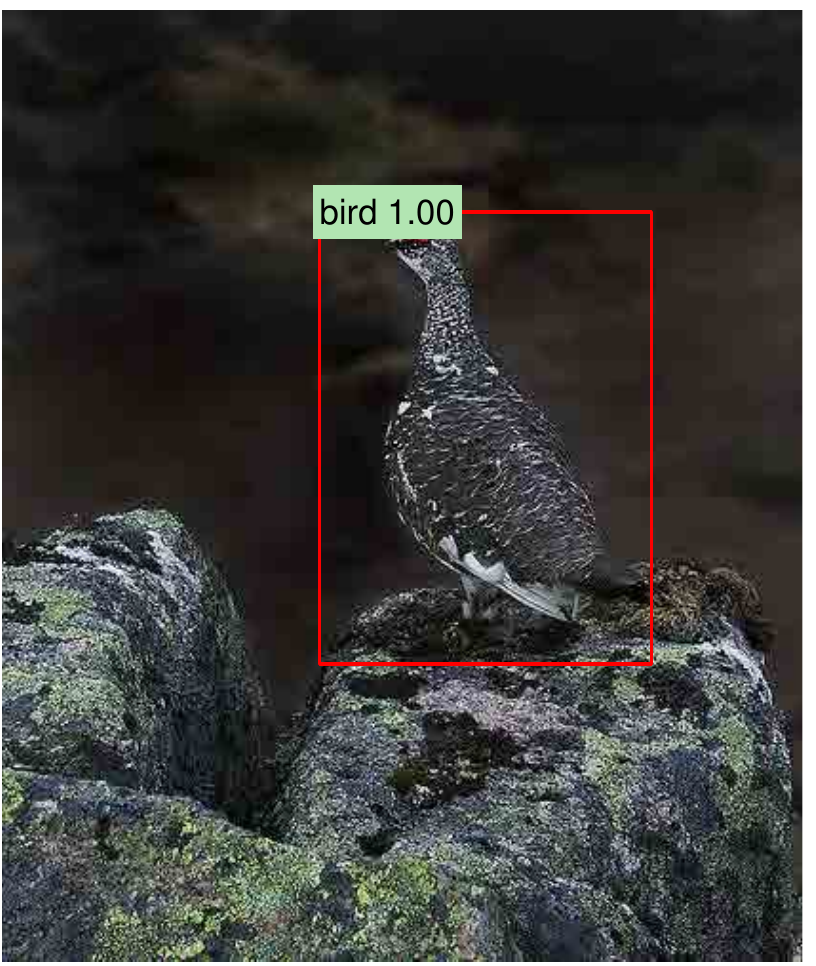}
\includegraphics[height=\sz]{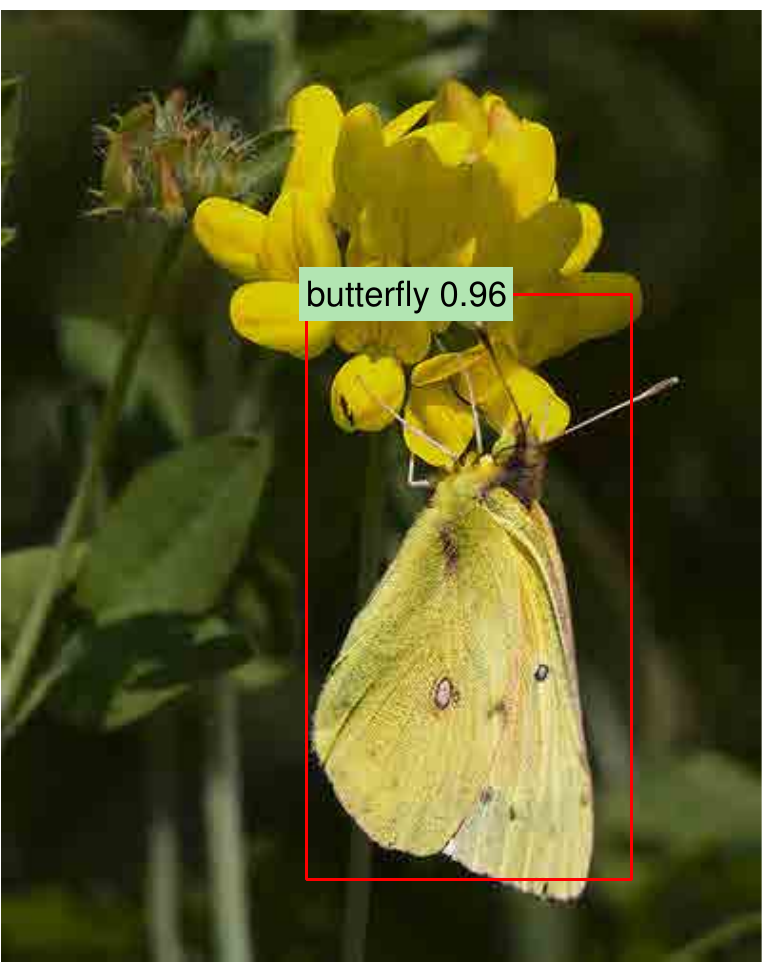}
\includegraphics[height=\sz]{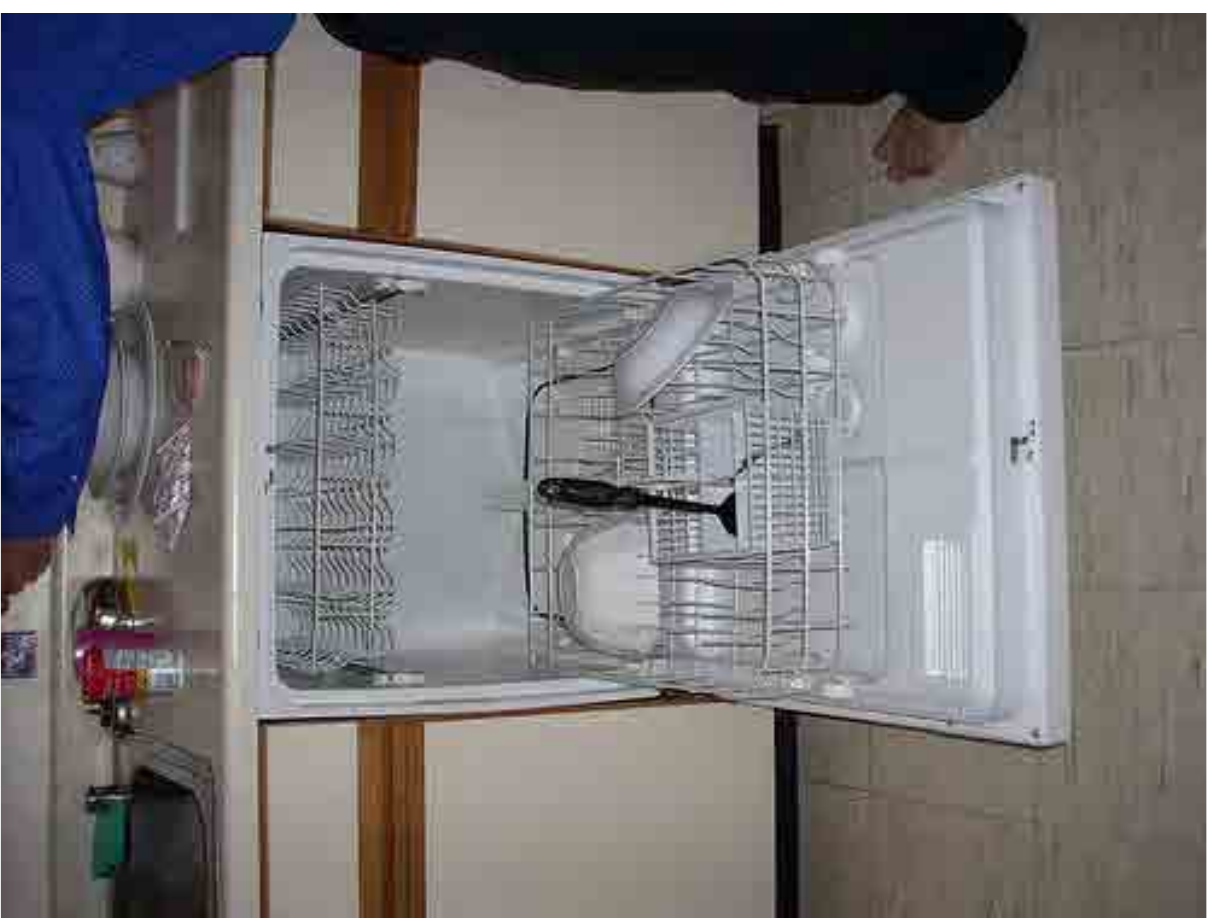}
\includegraphics[height=\sz]{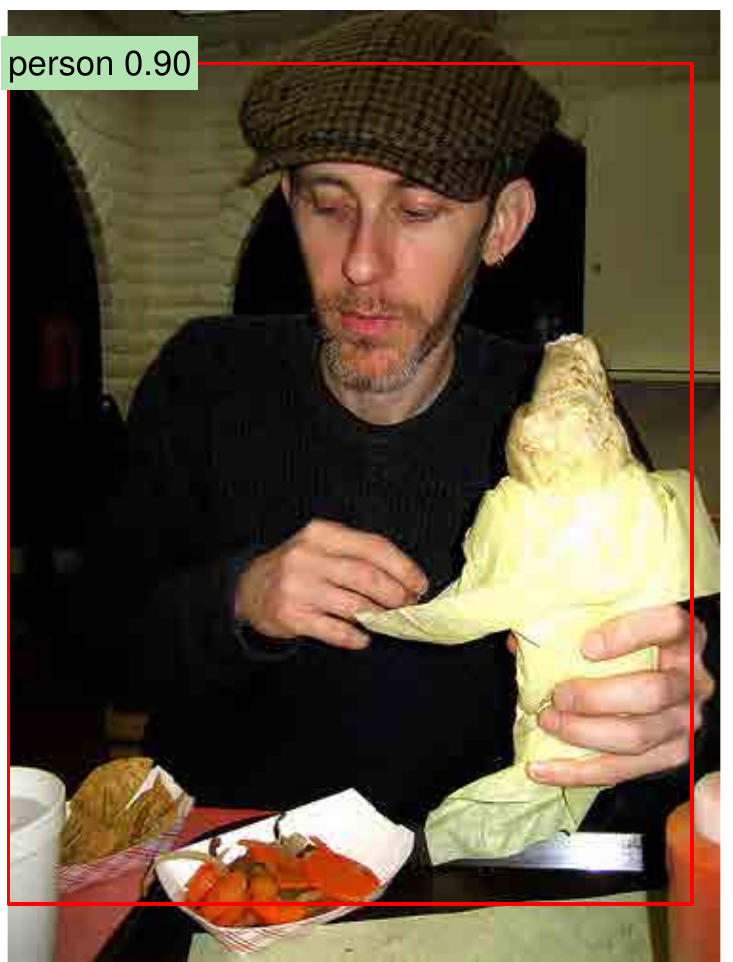}
\includegraphics[height=\sz]{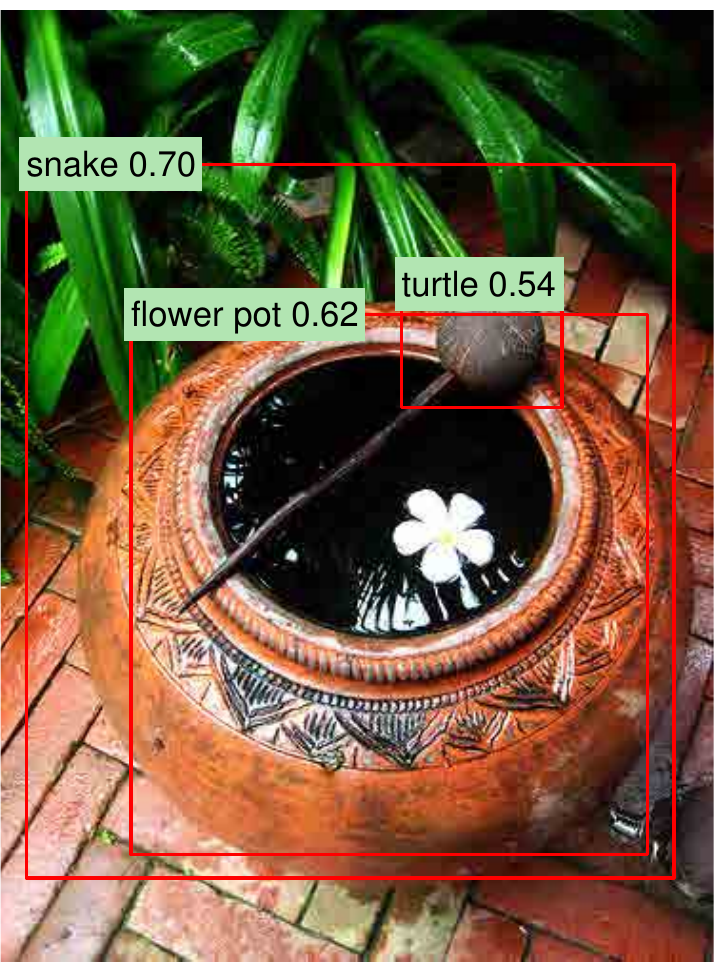}
\includegraphics[height=\sz]{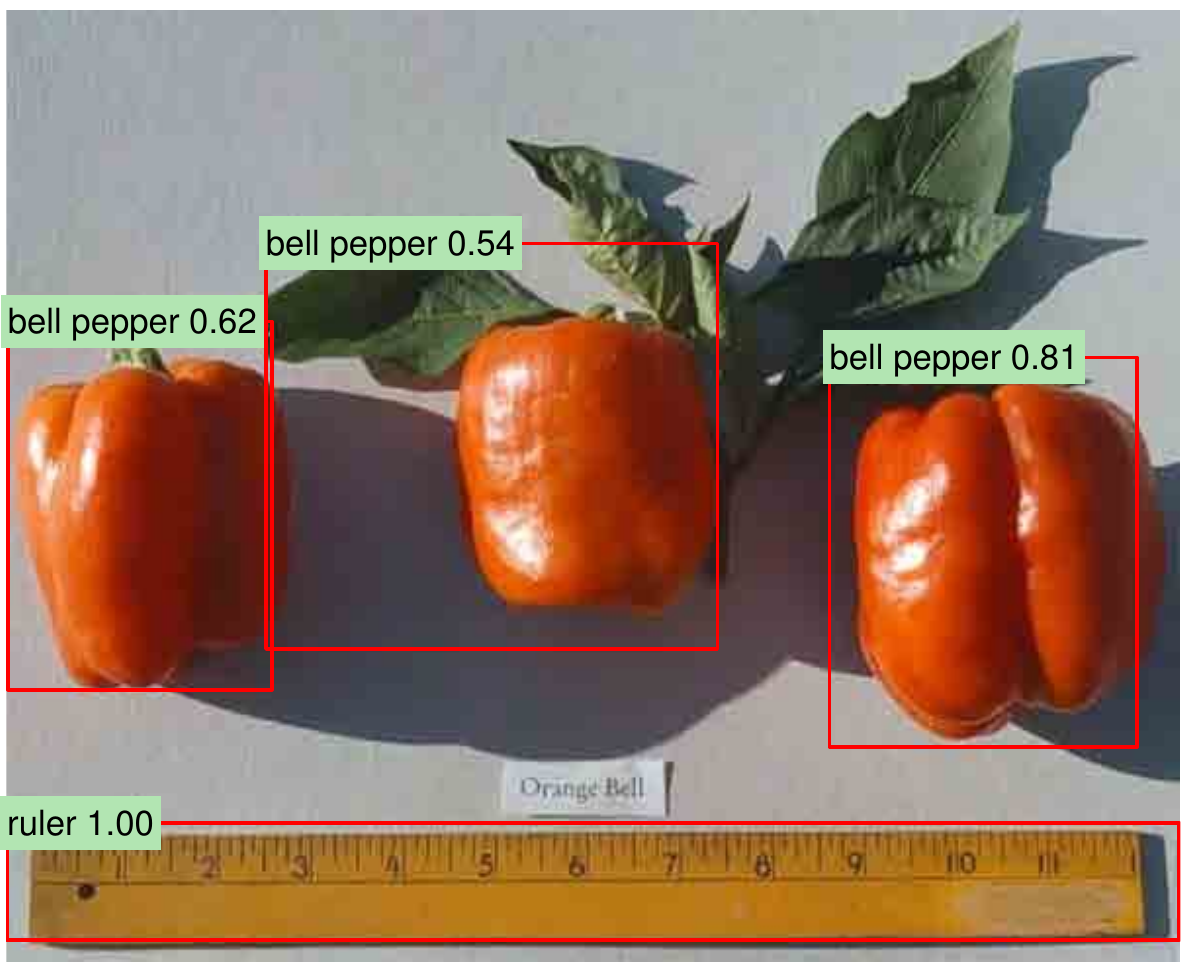}
\includegraphics[height=\sz]{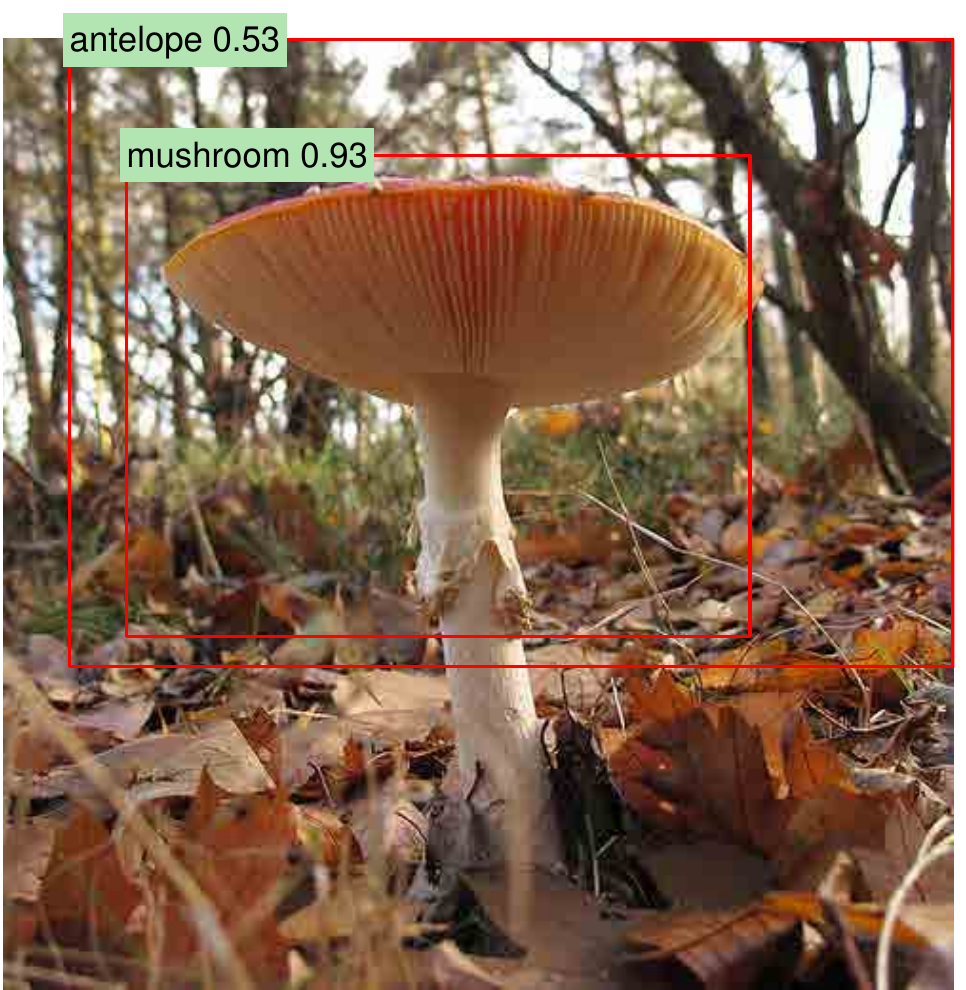}
\includegraphics[height=\sz]{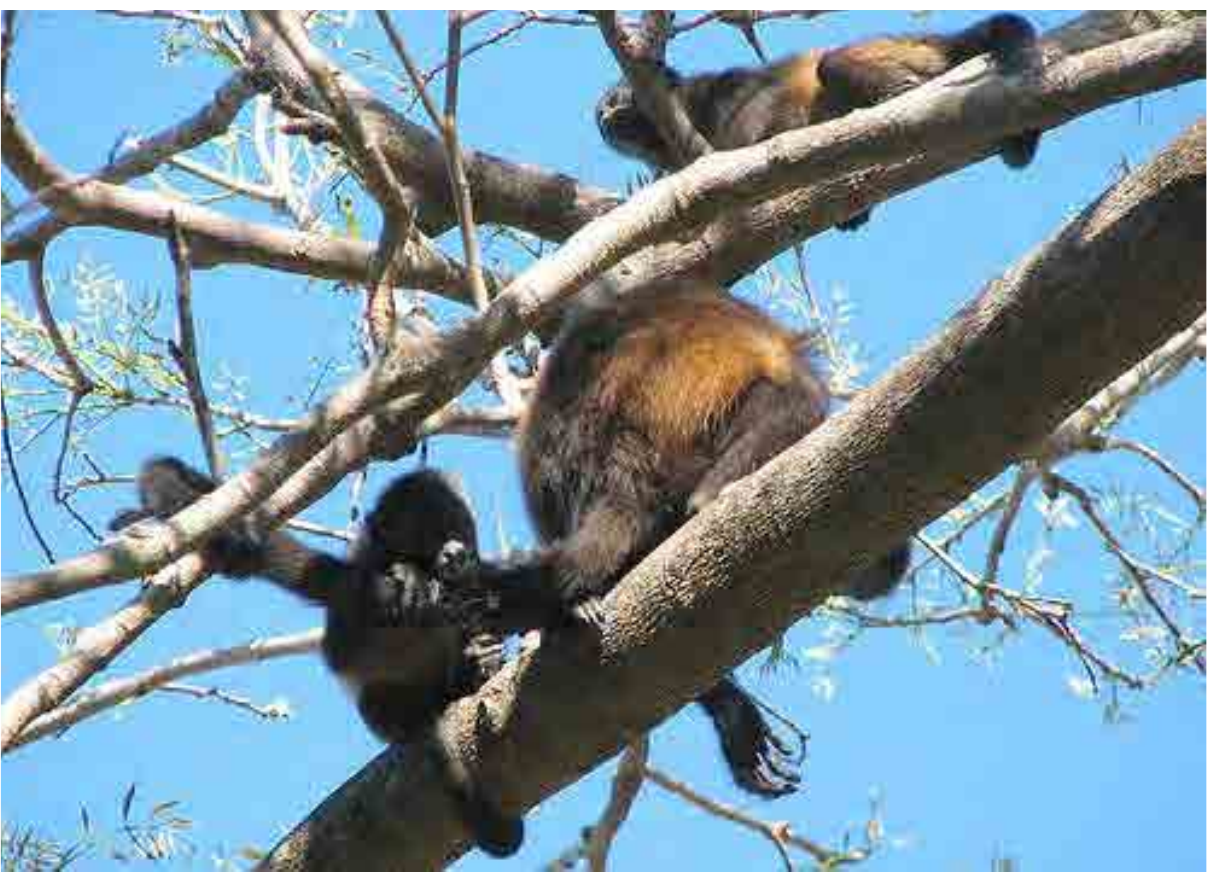}
\includegraphics[height=\sz]{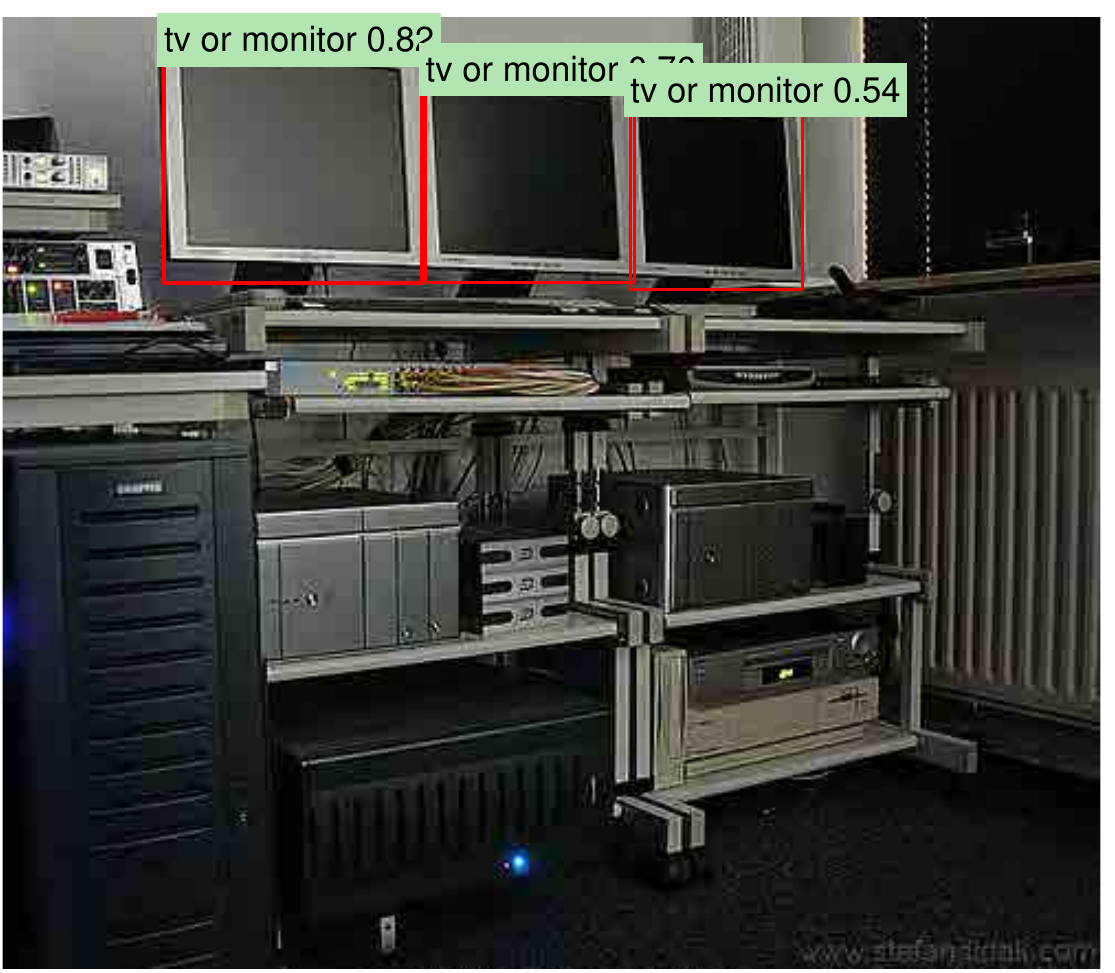}
\includegraphics[height=\sz]{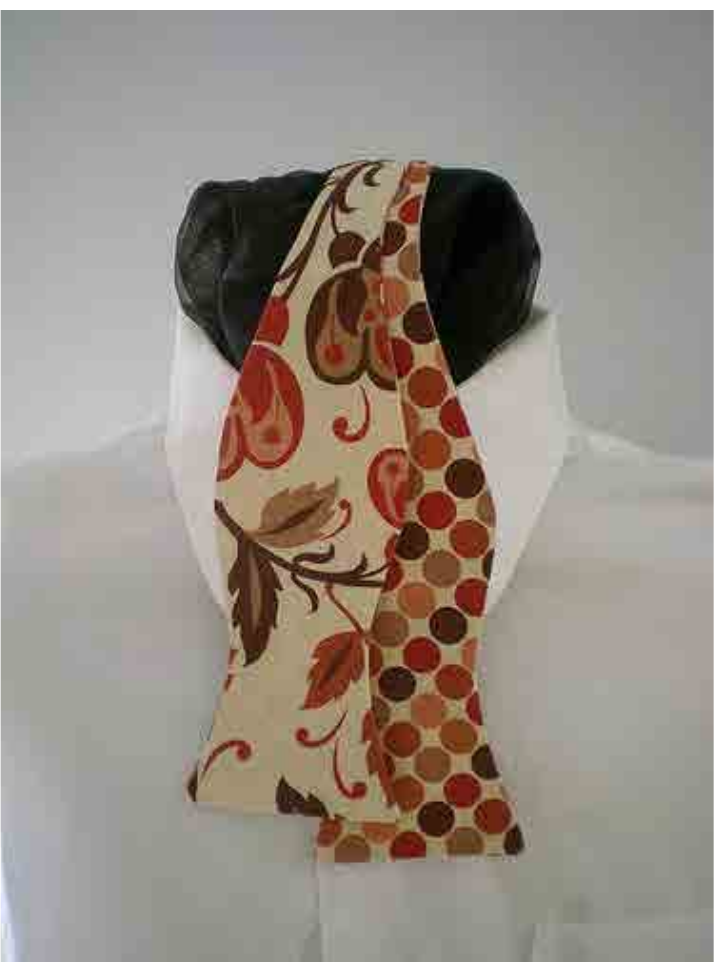}
\includegraphics[height=\sz]{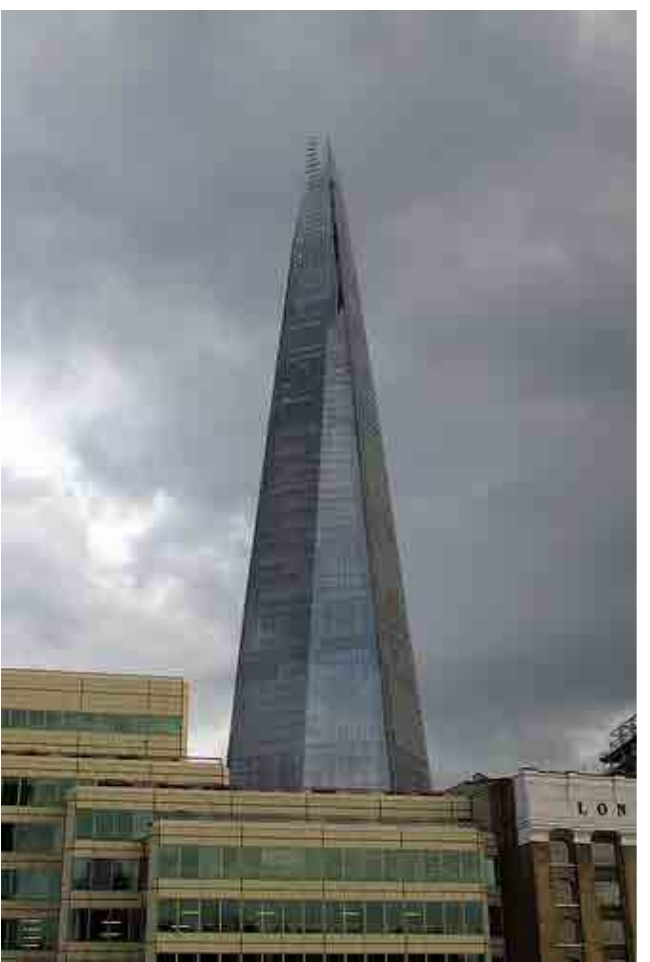}
\includegraphics[height=\sz]{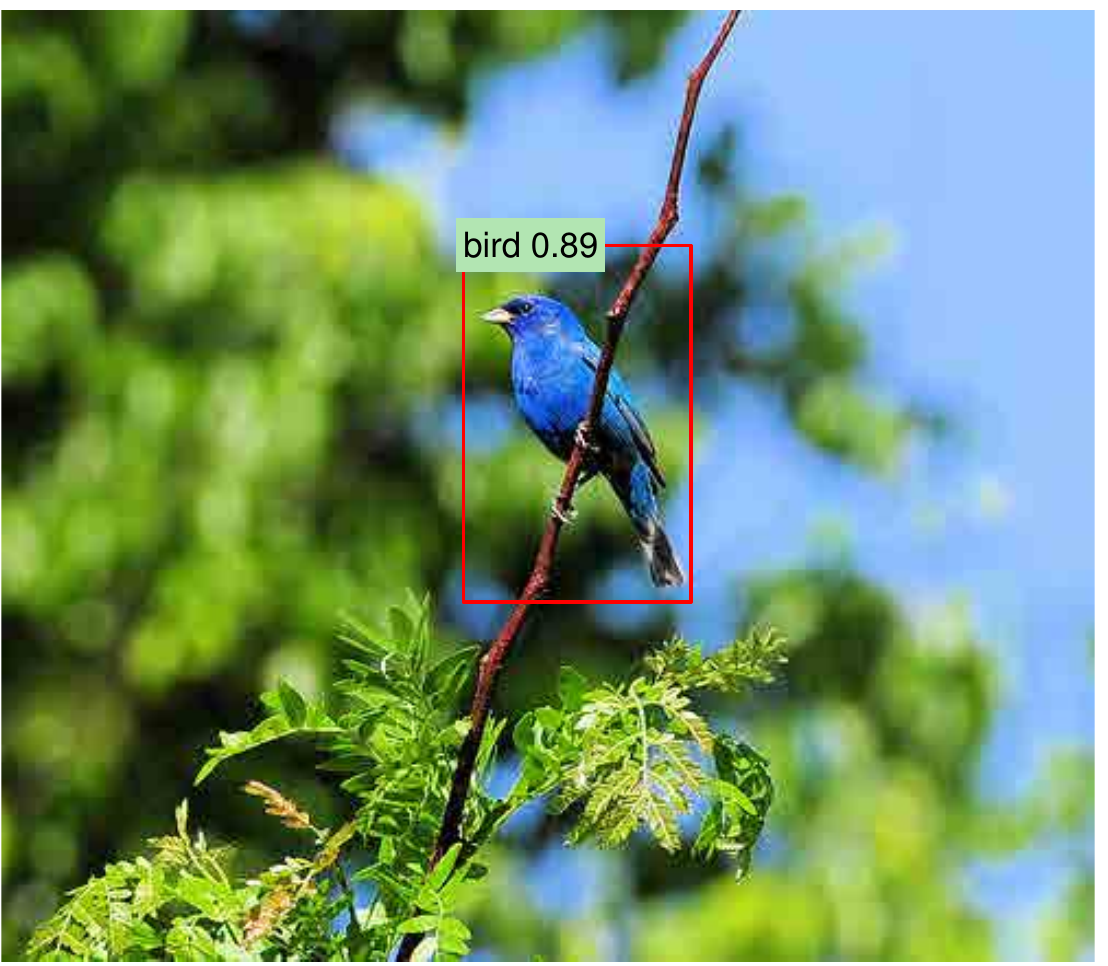}
\includegraphics[height=\sz]{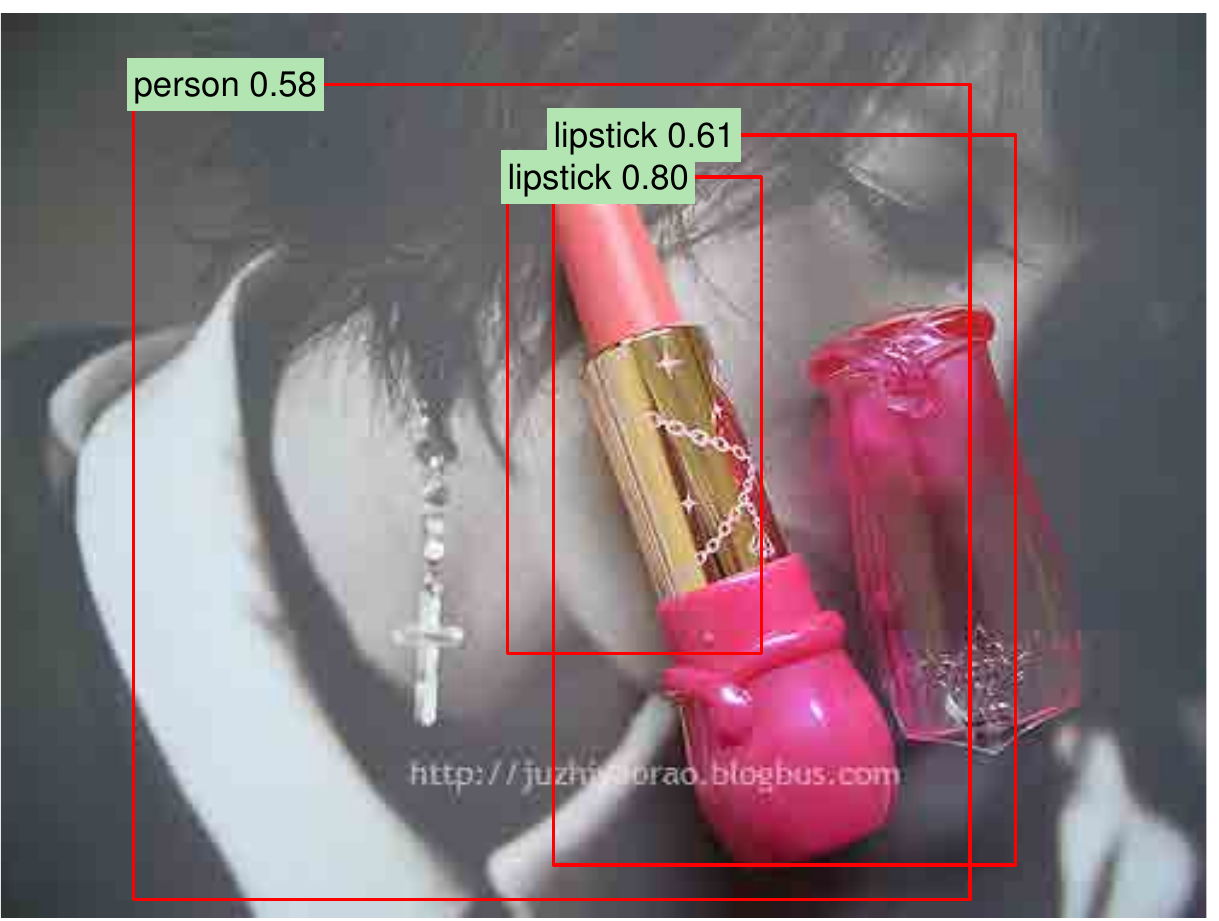}
\includegraphics[height=\sz]{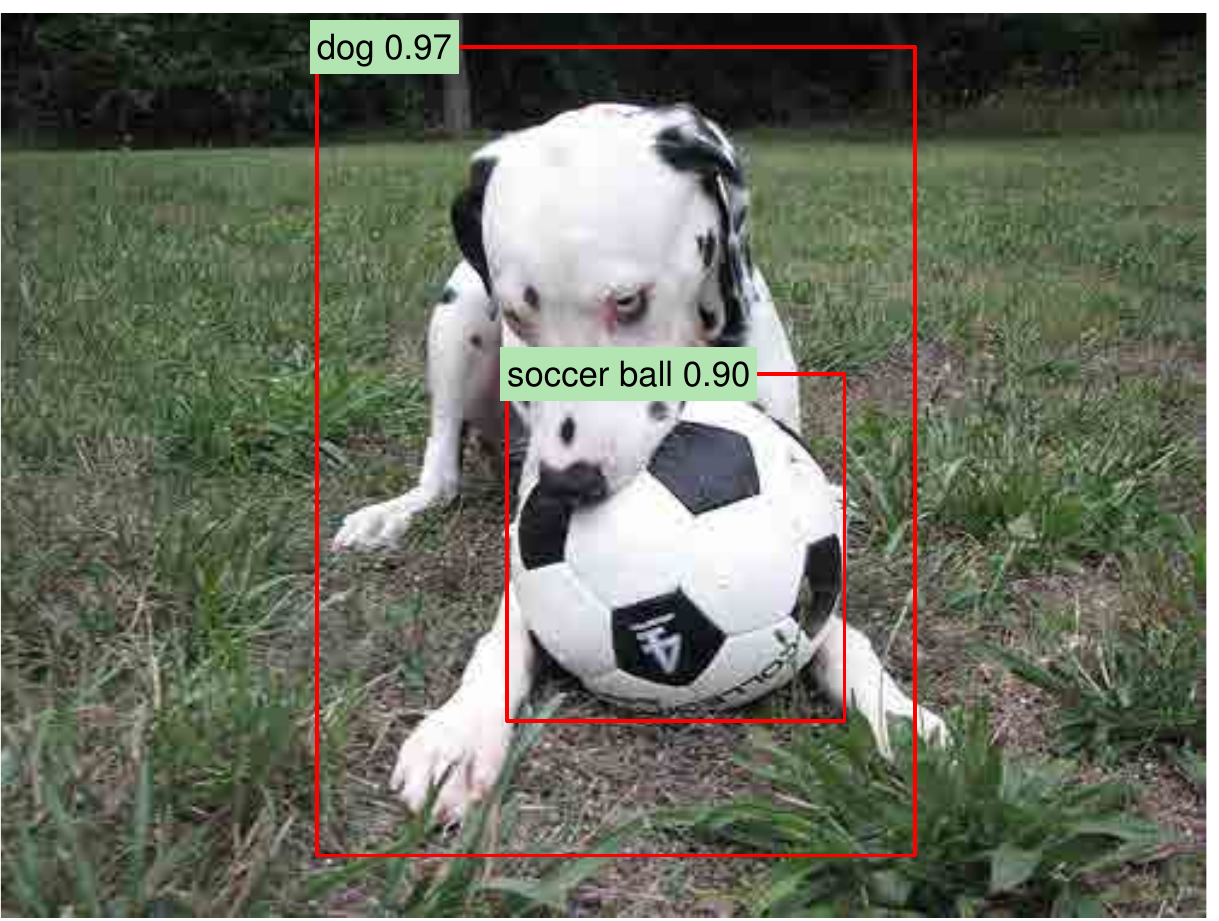}
\includegraphics[height=\sz]{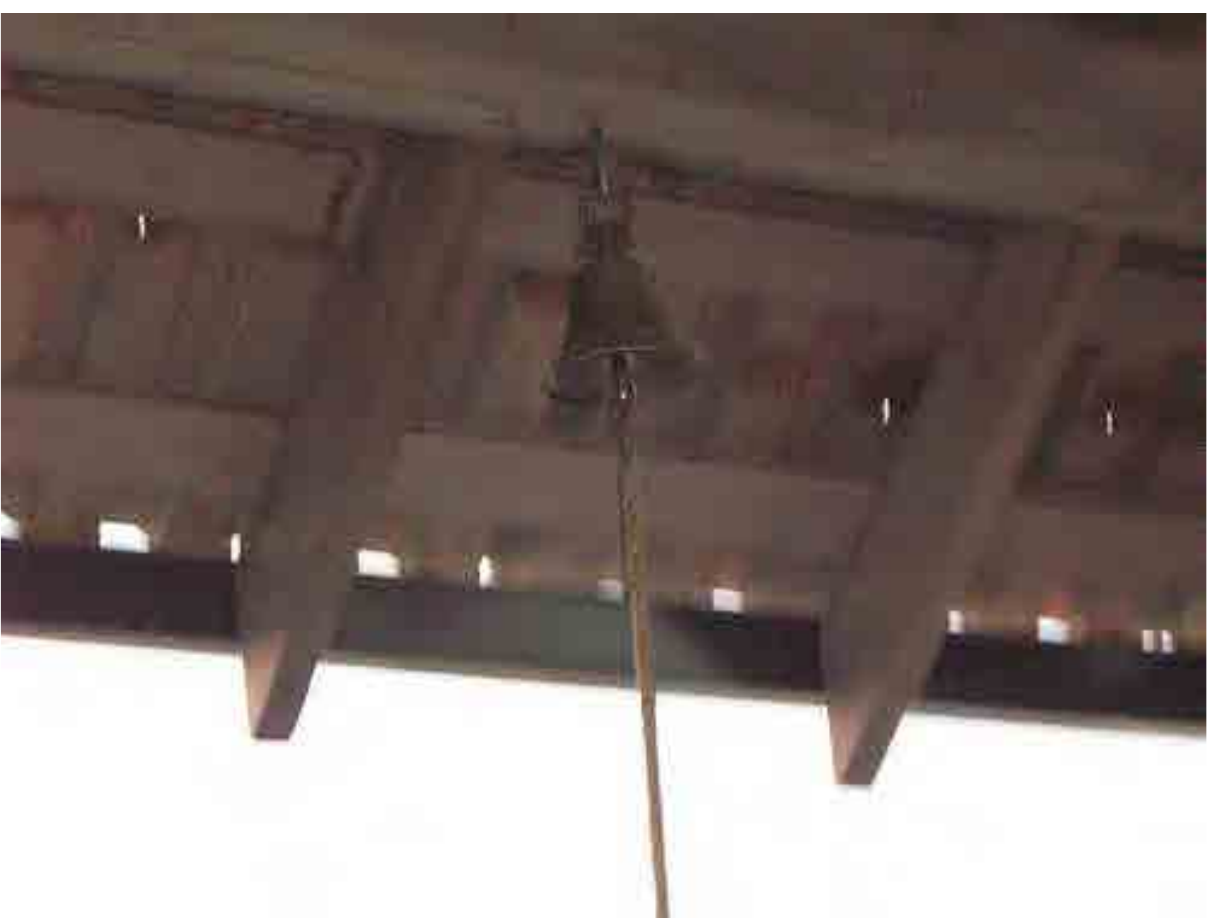}
\end{center}
\caption{Example detections on the \valb set from the configuration that achieved 31.0\% mAP on \valb.
Each image was sampled randomly (these are \emph{not} curated).
All detections at precision greater than 0.5 are shown.
Each detection is labeled with the predicted class and the precision value of that detection from the detector's precision-recall curve.
Viewing digitally with zoom is recommended.
}
\figlabel{examples1}
\end{figure*}

\begin{figure*}[t]
\begin{center}
\def \sz {0.99in}
\includegraphics[height=\sz]{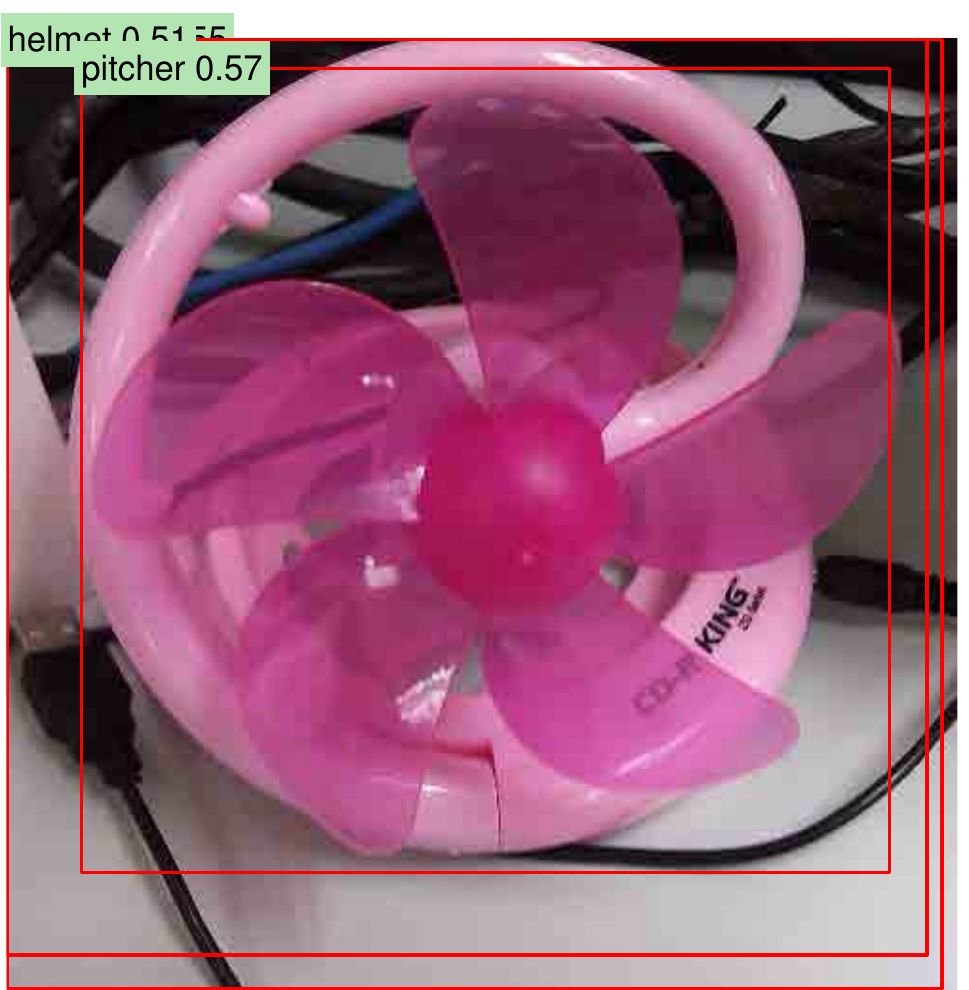}
\includegraphics[height=\sz]{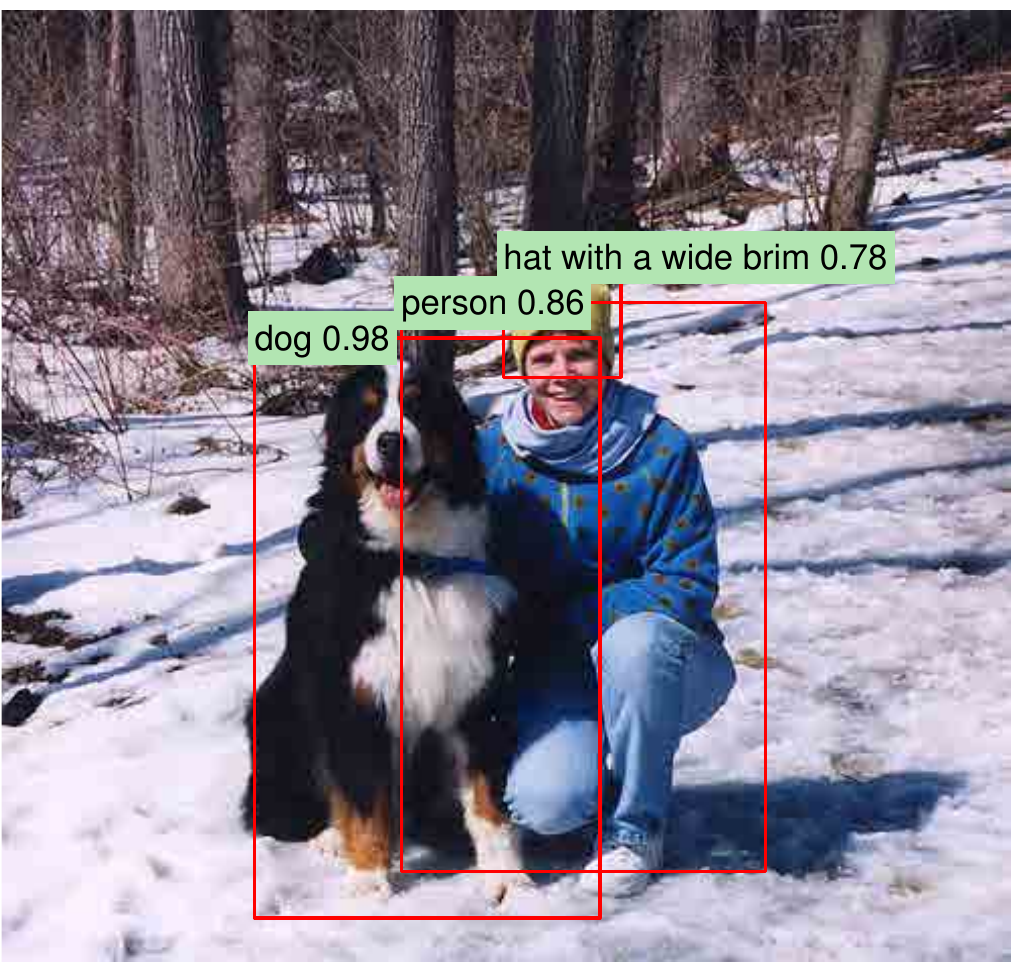}
\includegraphics[height=\sz]{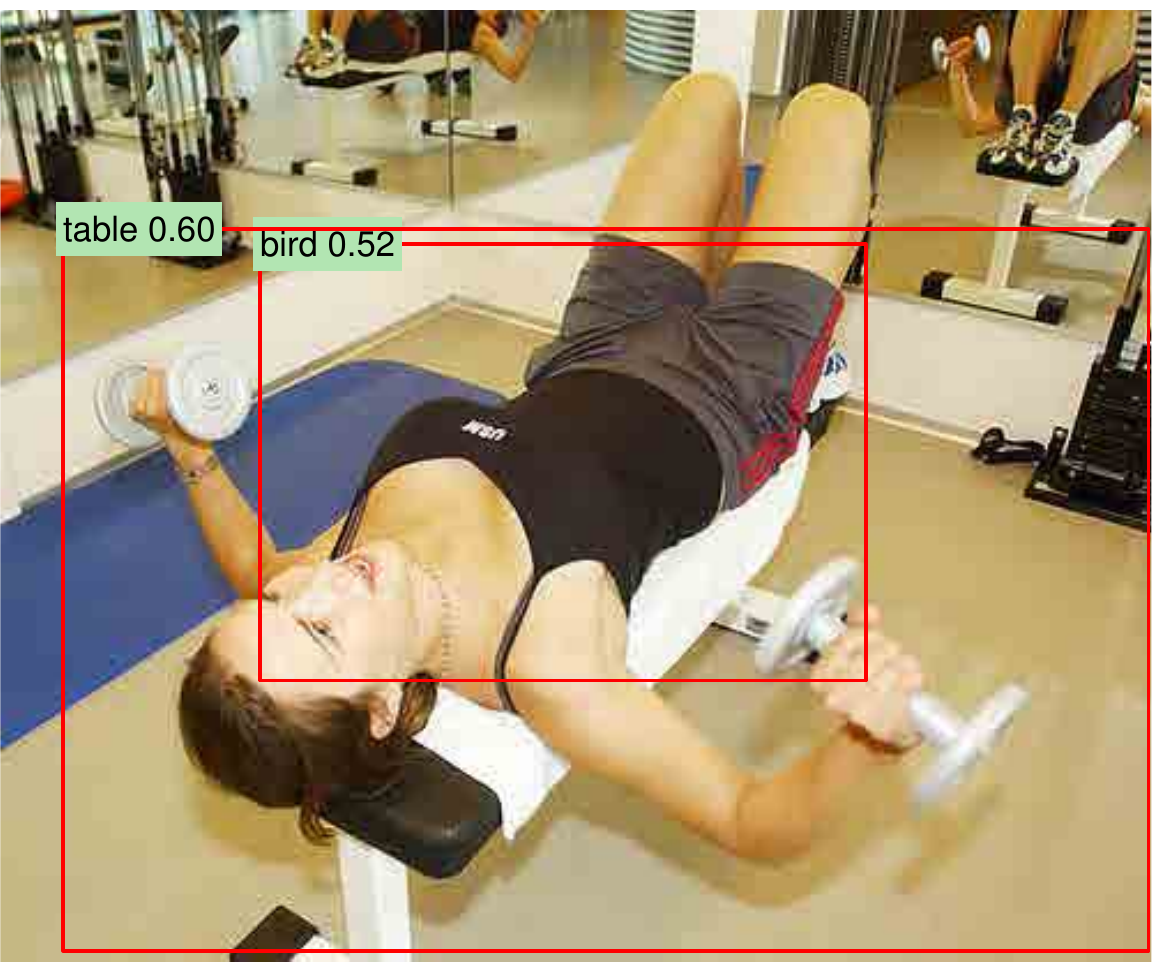}
\includegraphics[height=\sz]{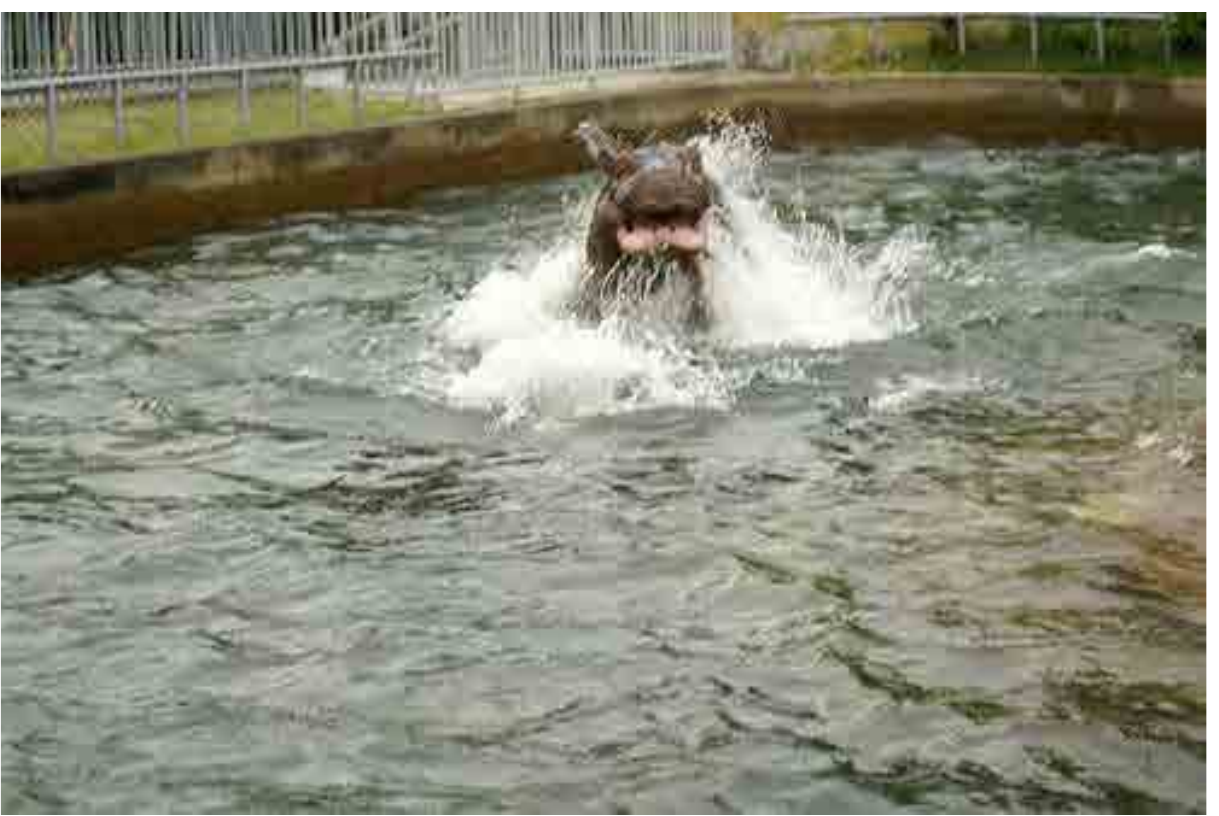}
\includegraphics[height=\sz]{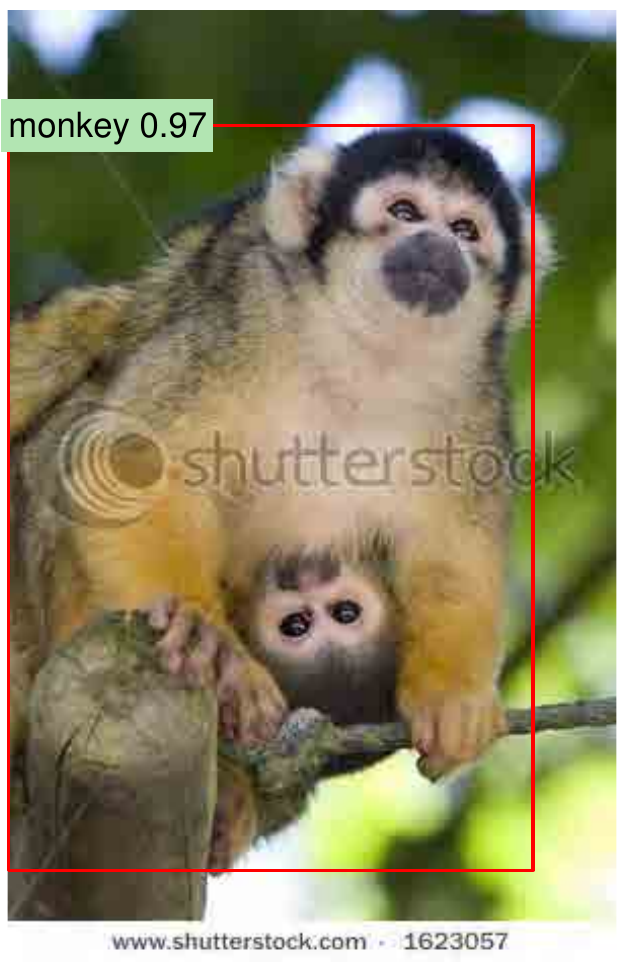}
\includegraphics[height=\sz]{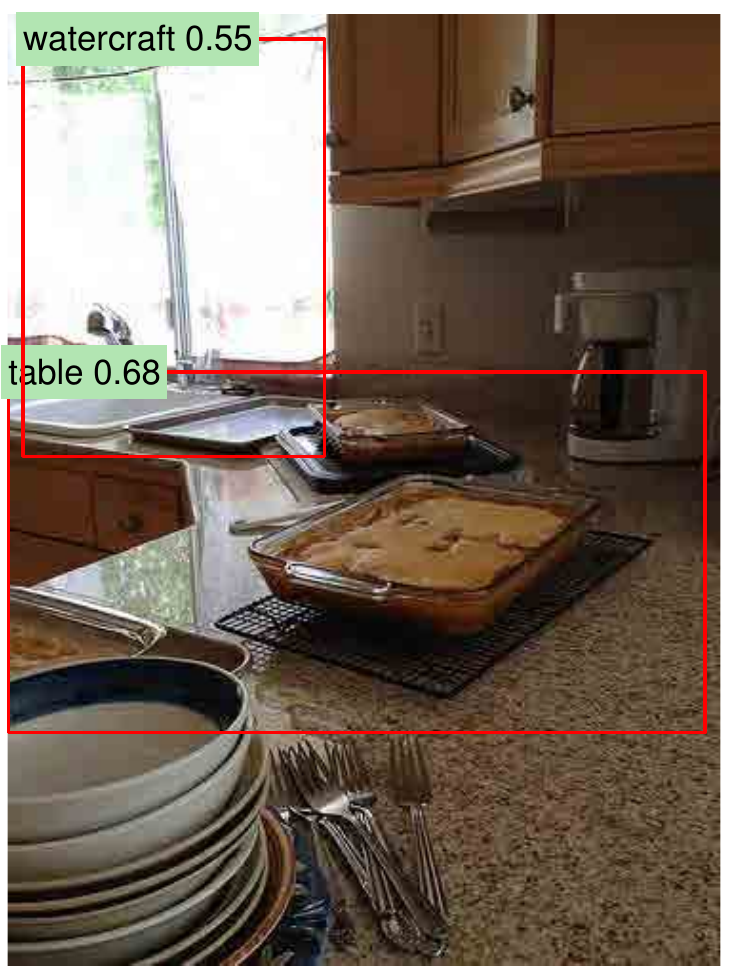}
\includegraphics[height=\sz]{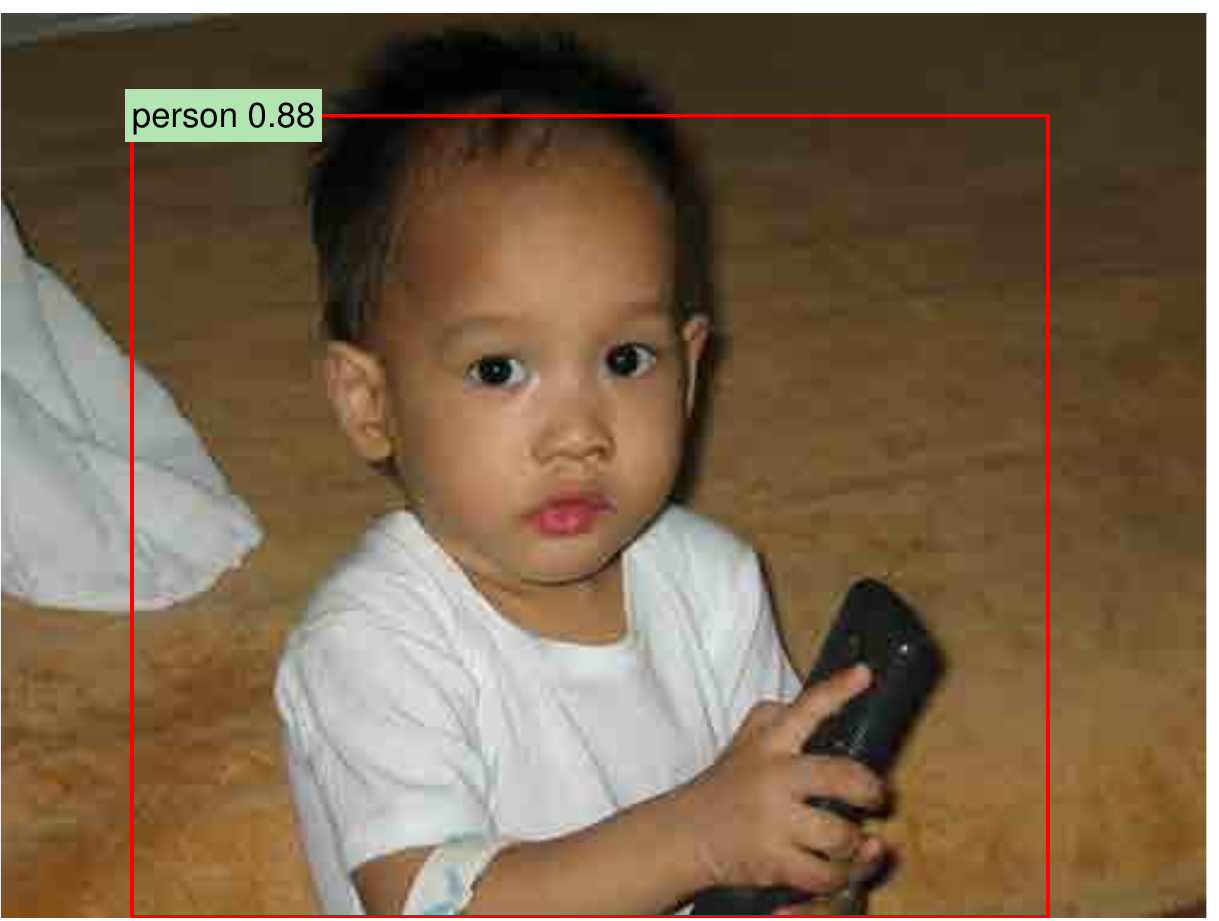}
\includegraphics[height=\sz]{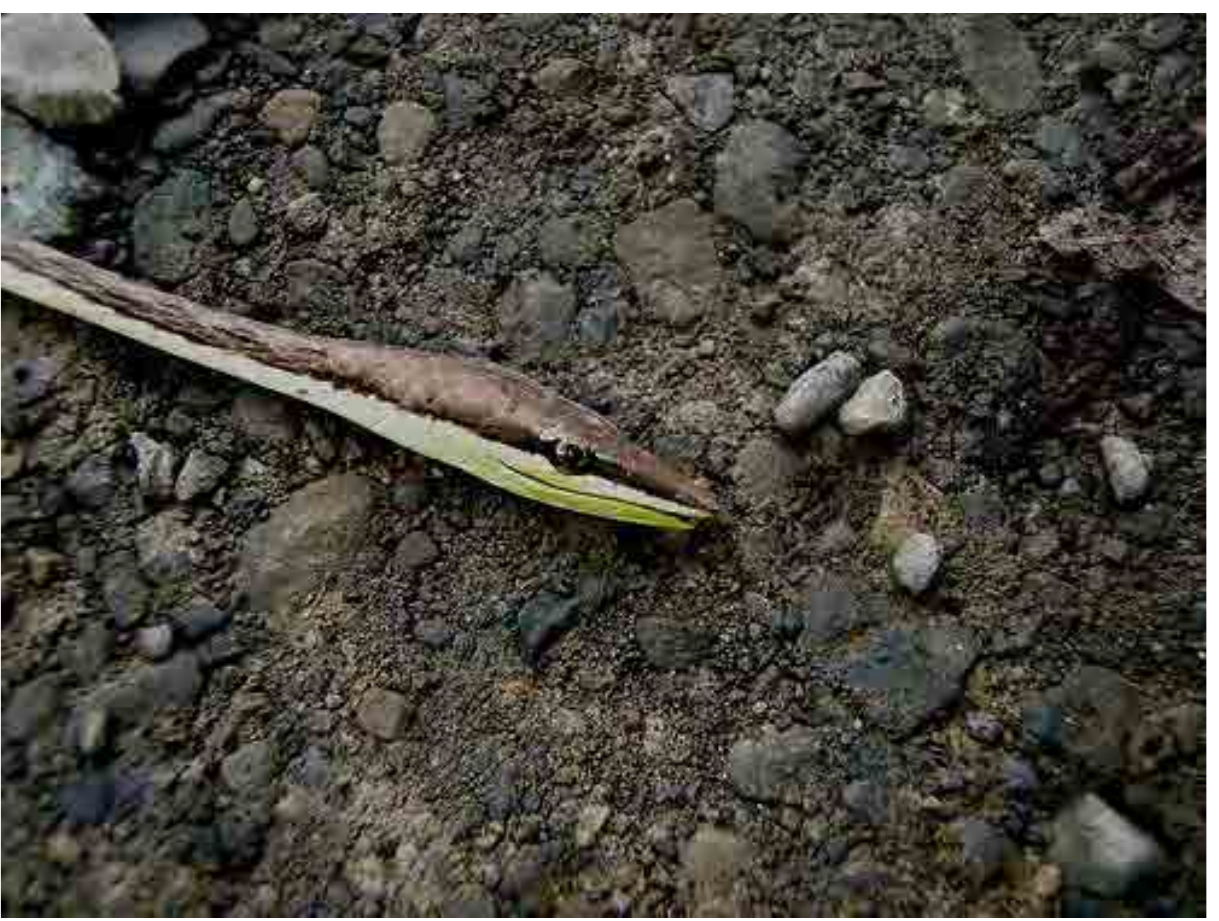}
\includegraphics[height=\sz]{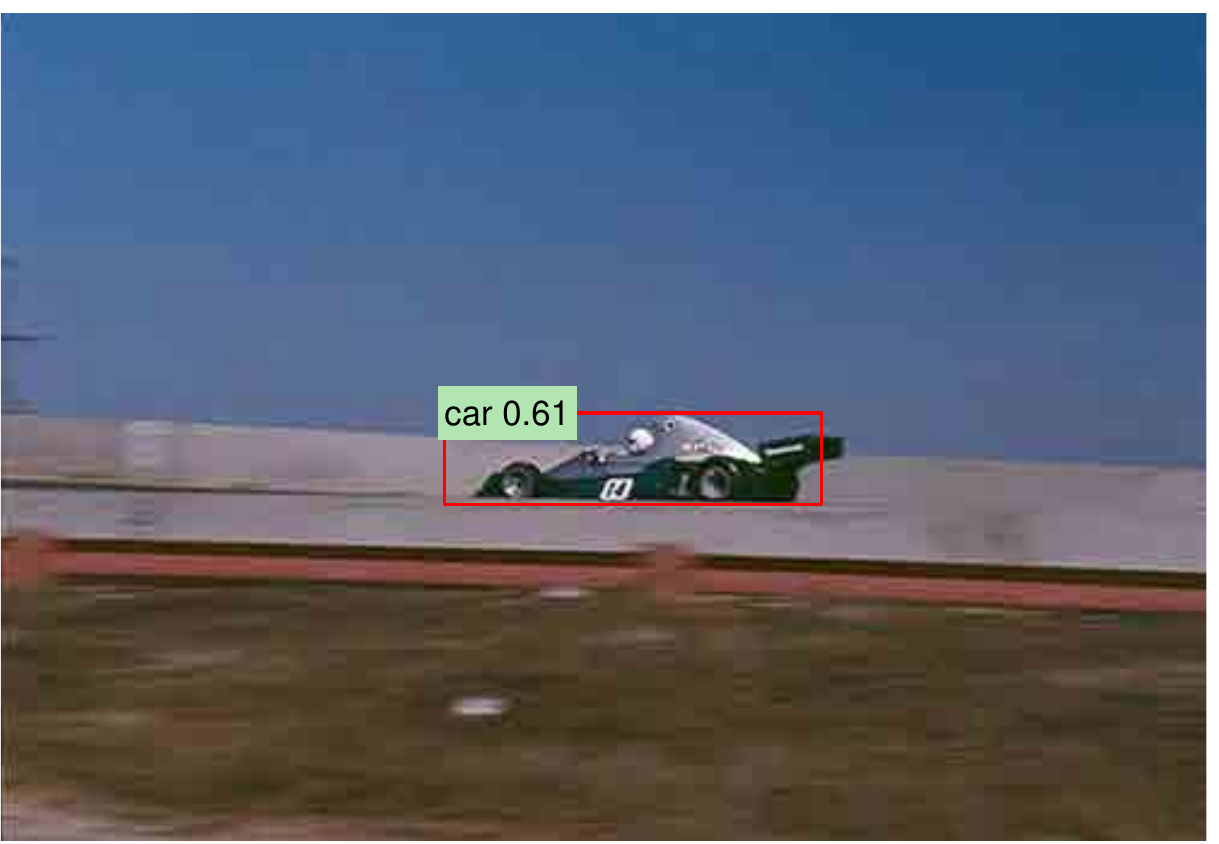}
\includegraphics[height=\sz]{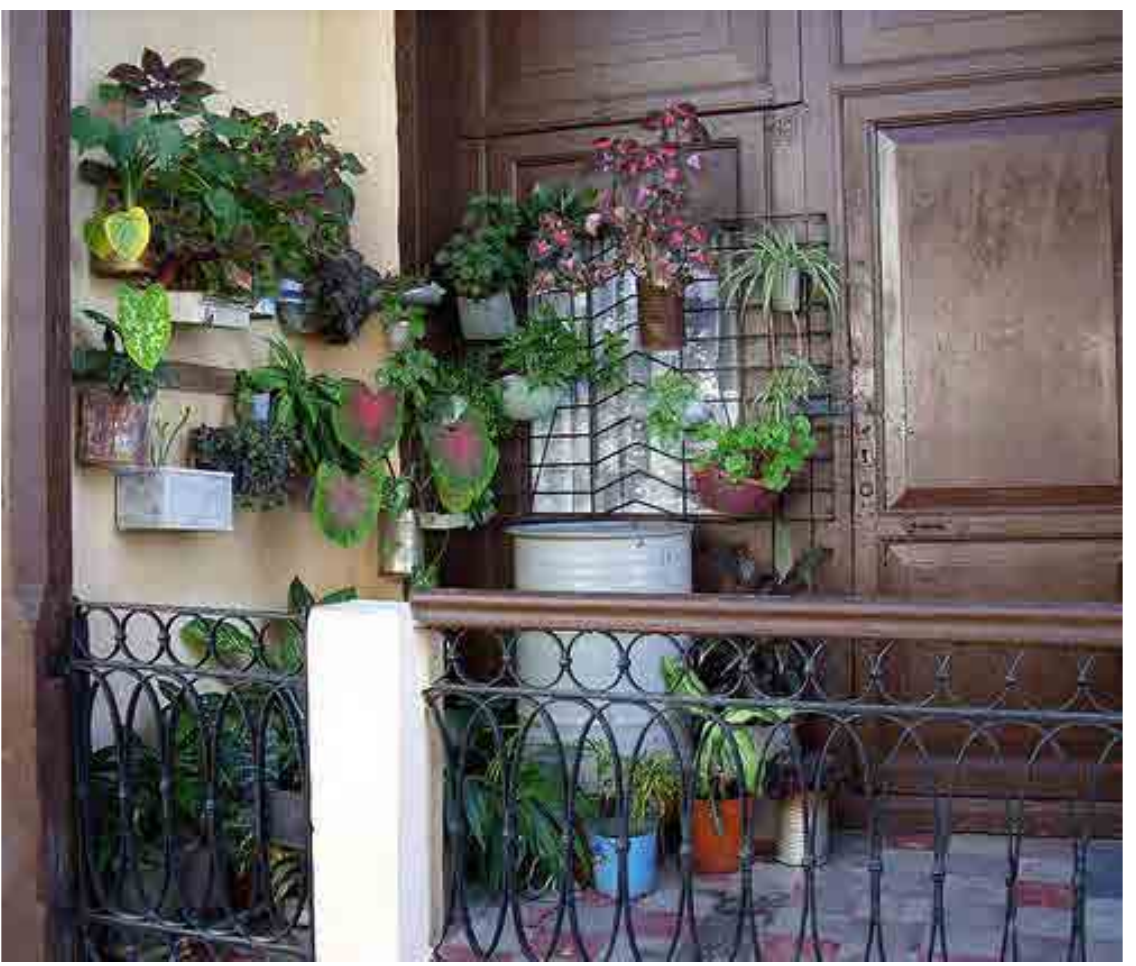}
\includegraphics[height=\sz]{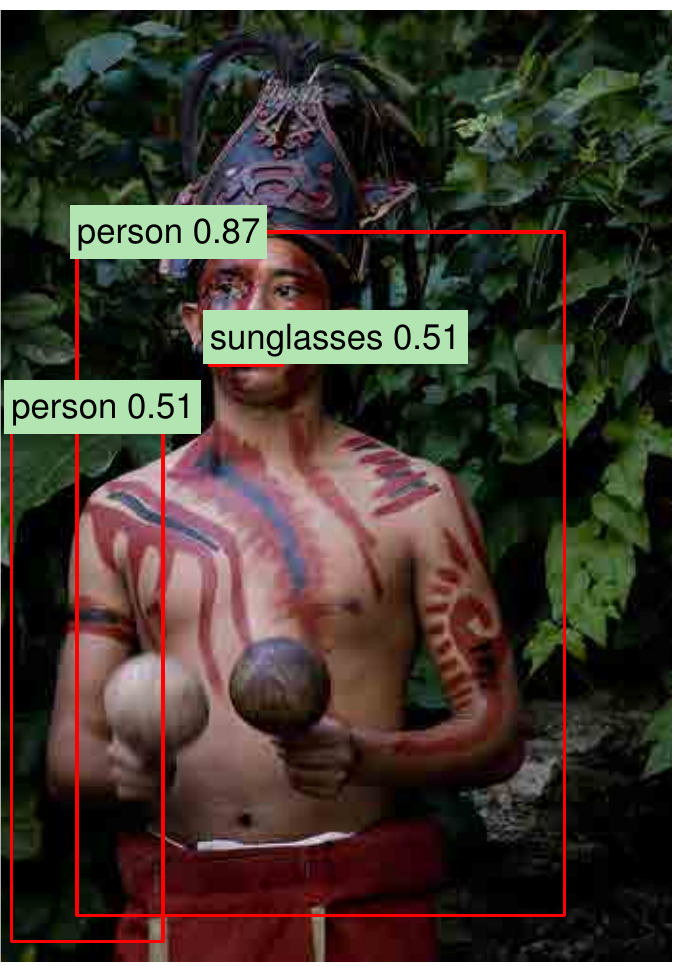}
\includegraphics[height=\sz]{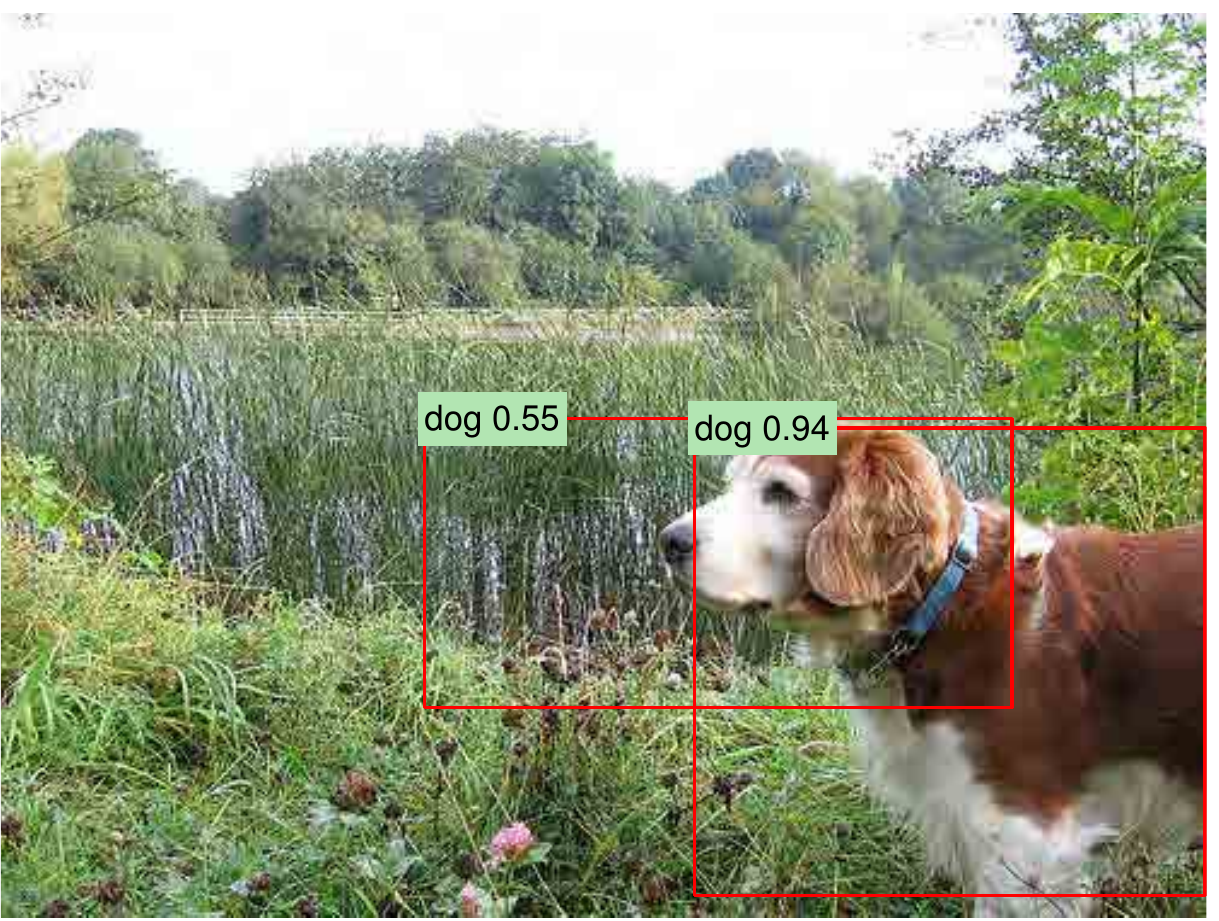}
\includegraphics[height=\sz]{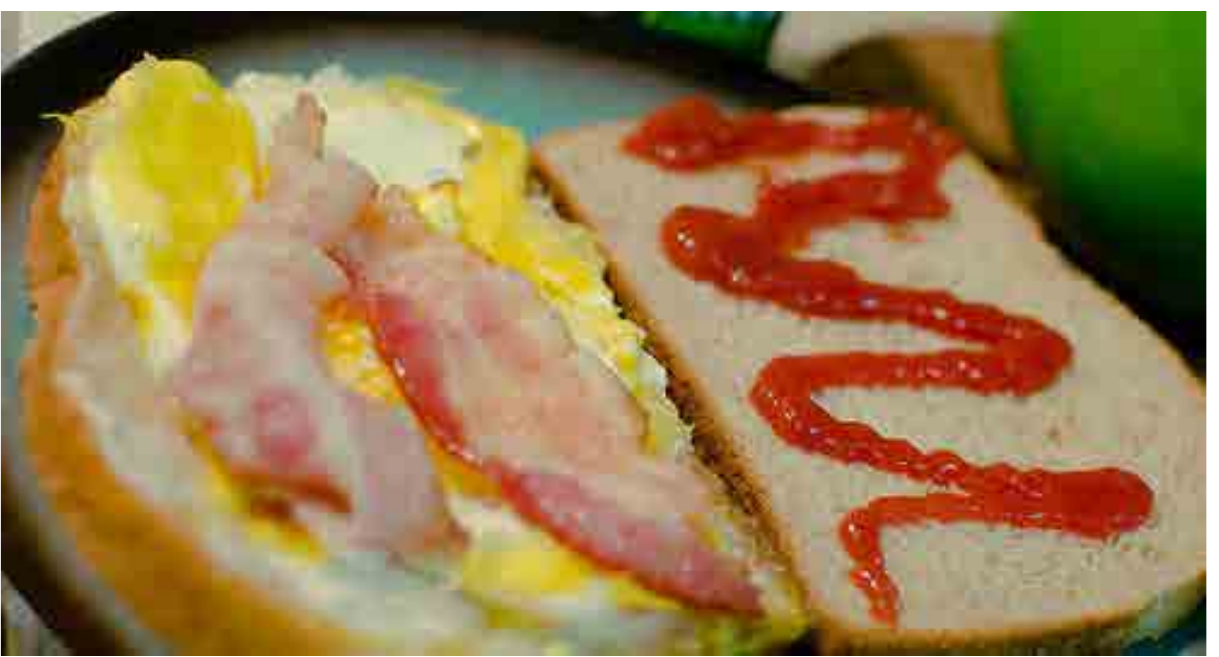}
\includegraphics[height=\sz]{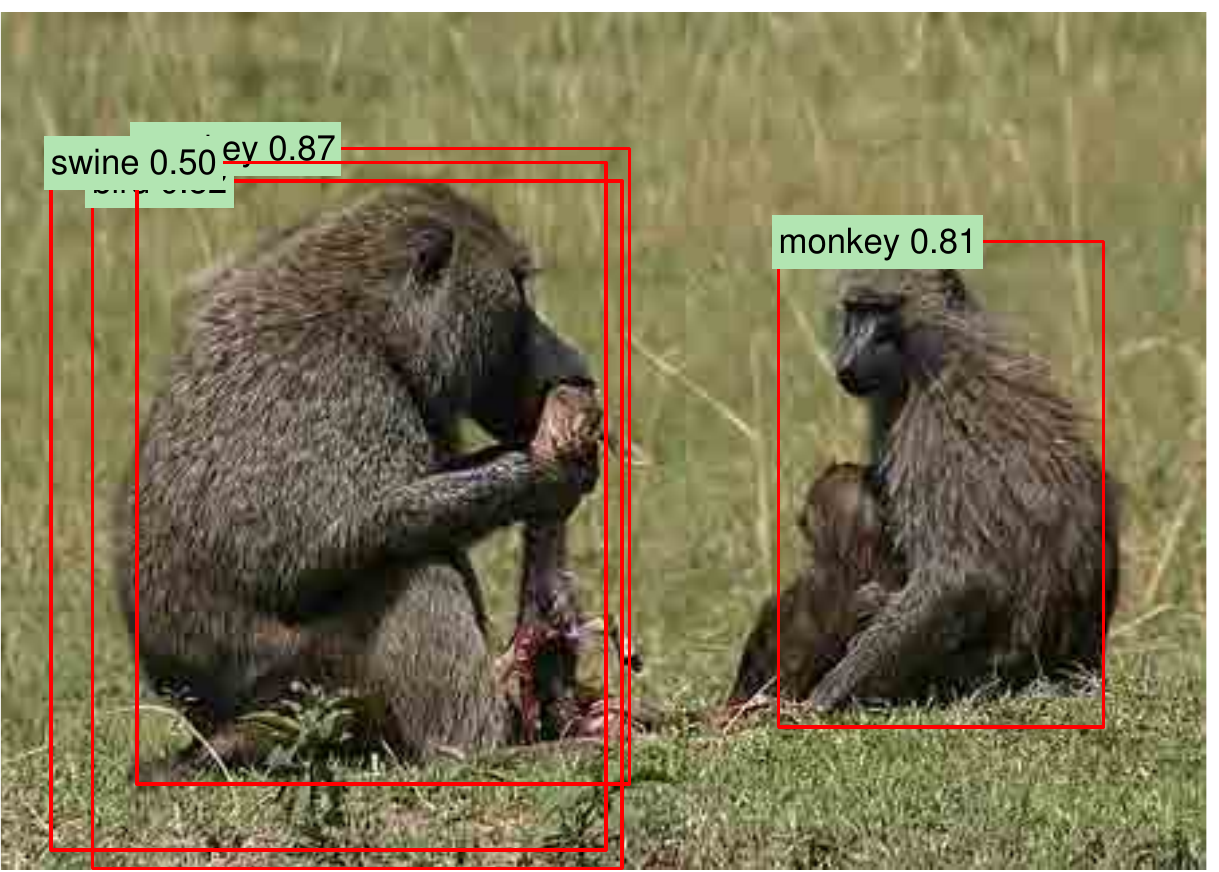}
\includegraphics[height=\sz]{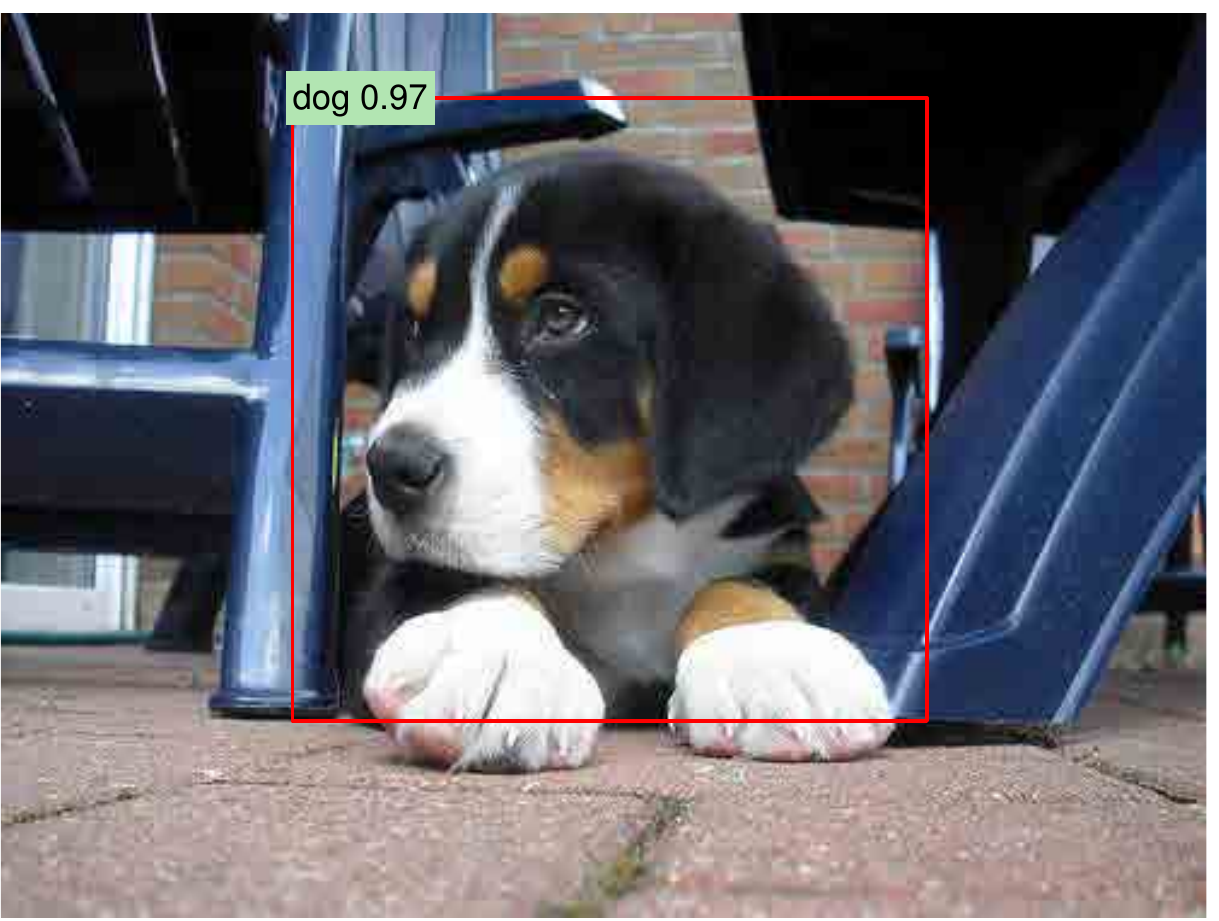}
\includegraphics[height=\sz]{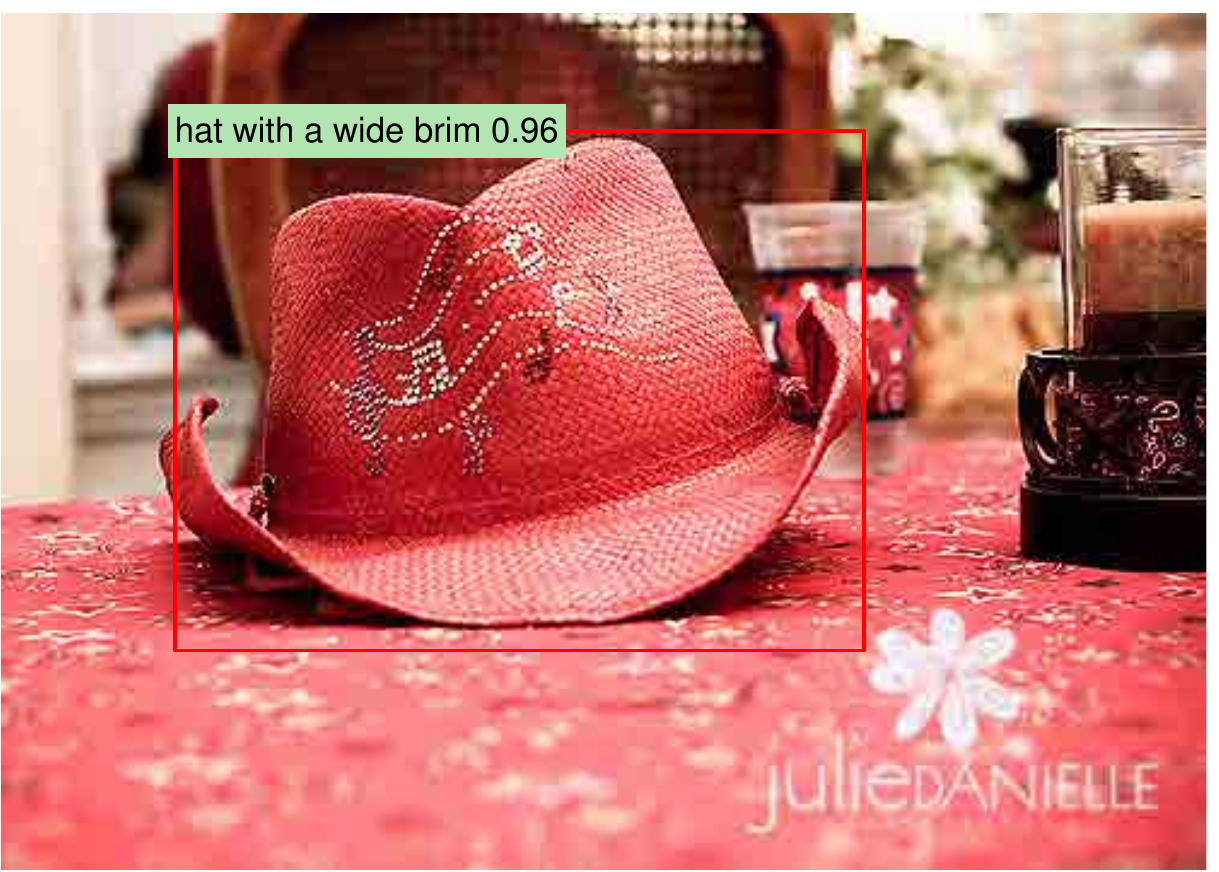}
\includegraphics[height=\sz]{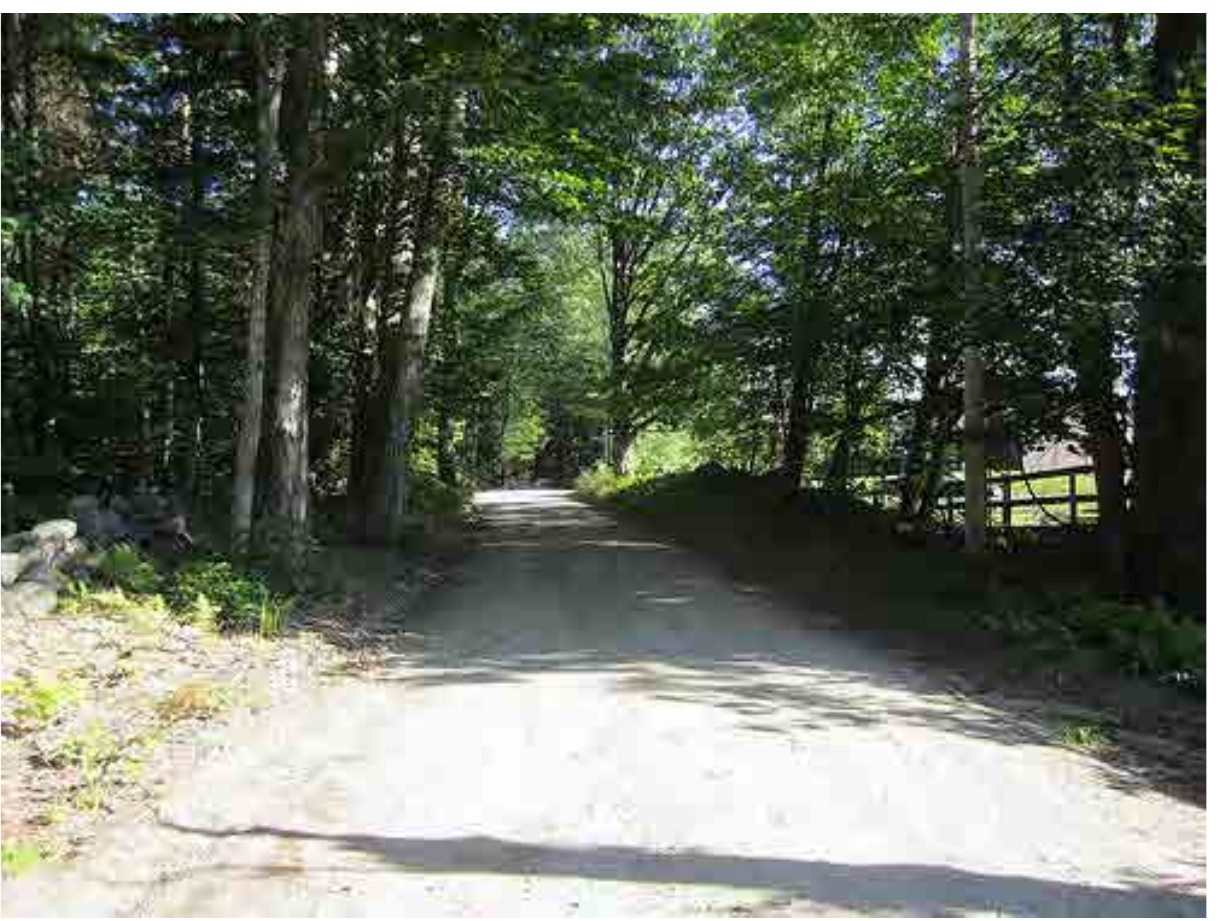}
\includegraphics[height=\sz]{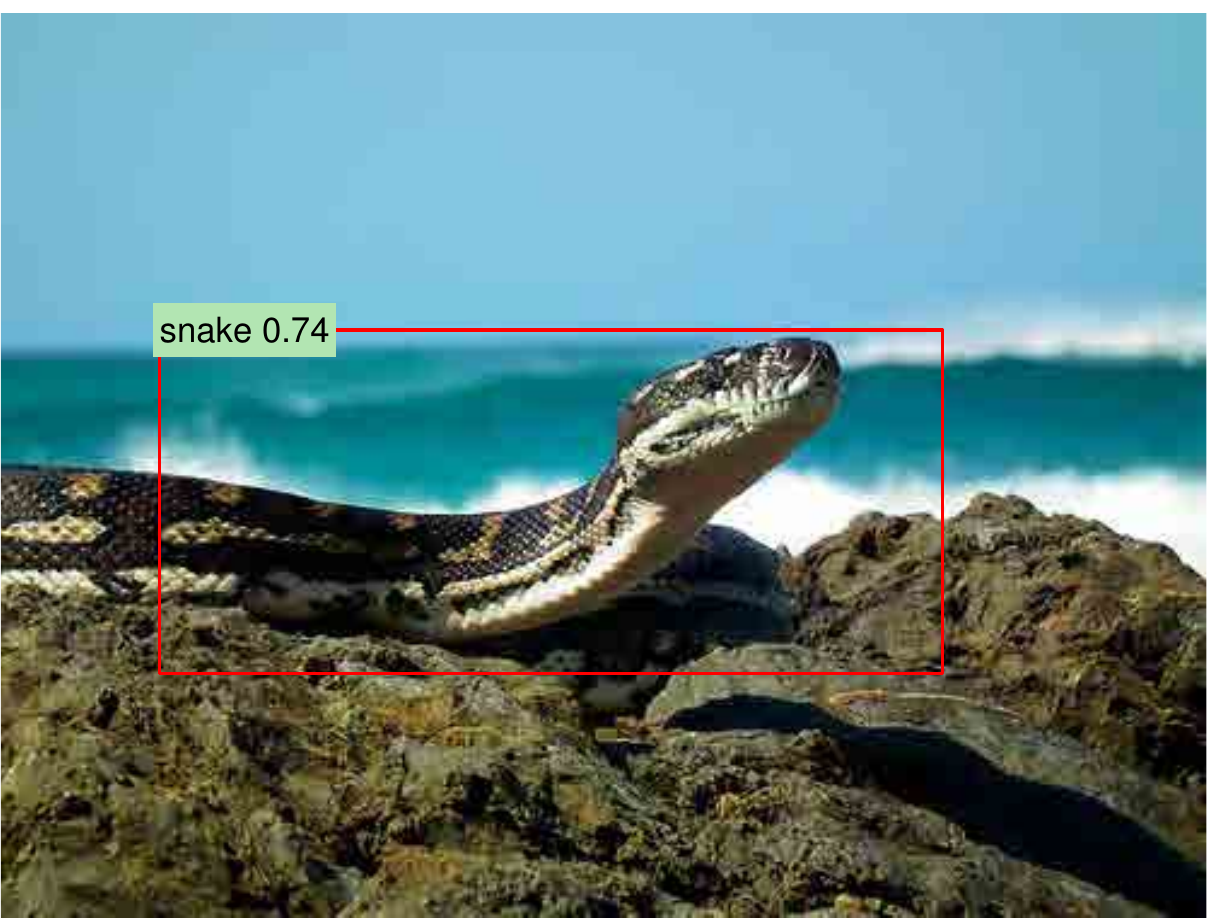}
\includegraphics[height=\sz]{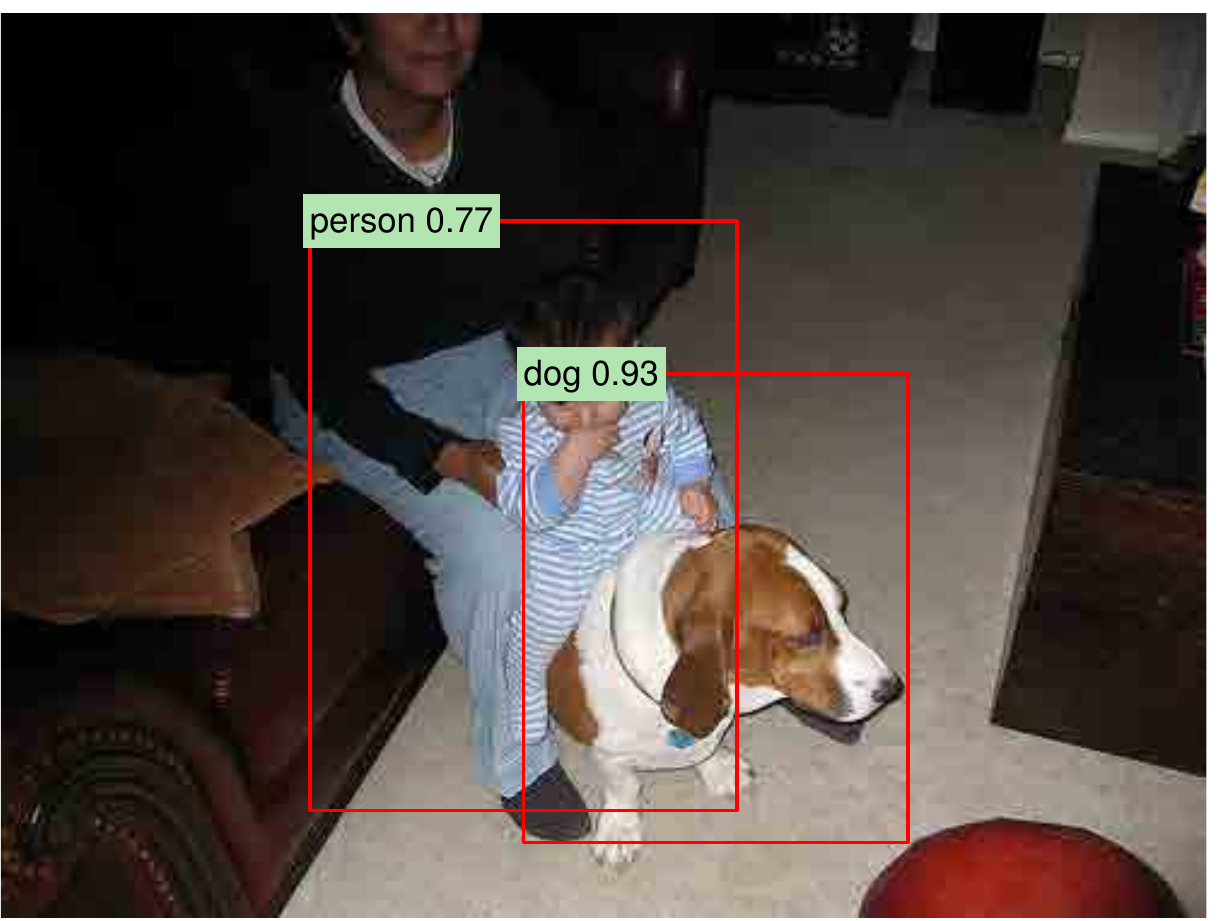}
\includegraphics[height=\sz]{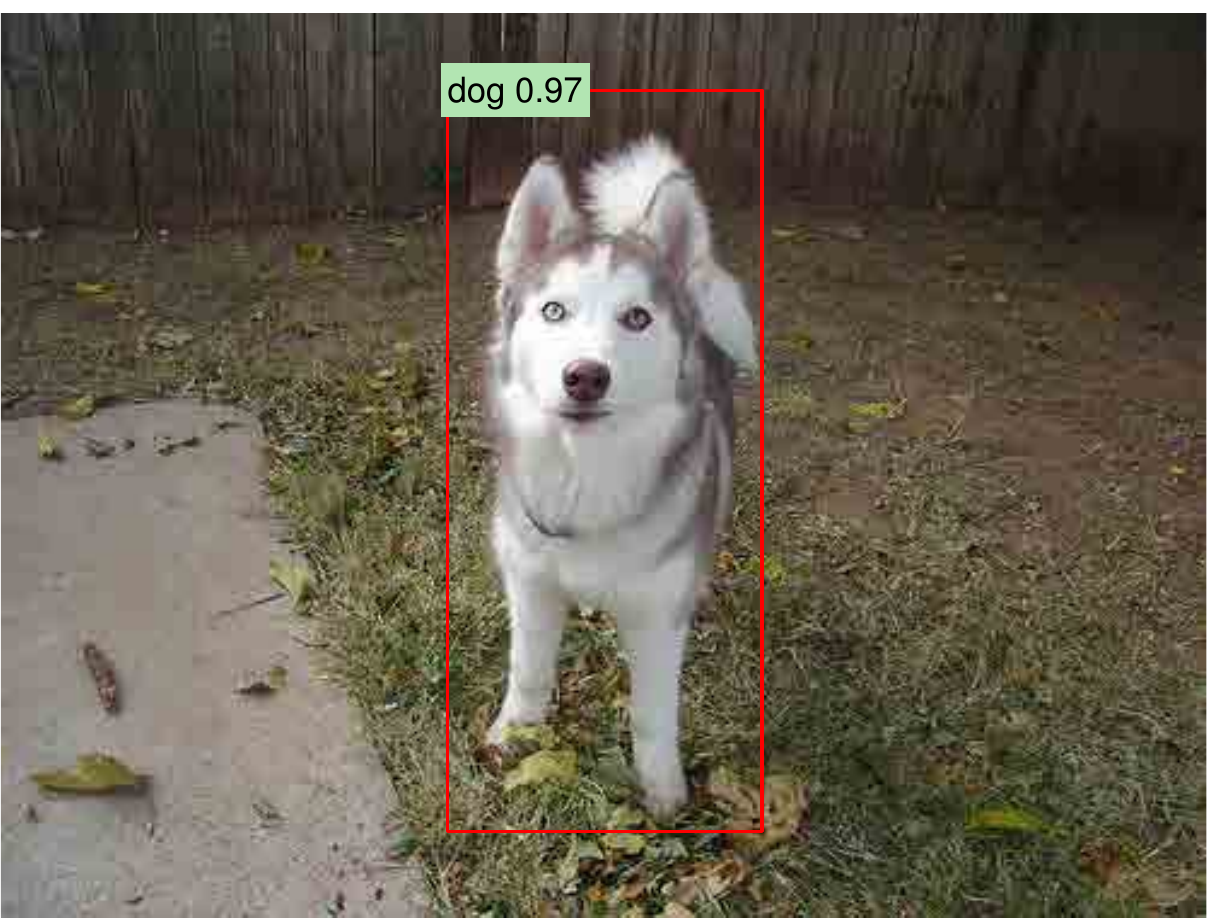}
\includegraphics[height=\sz]{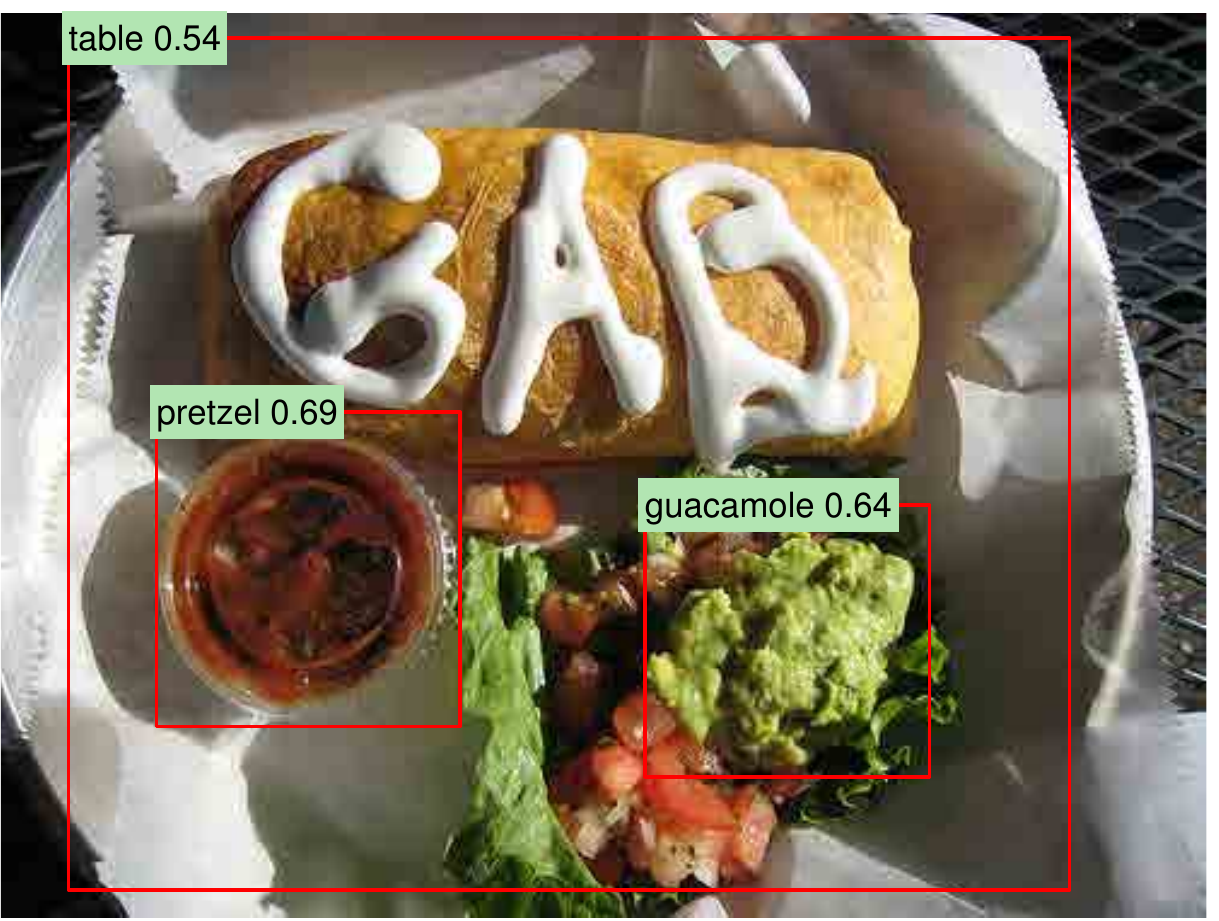}
\includegraphics[height=\sz]{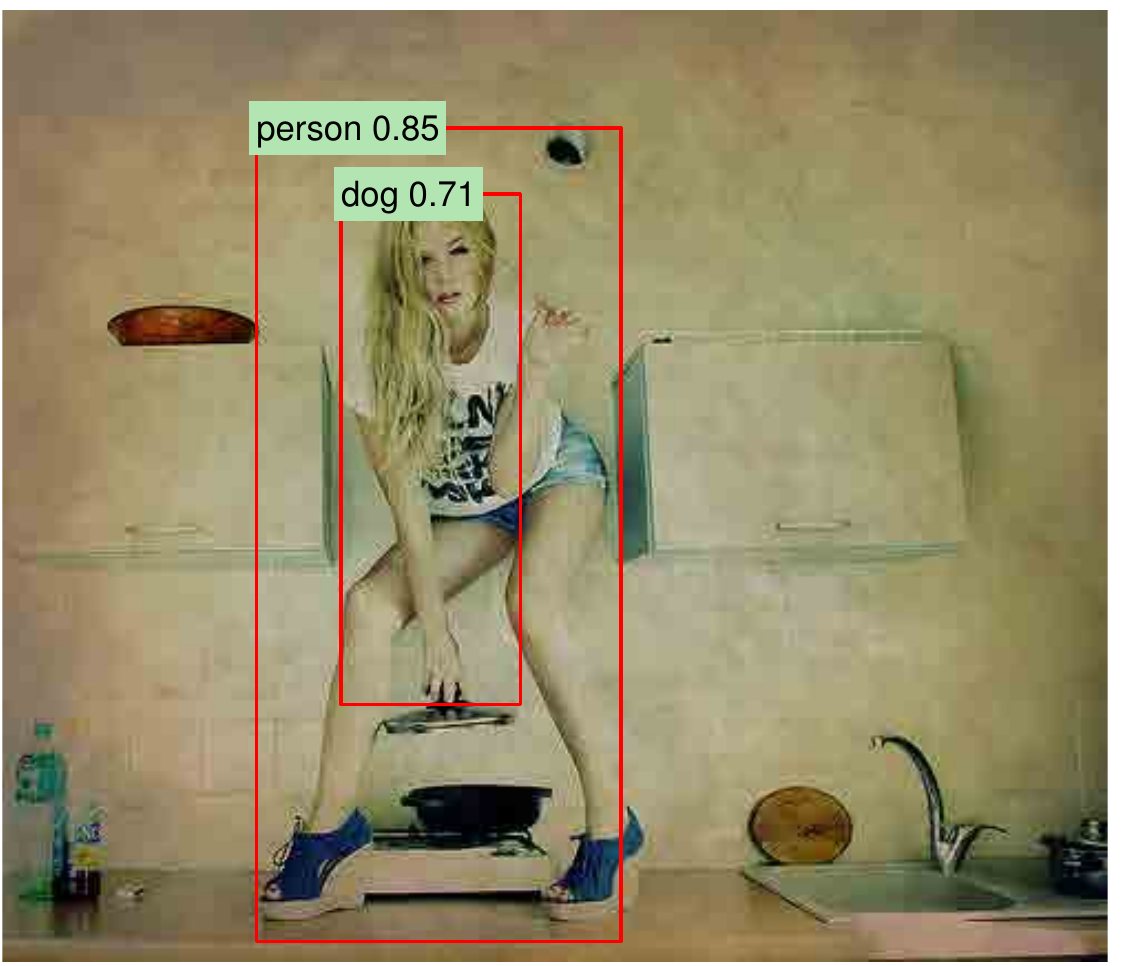}
\includegraphics[height=\sz]{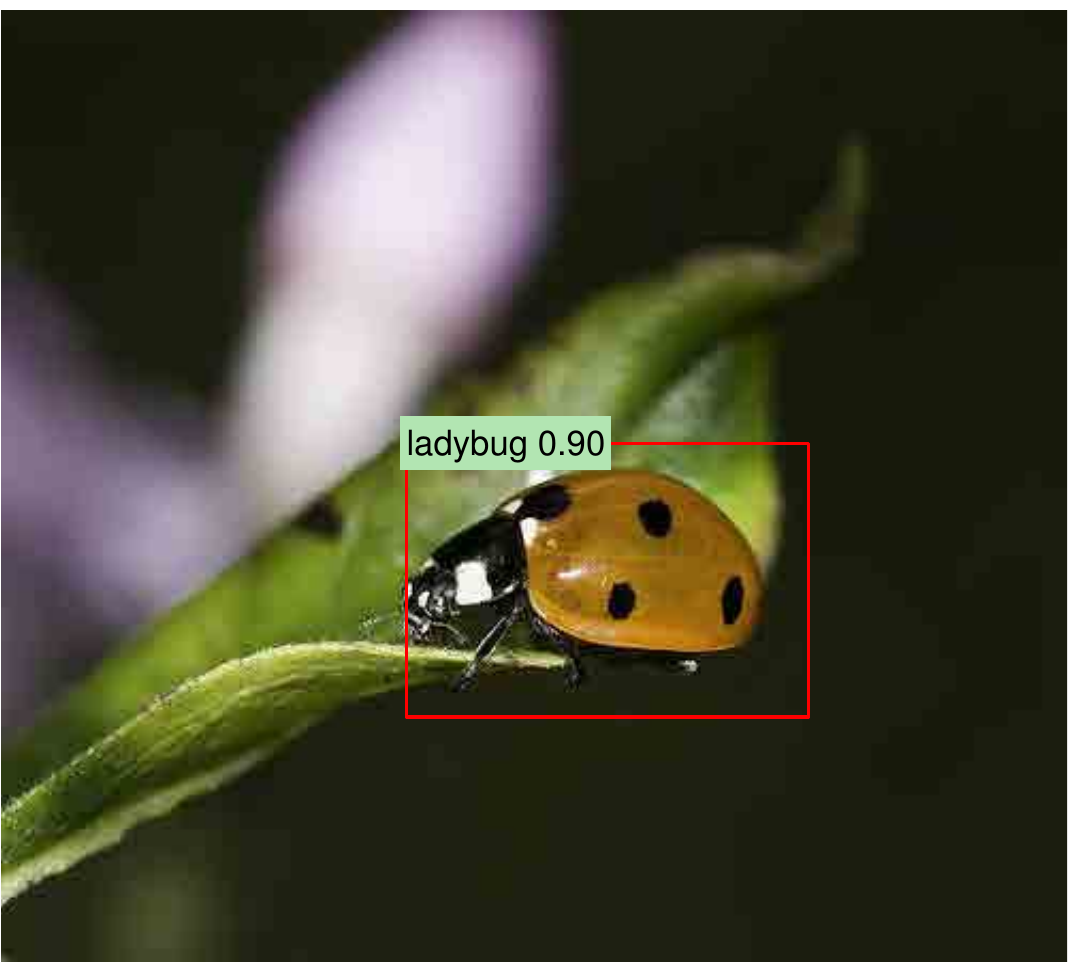}
\includegraphics[height=\sz]{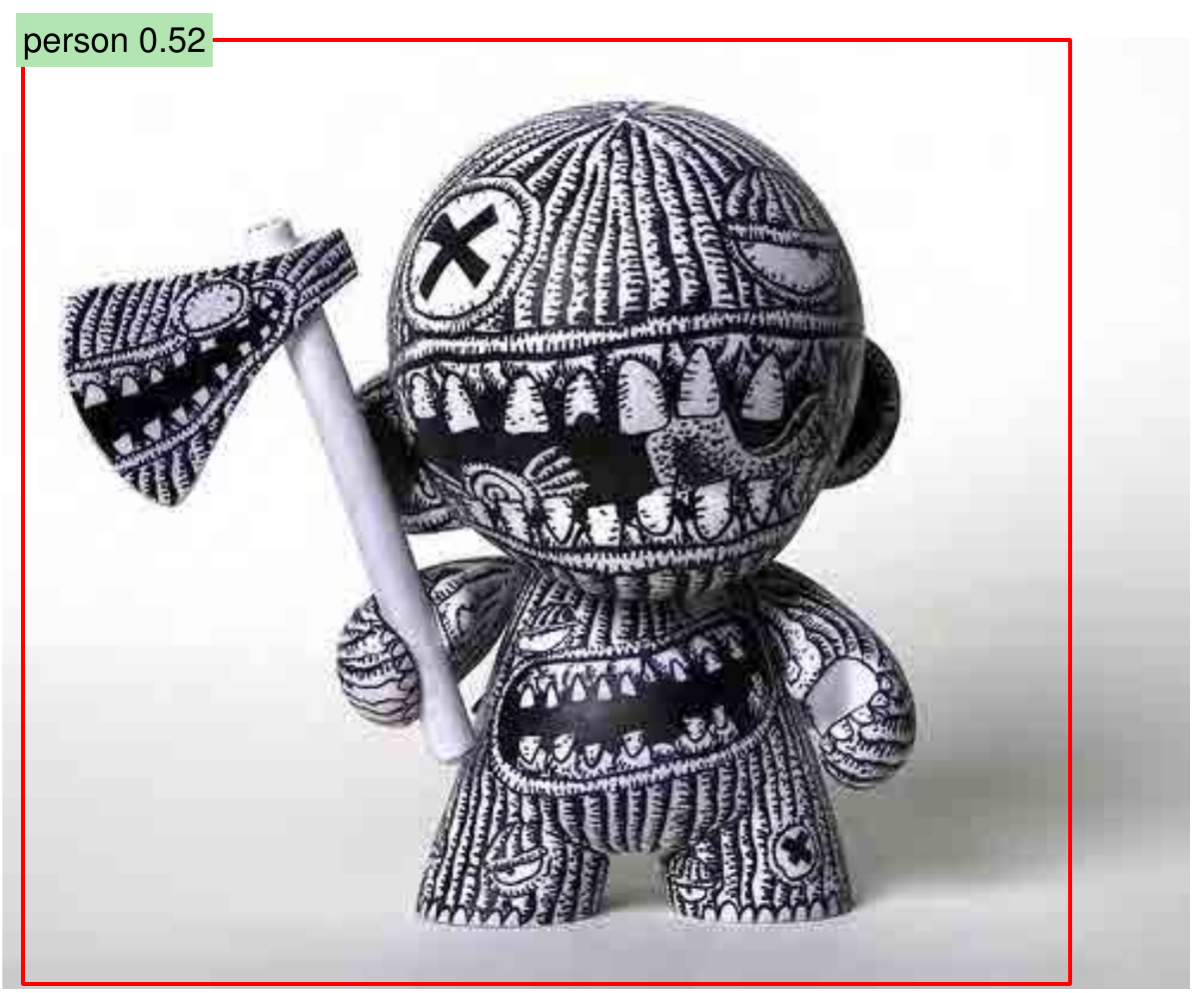}
\includegraphics[height=\sz]{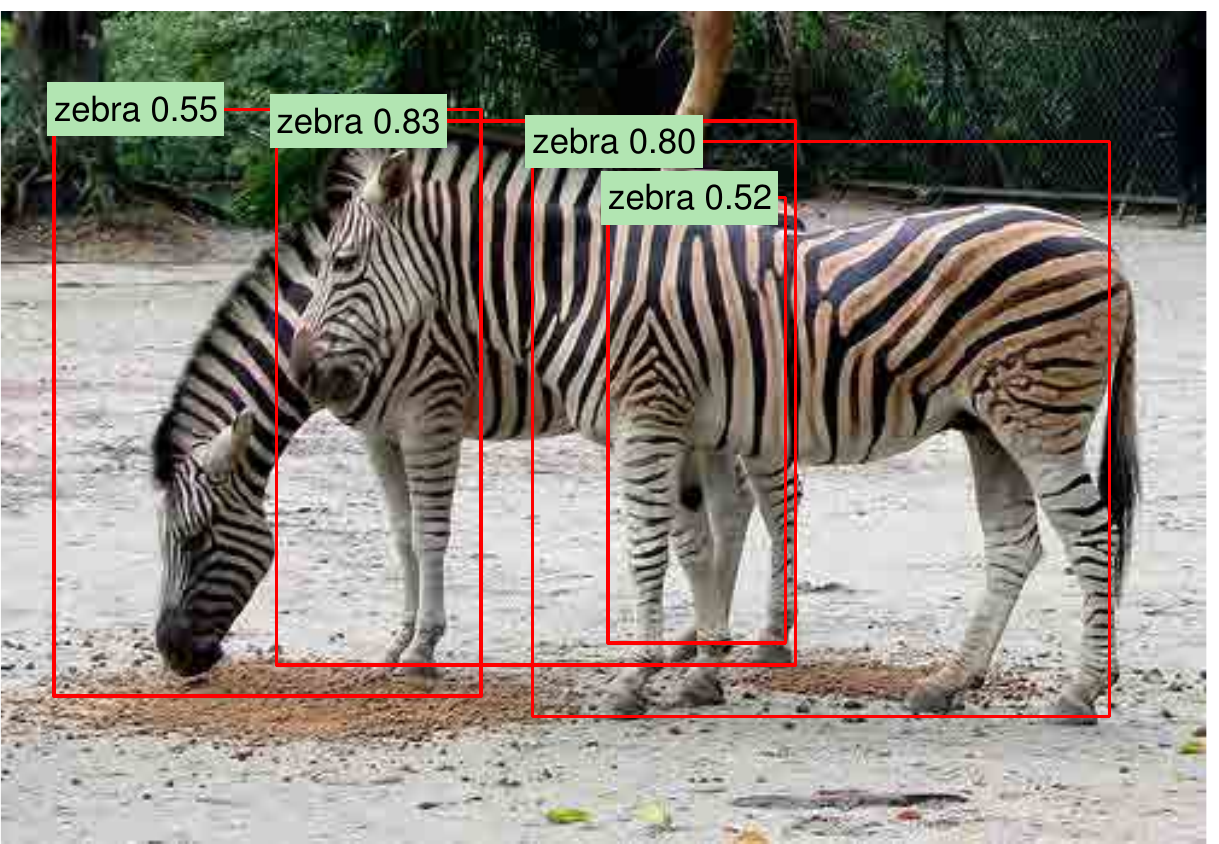}
\includegraphics[height=\sz]{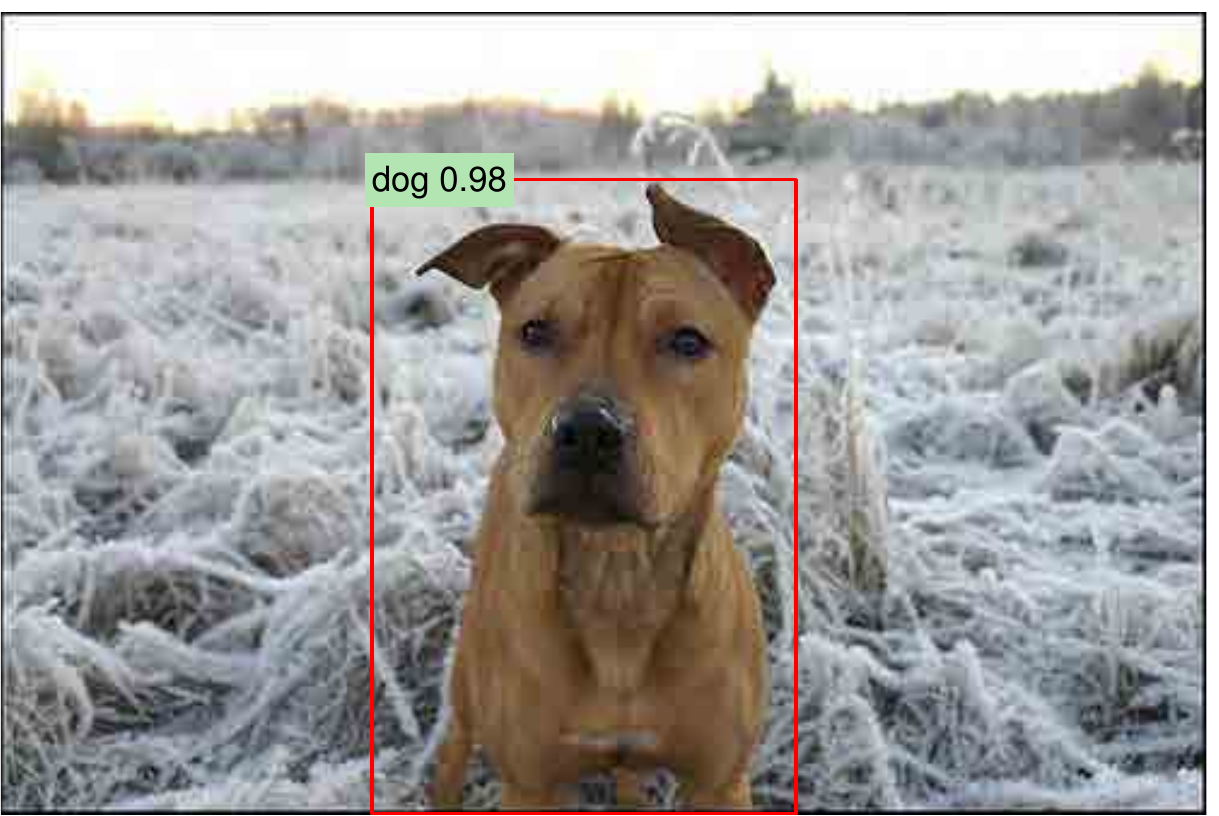}
\includegraphics[height=\sz]{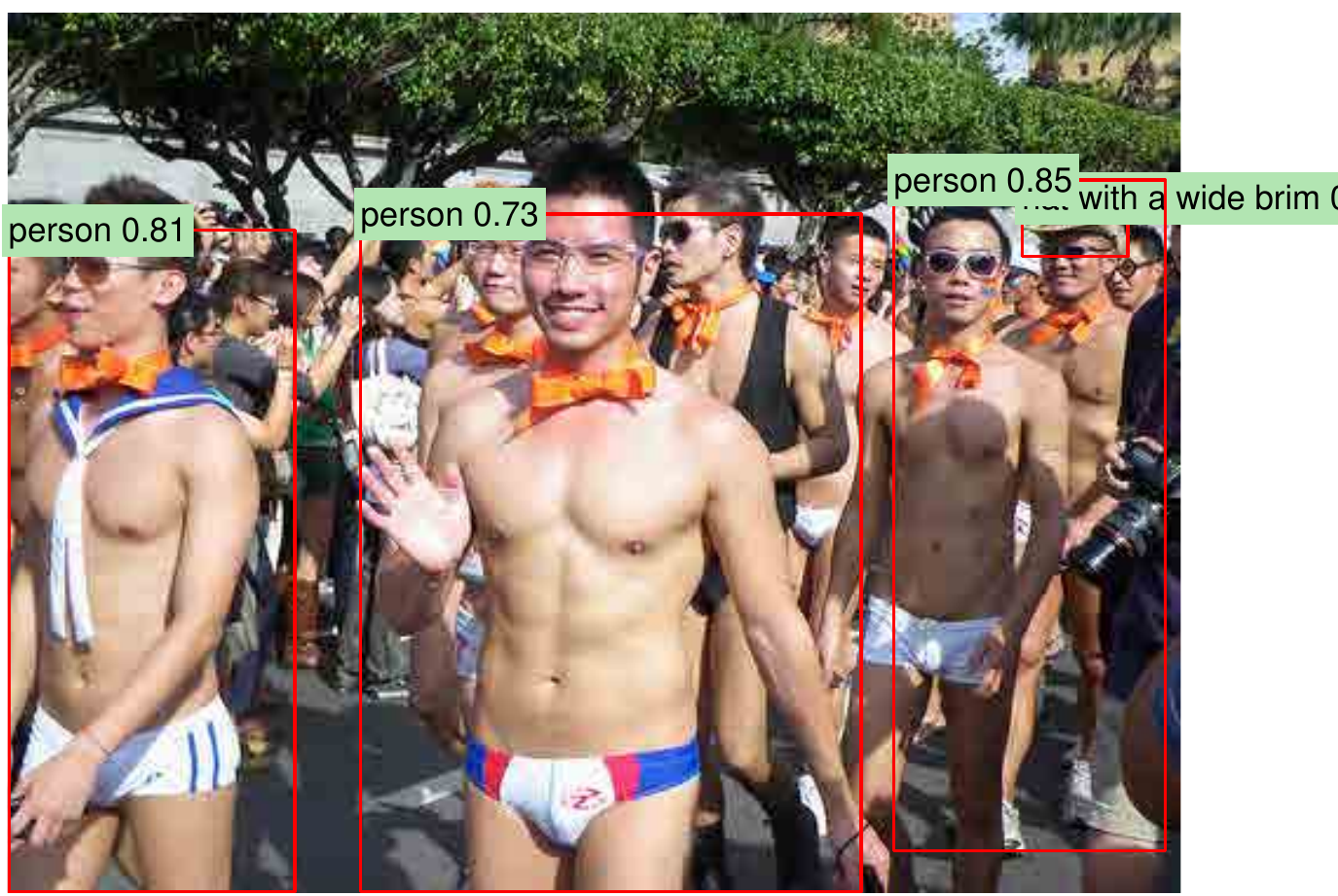}
\includegraphics[height=\sz]{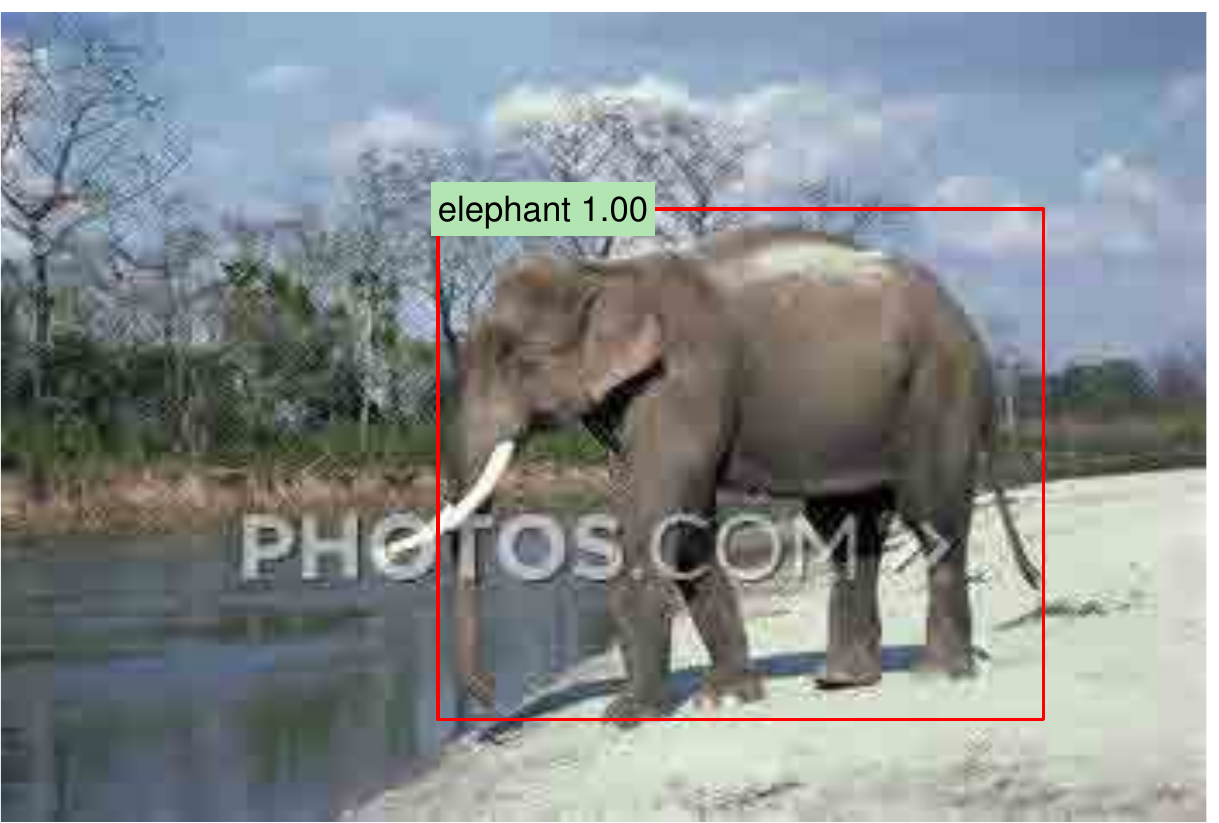}
\includegraphics[height=\sz]{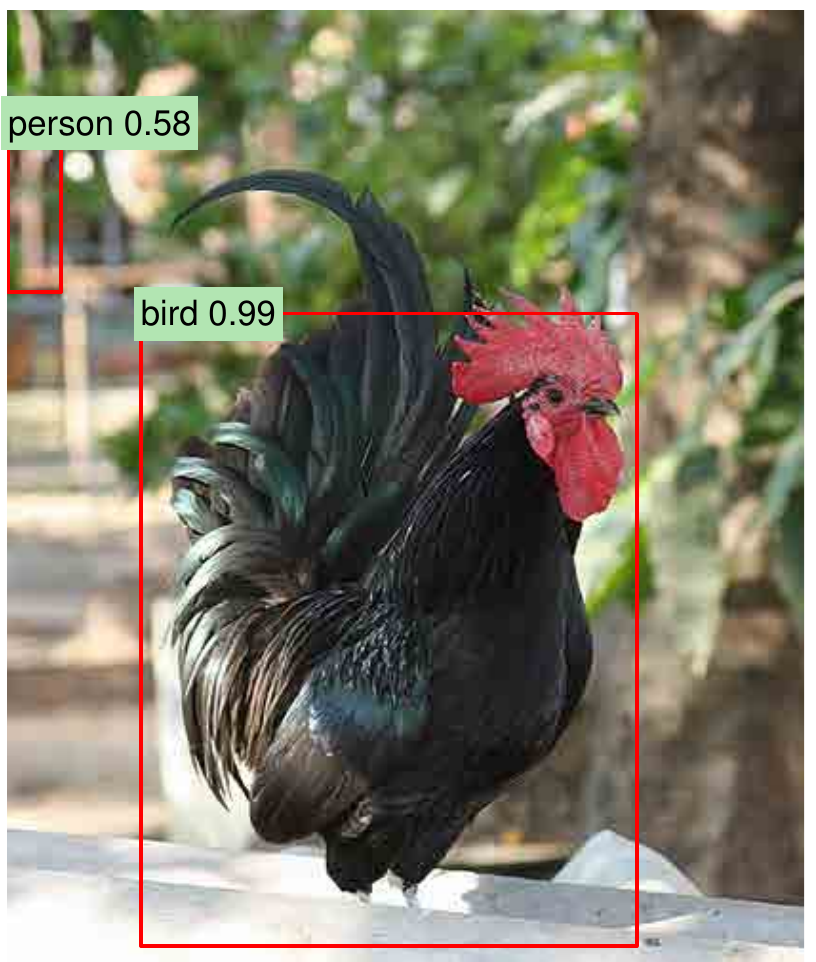}
\includegraphics[height=\sz]{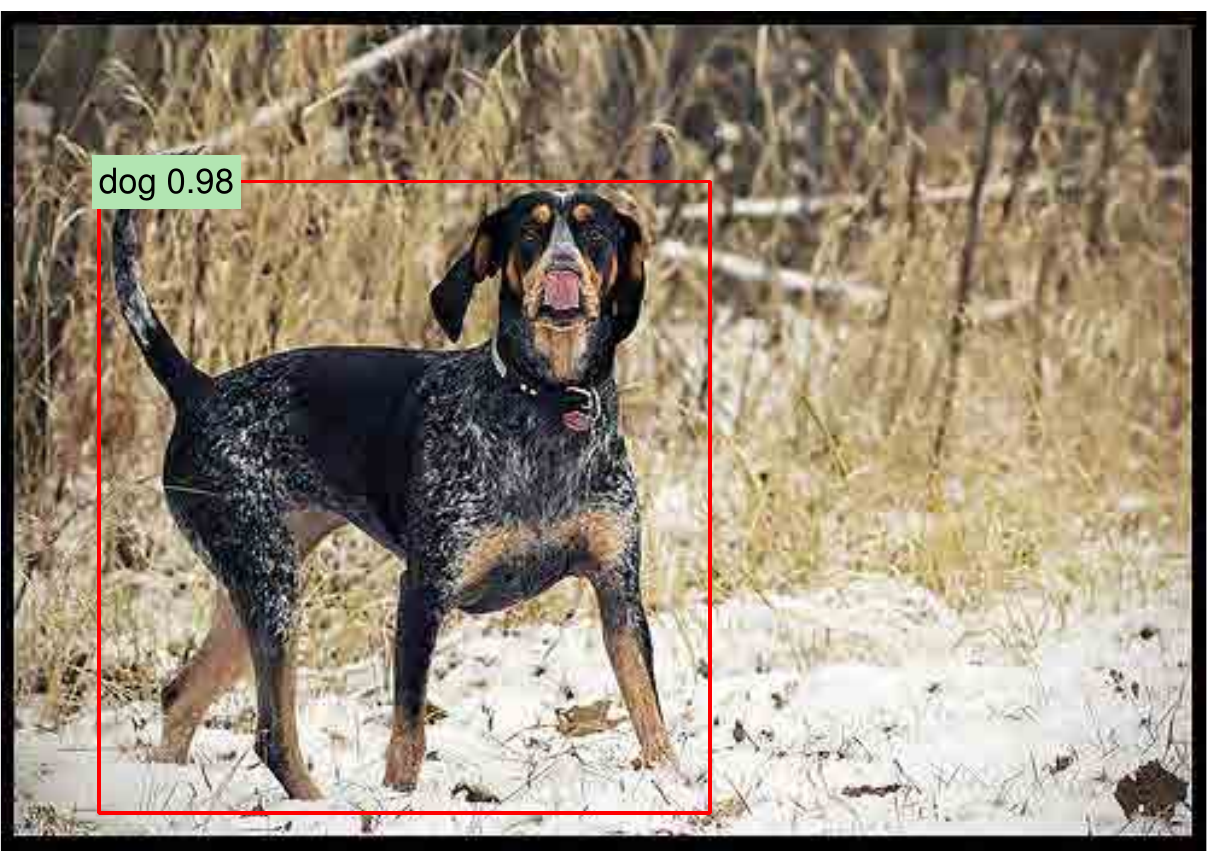}
\includegraphics[height=\sz]{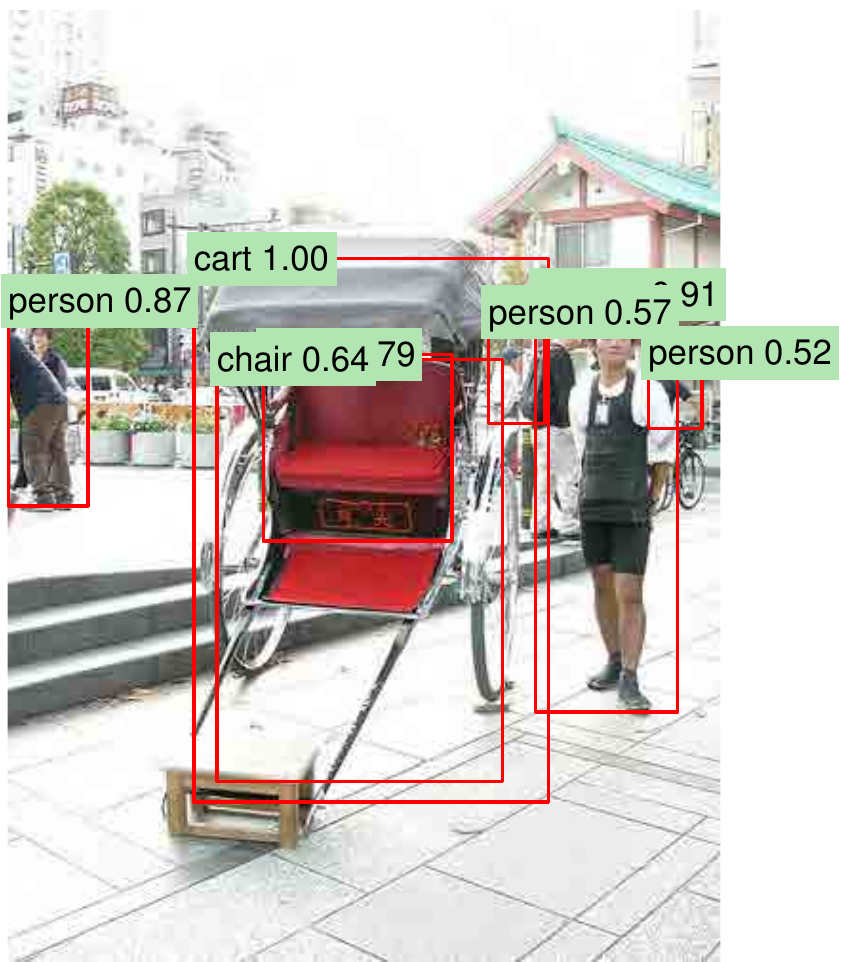}
\includegraphics[height=\sz]{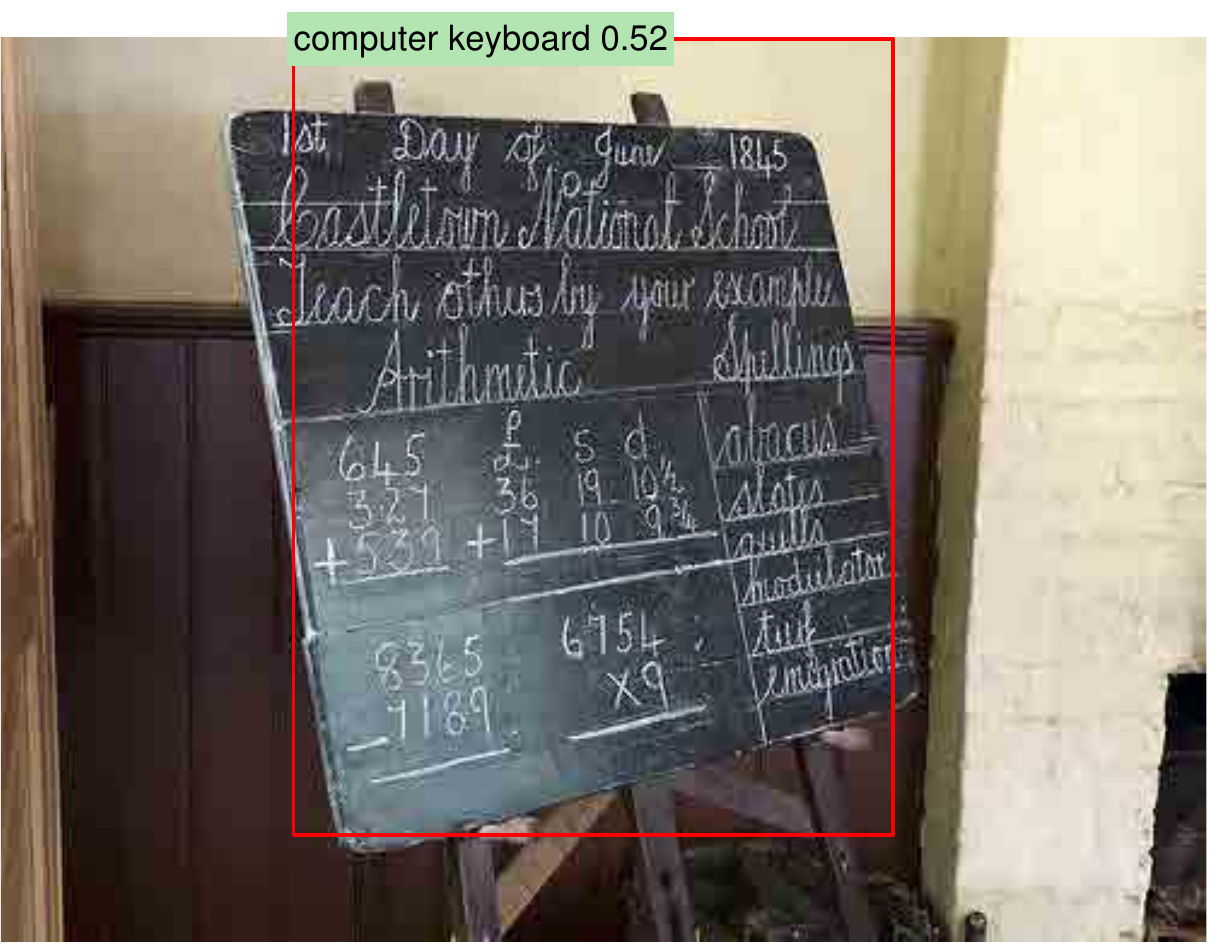}
\includegraphics[height=\sz]{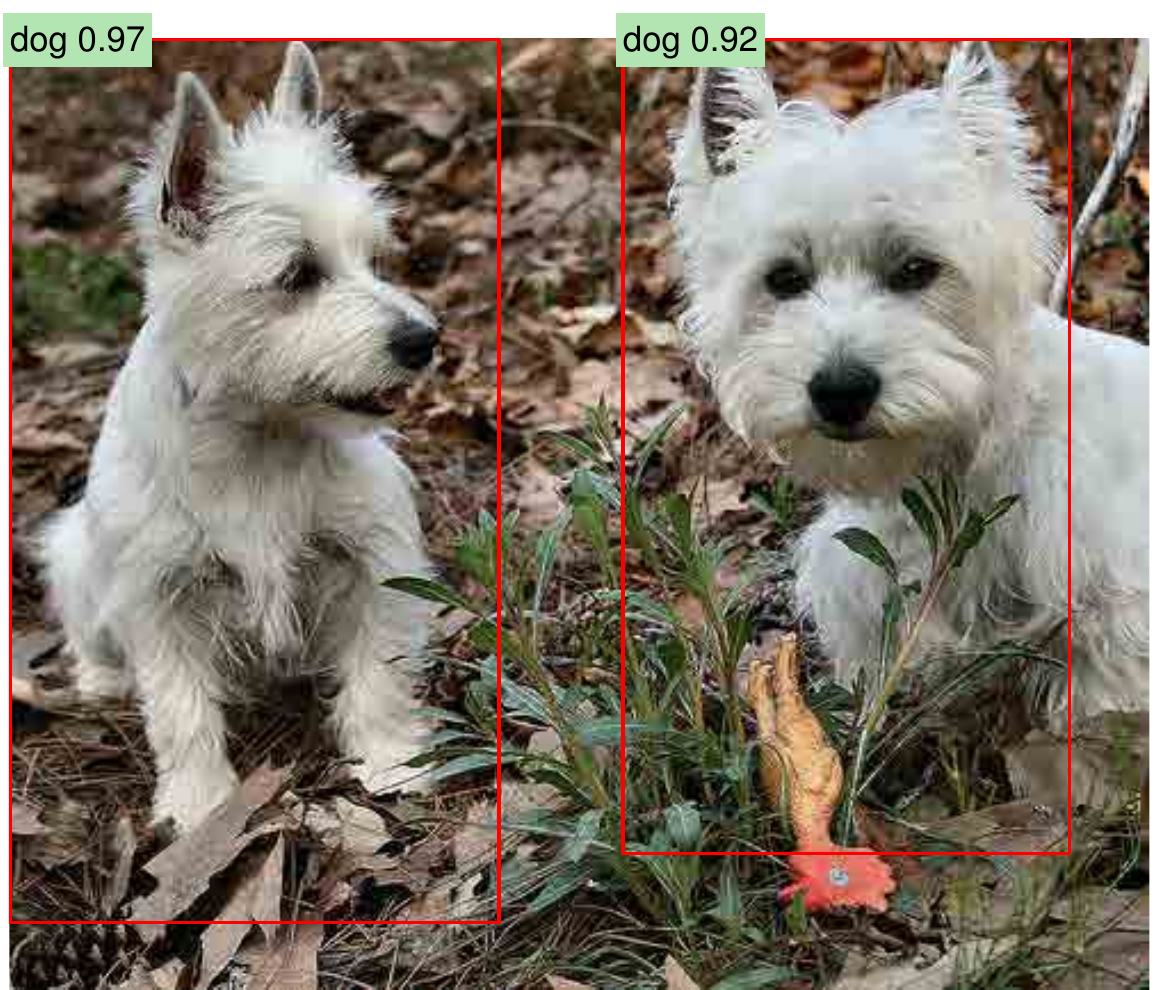}
\includegraphics[height=\sz]{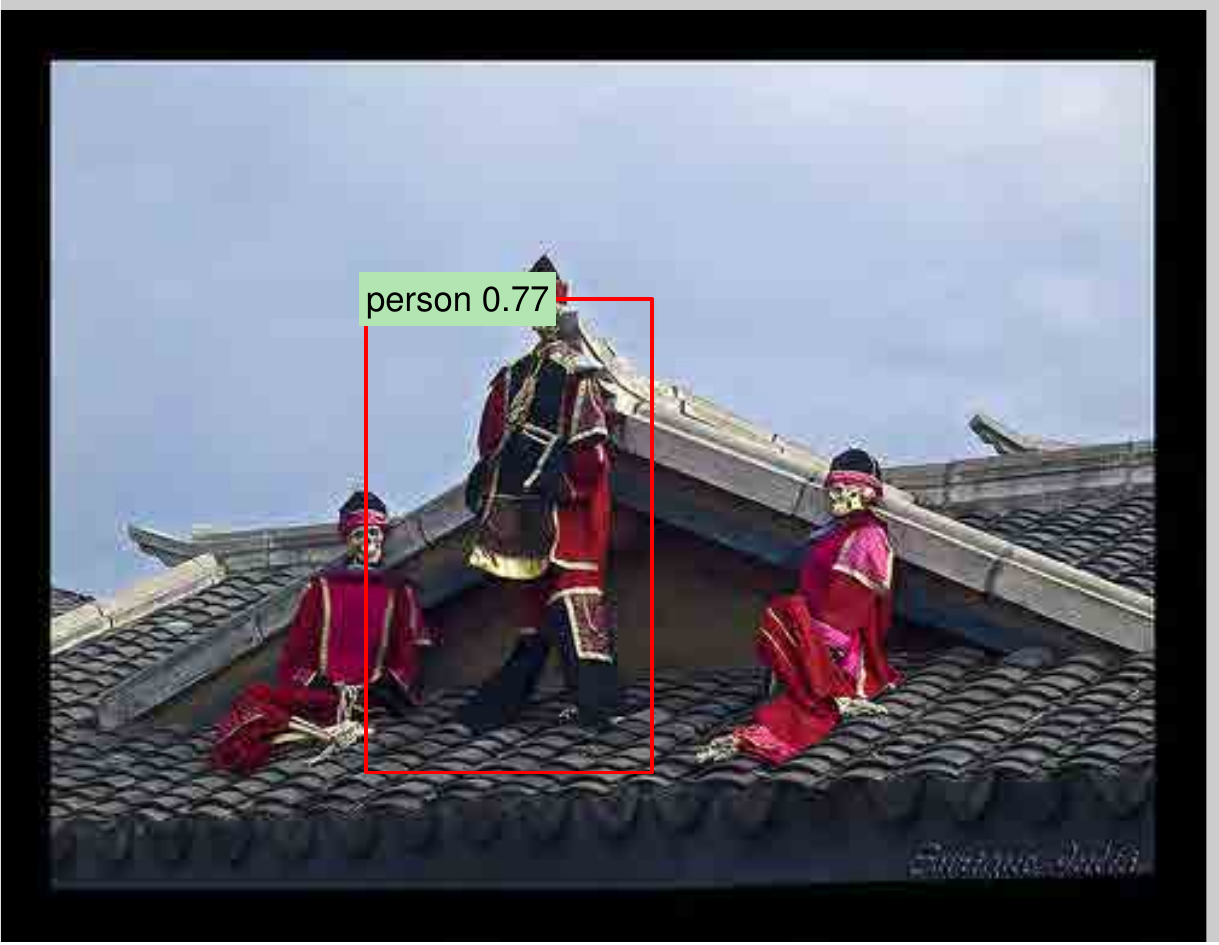}
\includegraphics[height=\sz]{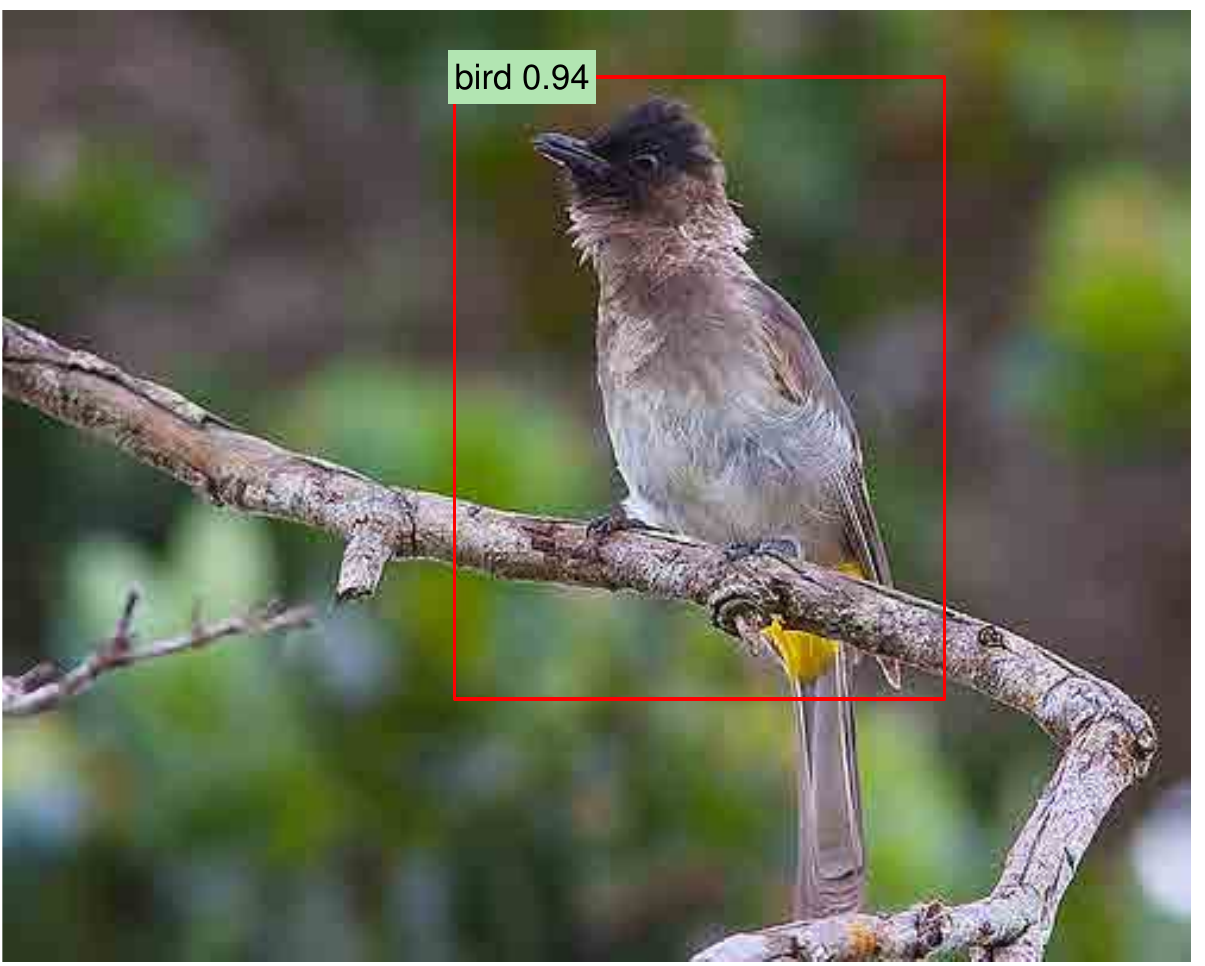}
\includegraphics[height=\sz]{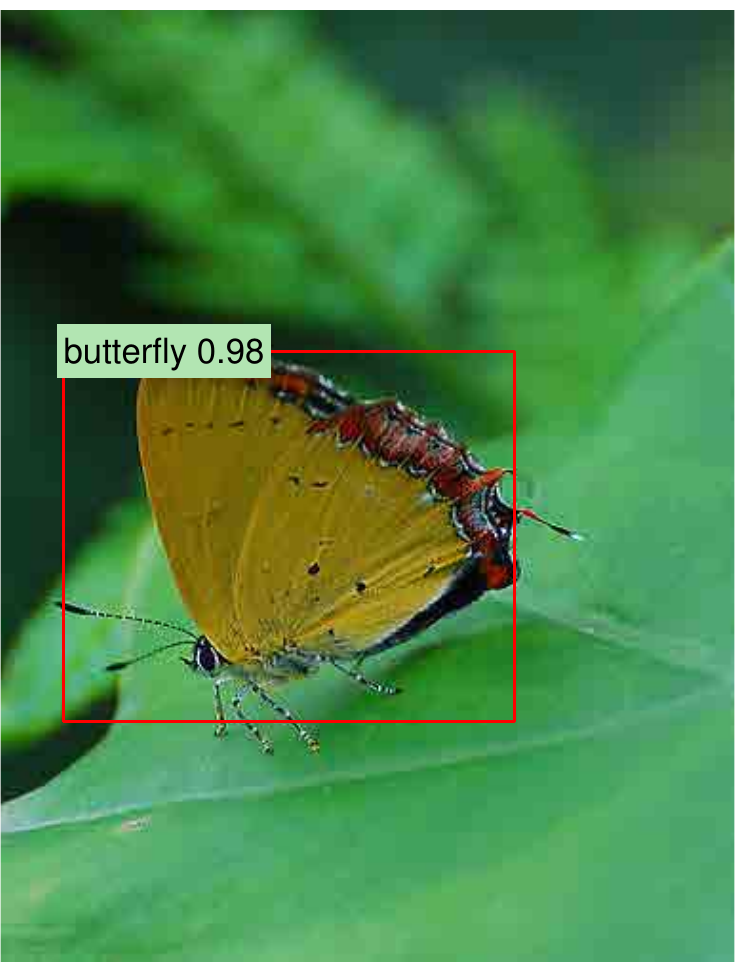}
\includegraphics[height=\sz]{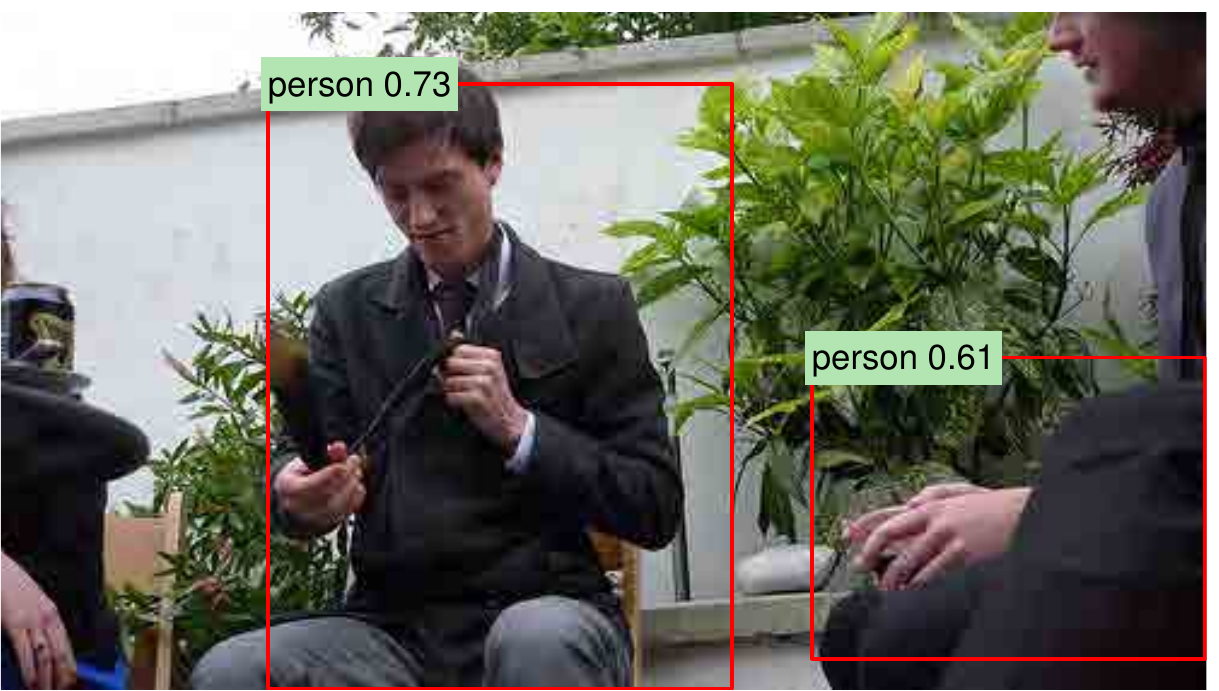}
\includegraphics[height=\sz]{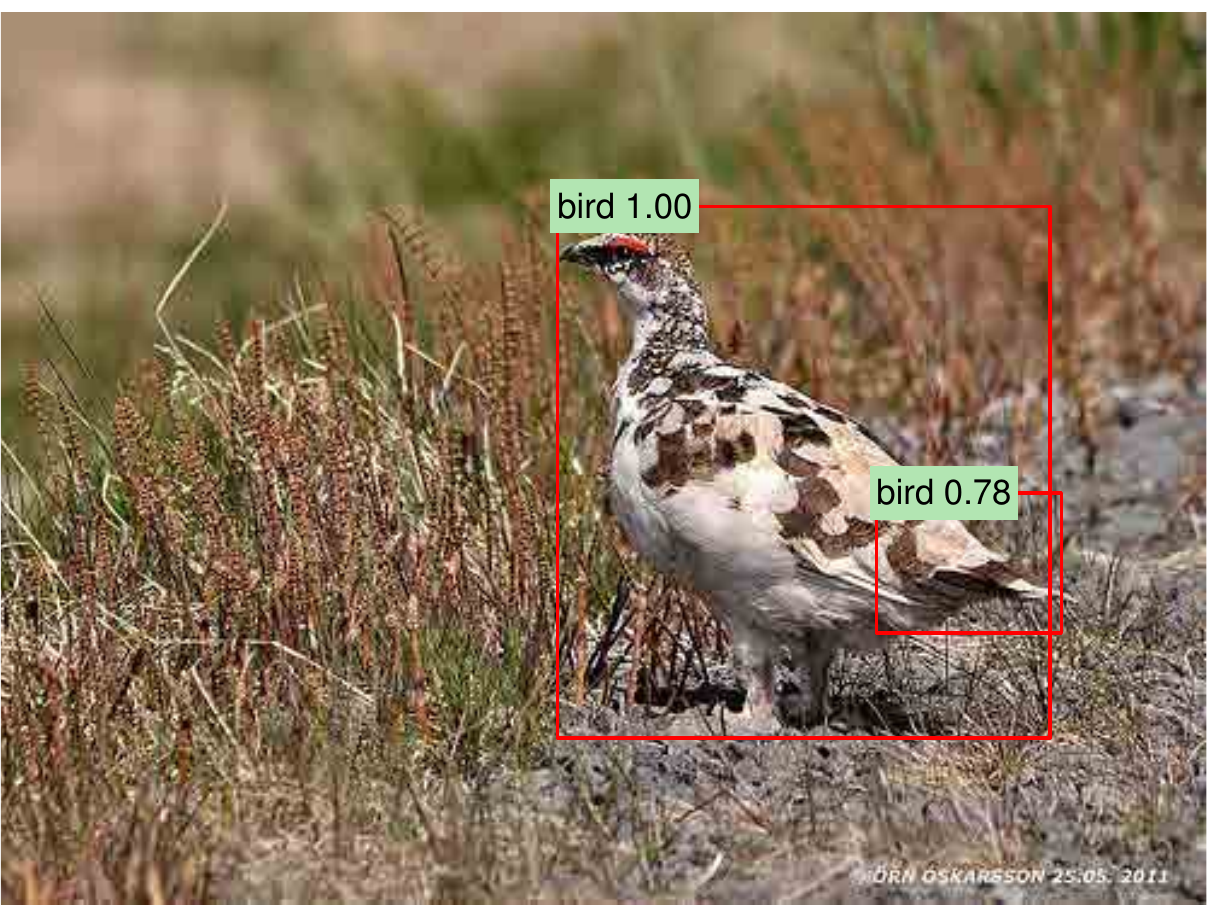}
\includegraphics[height=\sz]{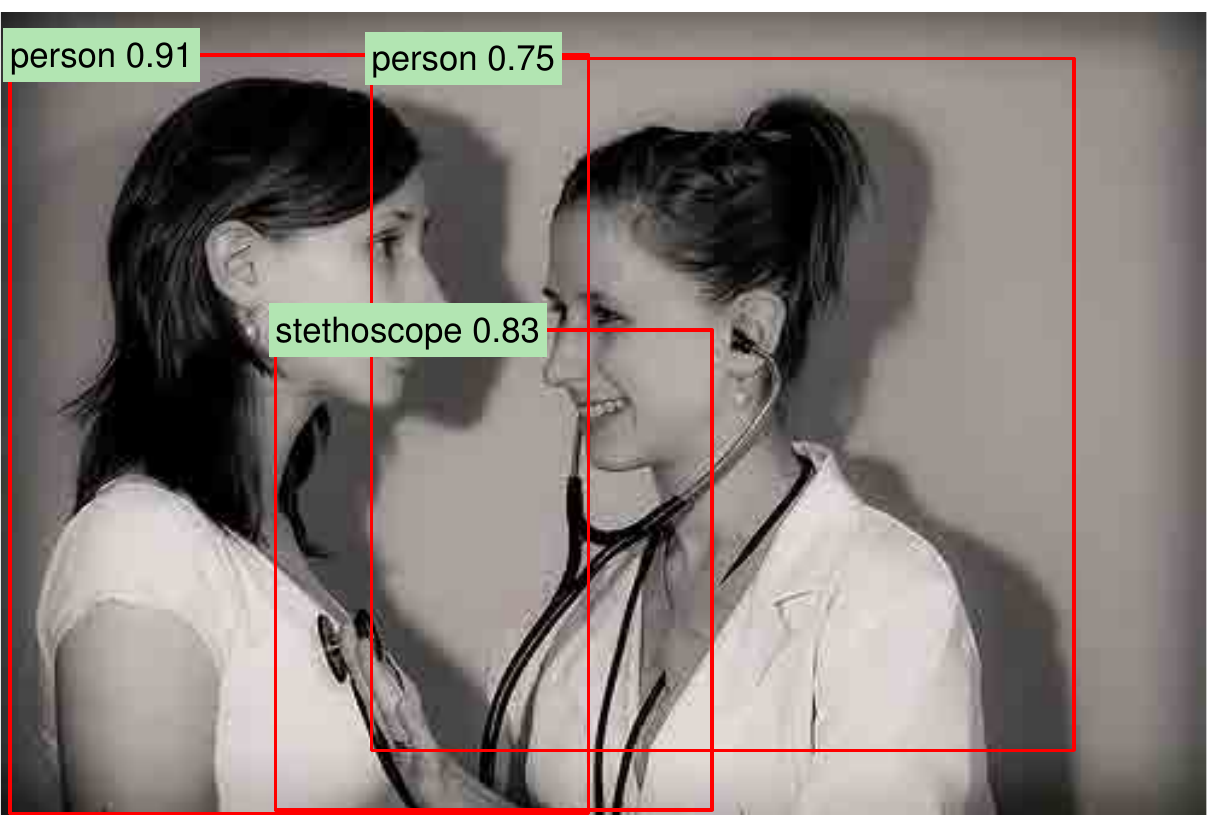}
\includegraphics[height=\sz]{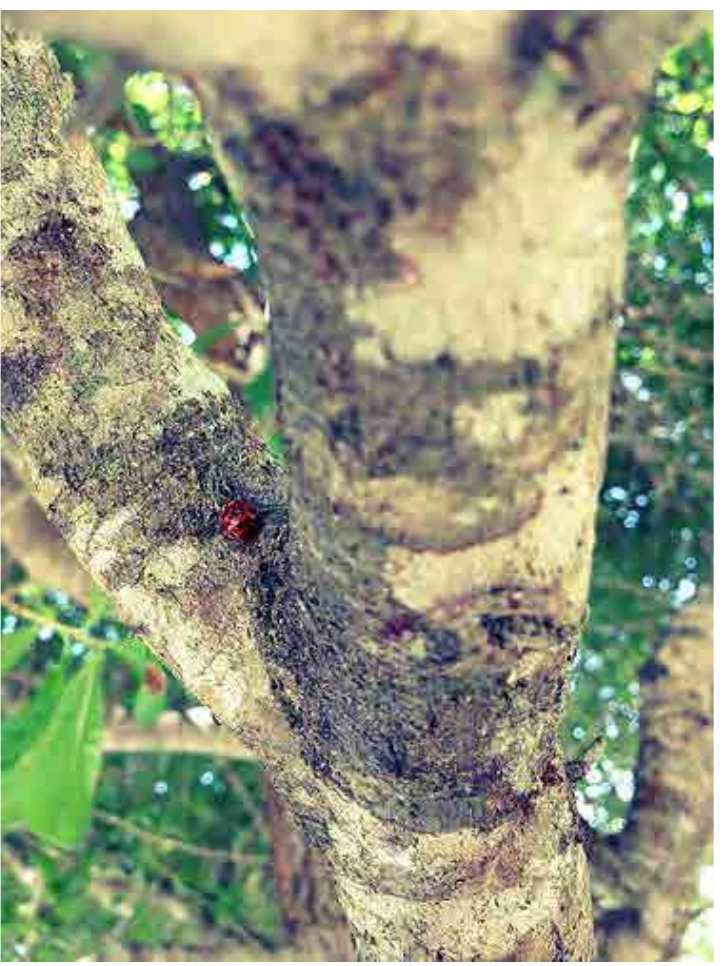}
\includegraphics[height=\sz]{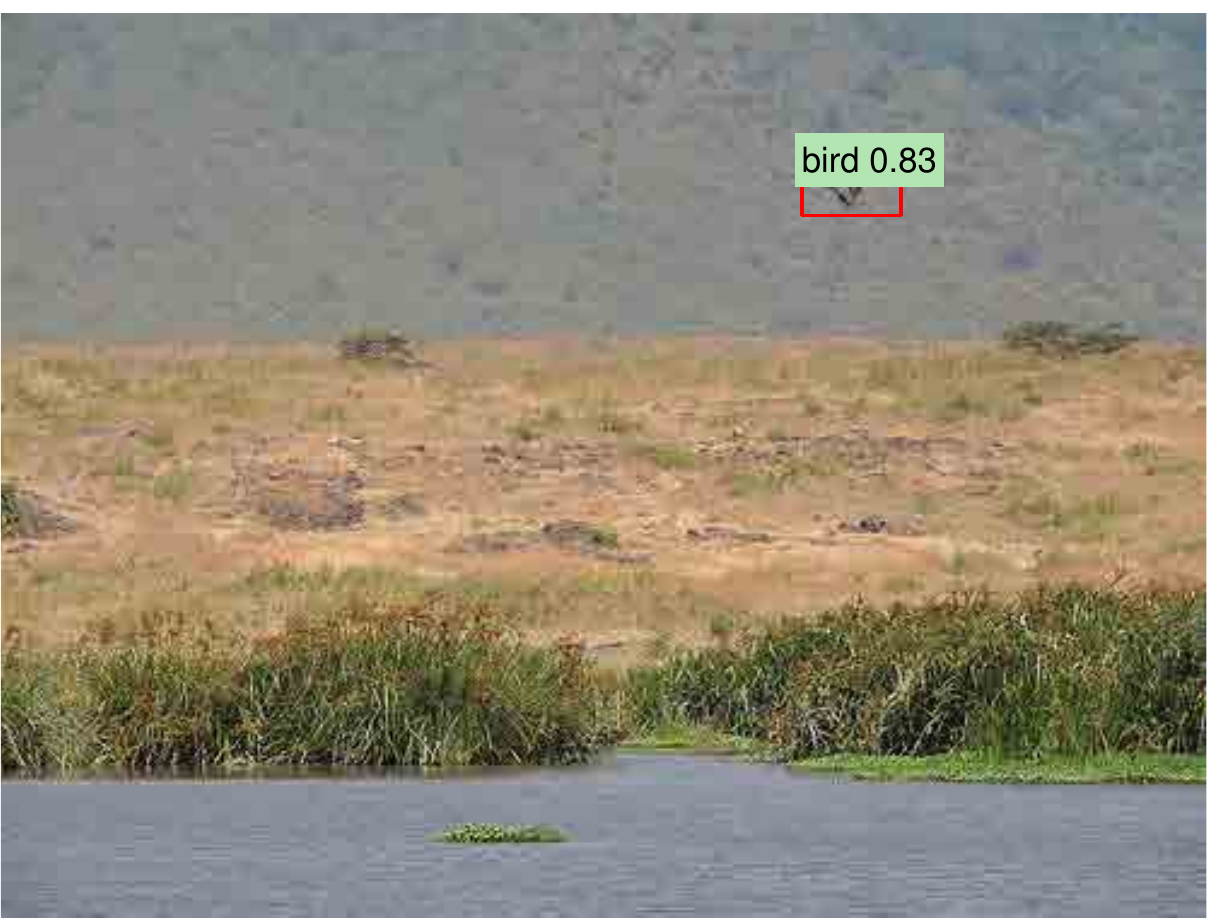}
\end{center}
\caption{More randomly selected examples. See \figref{examples1} caption for details.
Viewing digitally with zoom is recommended.
}
\figlabel{examples2}
\end{figure*}

\begin{figure*}[t]
\begin{center}
\def \sz {1in}
\includegraphics[height=\sz]{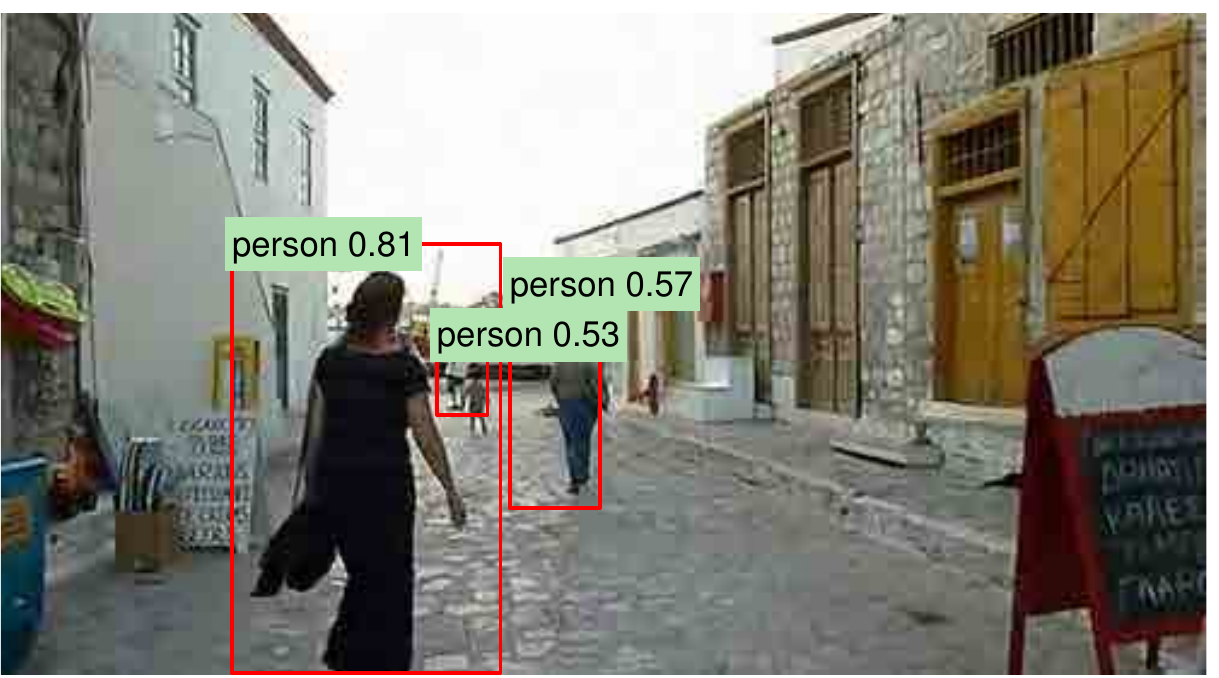}
\includegraphics[height=\sz]{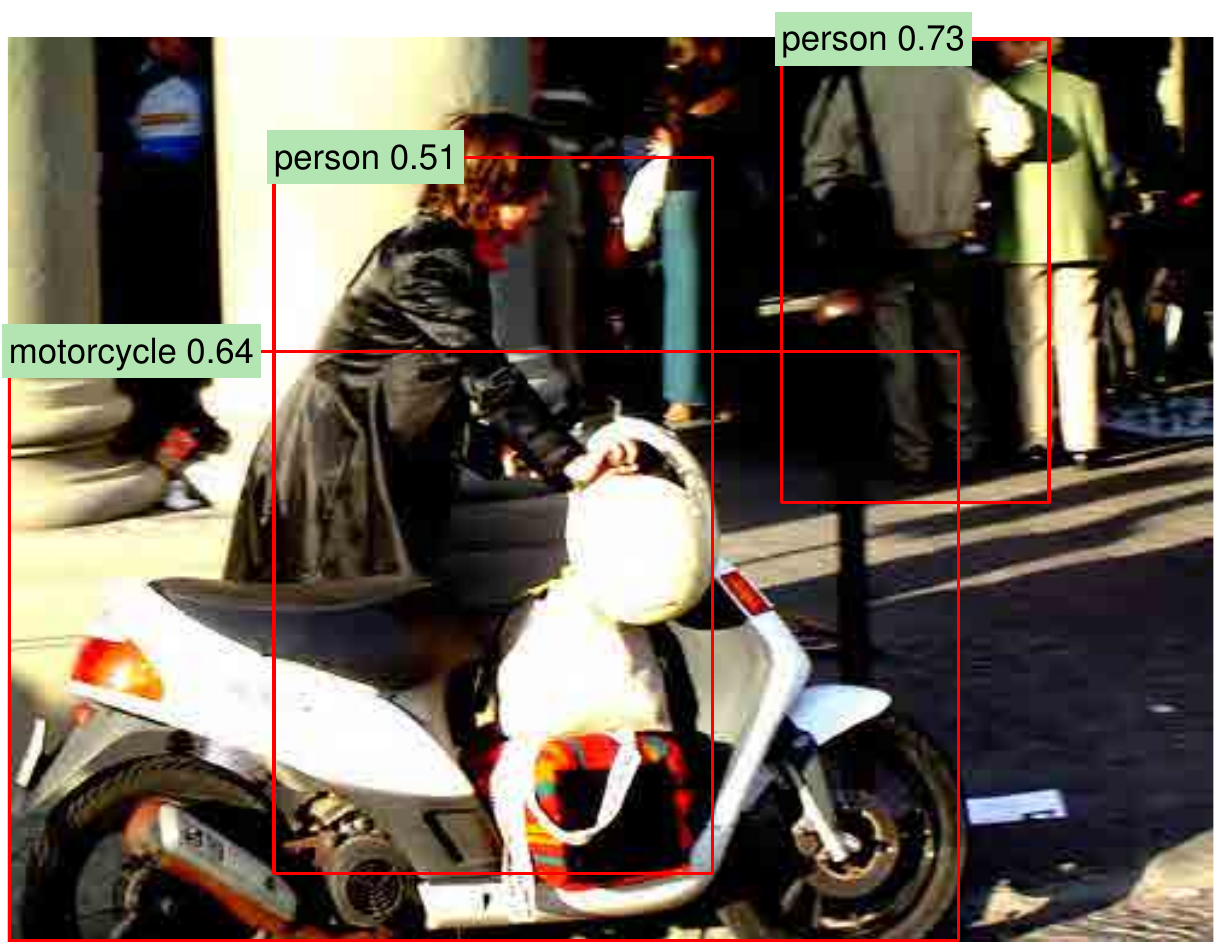}
\includegraphics[height=\sz]{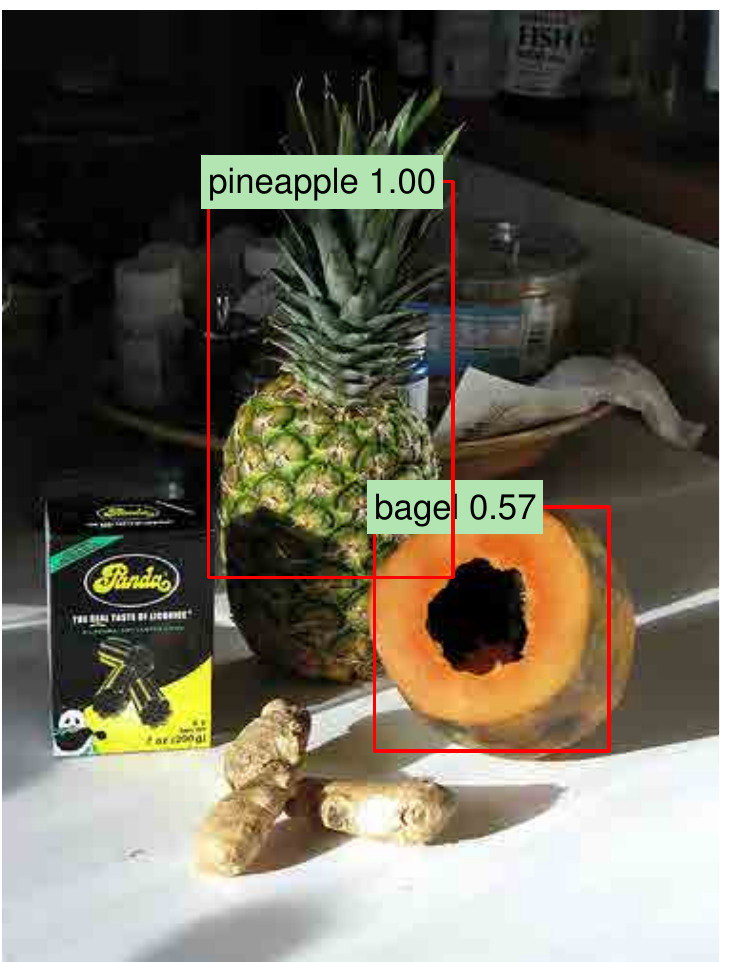}
\includegraphics[height=\sz]{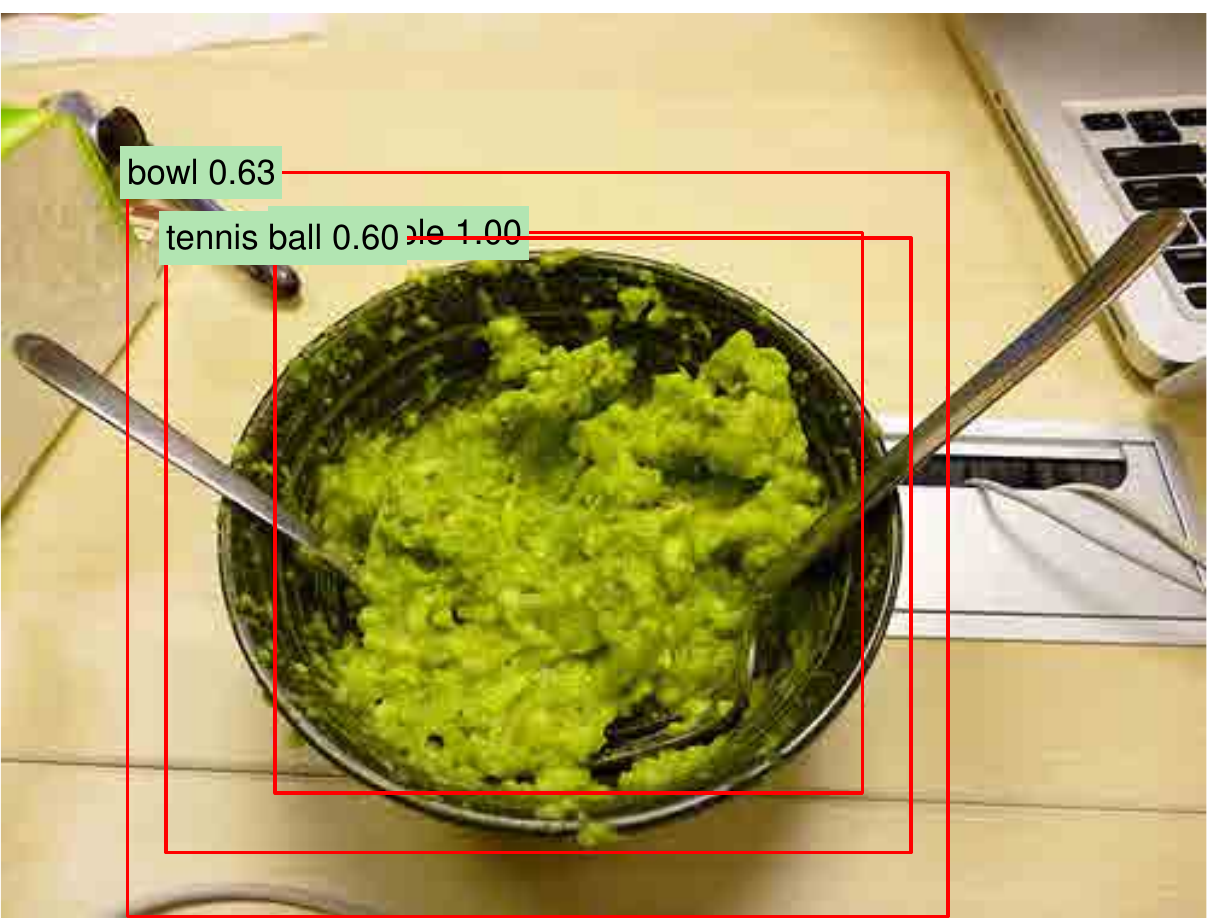}
\includegraphics[height=\sz]{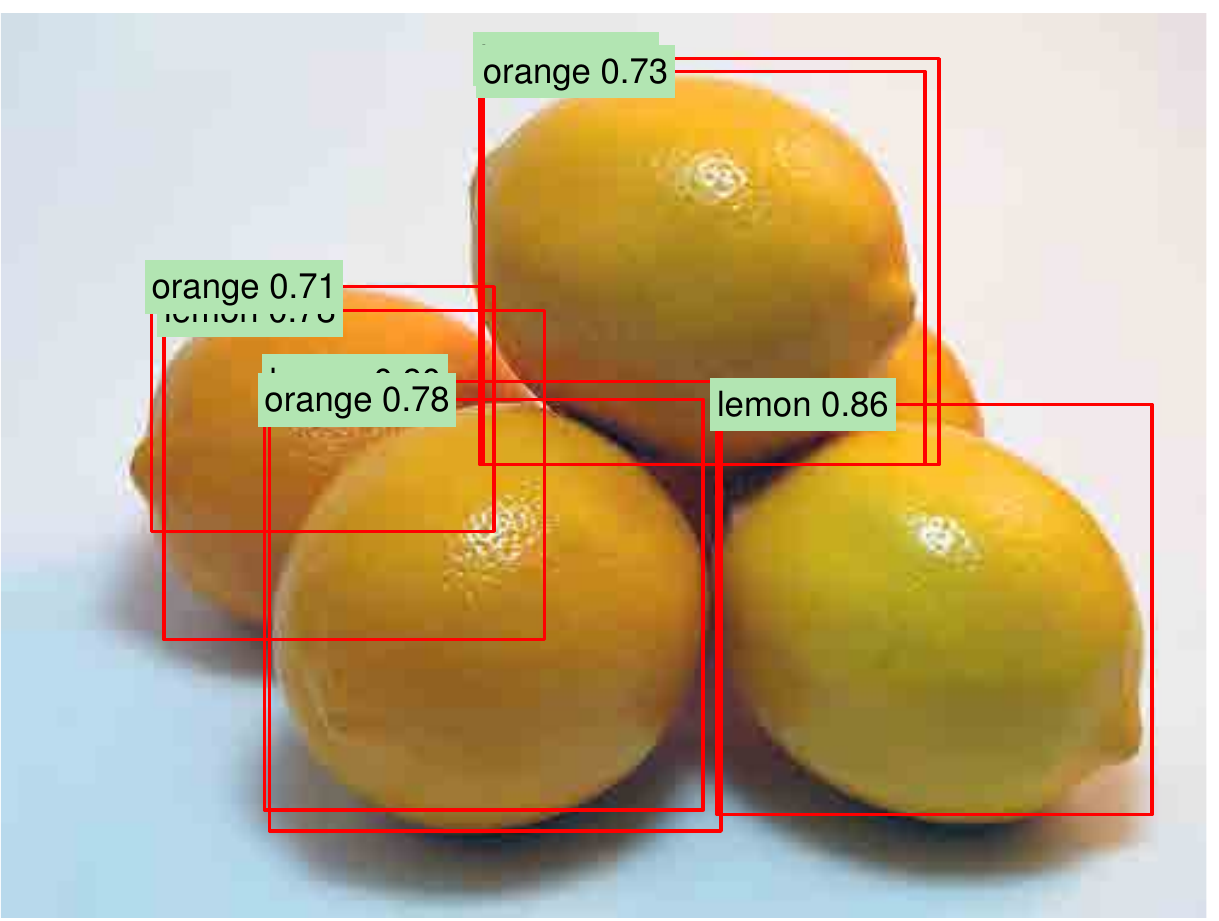}
\includegraphics[height=\sz]{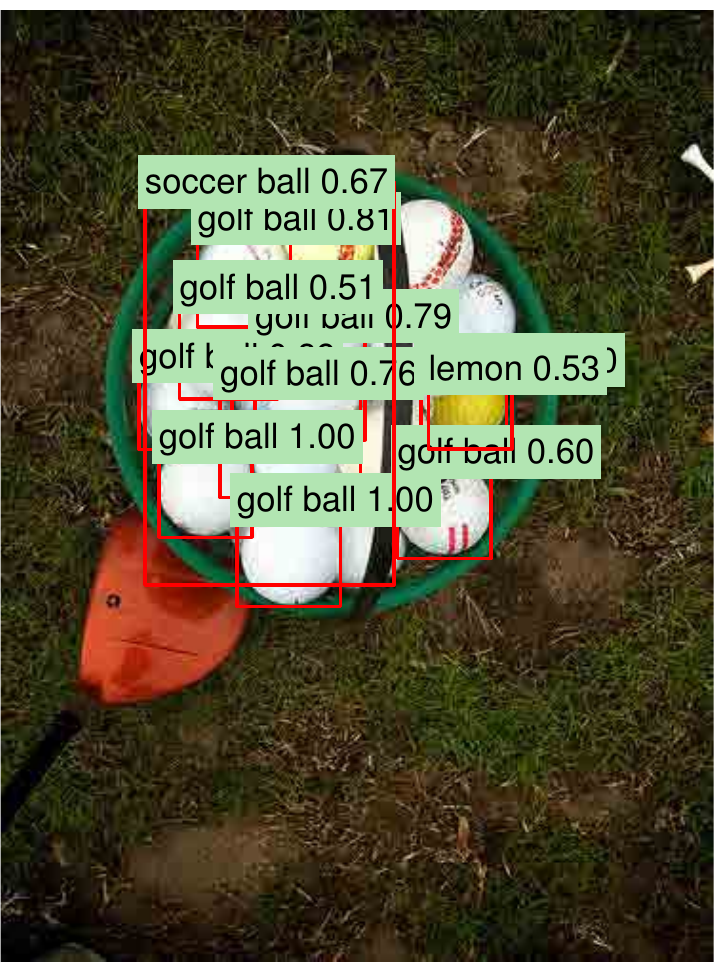}
\includegraphics[height=\sz]{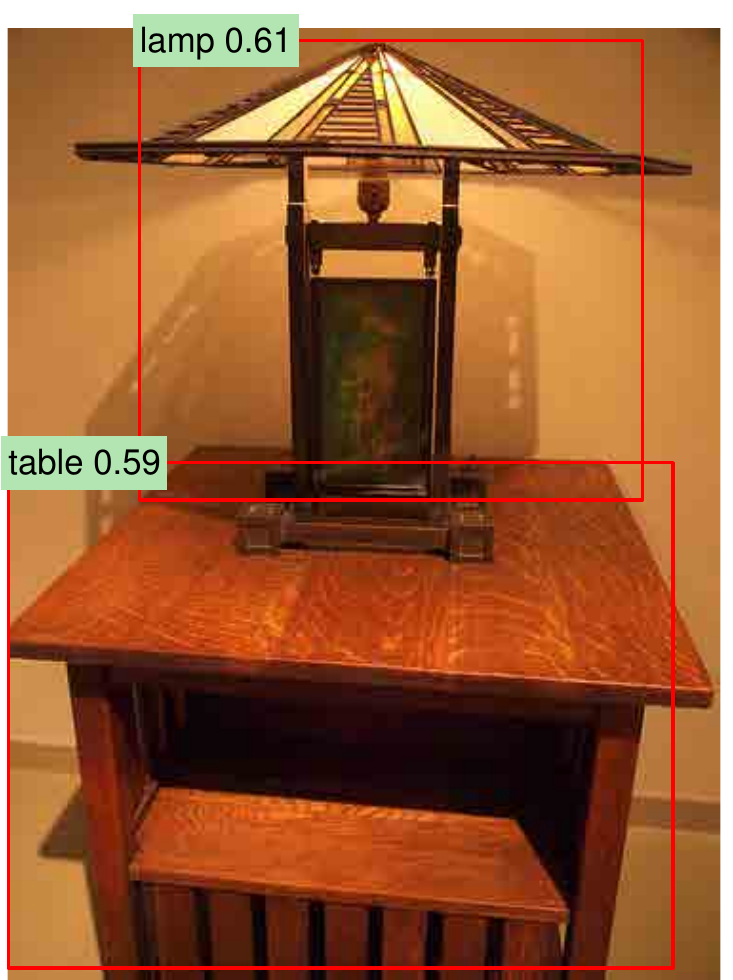}
\includegraphics[height=\sz]{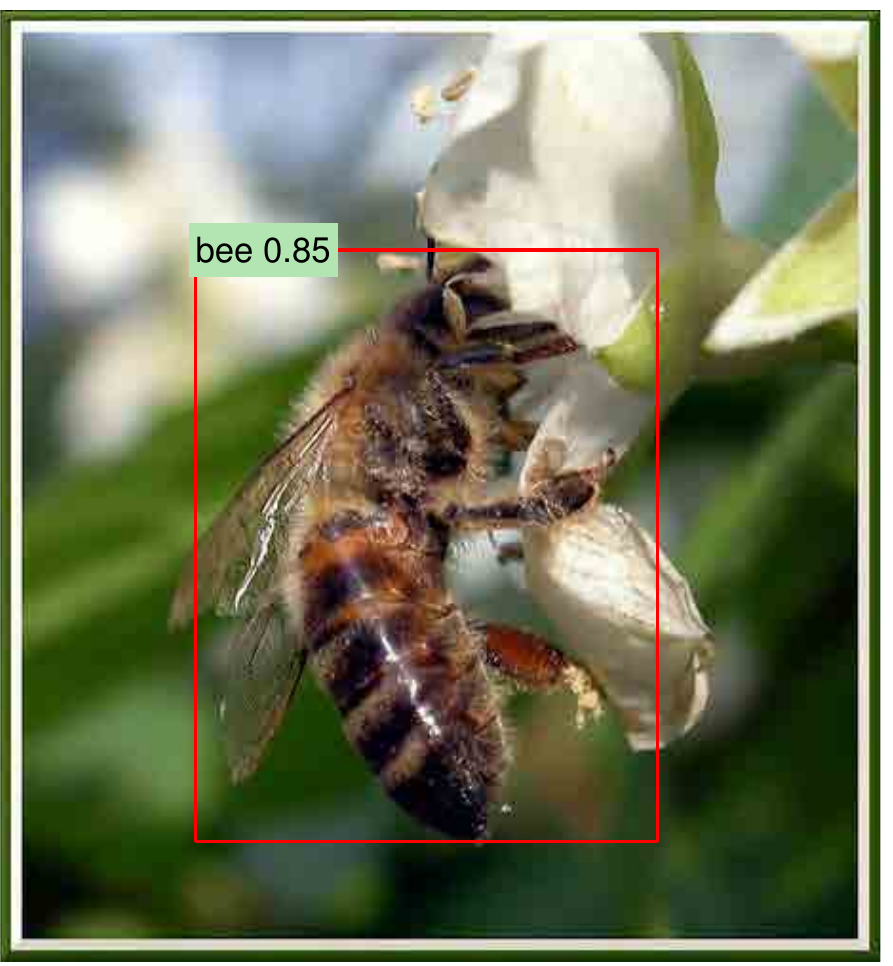}
\includegraphics[height=\sz]{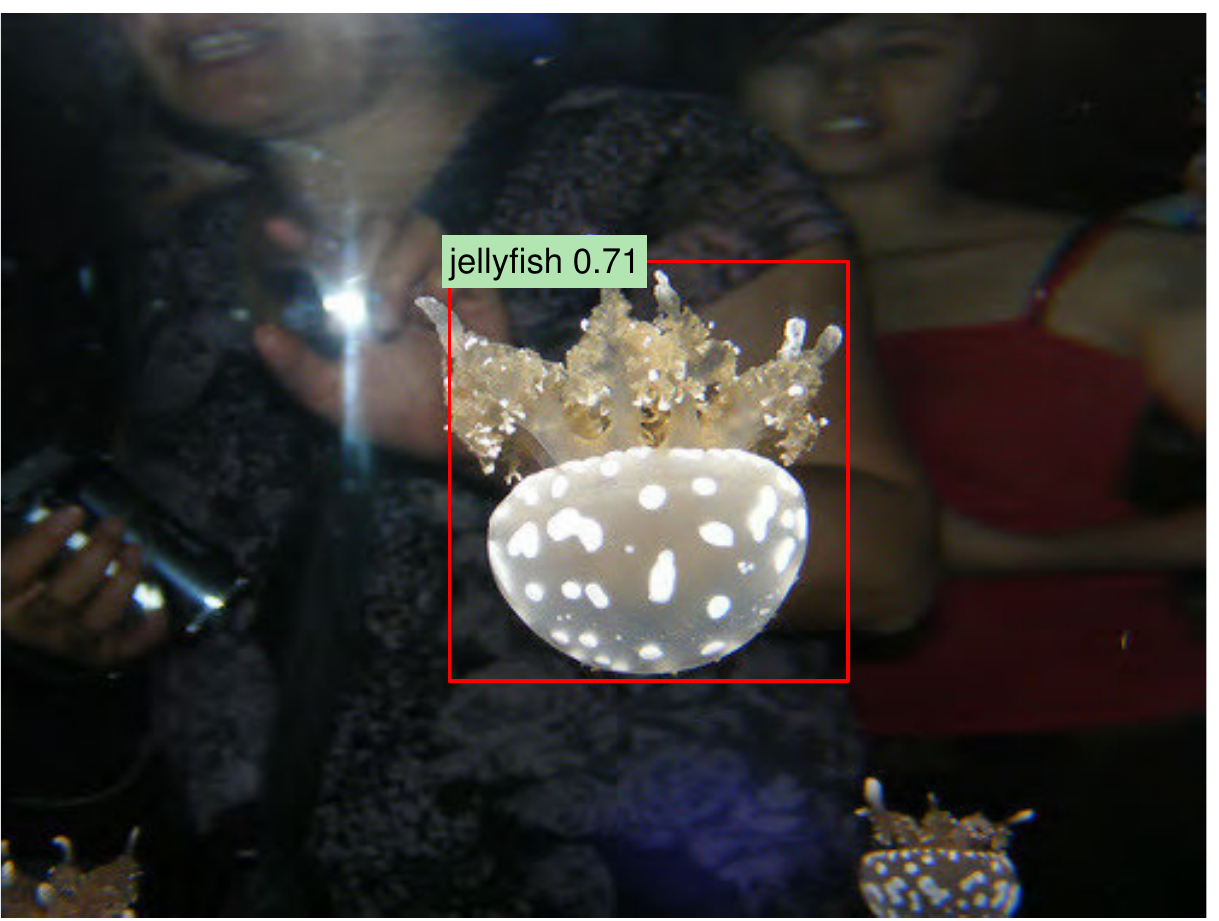}
\includegraphics[height=\sz]{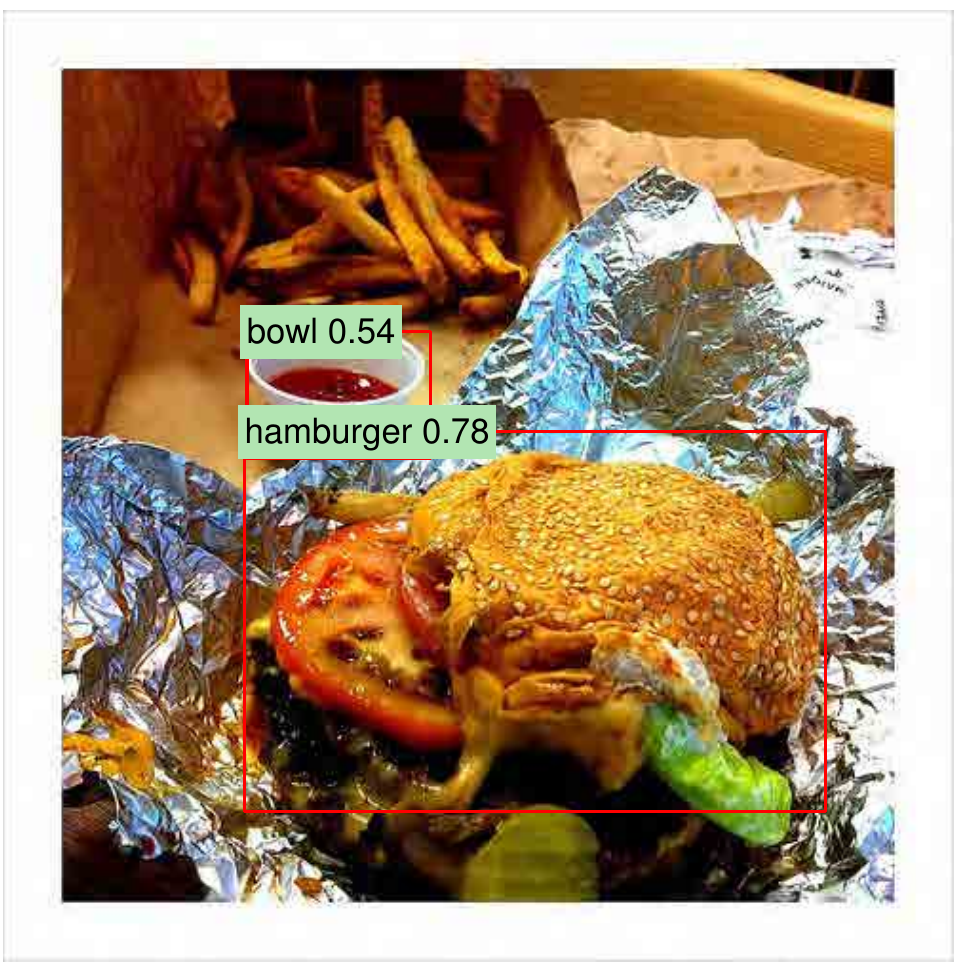}
\includegraphics[height=\sz]{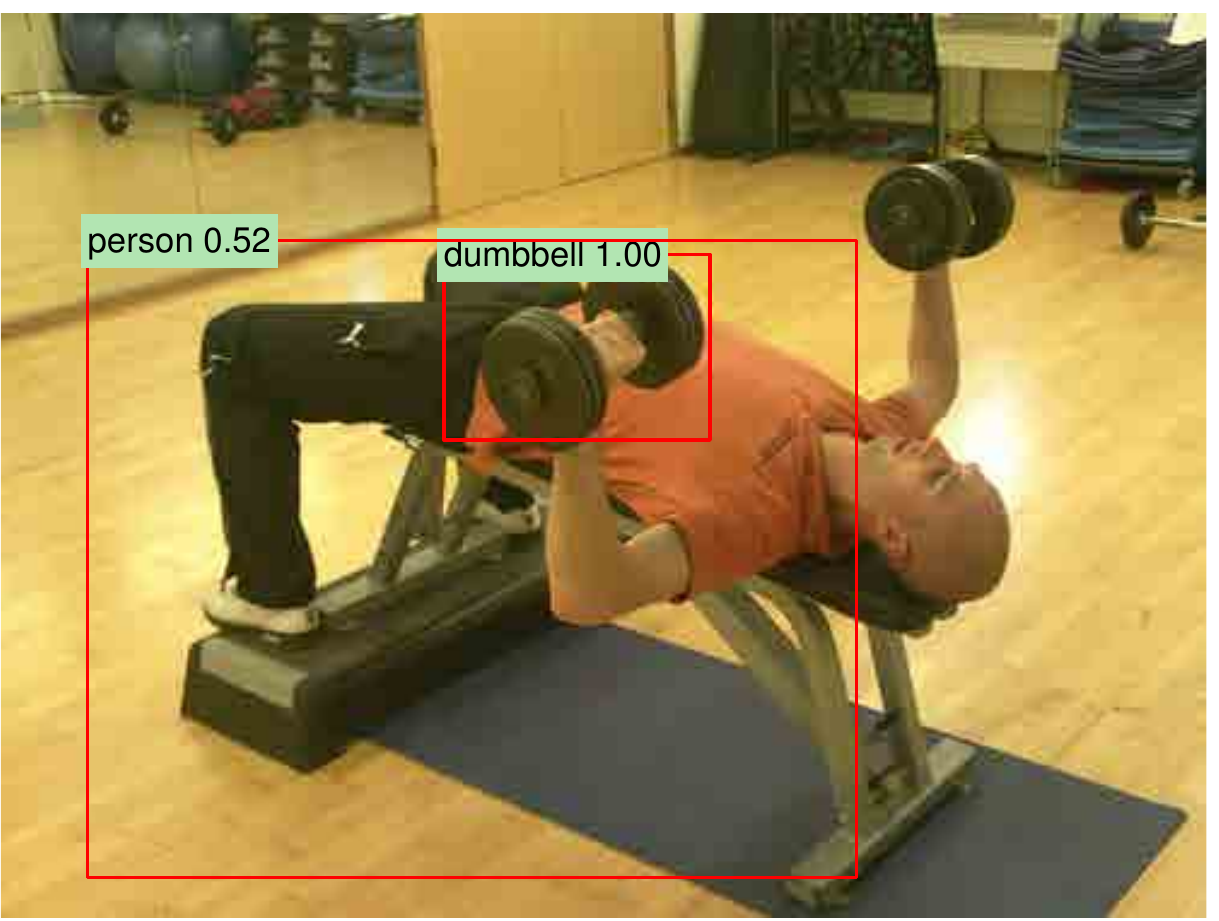}
\includegraphics[height=\sz]{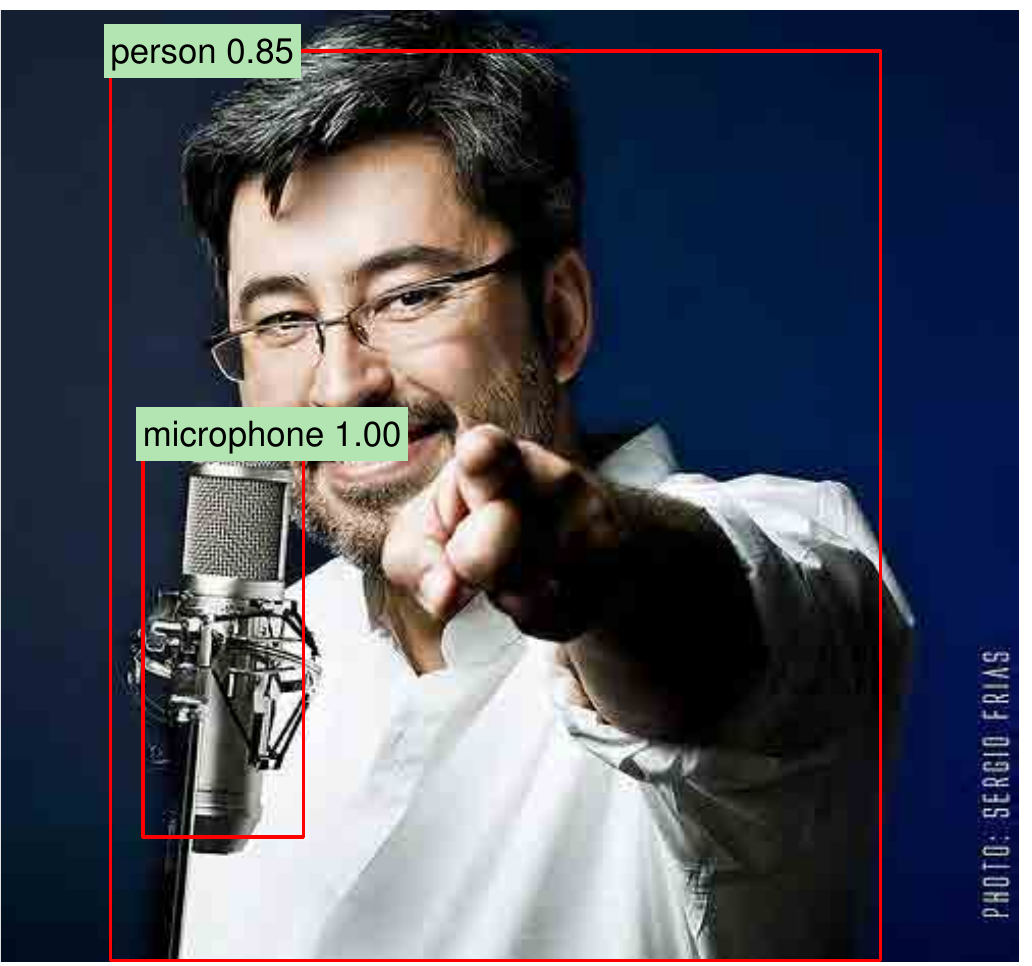}
\includegraphics[height=\sz]{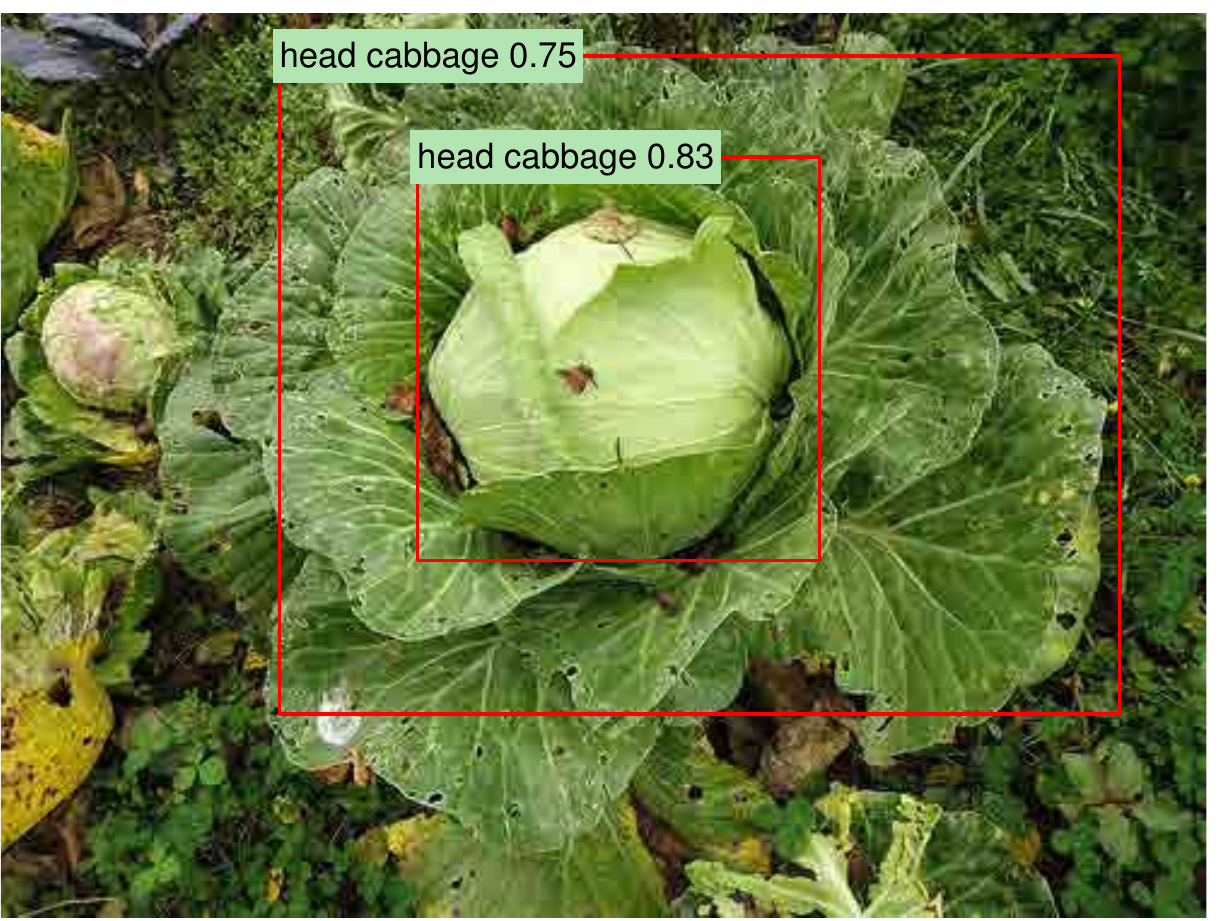}
\includegraphics[height=\sz]{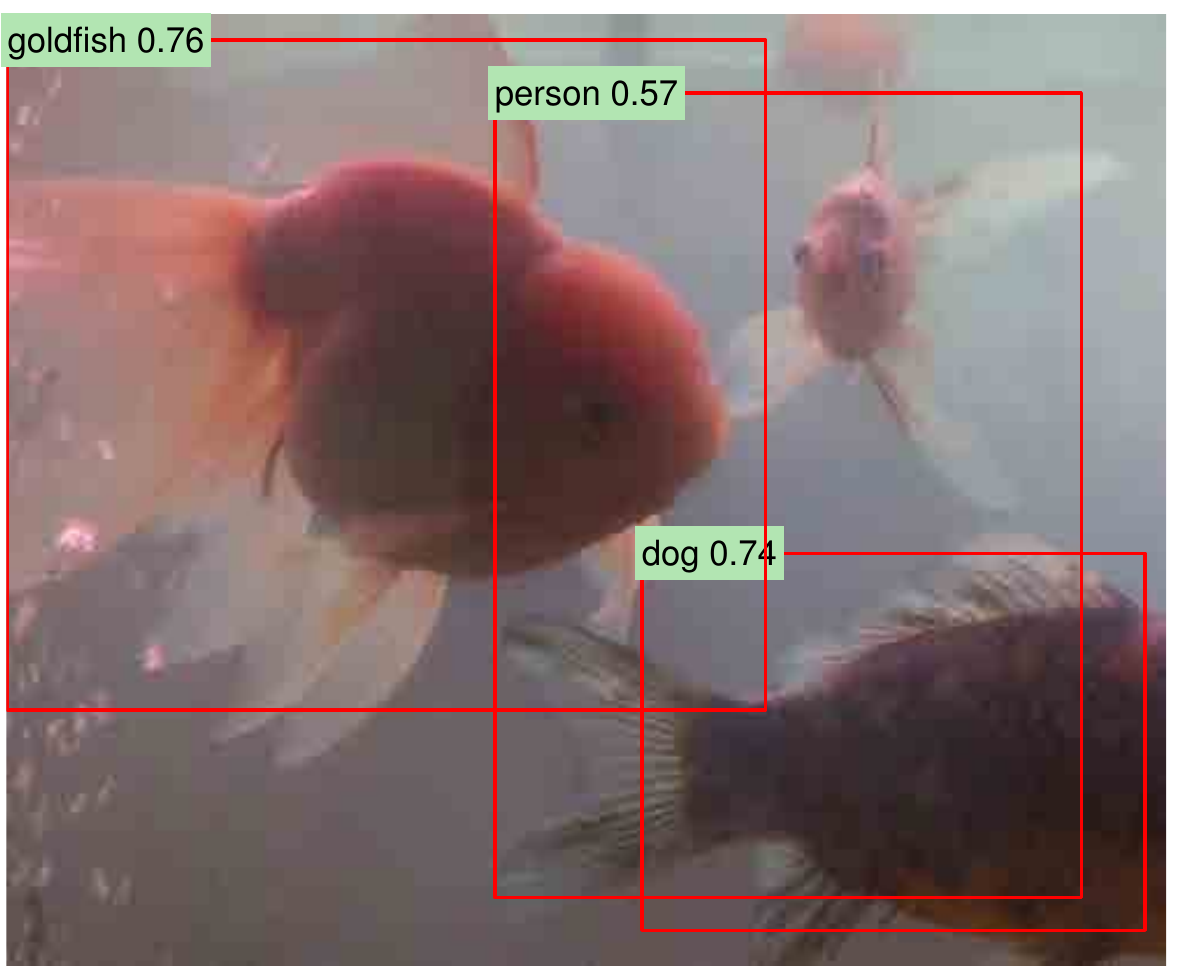}
\includegraphics[height=\sz]{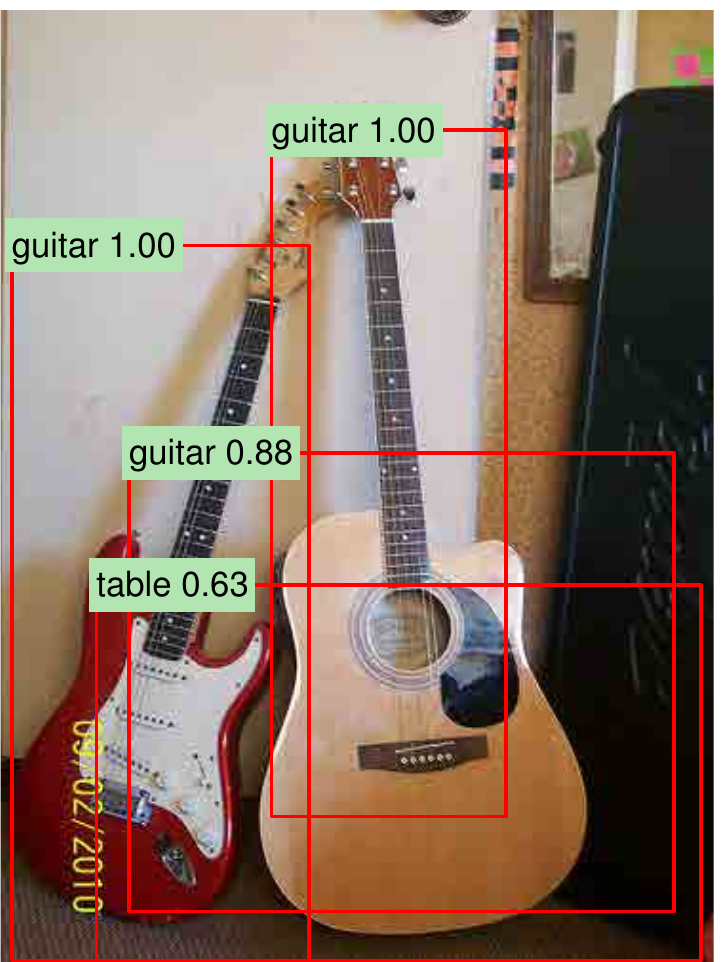}
\includegraphics[height=\sz]{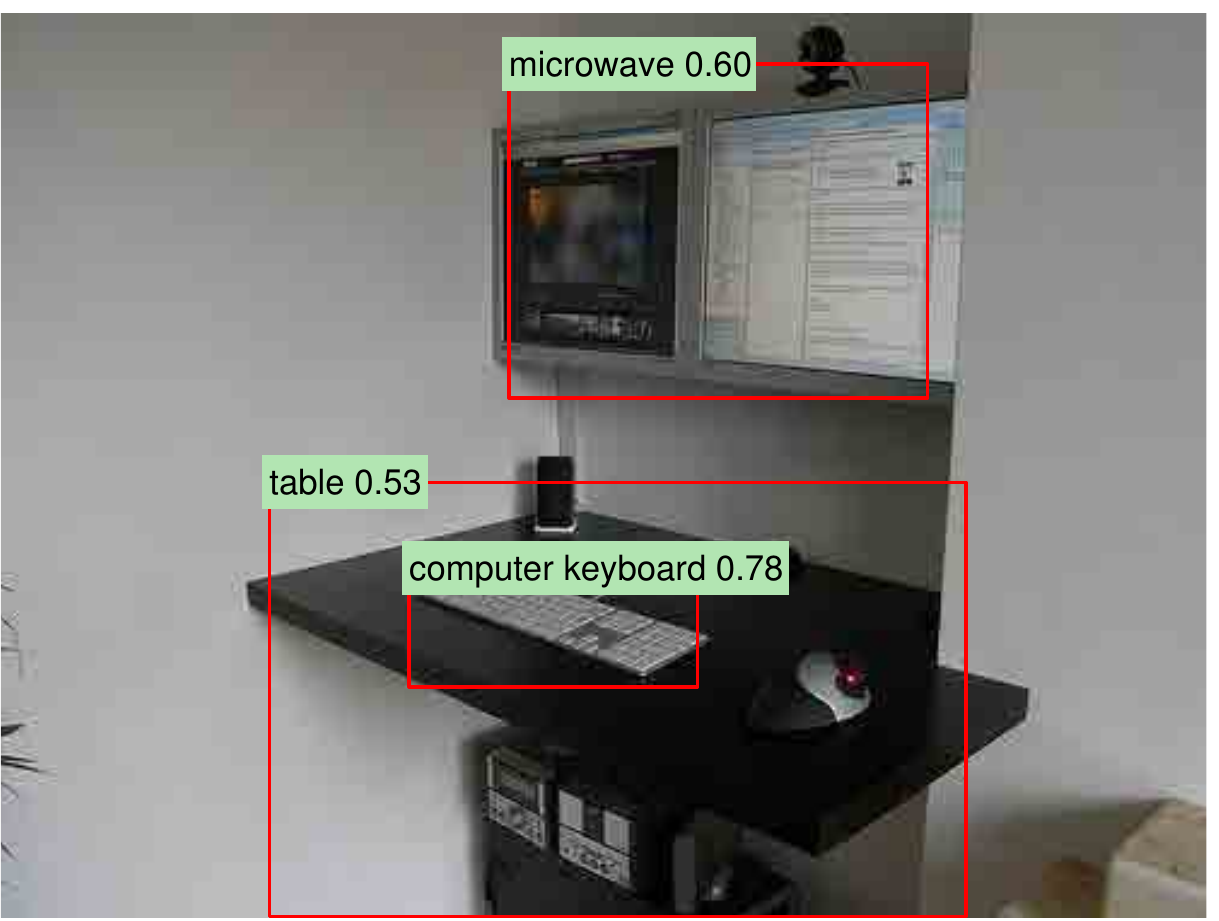}
\includegraphics[height=\sz]{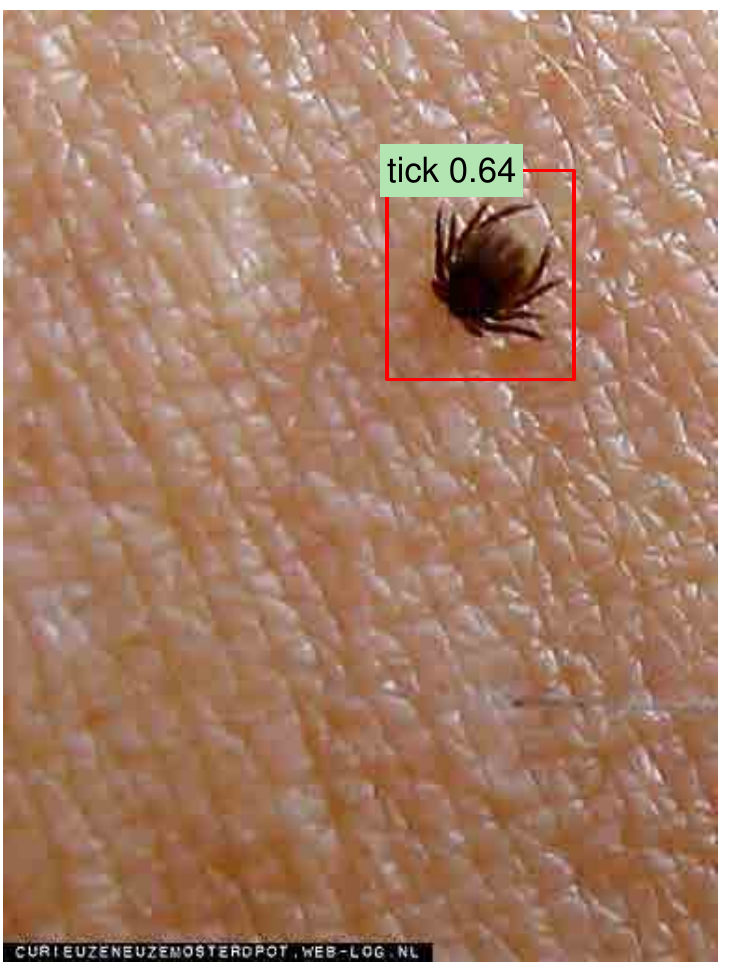}
\includegraphics[height=\sz]{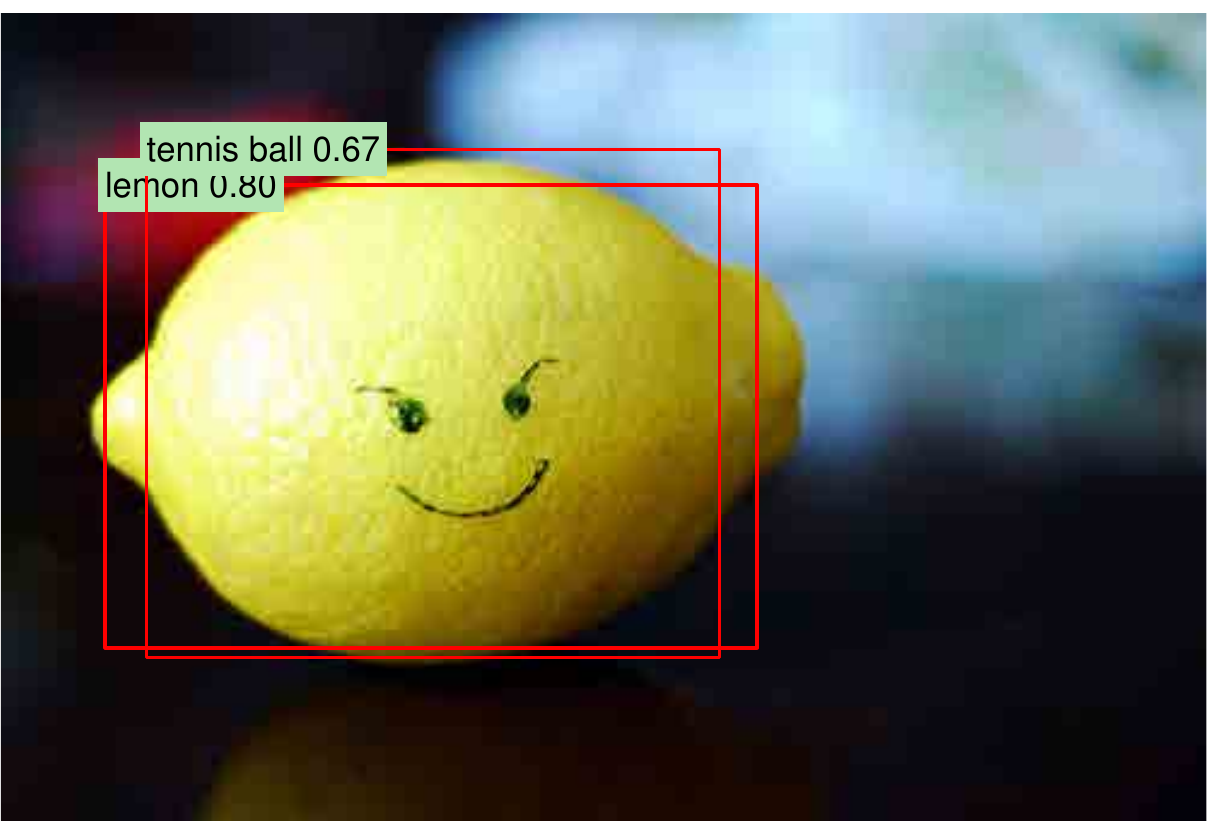}
\includegraphics[height=\sz]{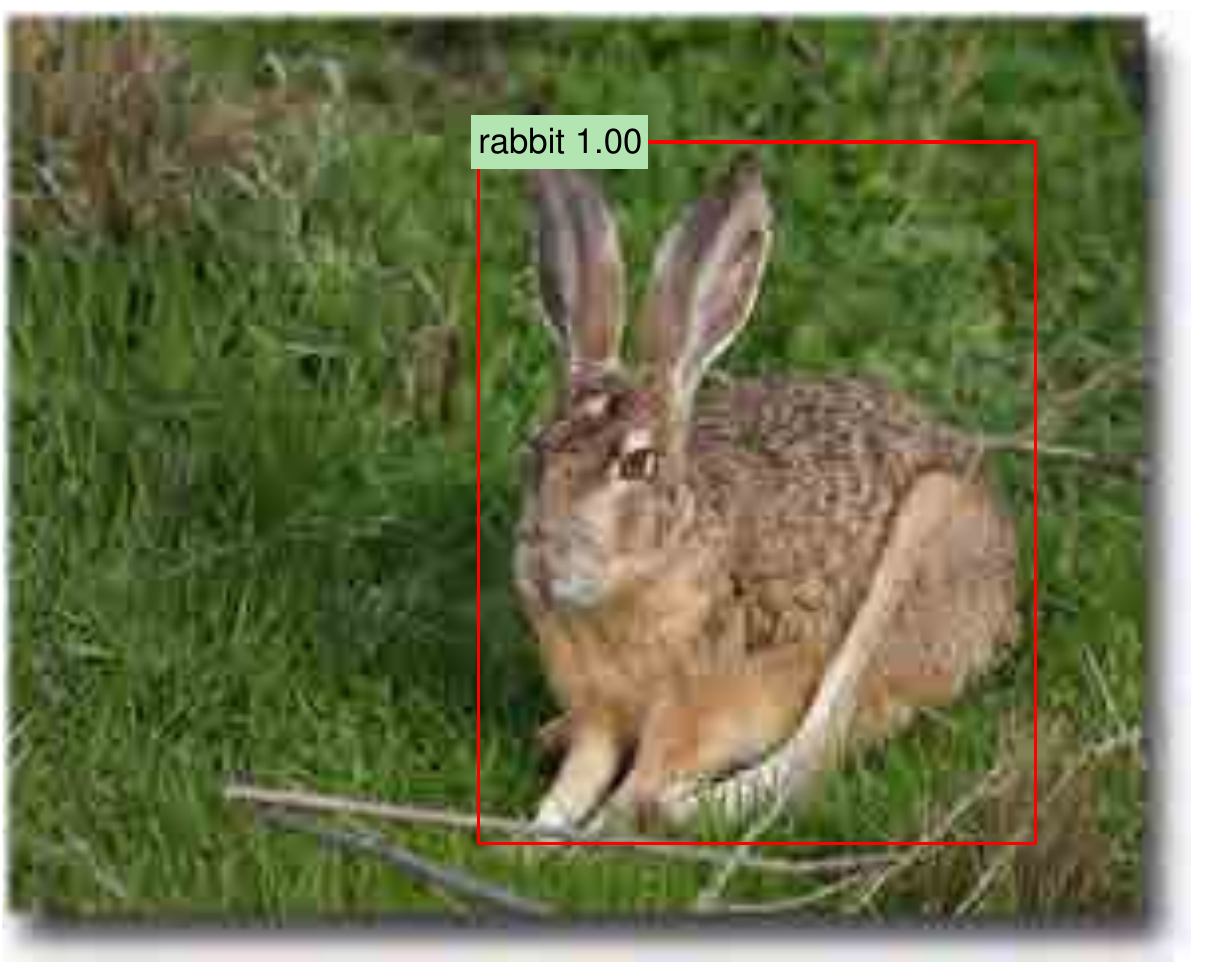}
\includegraphics[height=\sz]{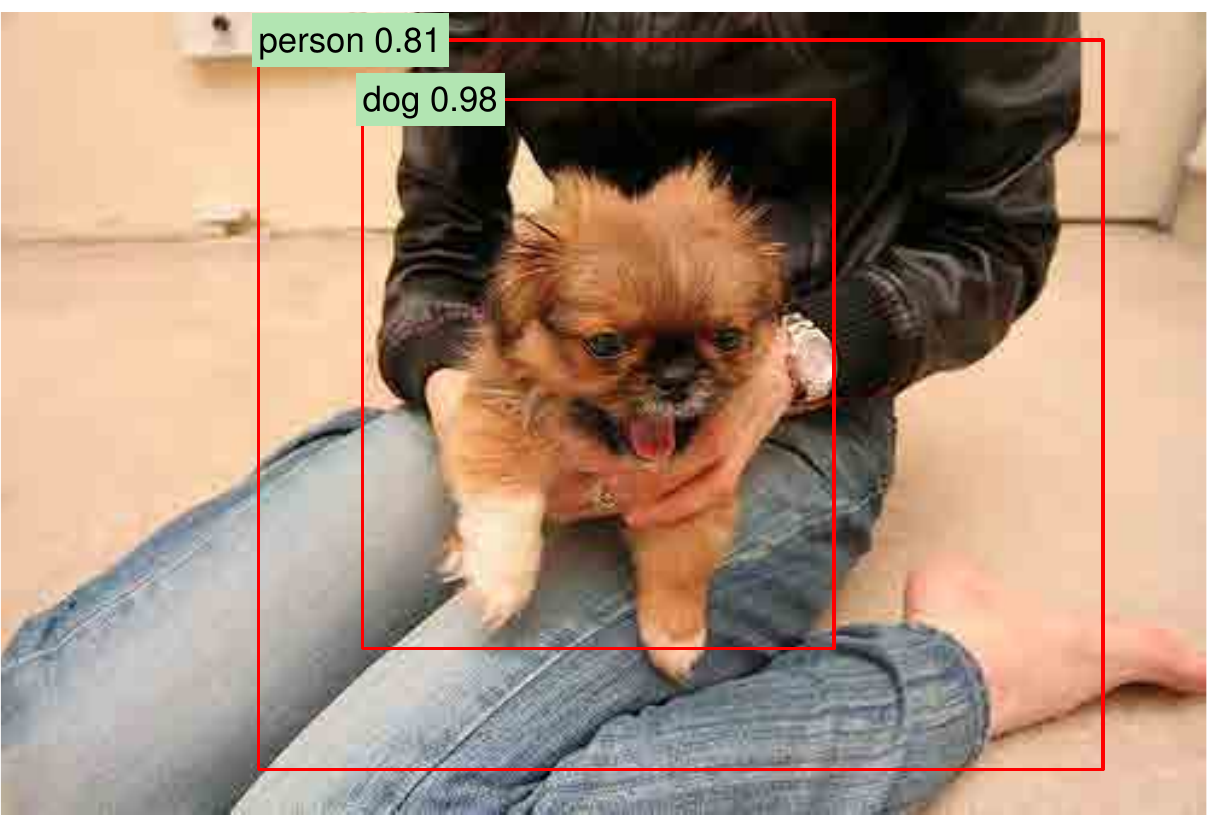}
\includegraphics[height=\sz]{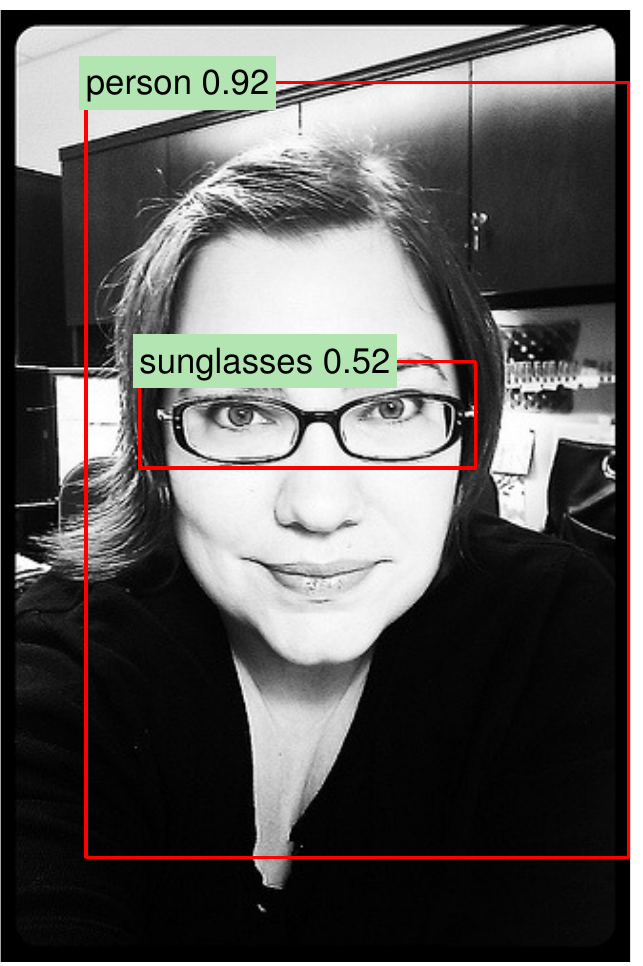}
\includegraphics[height=\sz]{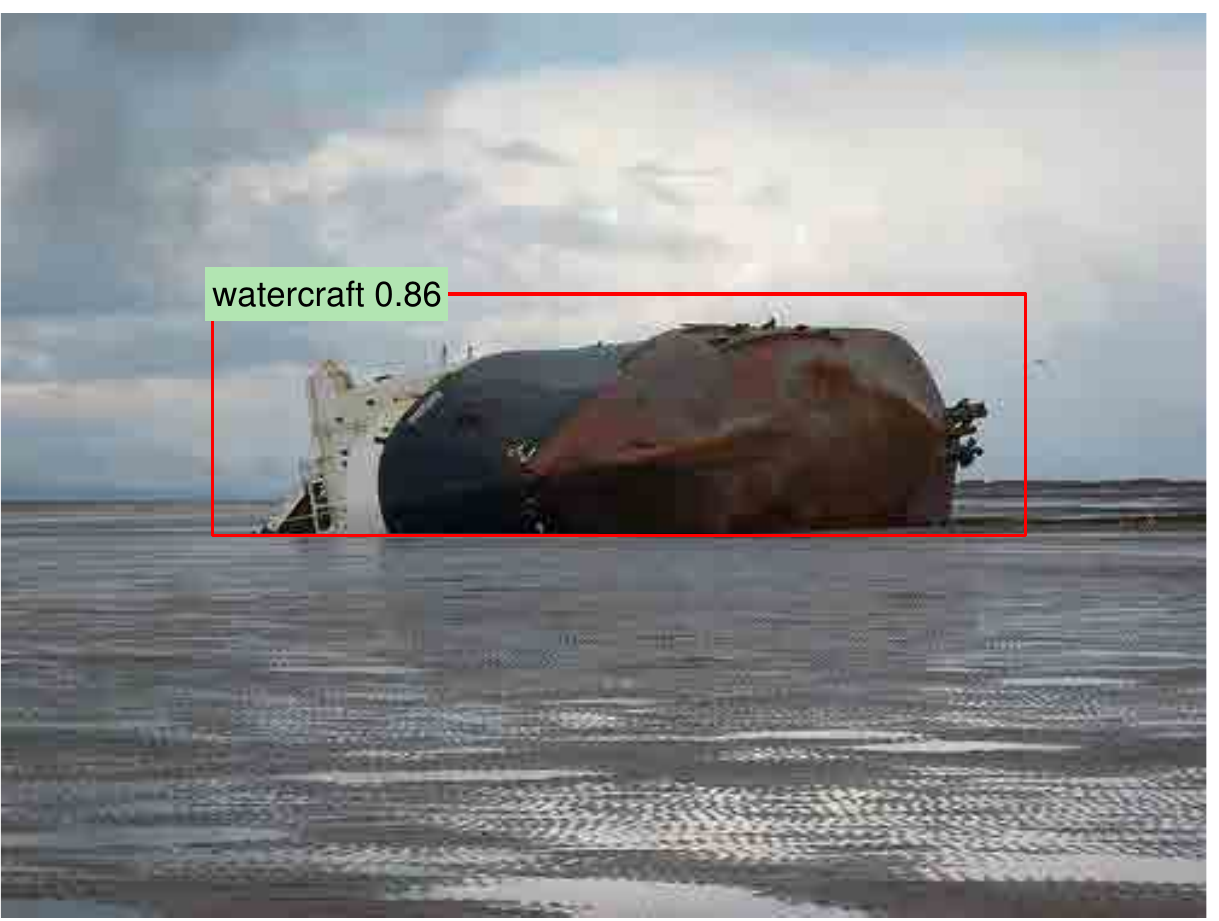}
\includegraphics[height=\sz]{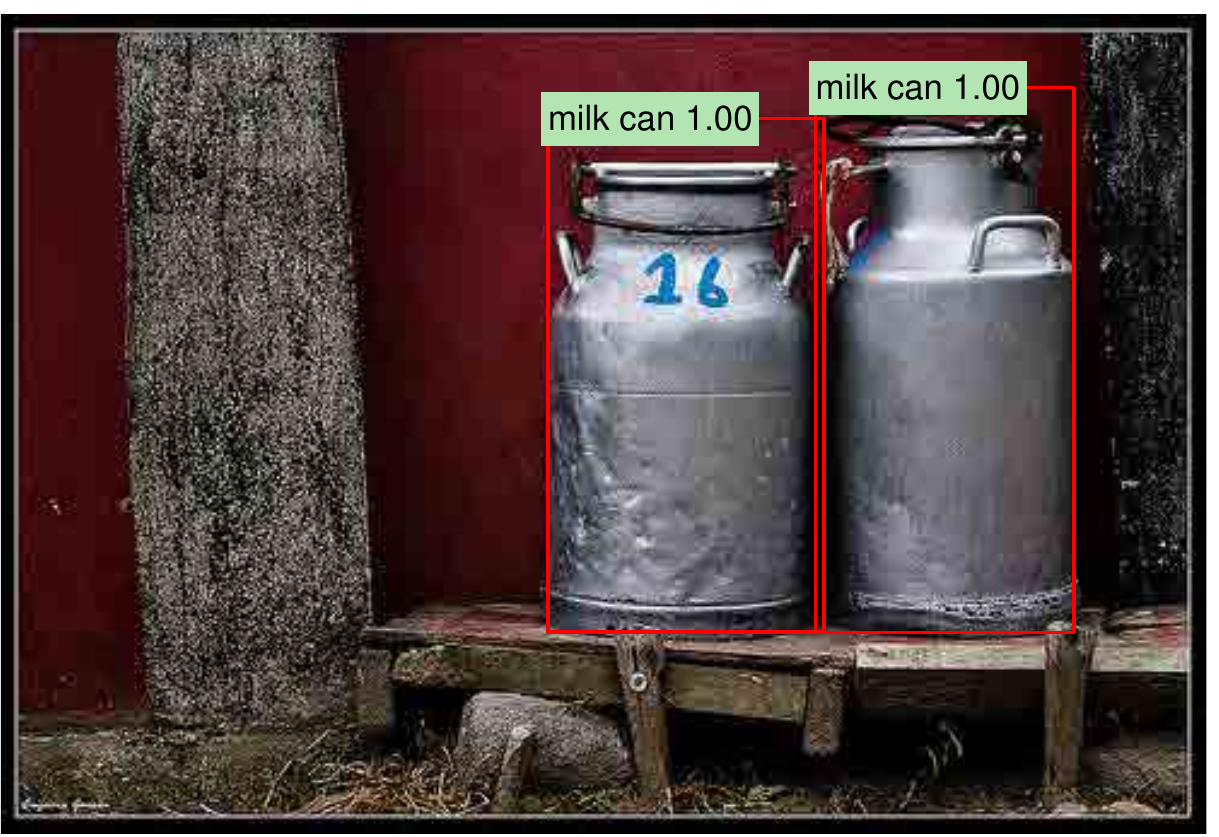}
\includegraphics[height=\sz]{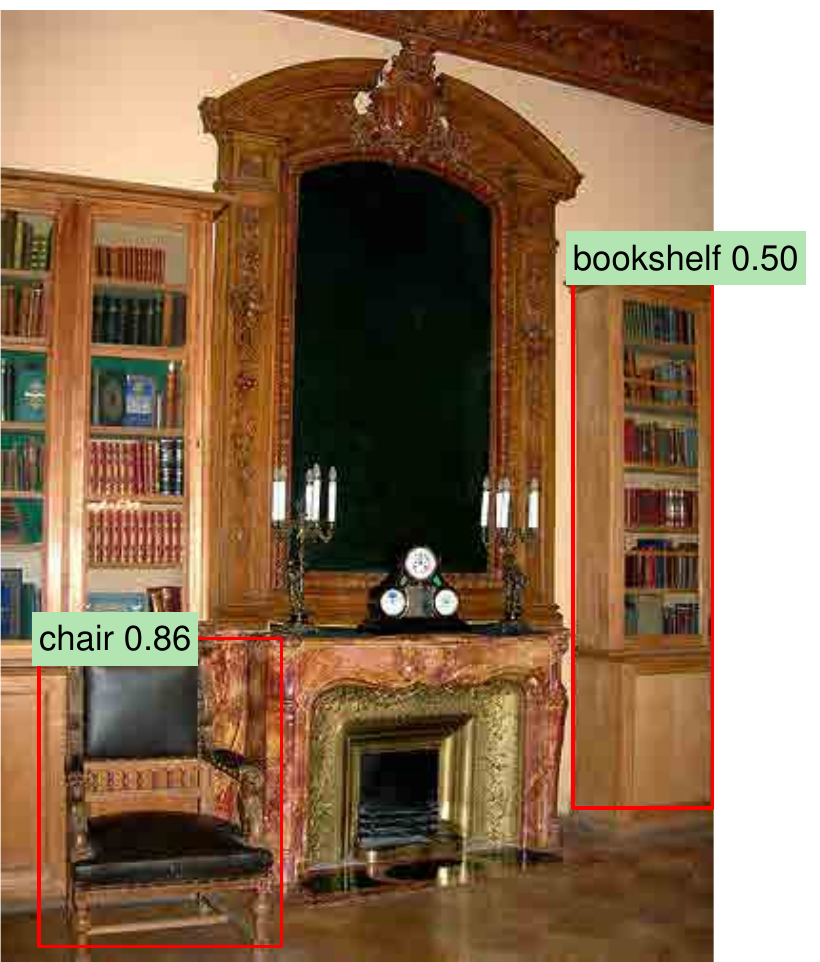}
\includegraphics[height=\sz]{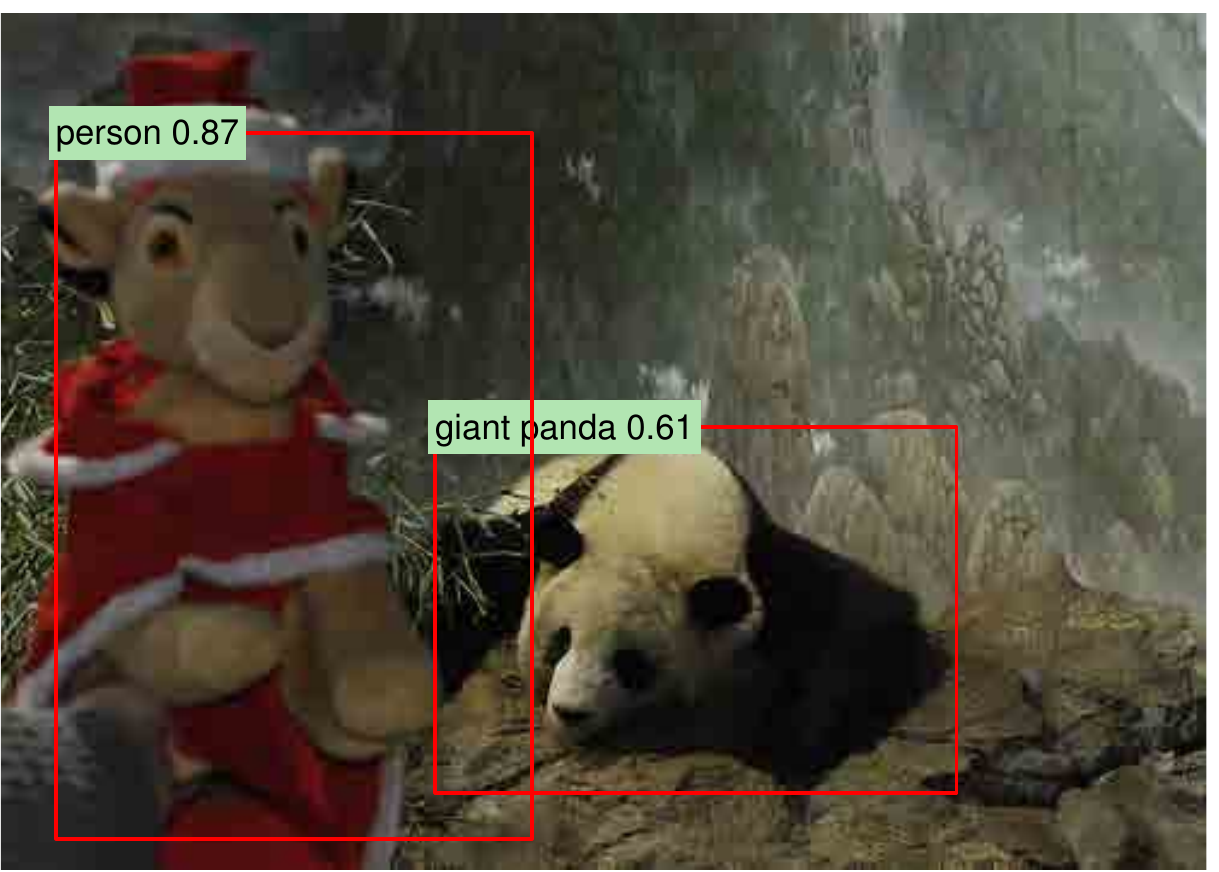}
\includegraphics[height=\sz]{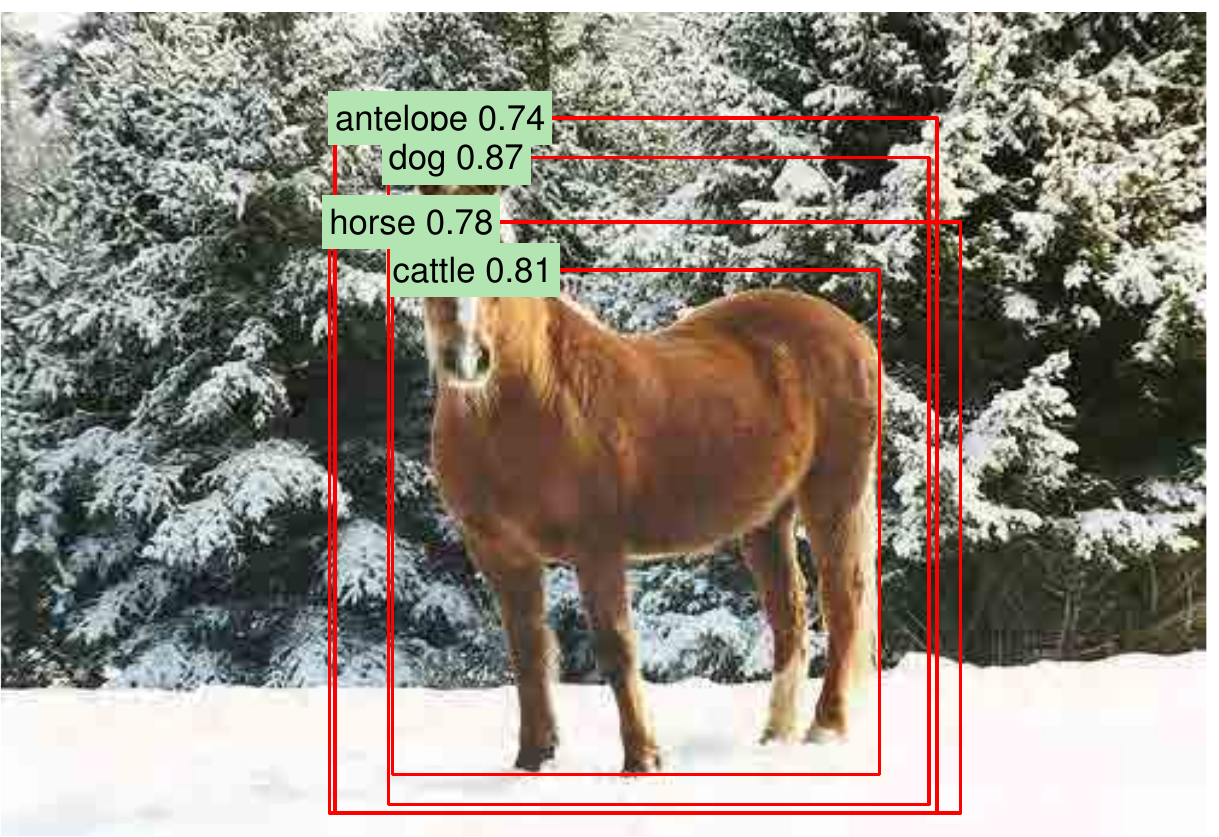}
\includegraphics[height=\sz]{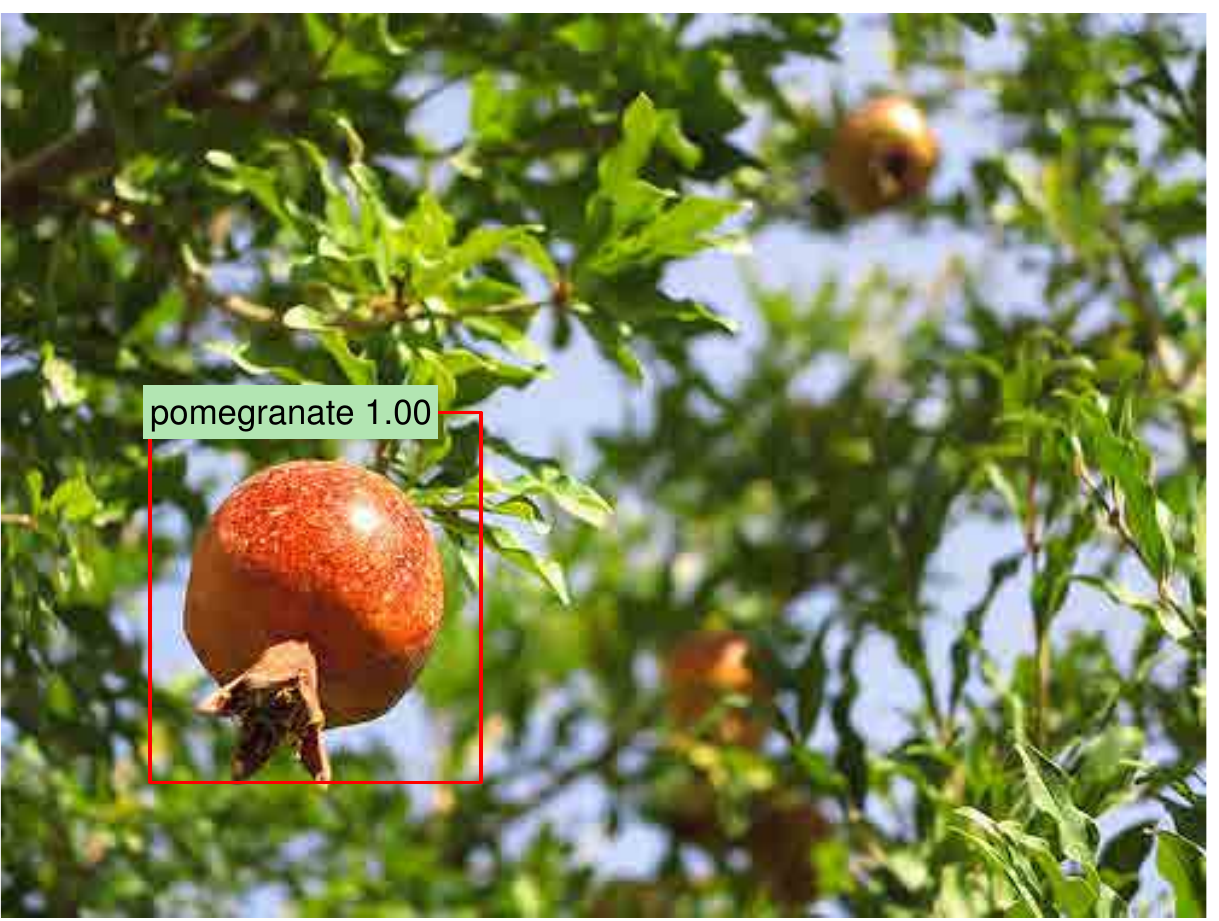}
\includegraphics[height=\sz]{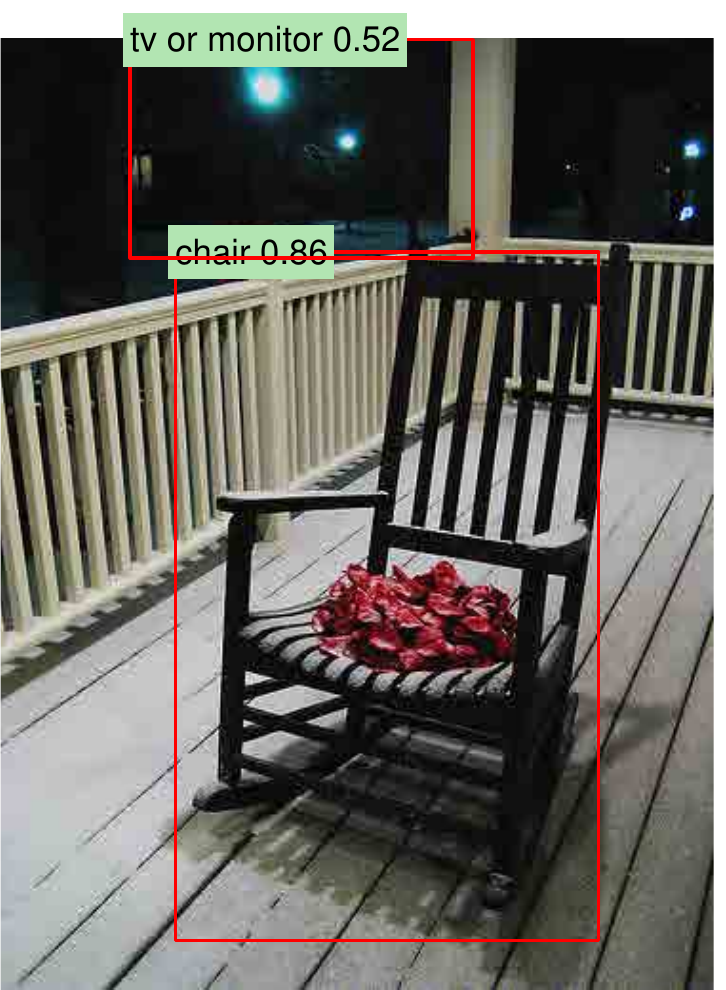}
\includegraphics[height=\sz]{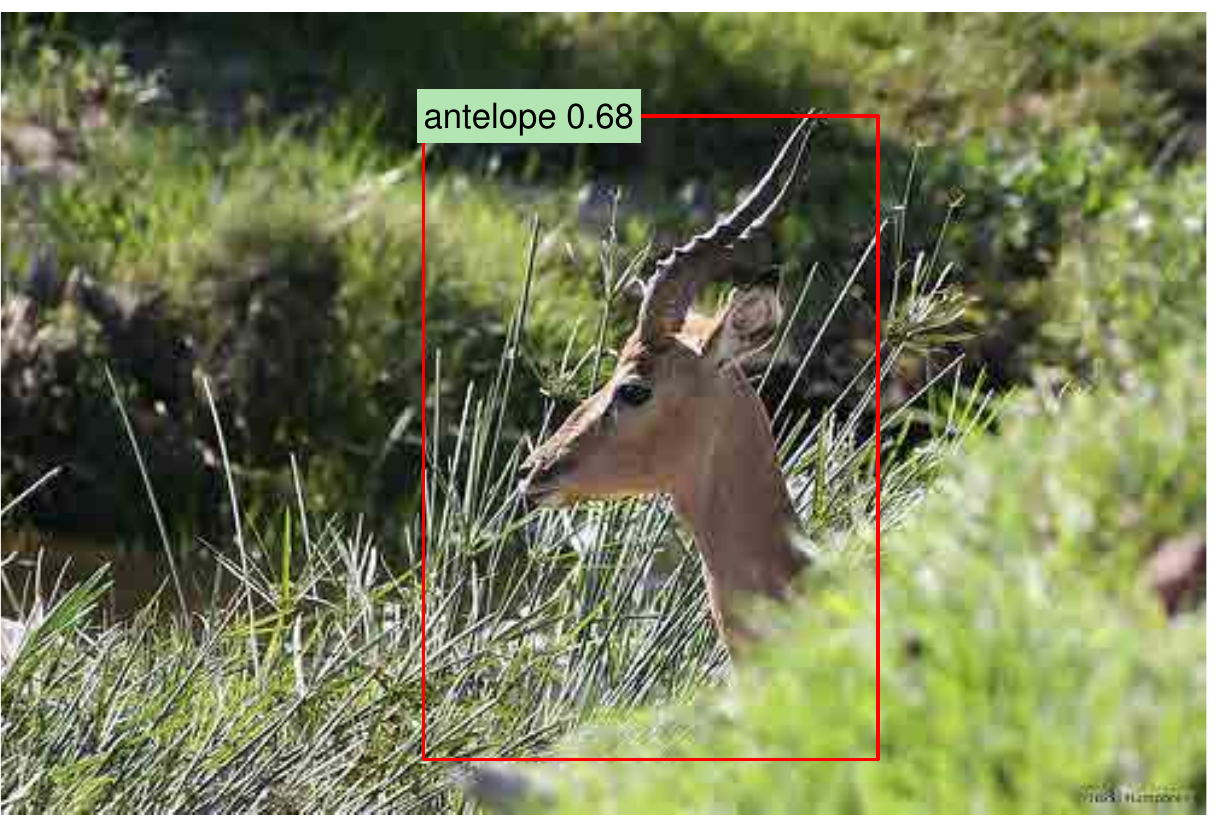}
\includegraphics[height=\sz]{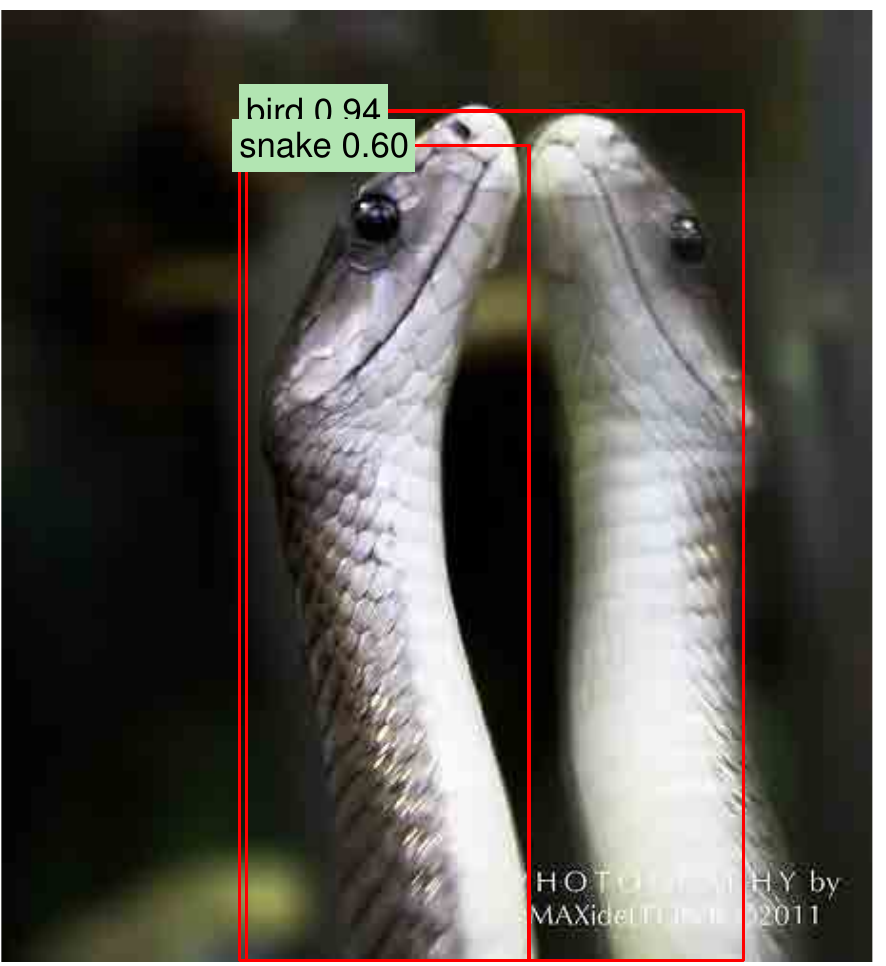}
\includegraphics[height=\sz]{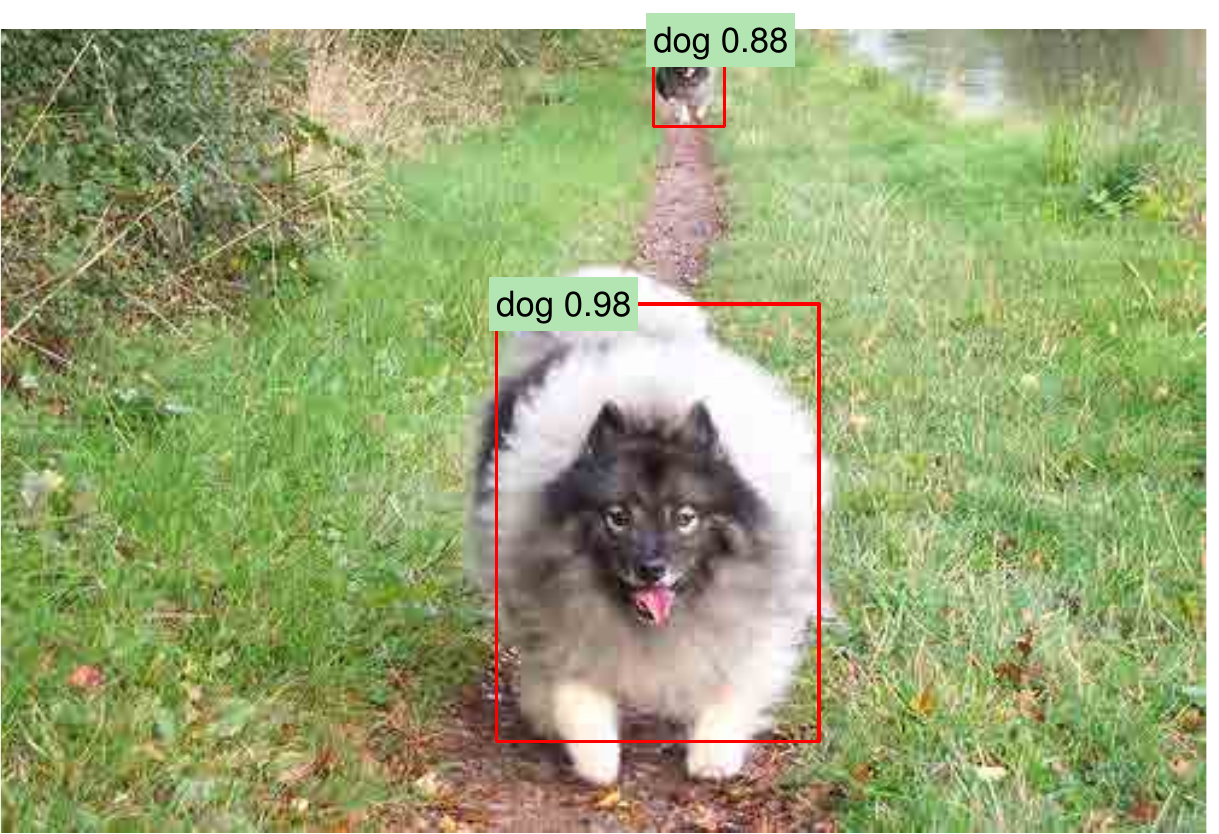}
\includegraphics[height=\sz]{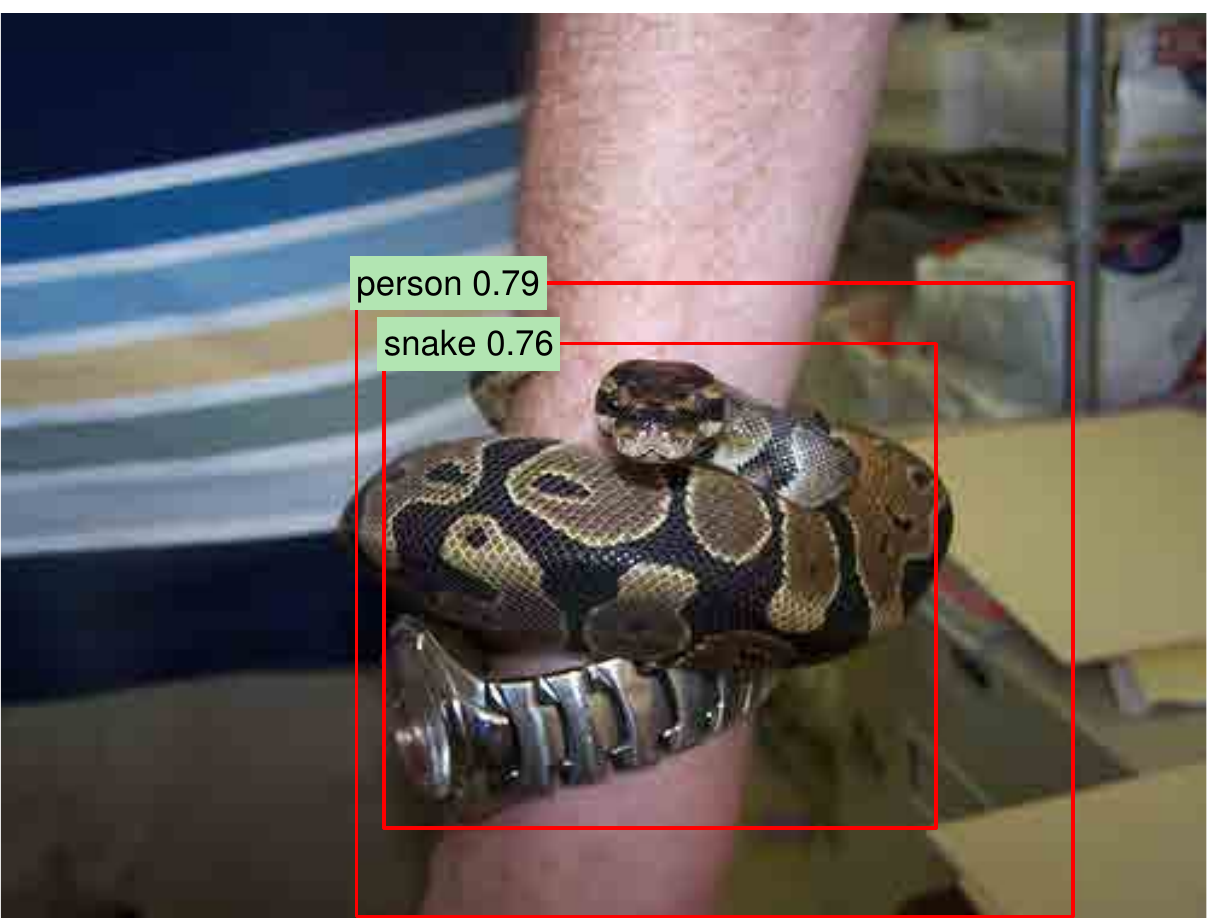}
\includegraphics[height=\sz]{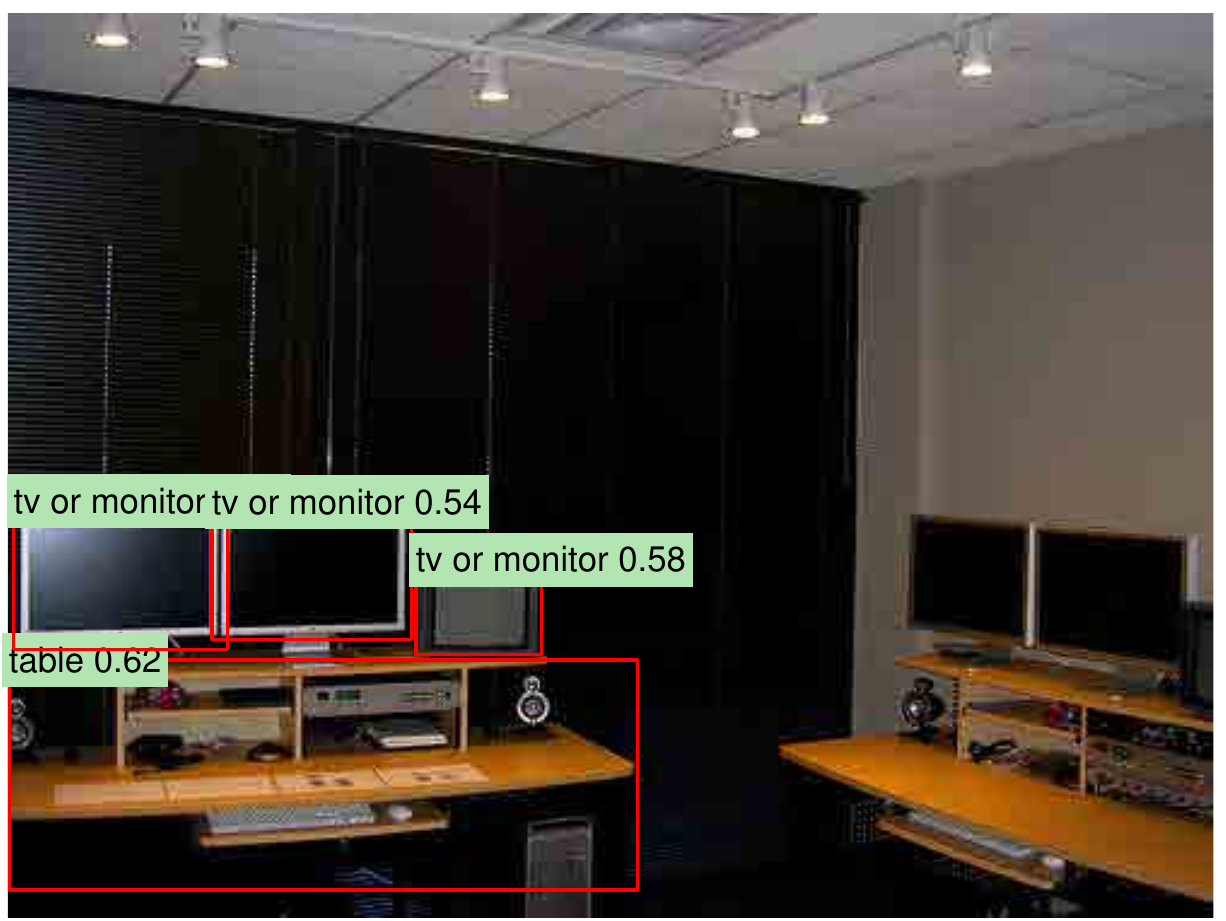}
\includegraphics[height=\sz]{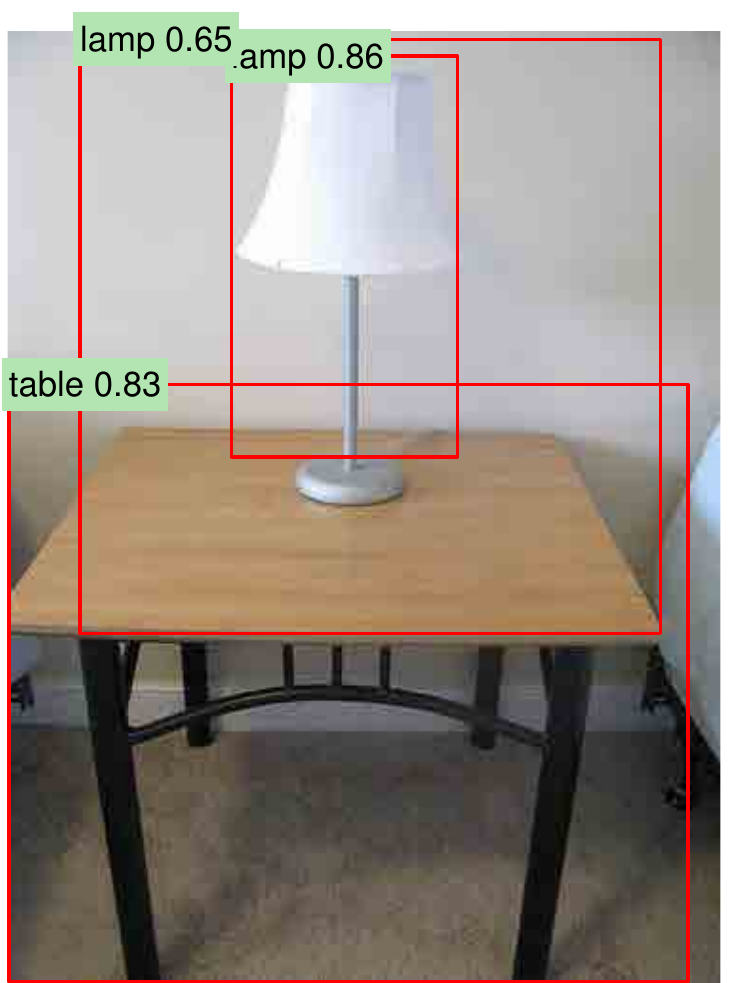}
\includegraphics[height=\sz]{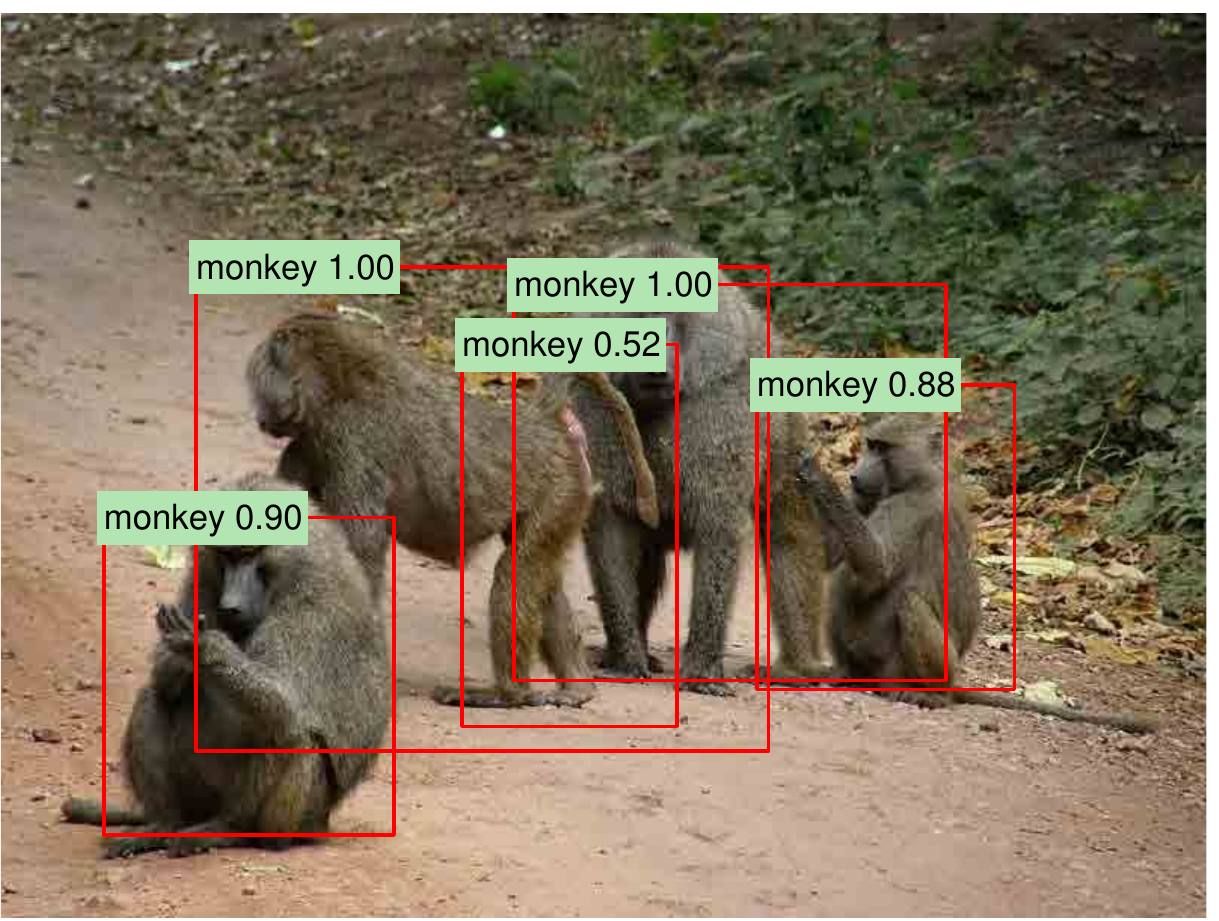}
\includegraphics[height=\sz]{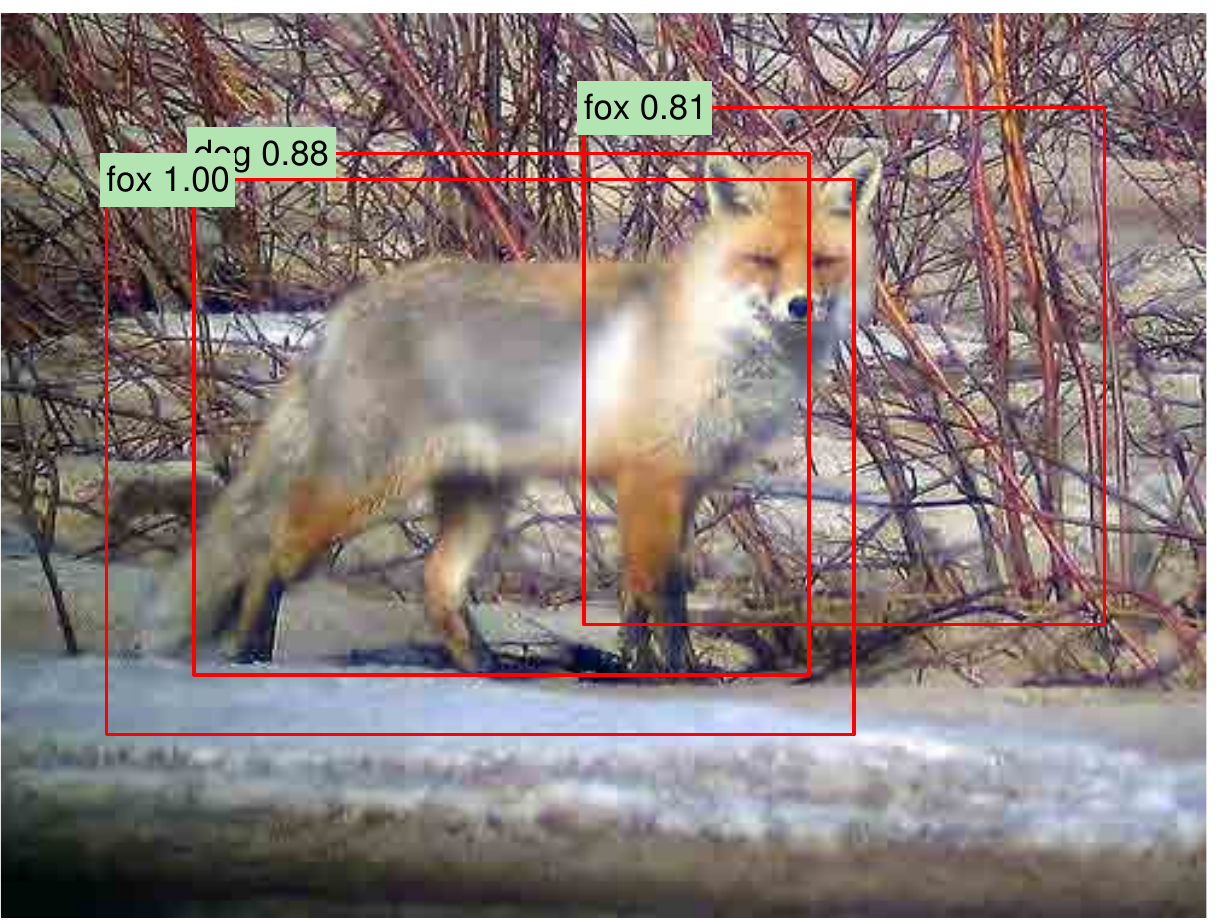}
\includegraphics[height=\sz]{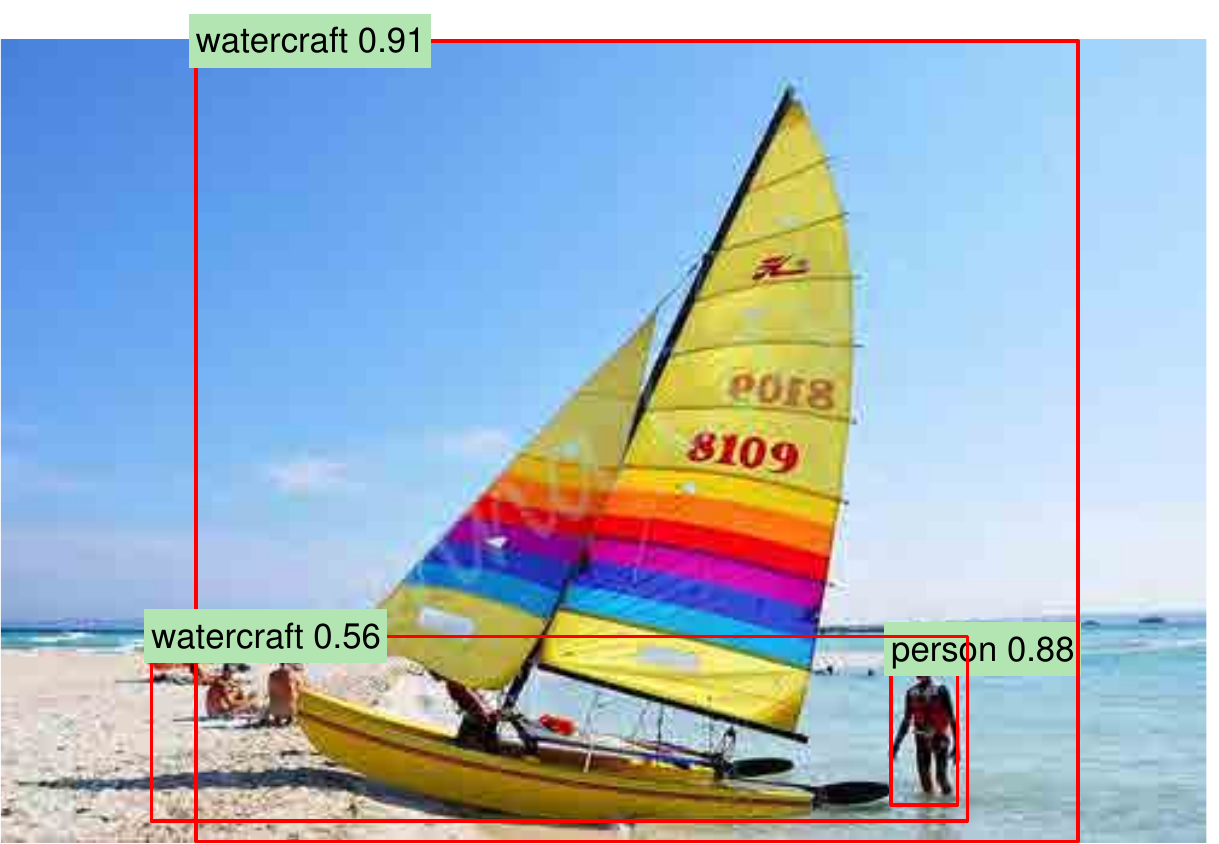}
\includegraphics[height=\sz]{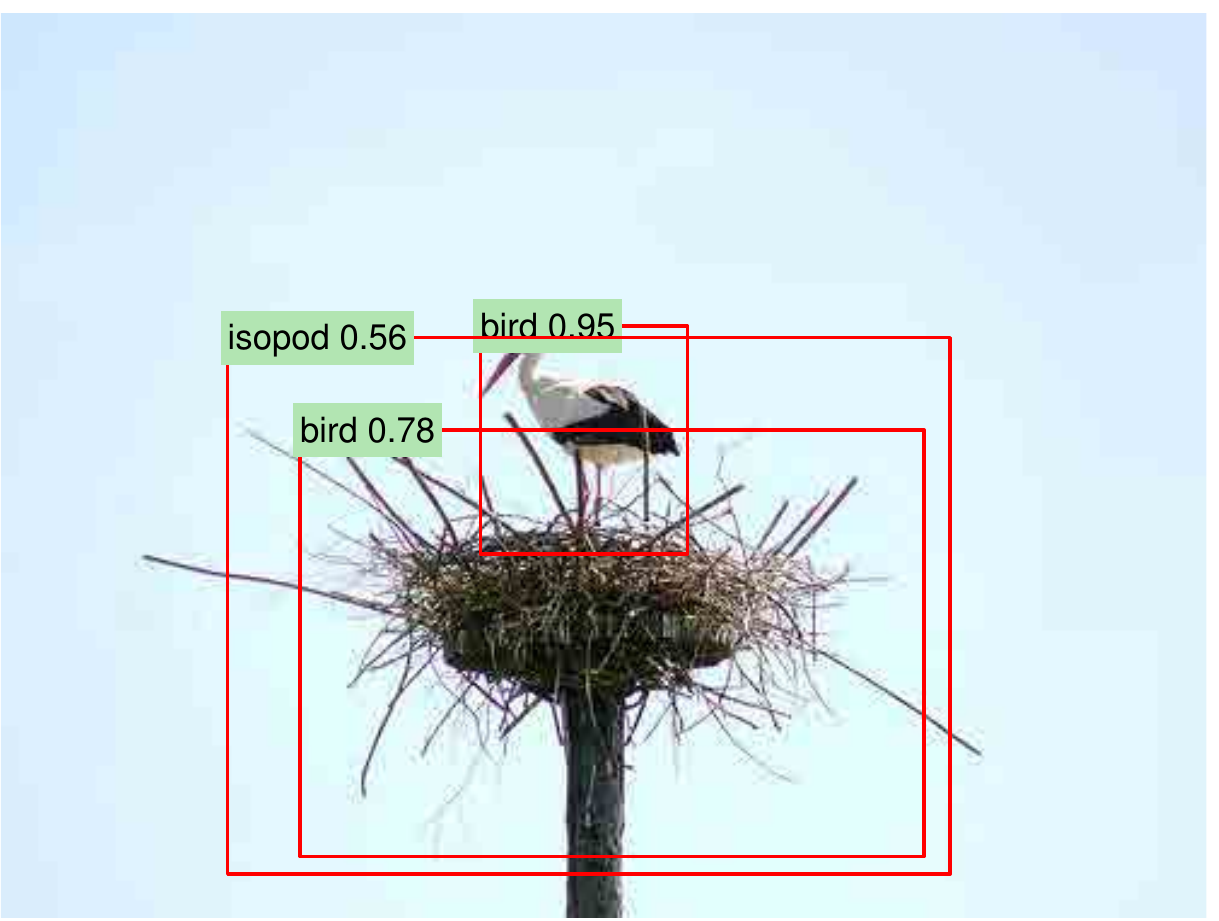}
\includegraphics[height=\sz]{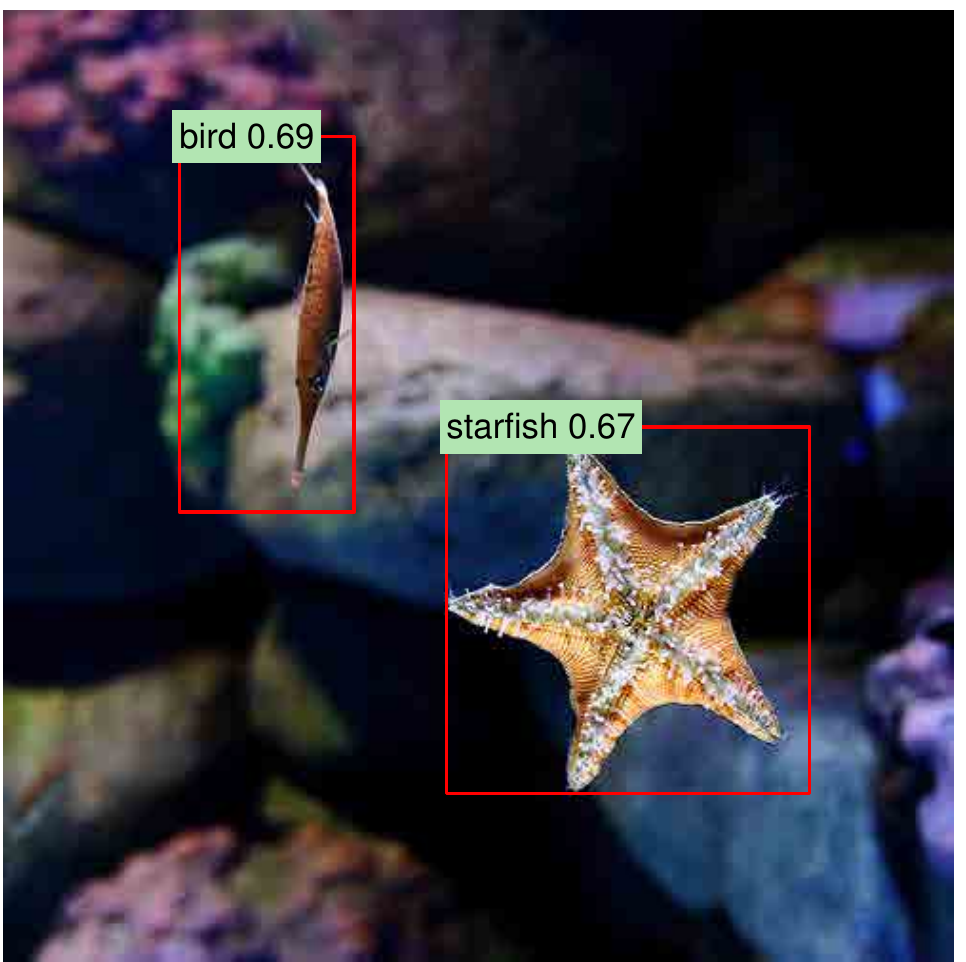}
\includegraphics[height=\sz]{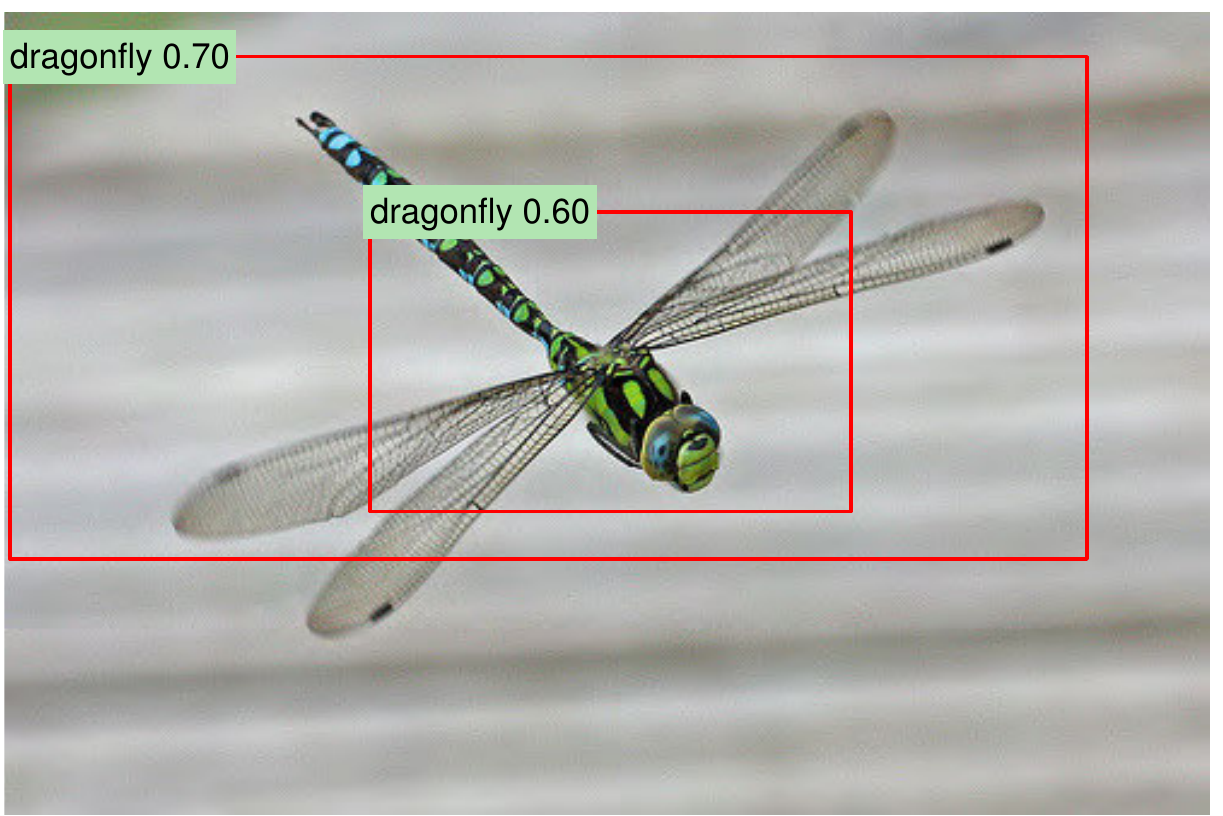}
\includegraphics[height=\sz]{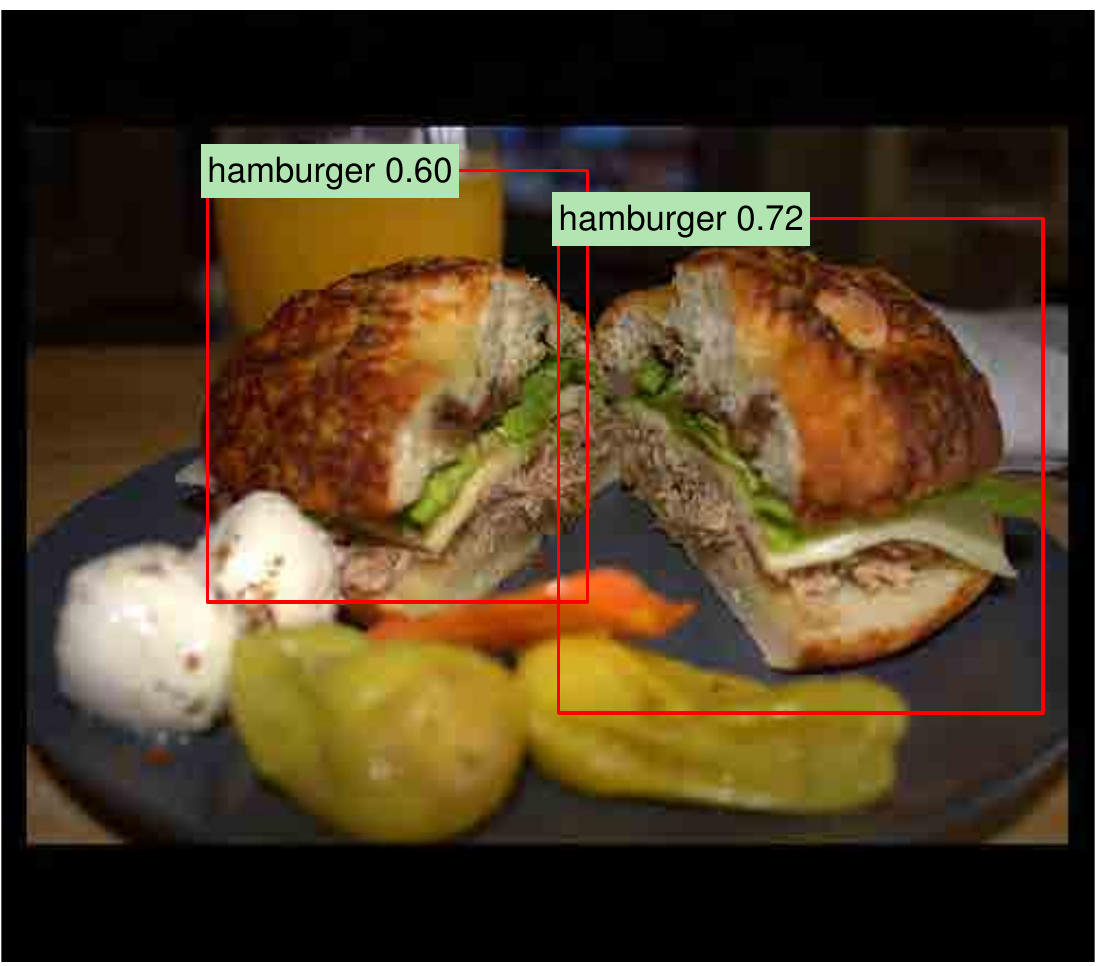}
\includegraphics[height=\sz]{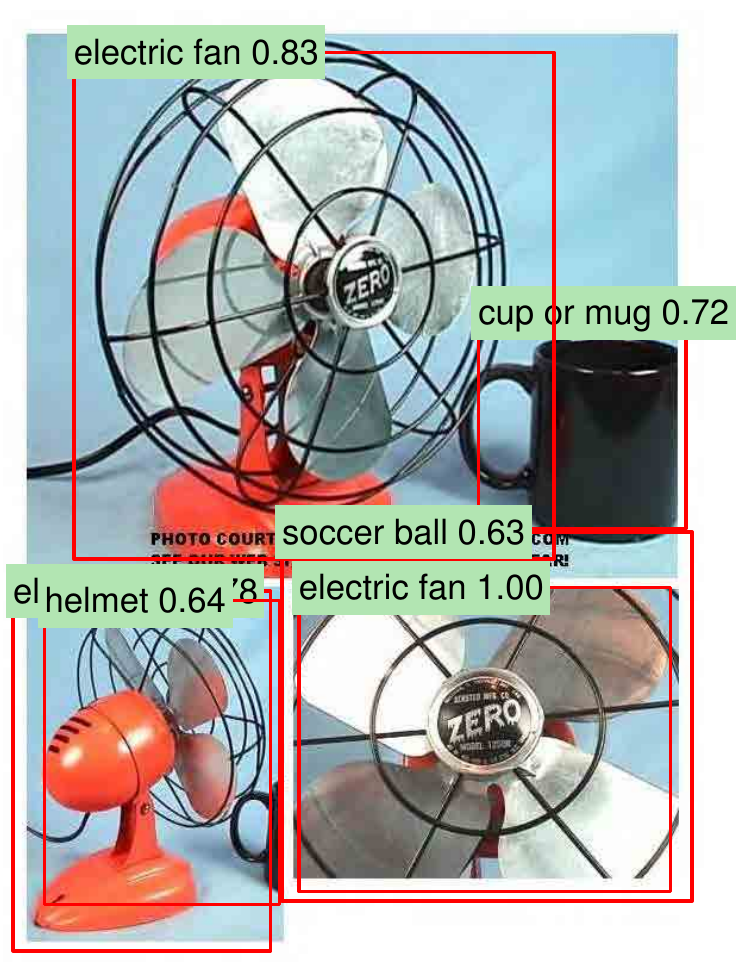}
\end{center}
\caption{Curated examples. Each image was selected because we found it impressive, surprising, interesting, or amusing.
Viewing digitally with zoom is recommended.
}
\figlabel{cexamples1}
\end{figure*}

\begin{figure*}[t]
\begin{center}
\def \sz {0.95in}
\includegraphics[height=\sz]{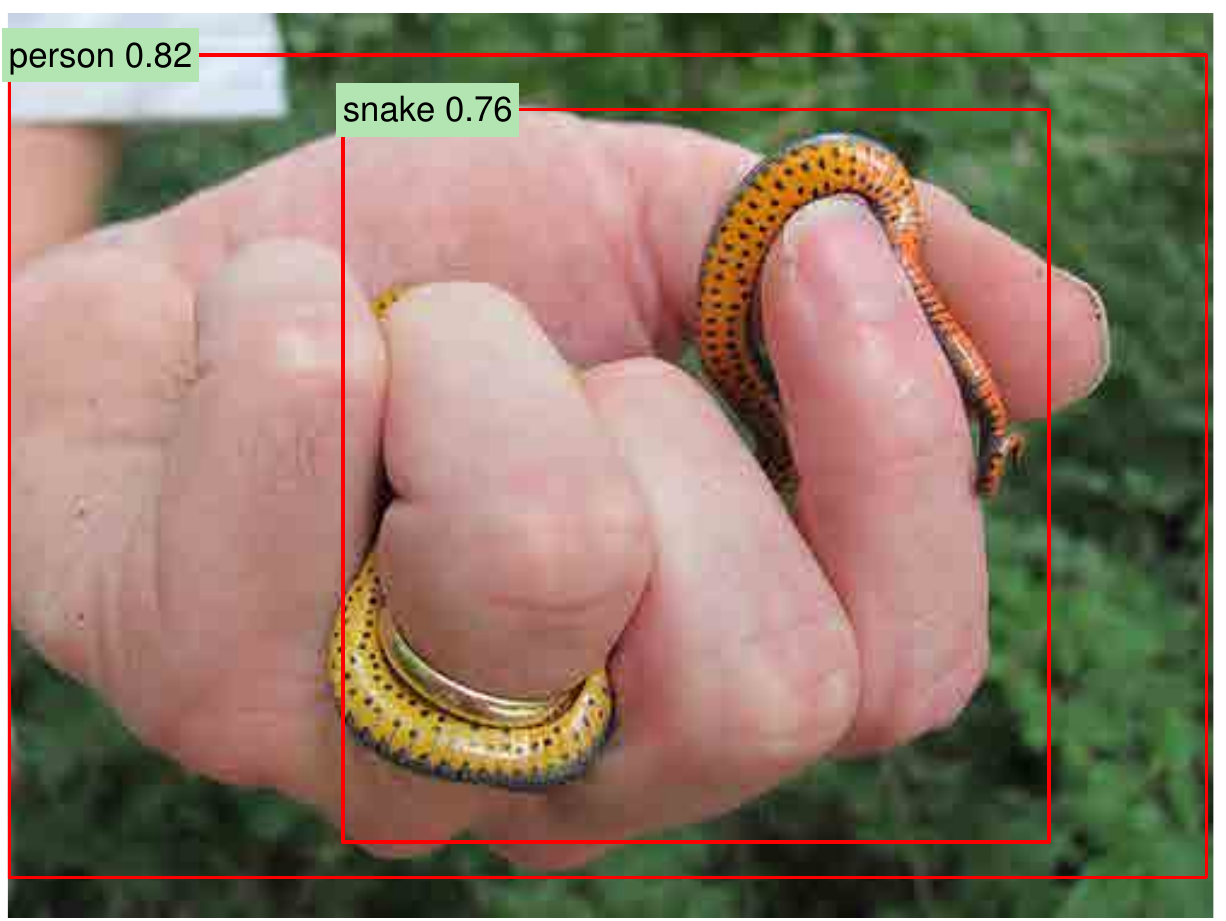}
\includegraphics[height=\sz]{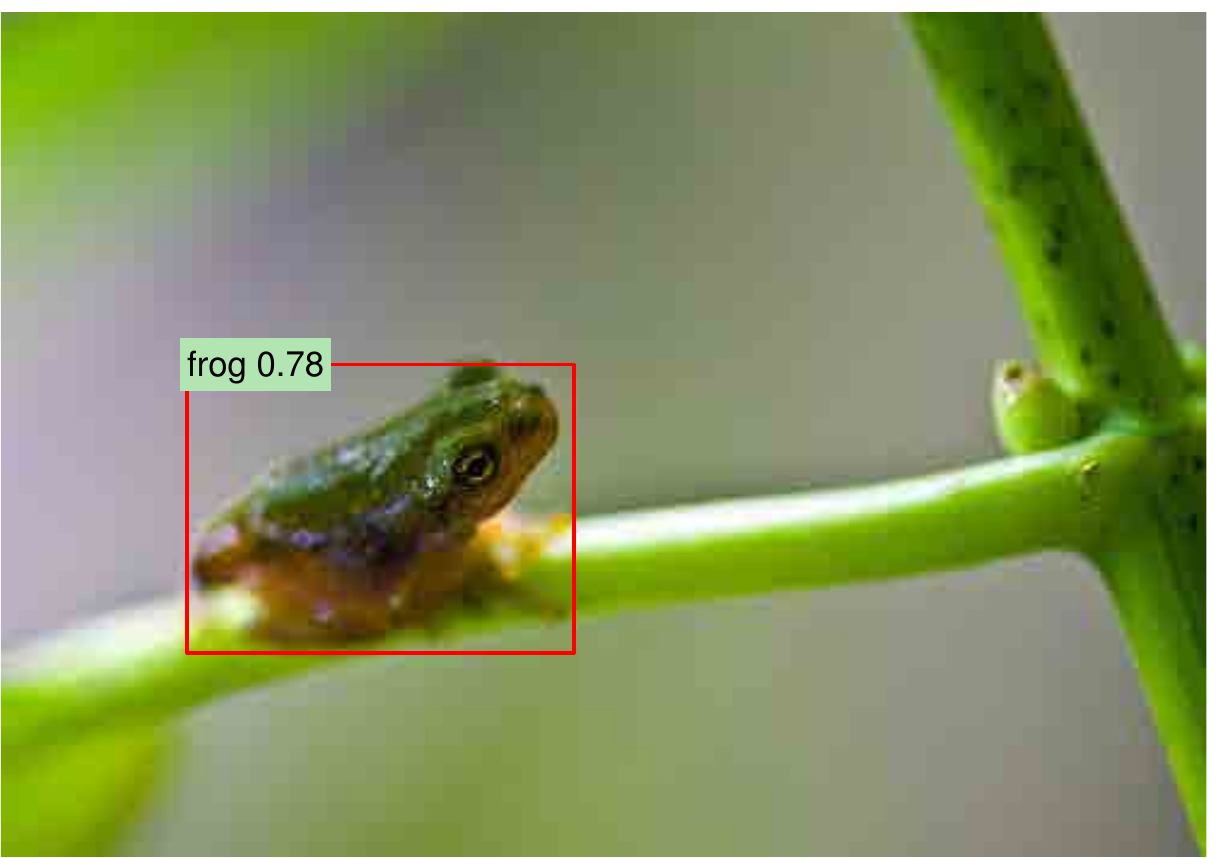}
\includegraphics[height=\sz]{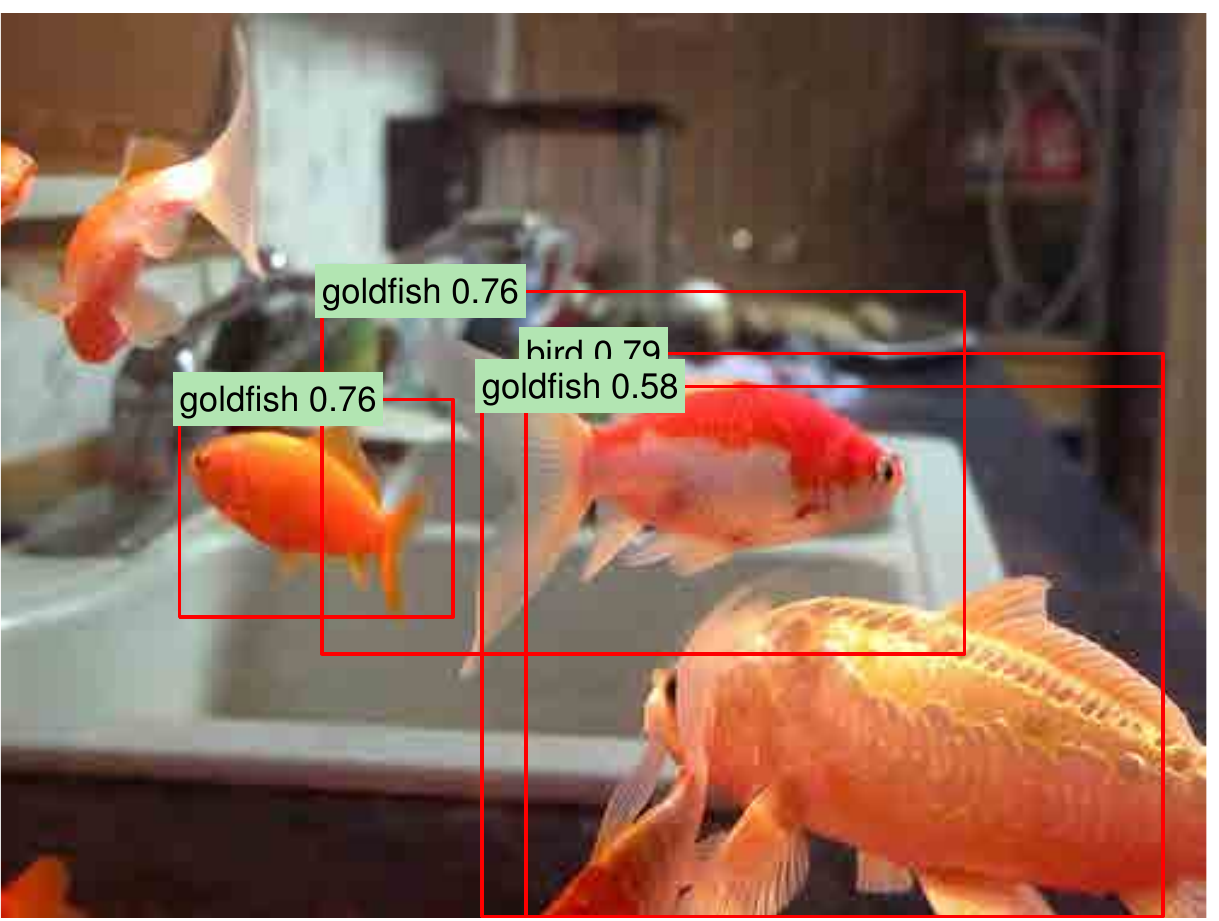}
\includegraphics[height=\sz]{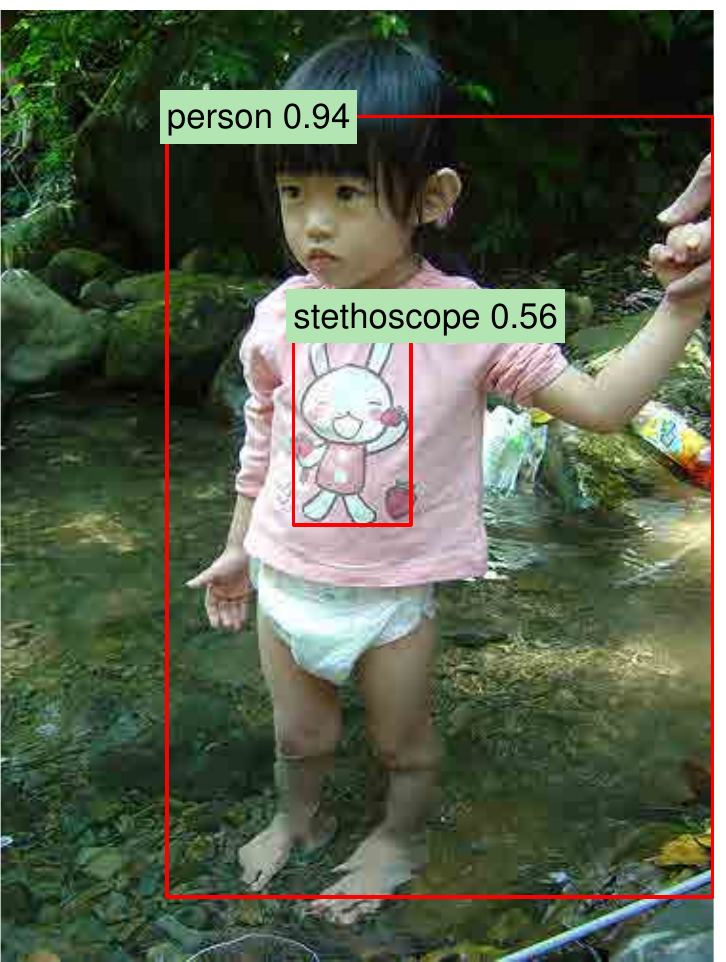}
\includegraphics[height=\sz]{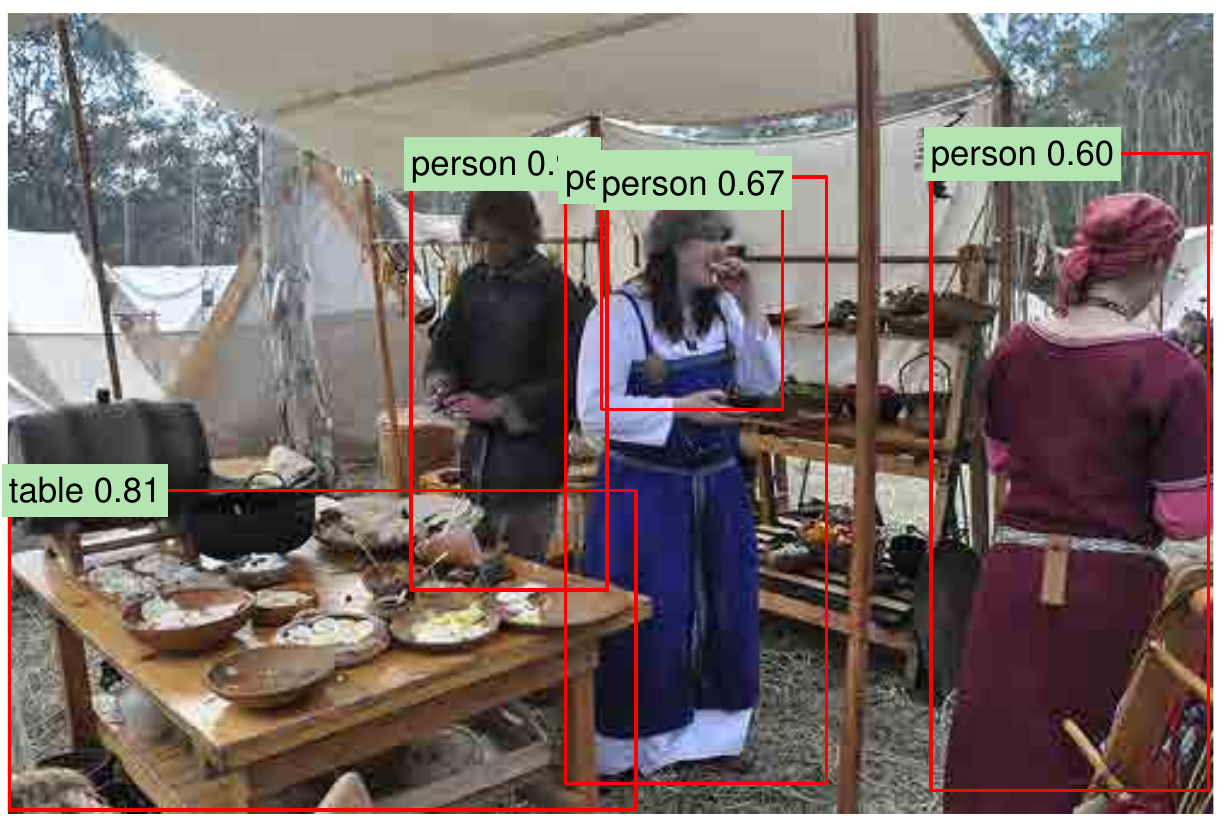}
\includegraphics[height=\sz]{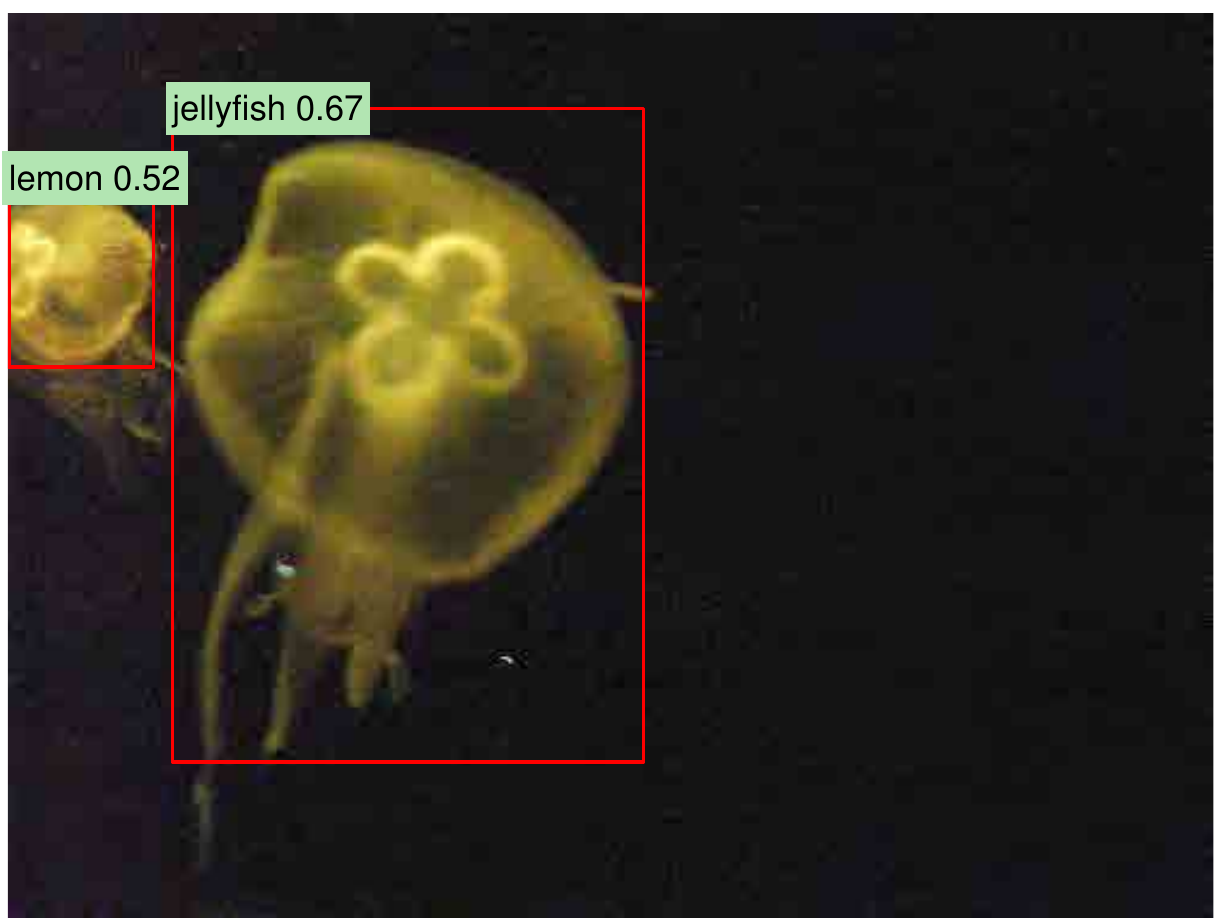}
\includegraphics[height=\sz]{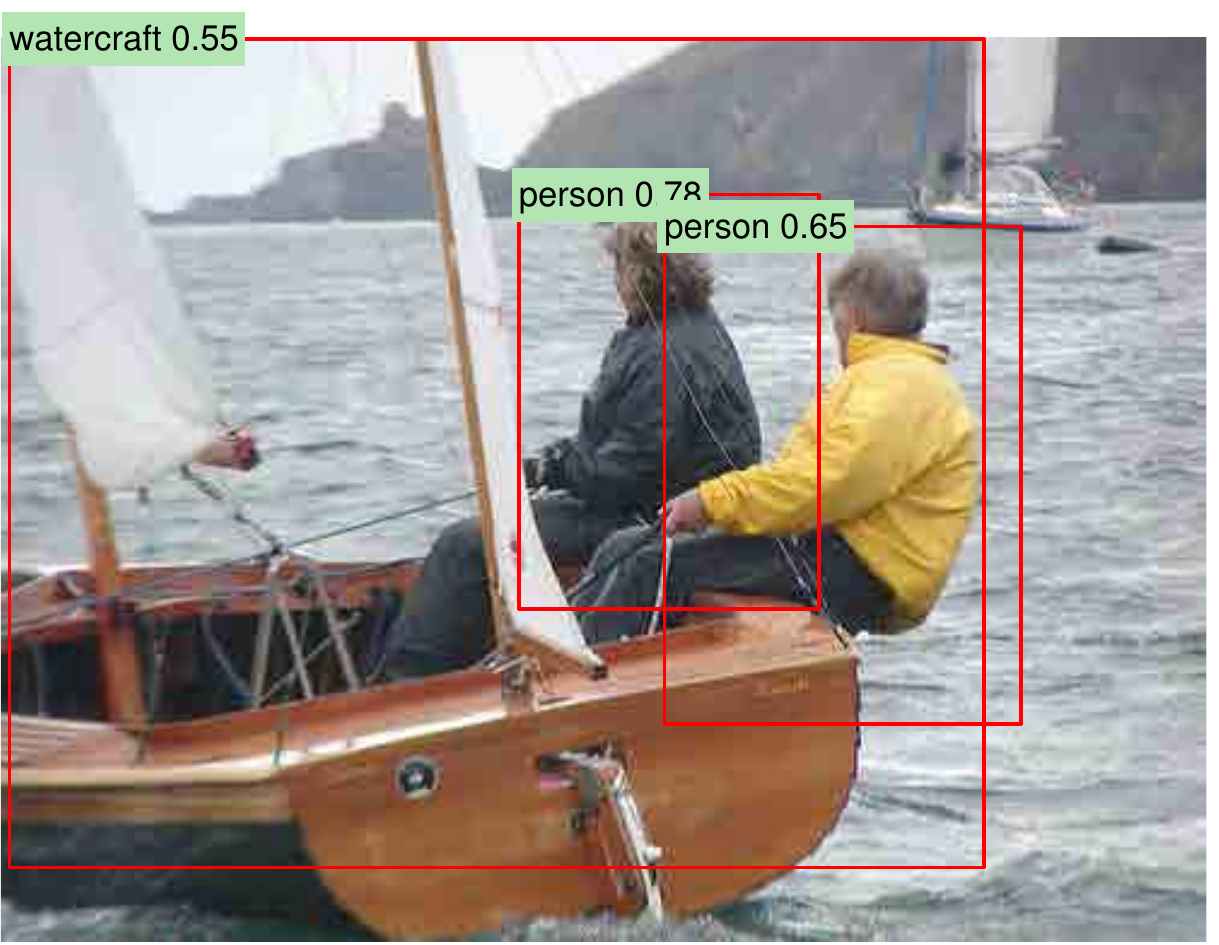}
\includegraphics[height=\sz]{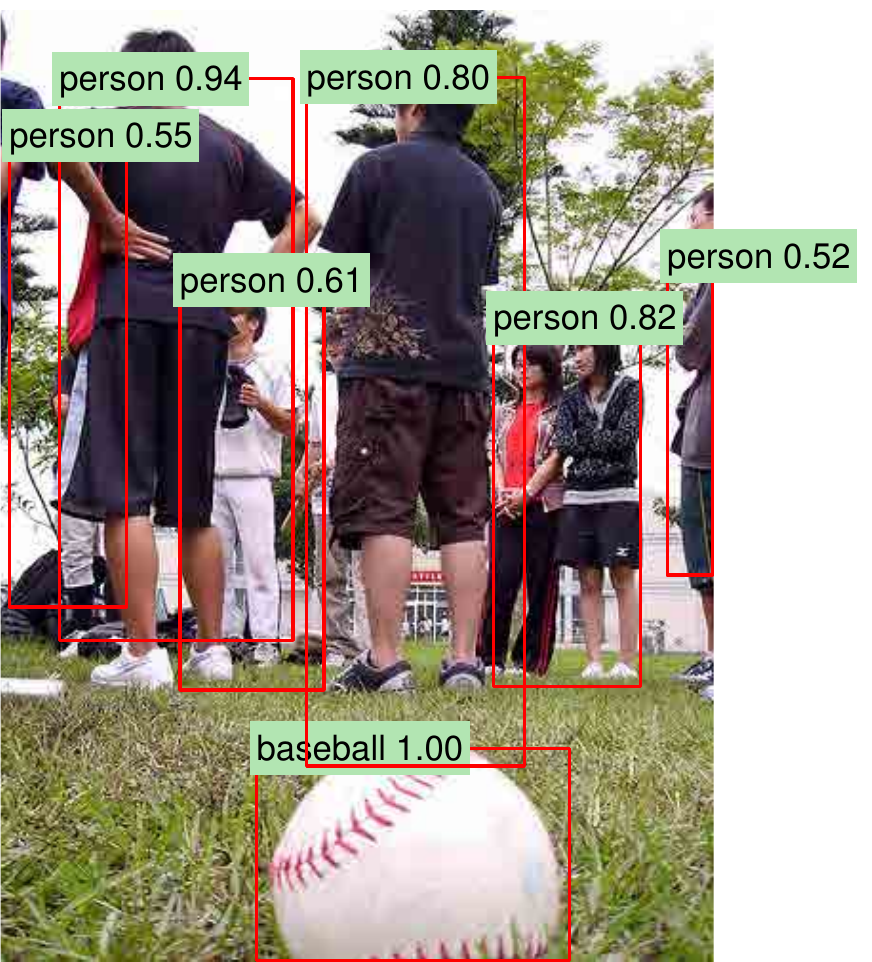}
\includegraphics[height=\sz]{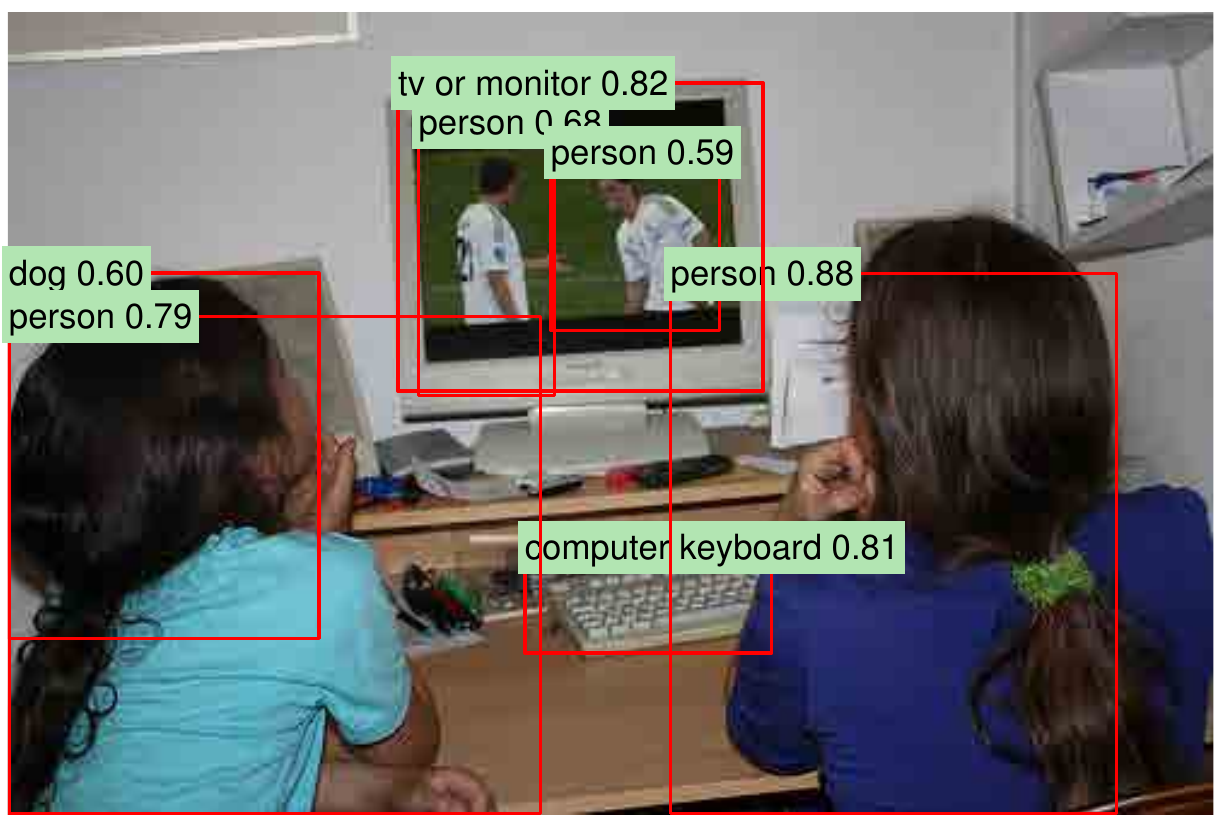}
\includegraphics[height=\sz]{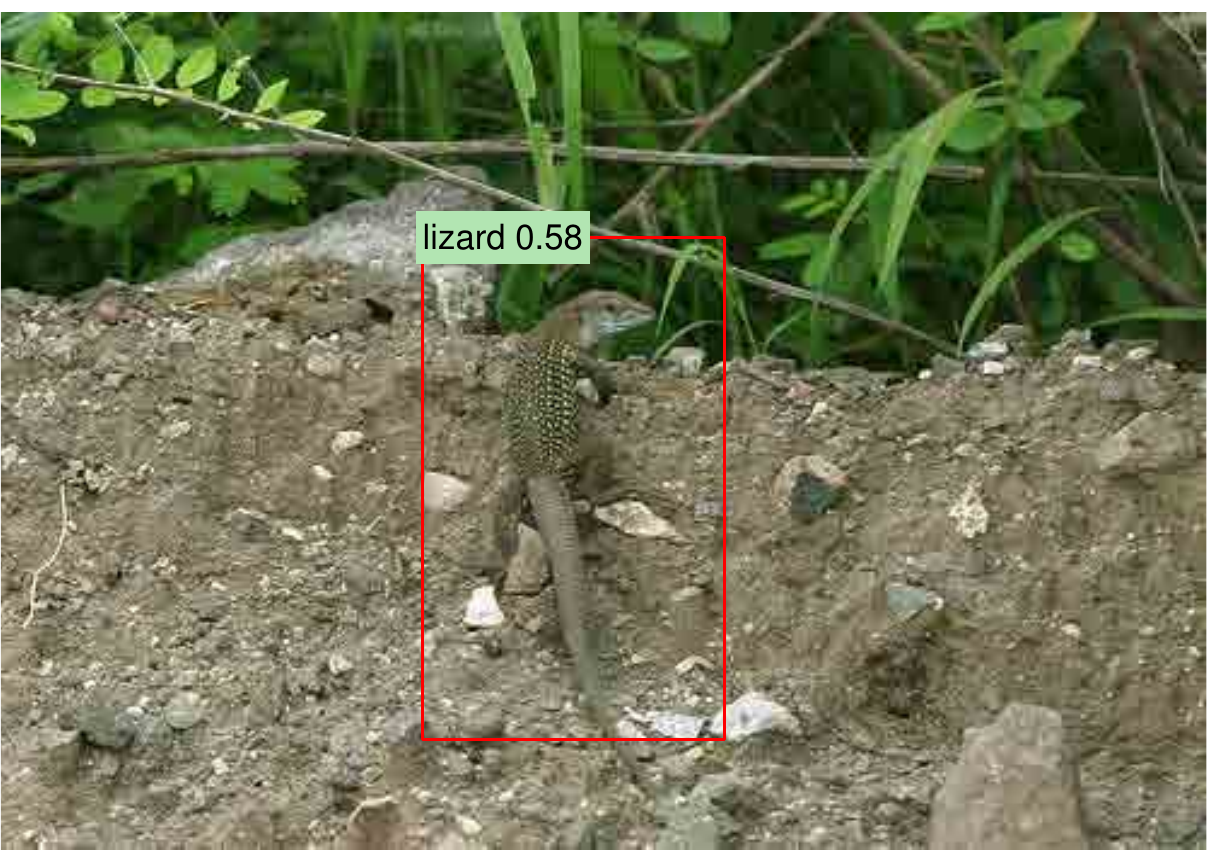}
\includegraphics[height=\sz]{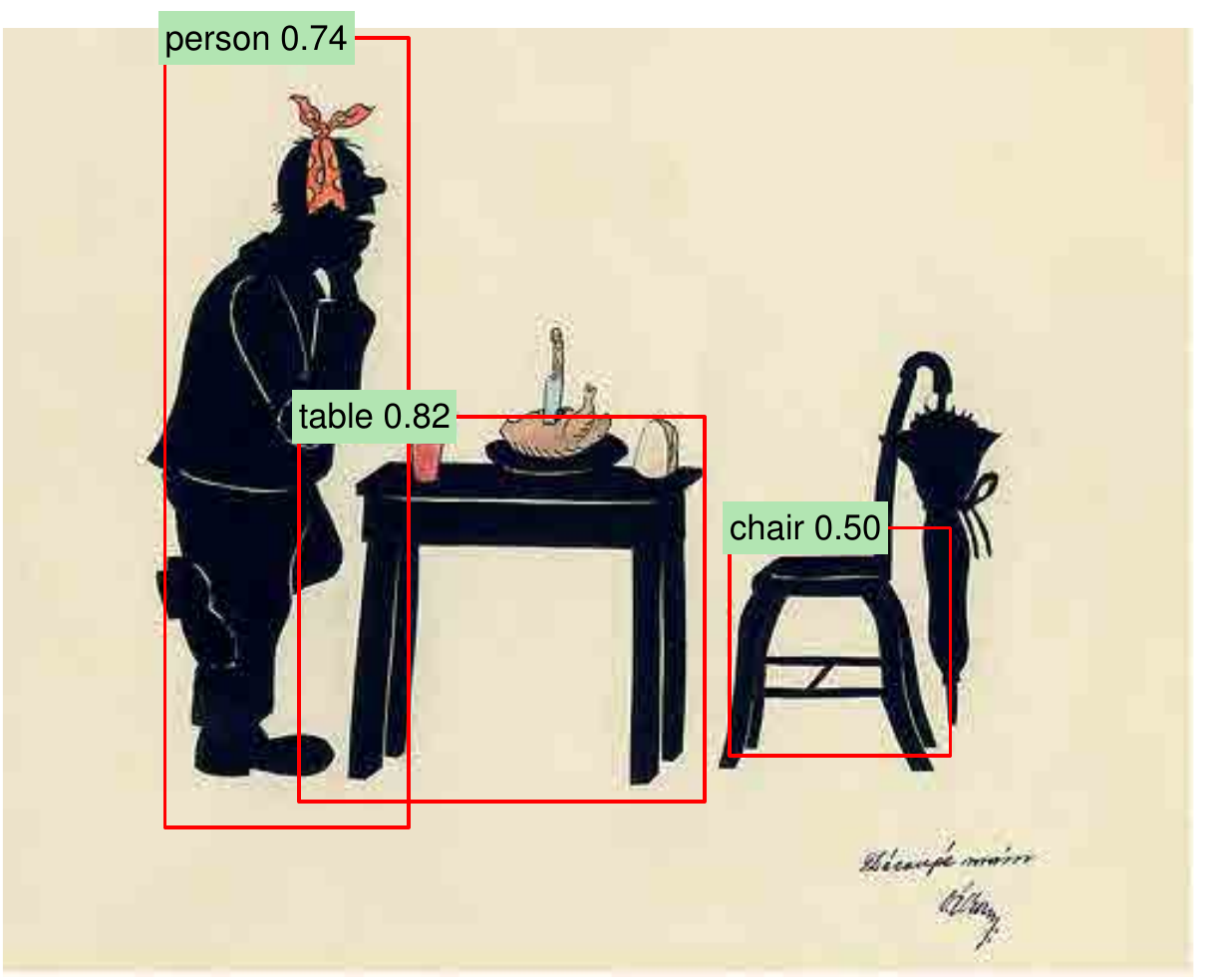}
\includegraphics[height=\sz]{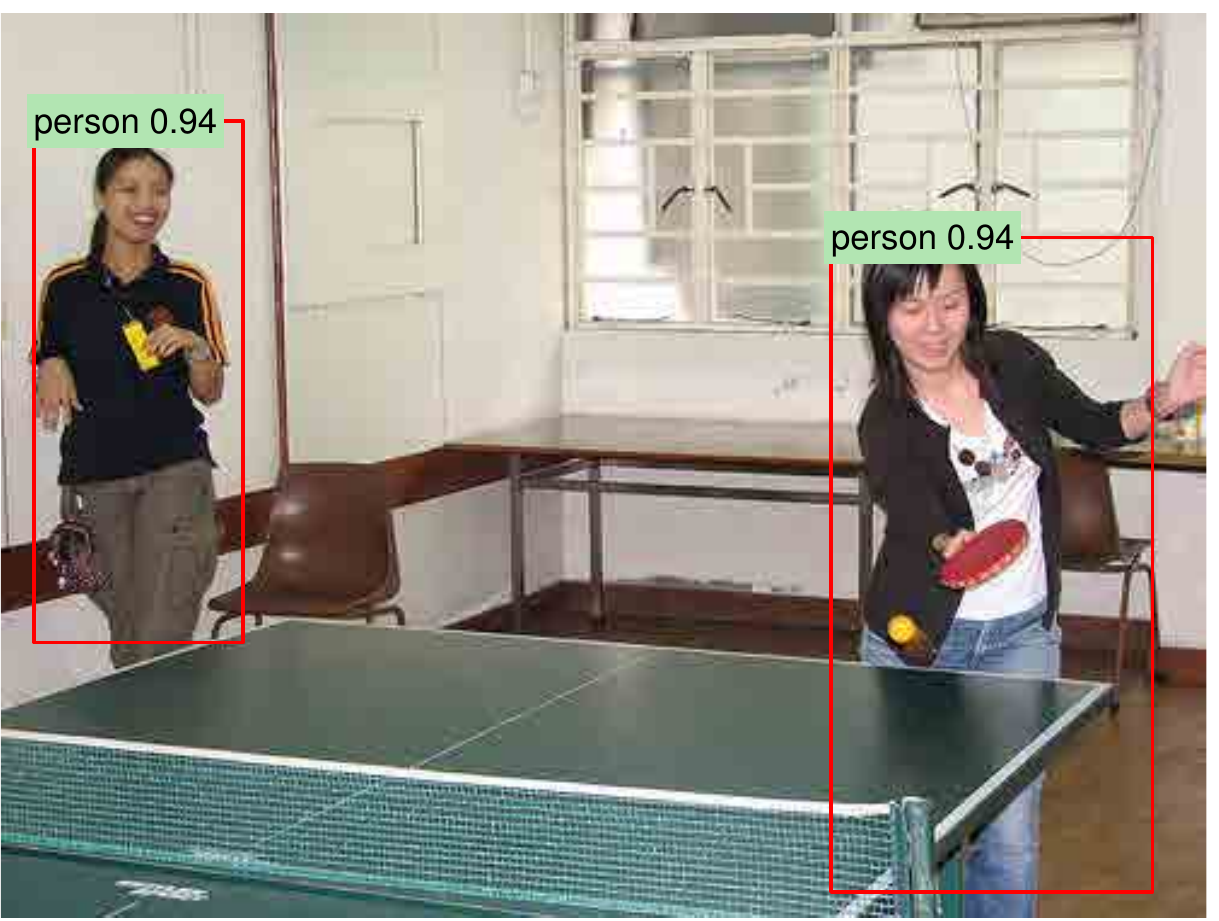}
\includegraphics[height=\sz]{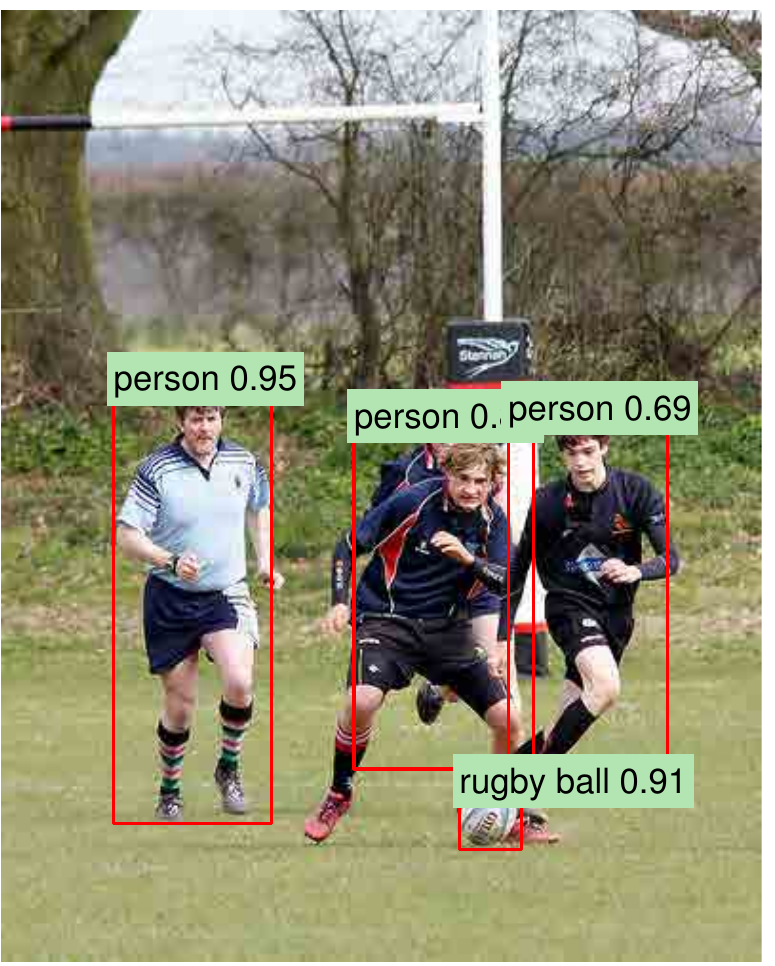}
\includegraphics[height=\sz]{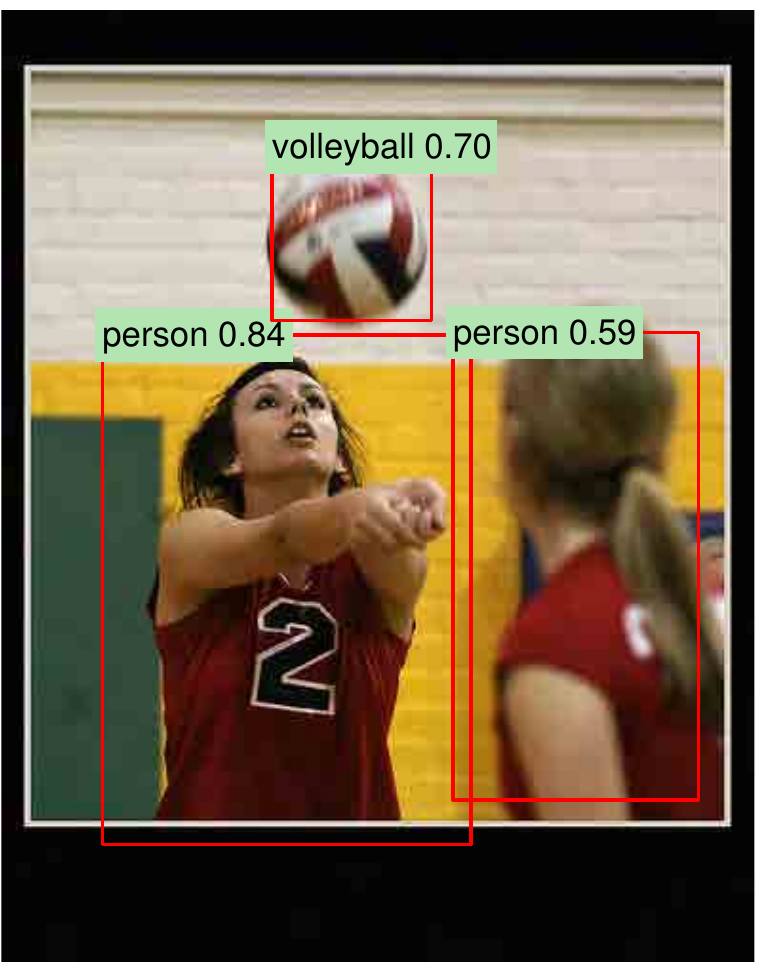}
\includegraphics[height=\sz]{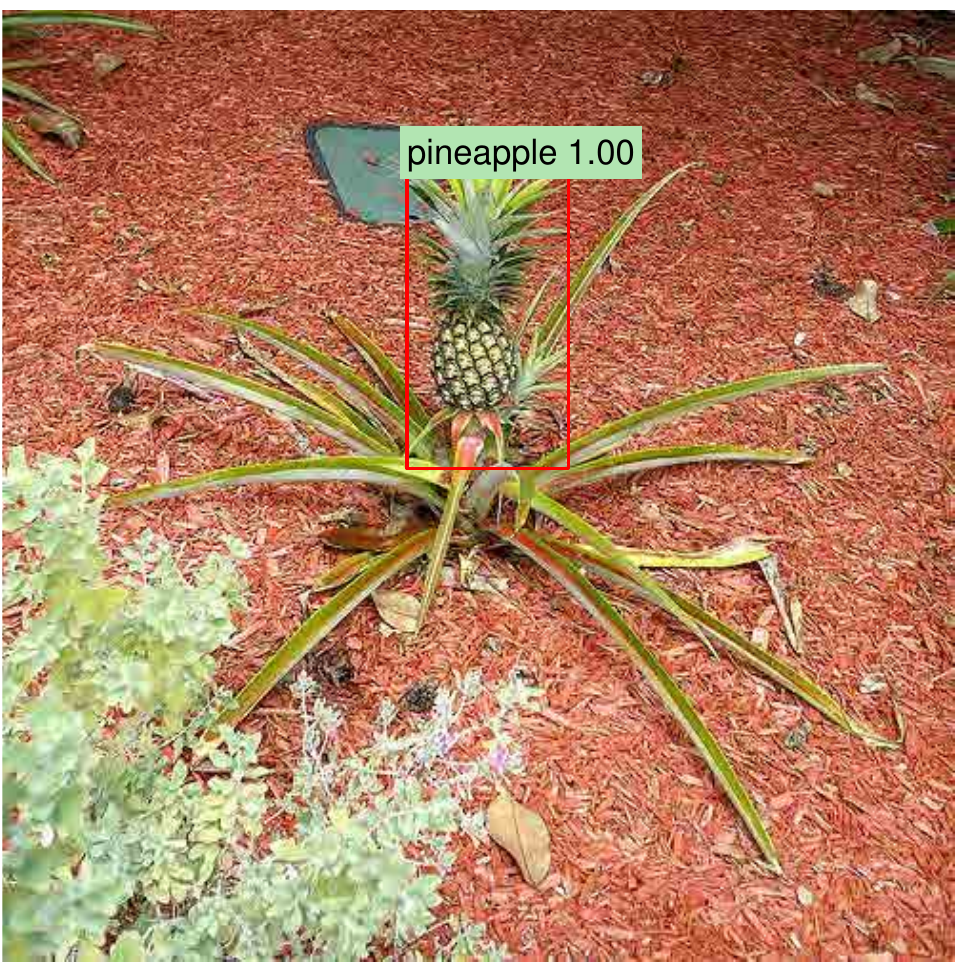}
\includegraphics[height=\sz]{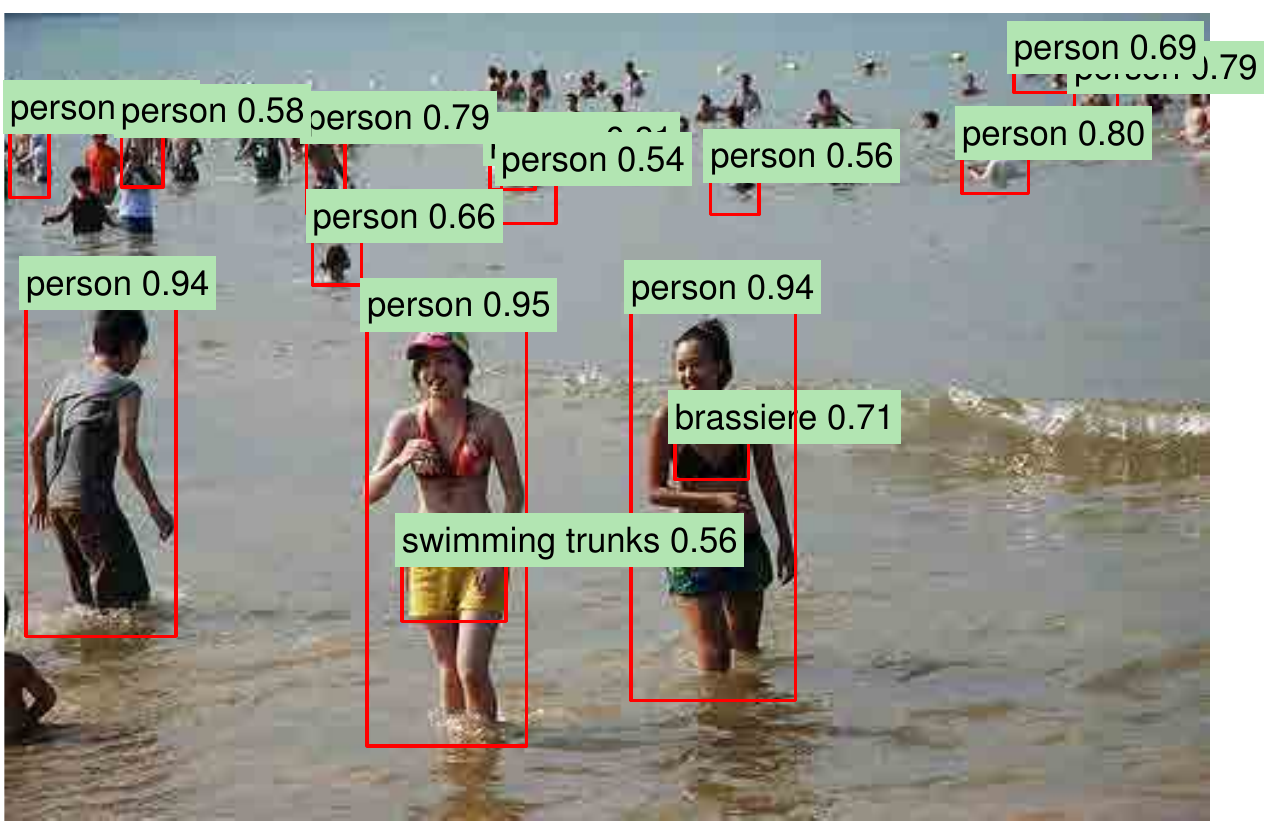}
\includegraphics[height=\sz]{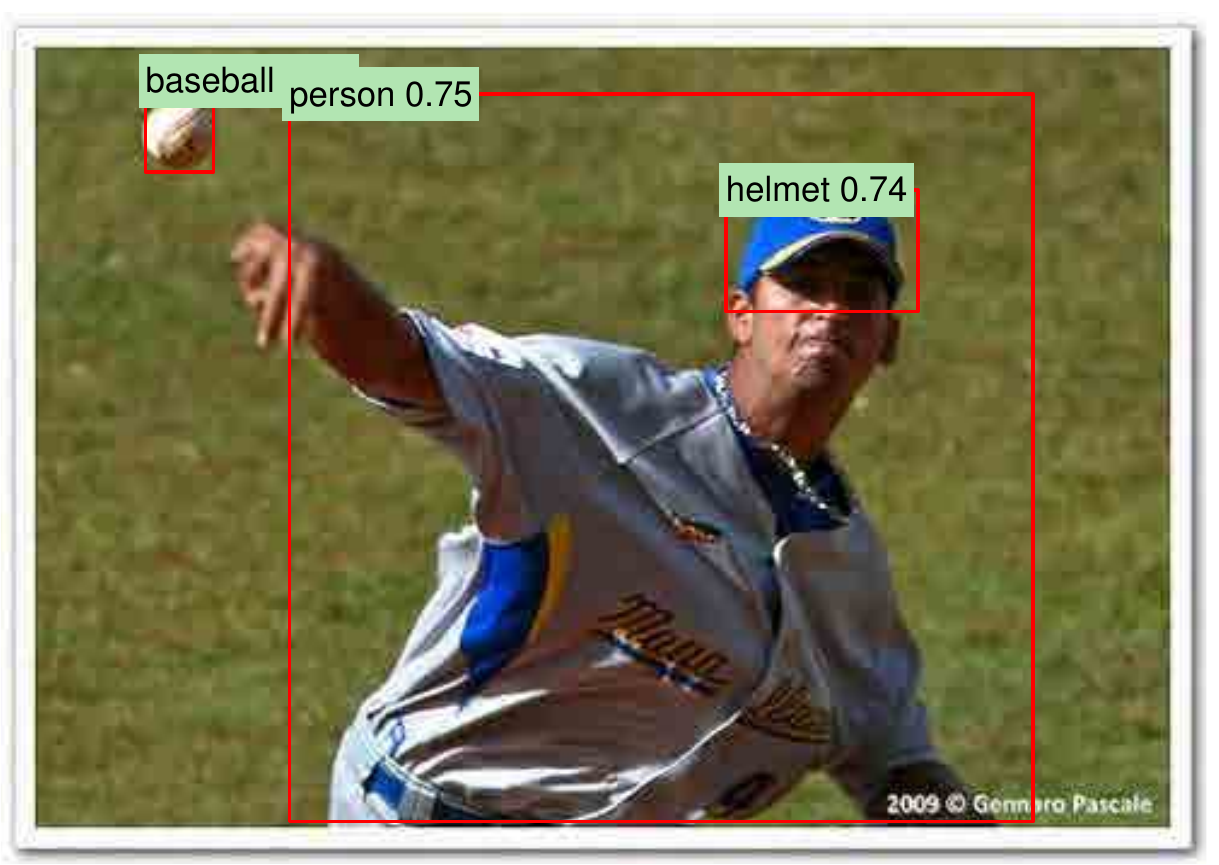}
\includegraphics[height=\sz]{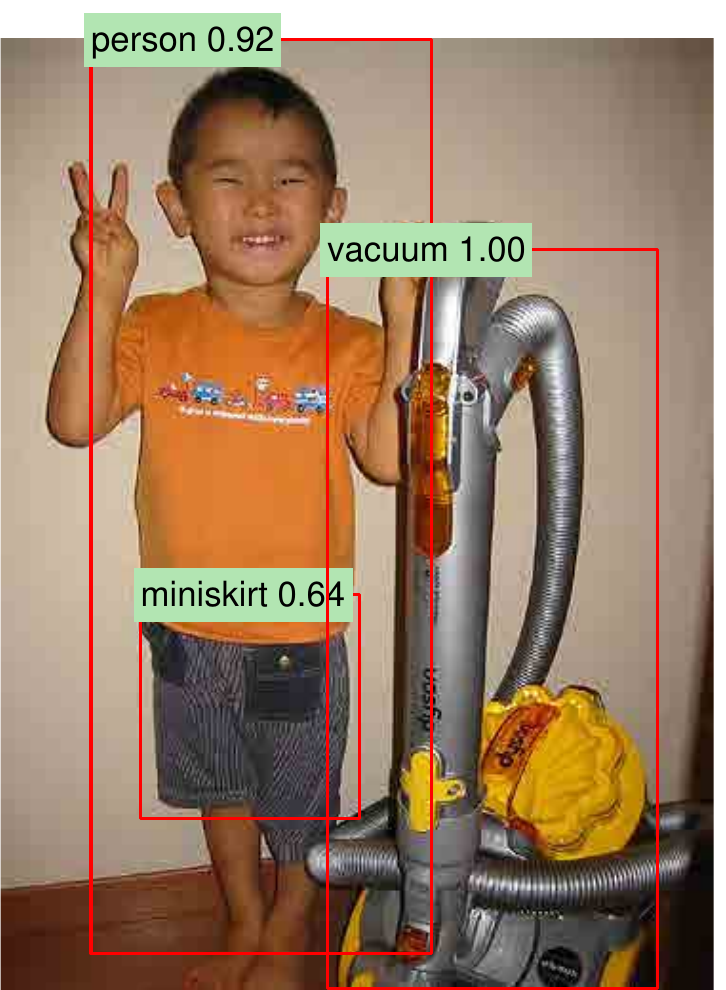}
\includegraphics[height=\sz]{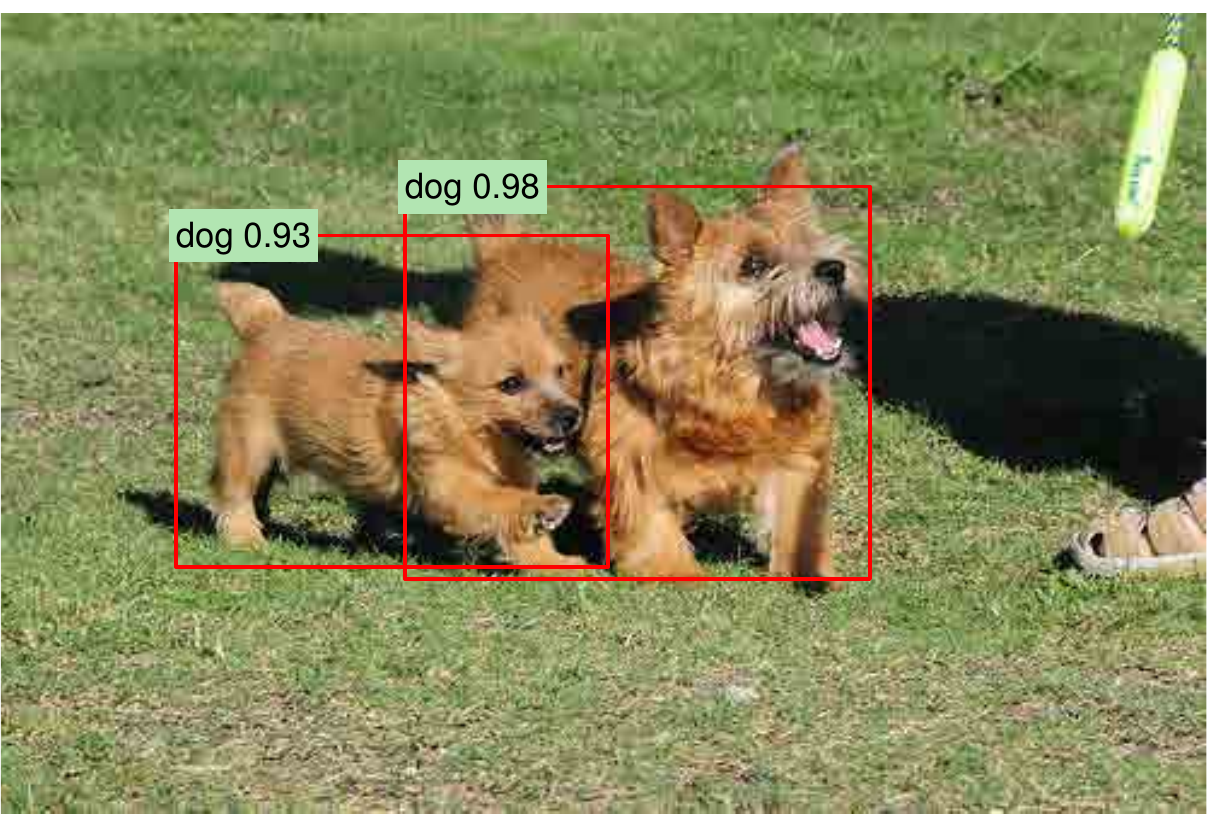}
\includegraphics[height=\sz]{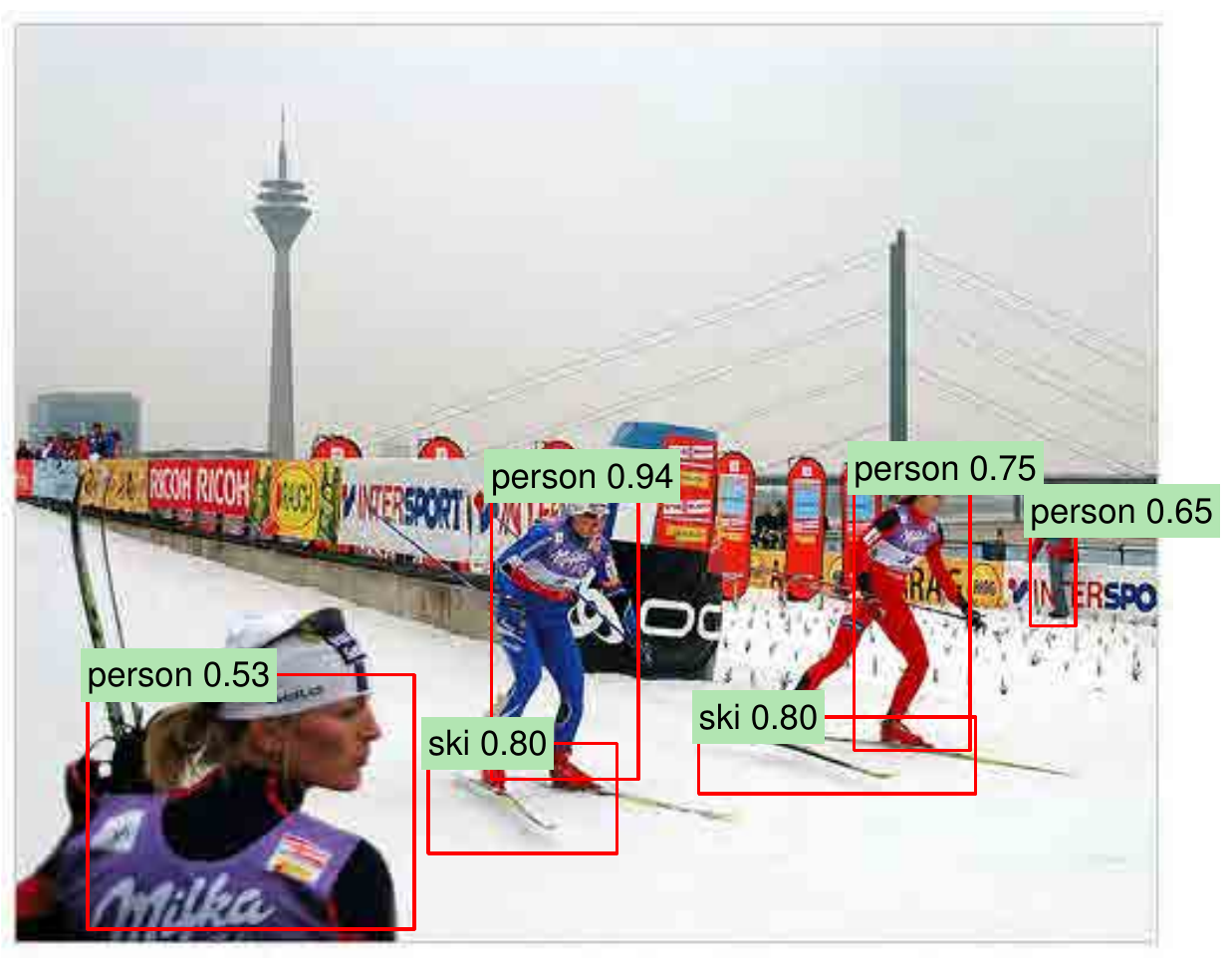}
\includegraphics[height=\sz]{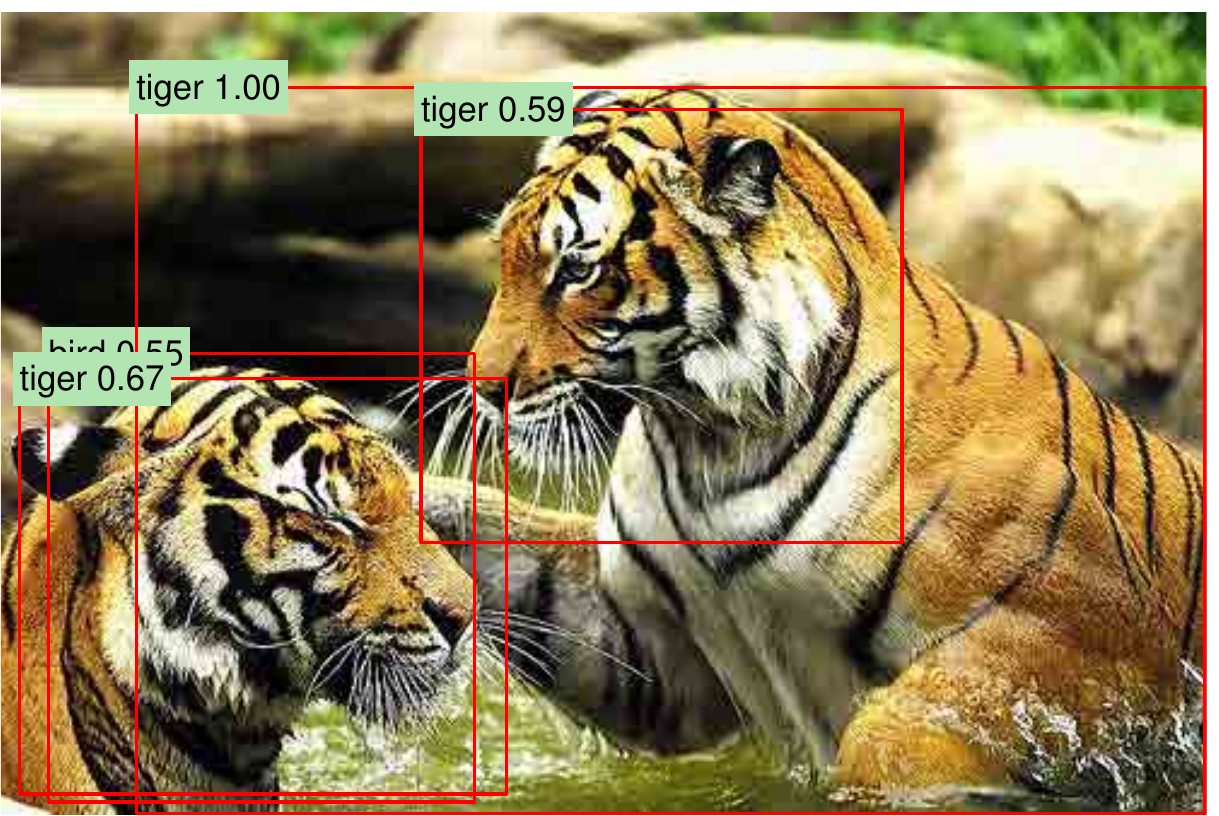}
\includegraphics[height=\sz]{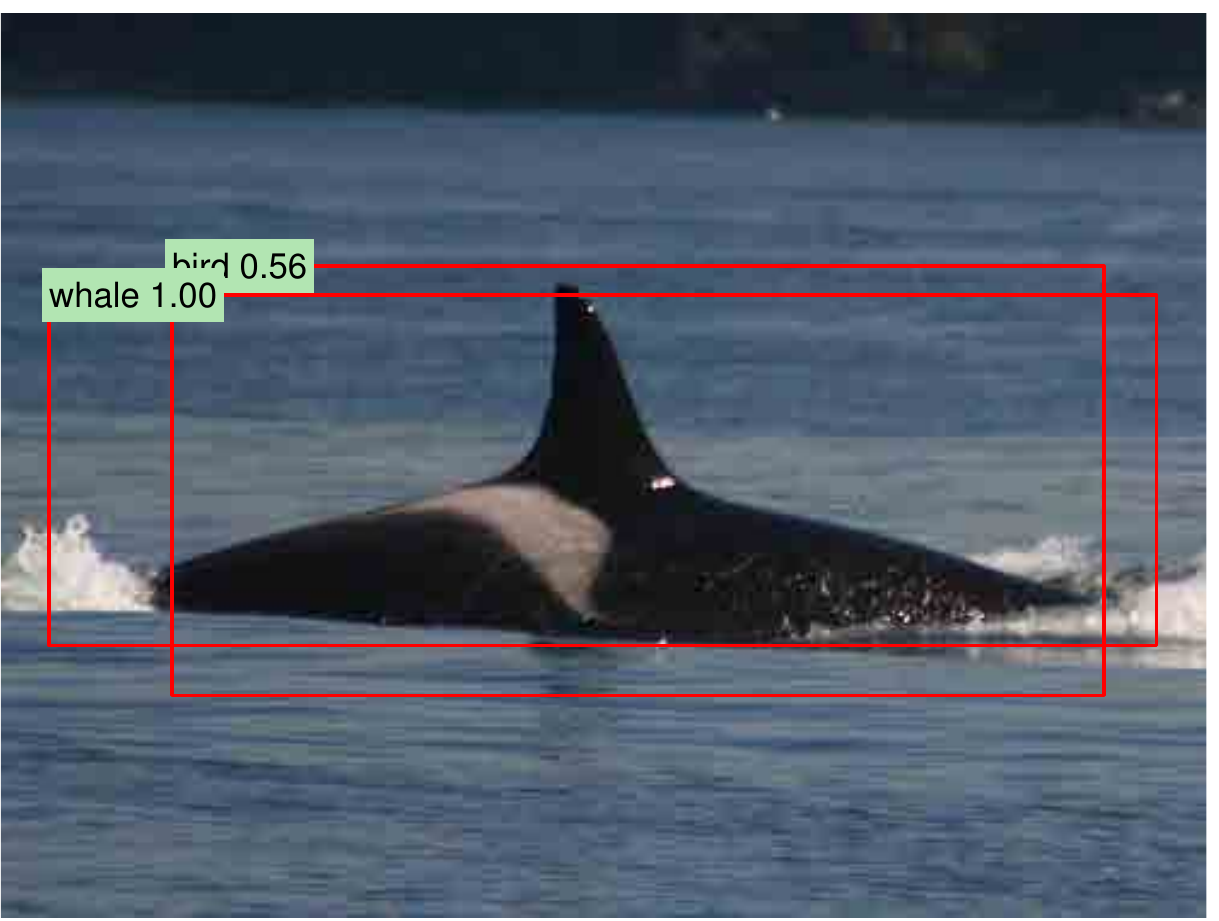}
\includegraphics[height=\sz]{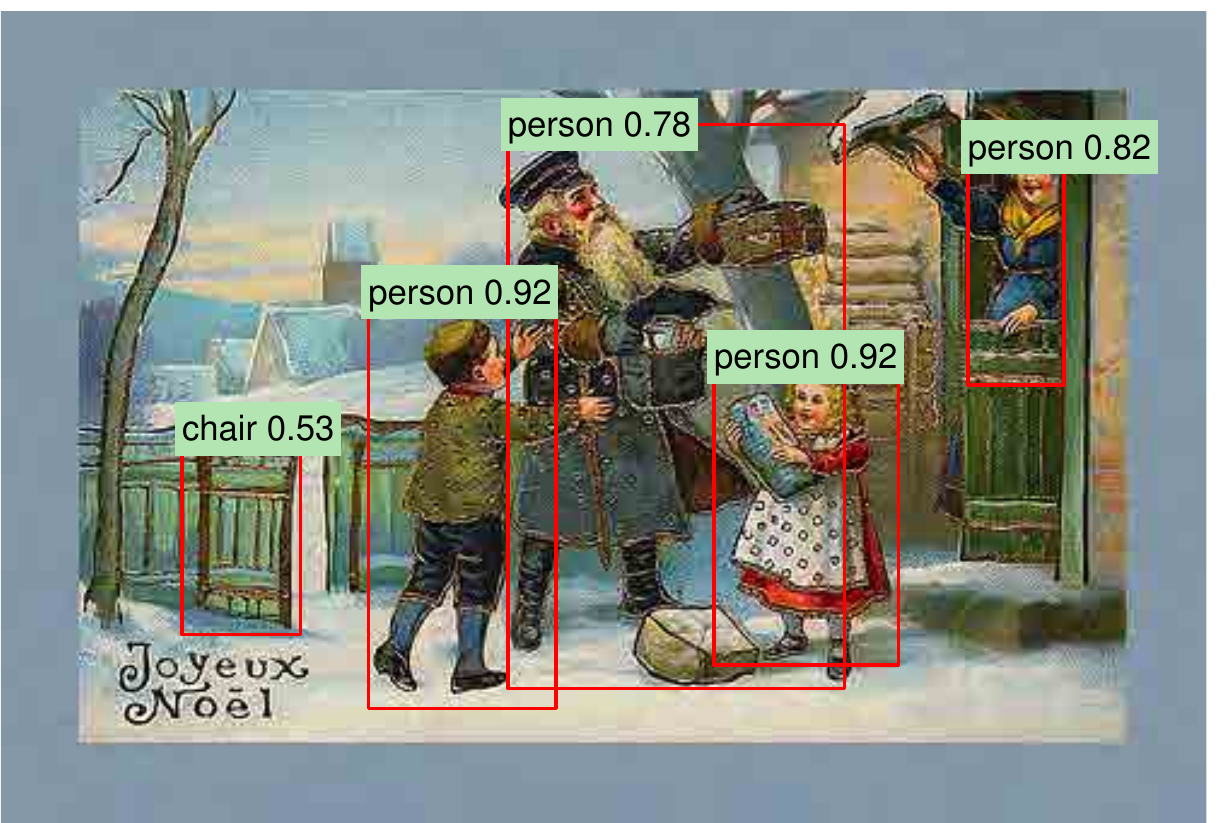}
\includegraphics[height=\sz]{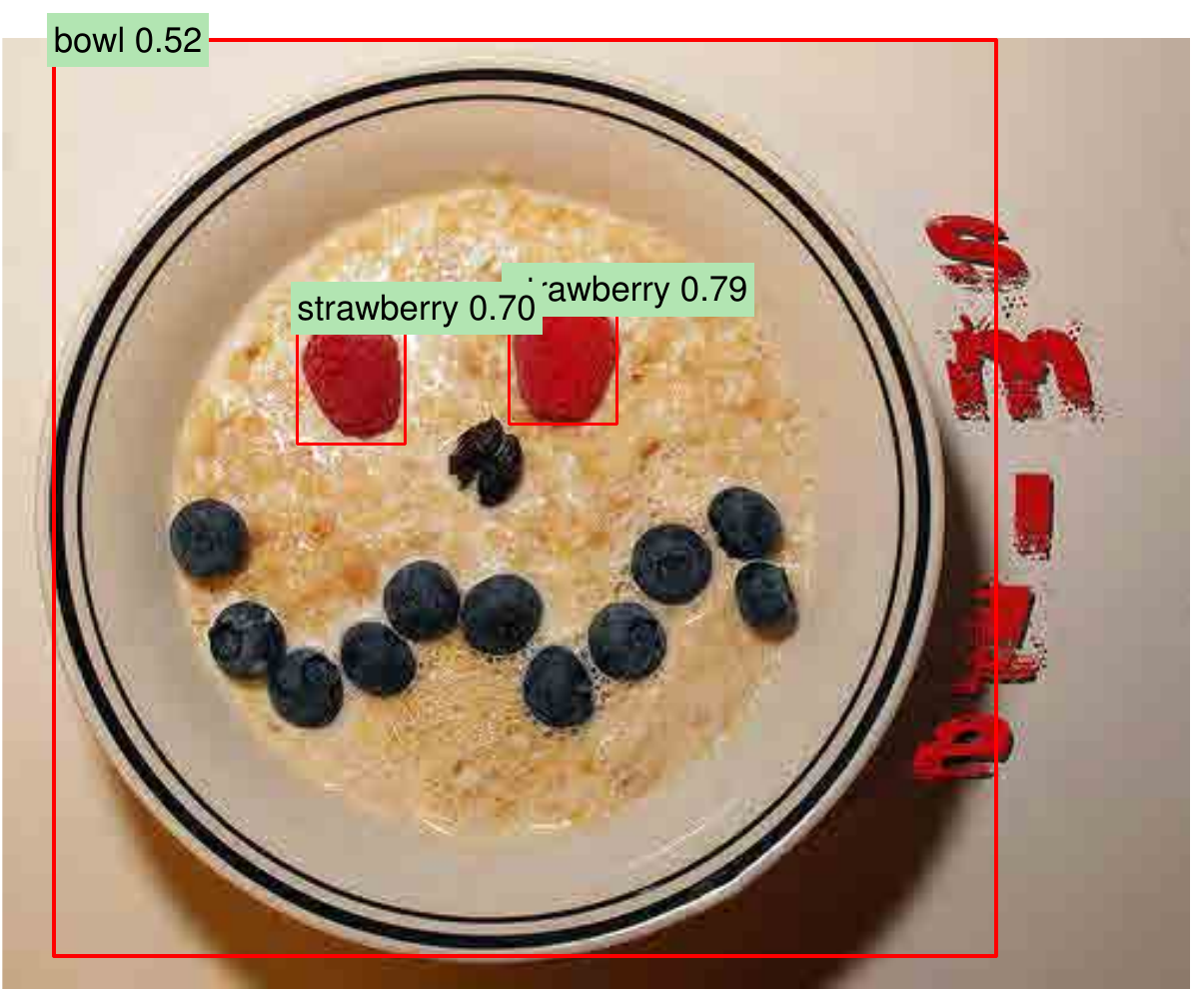}
\includegraphics[height=\sz]{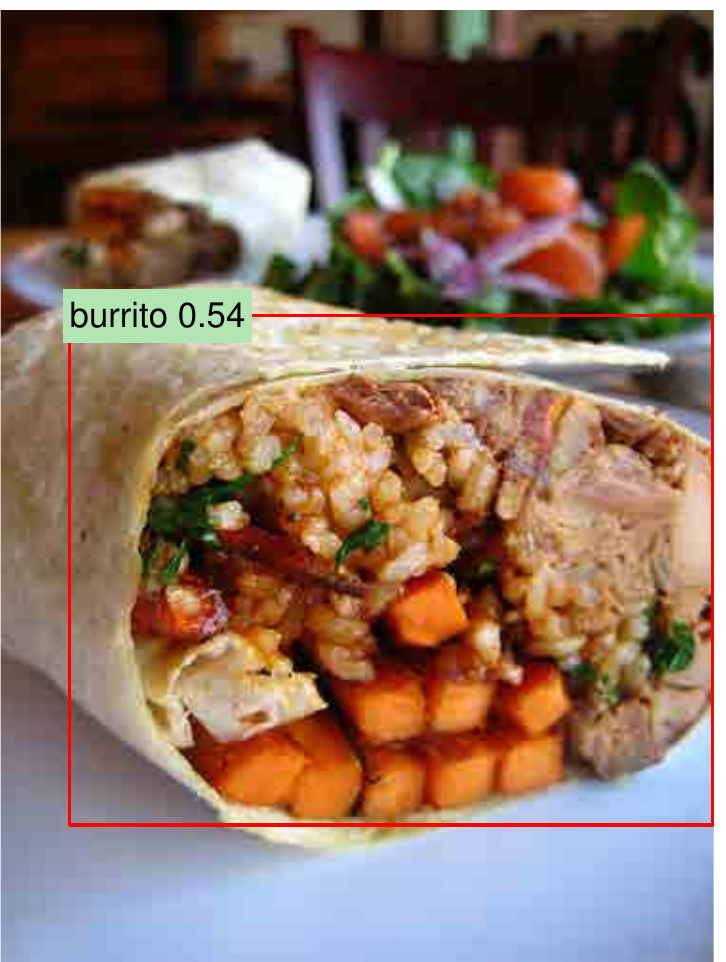}
\includegraphics[height=\sz]{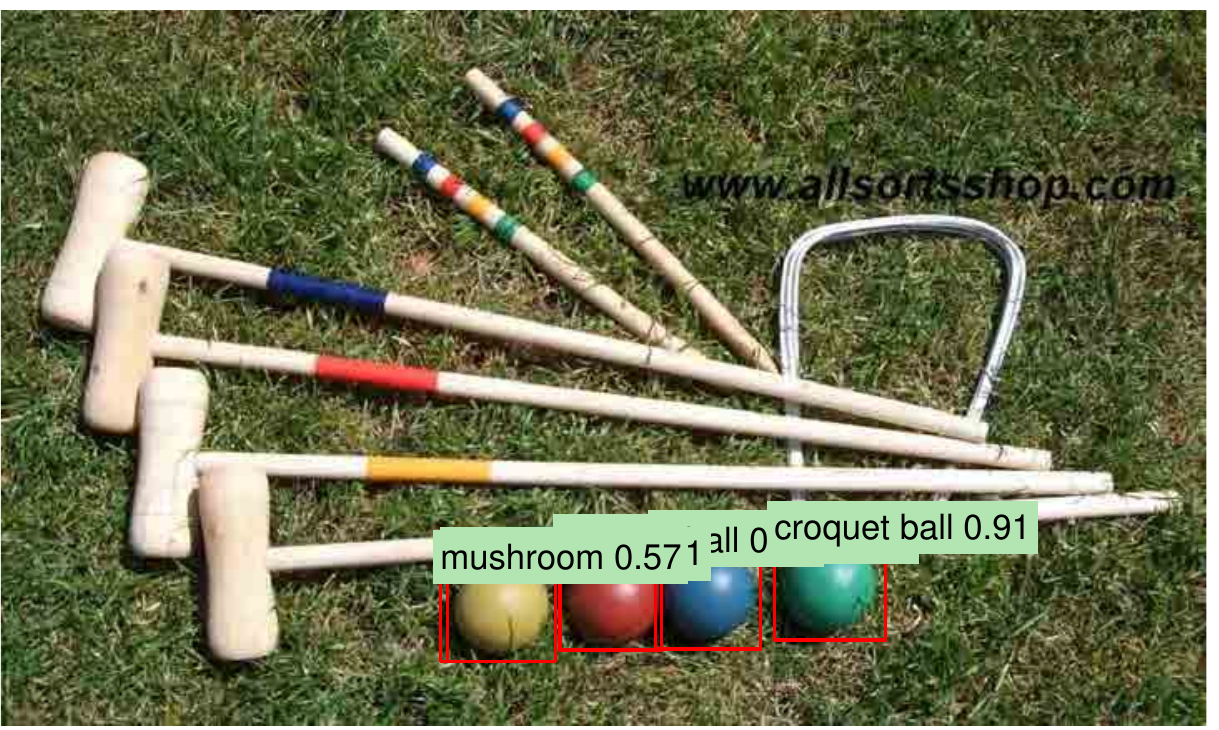}
\includegraphics[height=\sz]{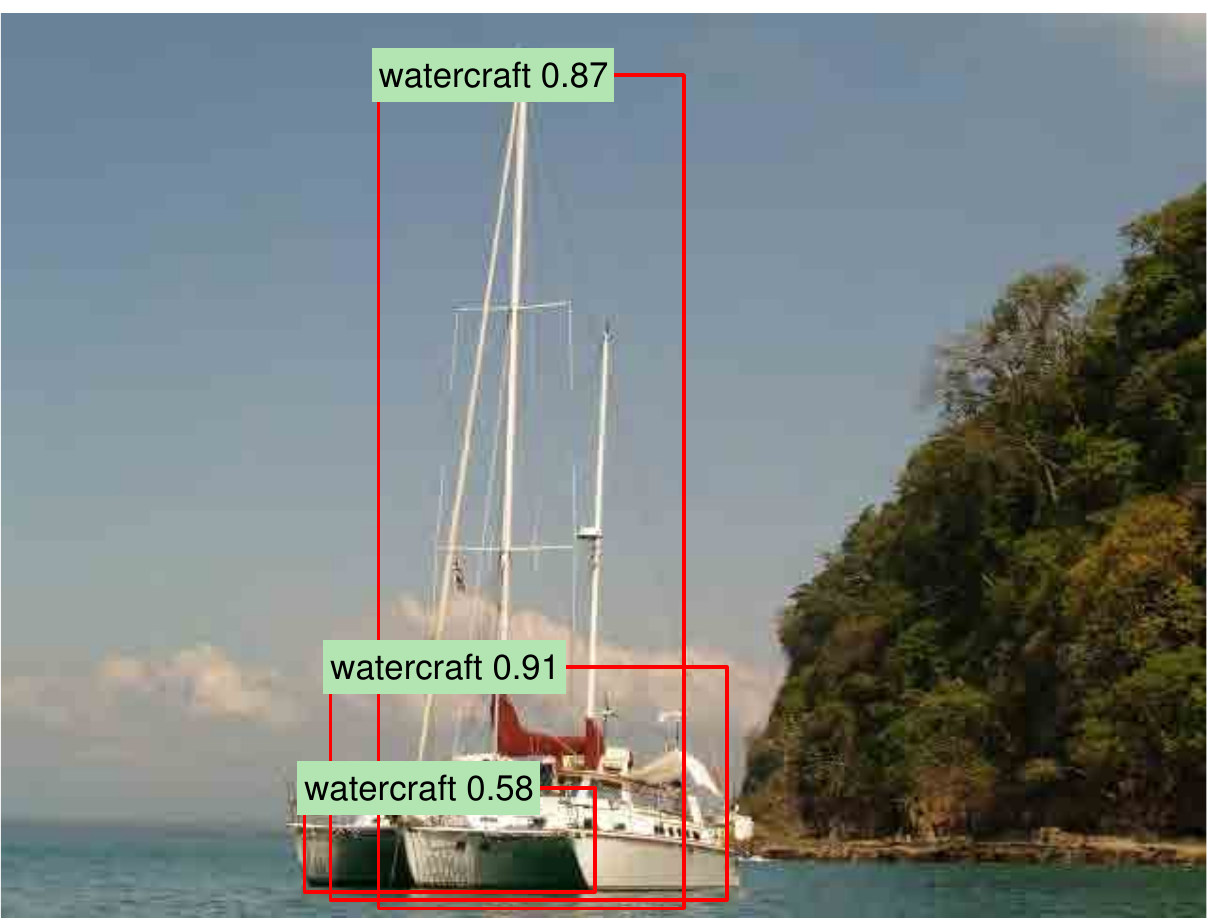}
\includegraphics[height=\sz]{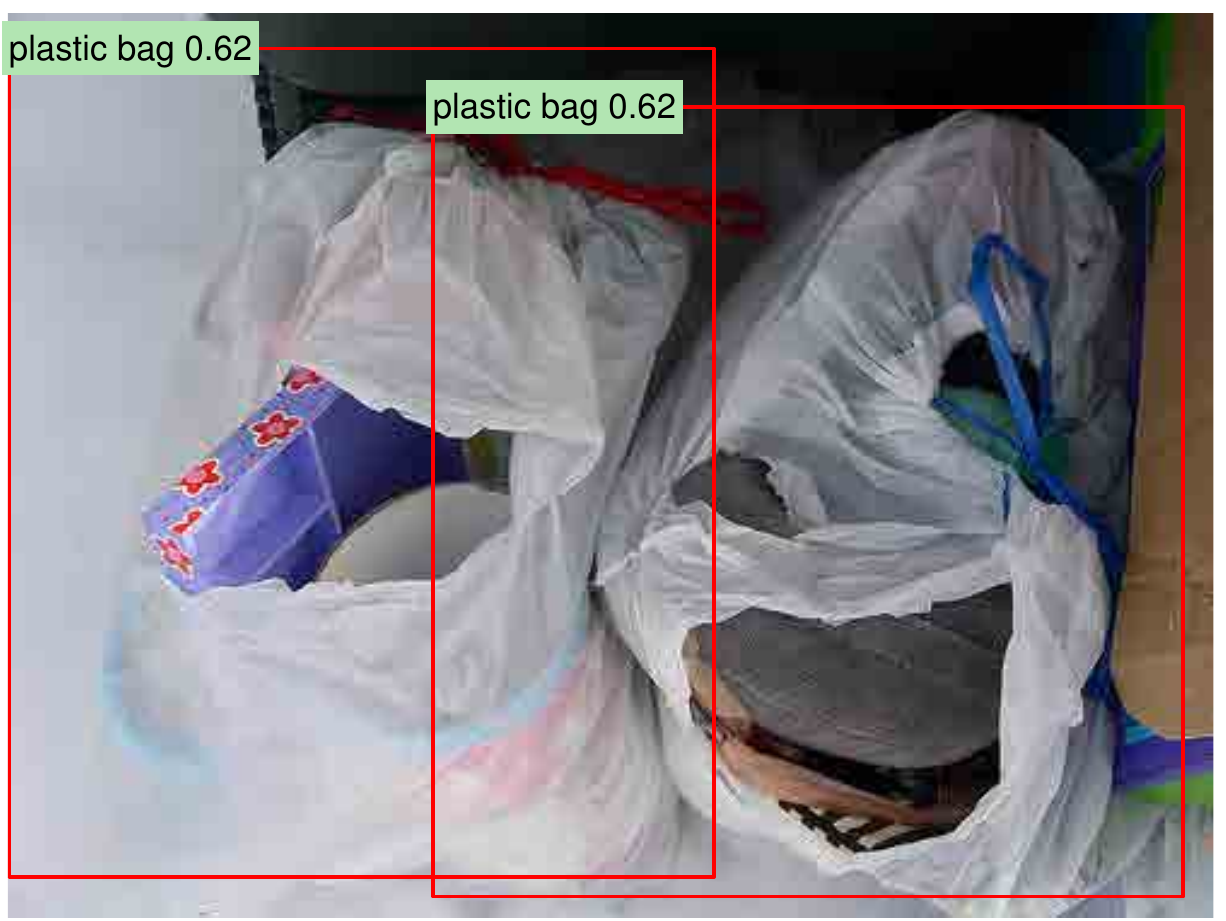}
\includegraphics[height=\sz]{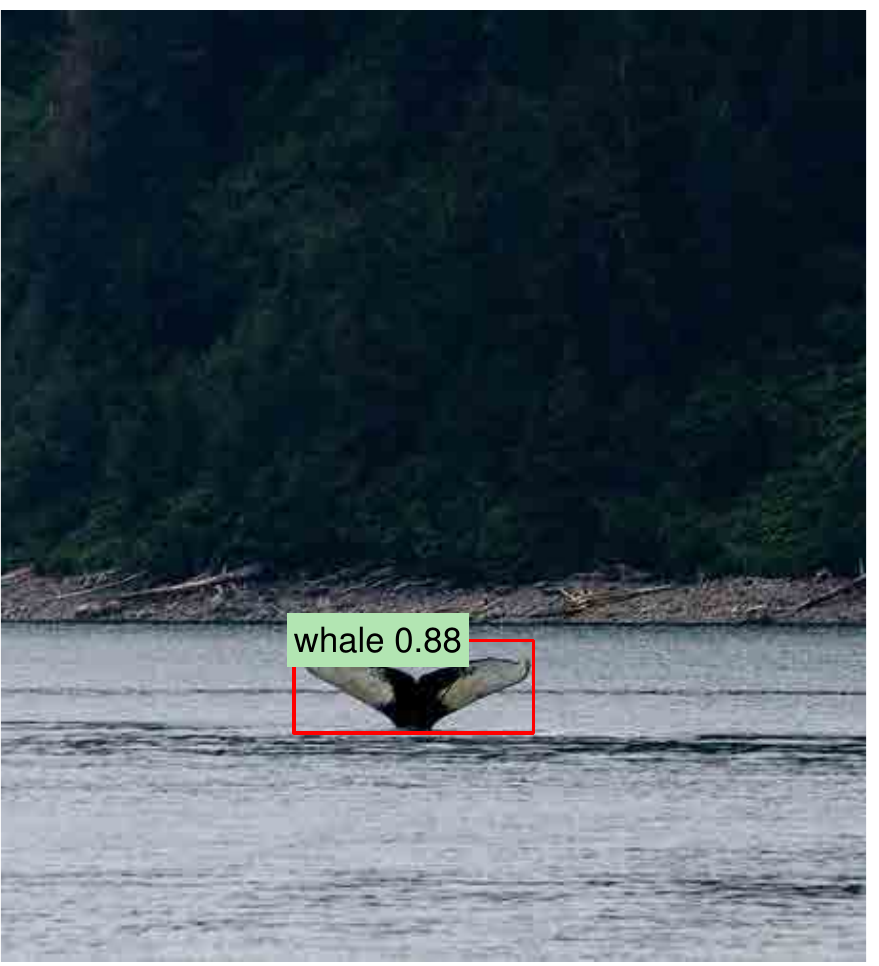}
\includegraphics[height=\sz]{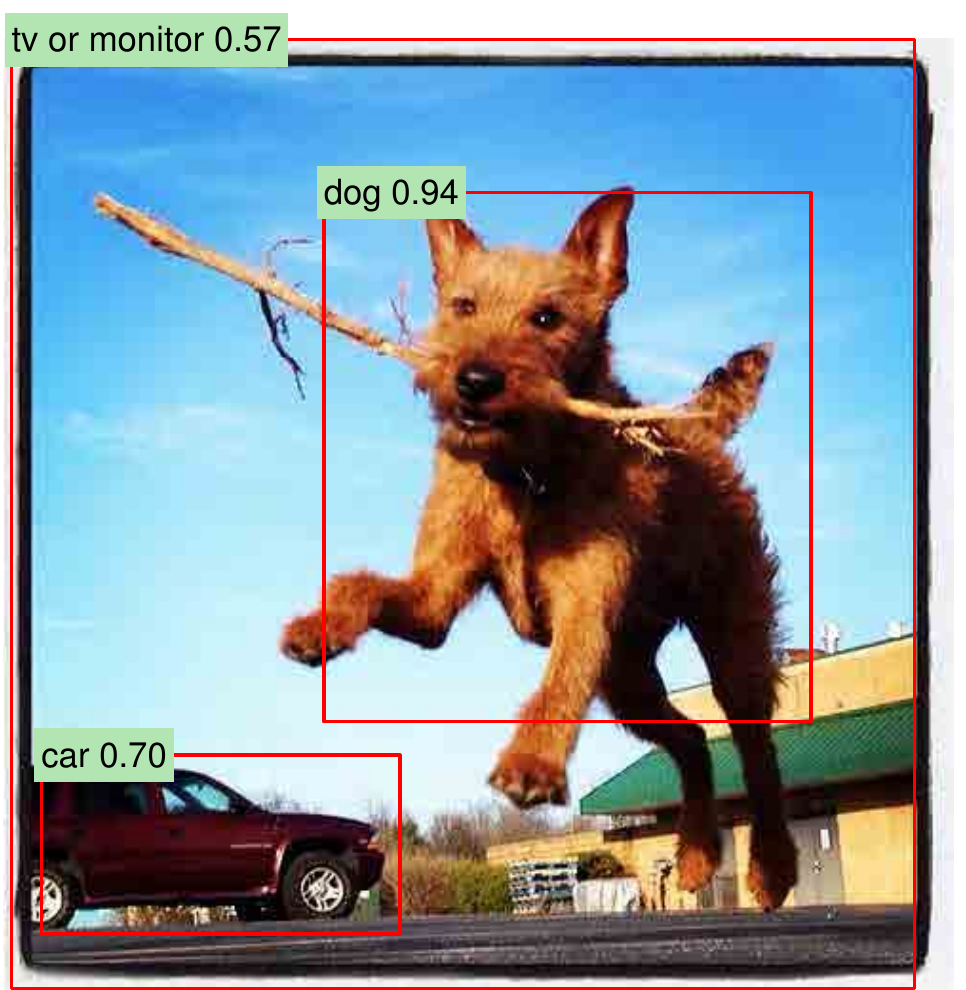}
\includegraphics[height=\sz]{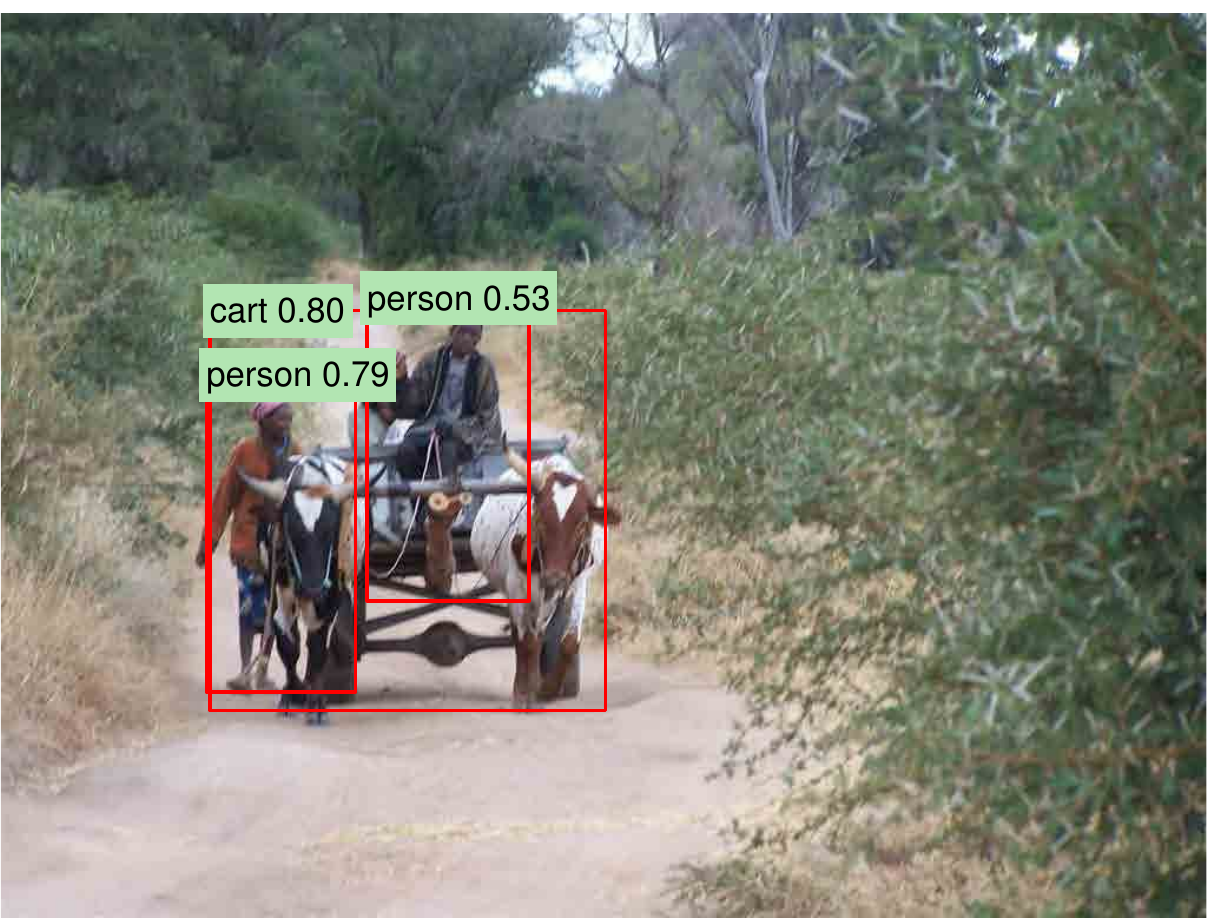}
\includegraphics[height=\sz]{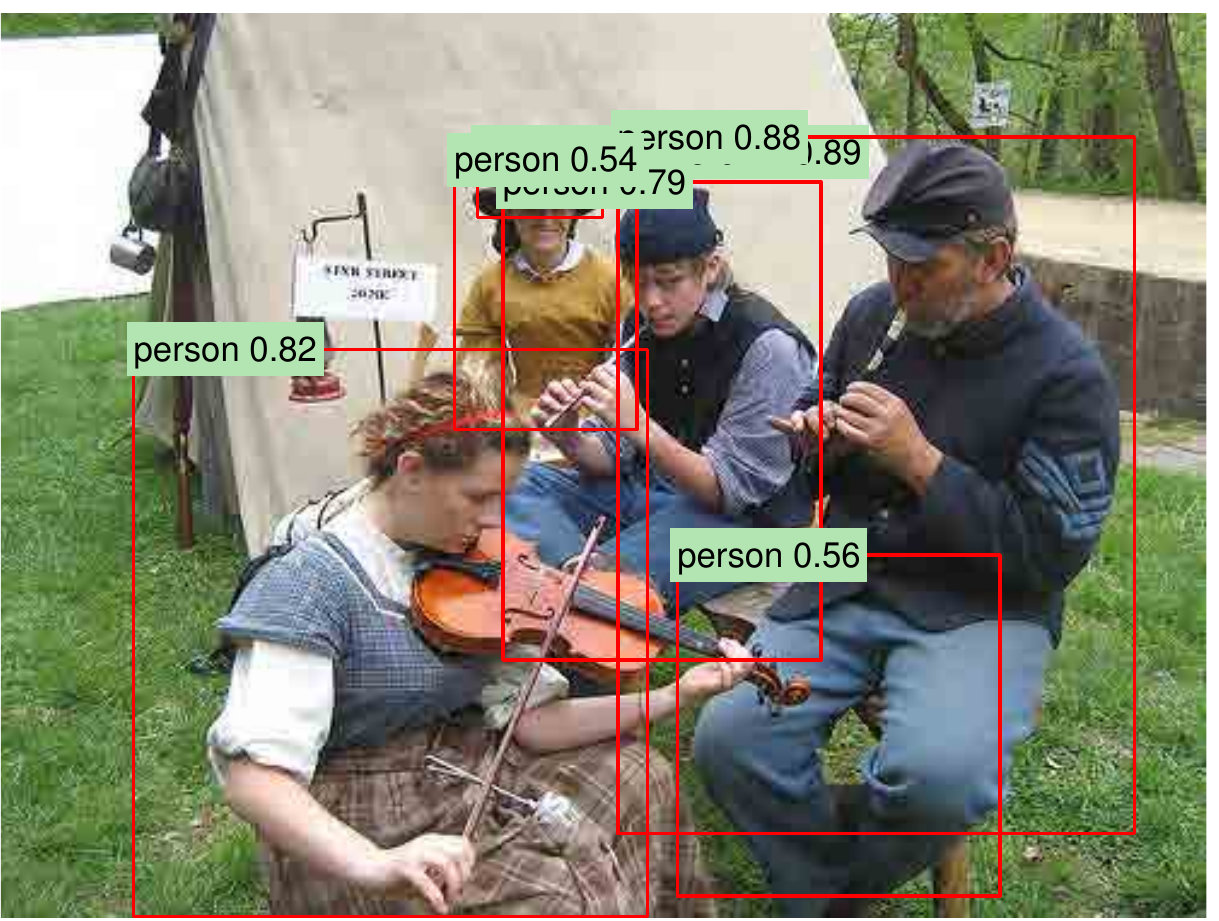}
\includegraphics[height=\sz]{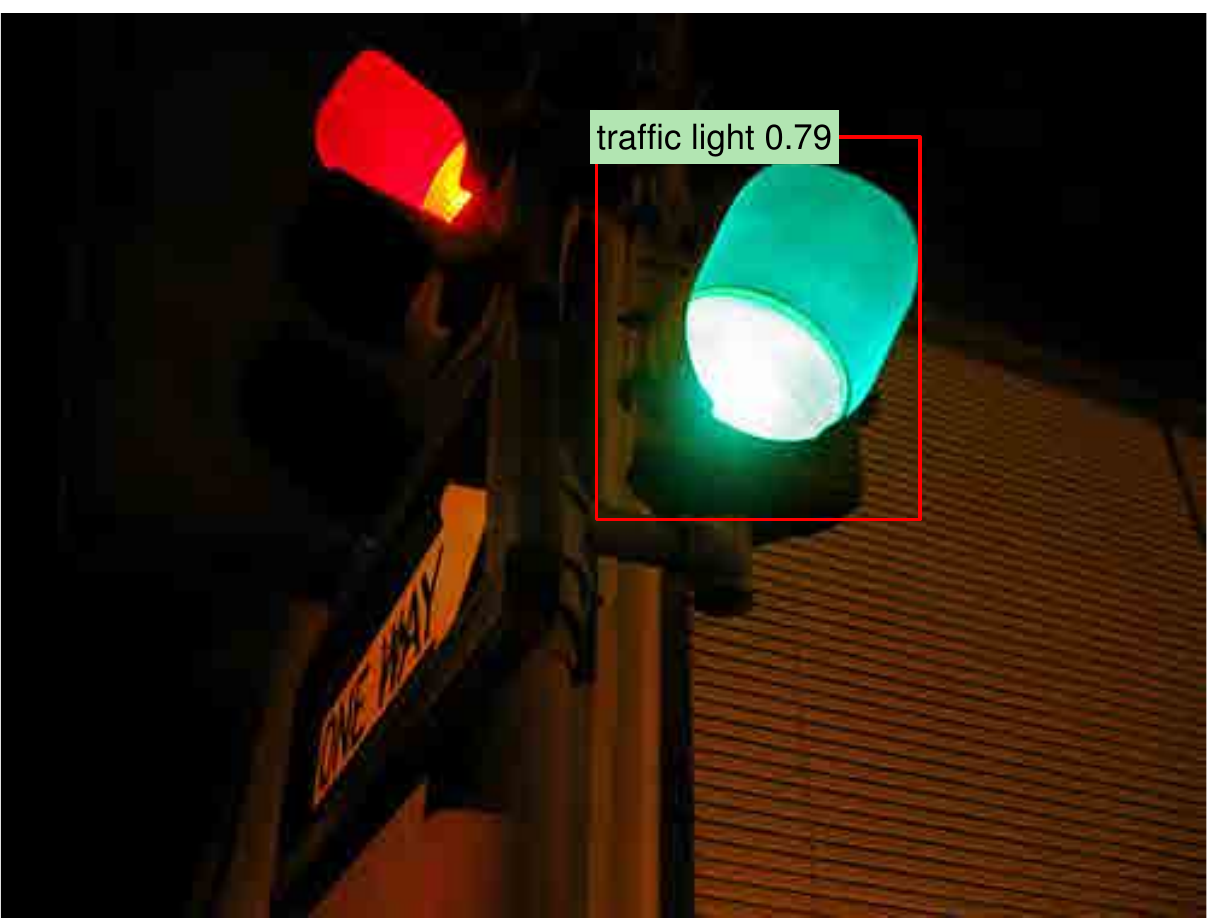}
\includegraphics[height=\sz]{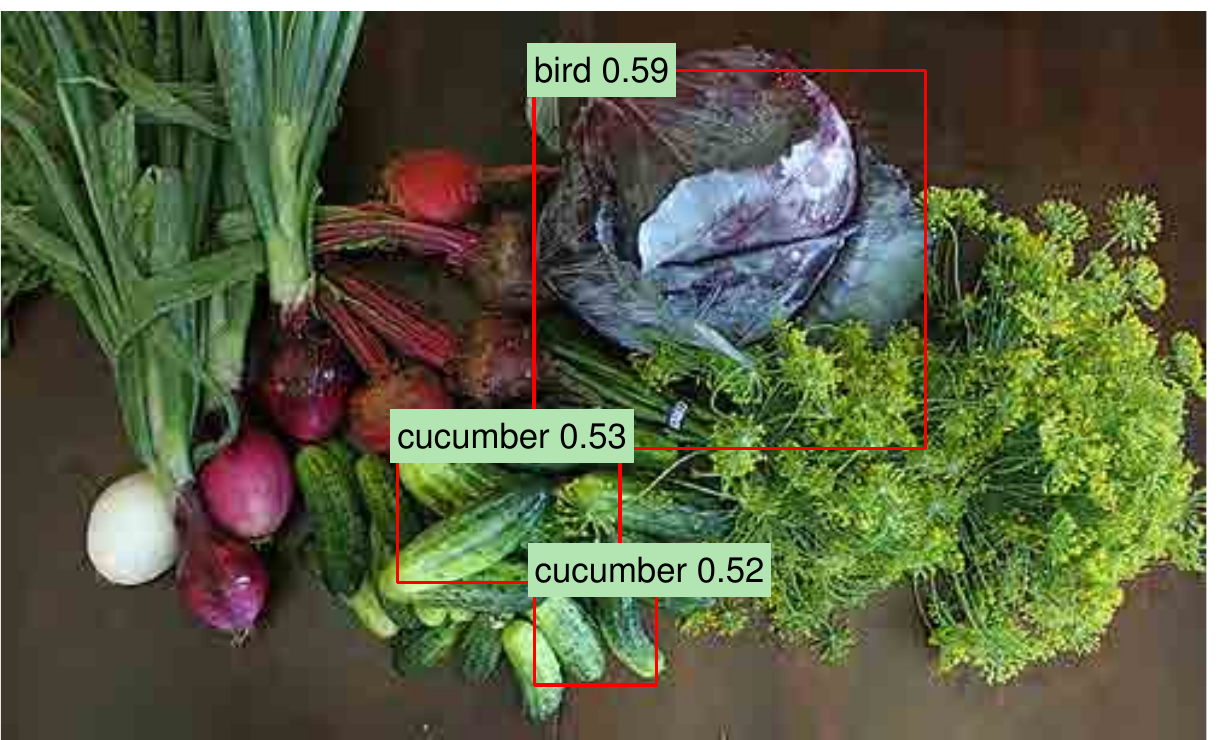}
\includegraphics[height=\sz]{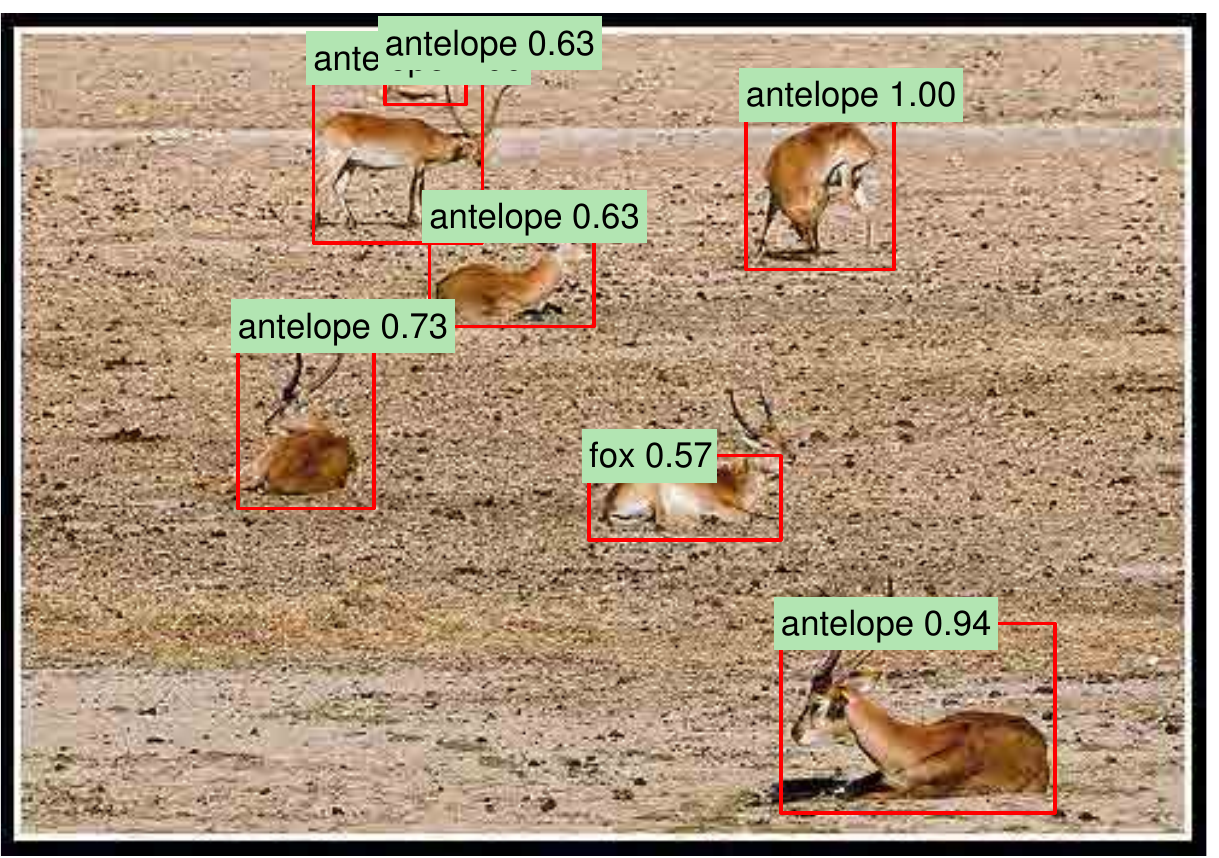}
\includegraphics[height=\sz]{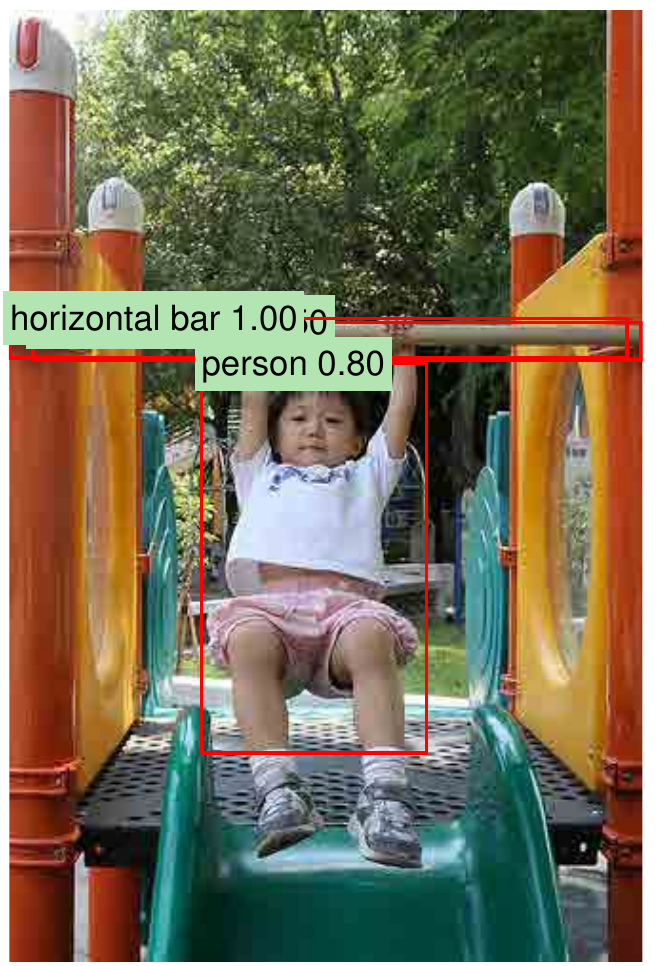}
\includegraphics[height=\sz]{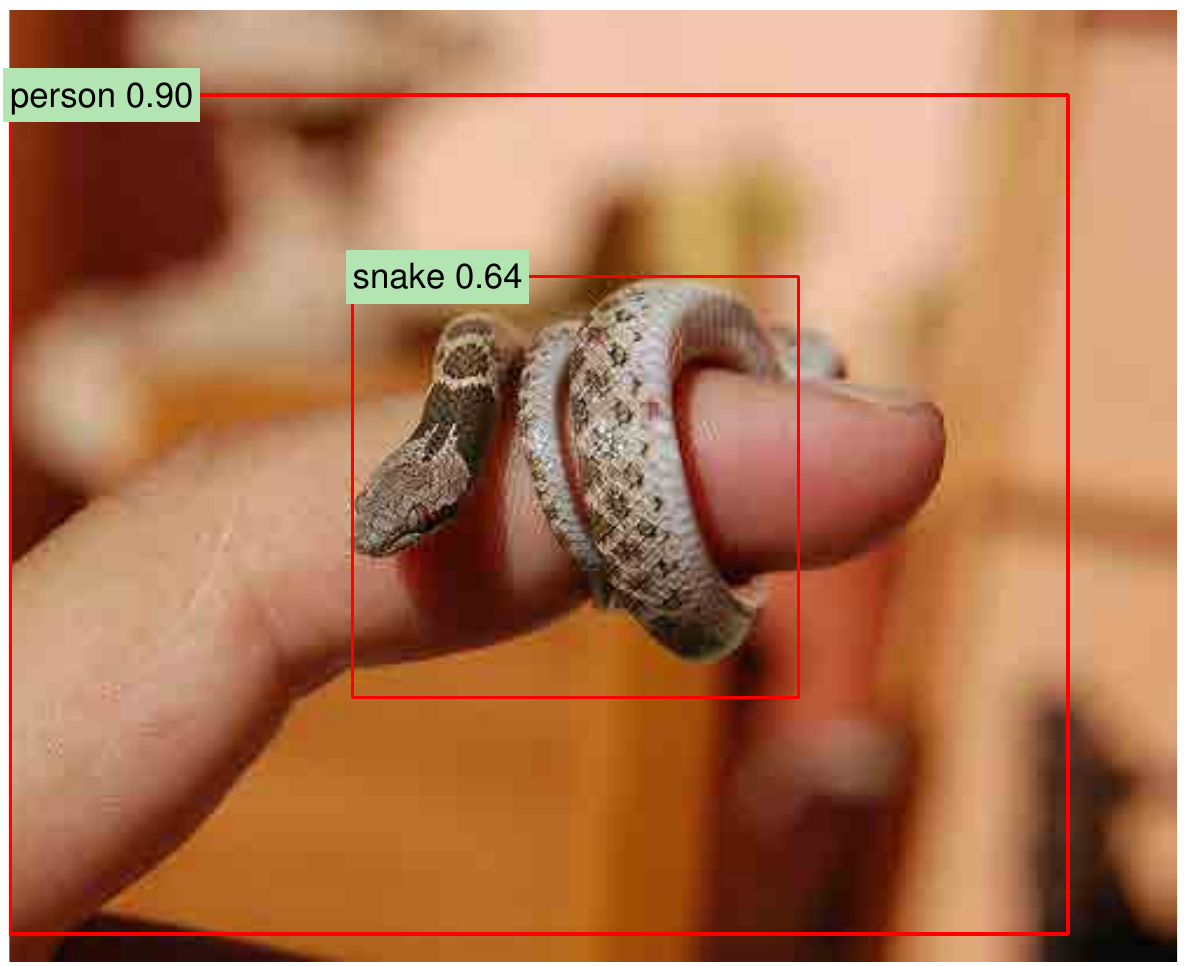}
\includegraphics[height=\sz]{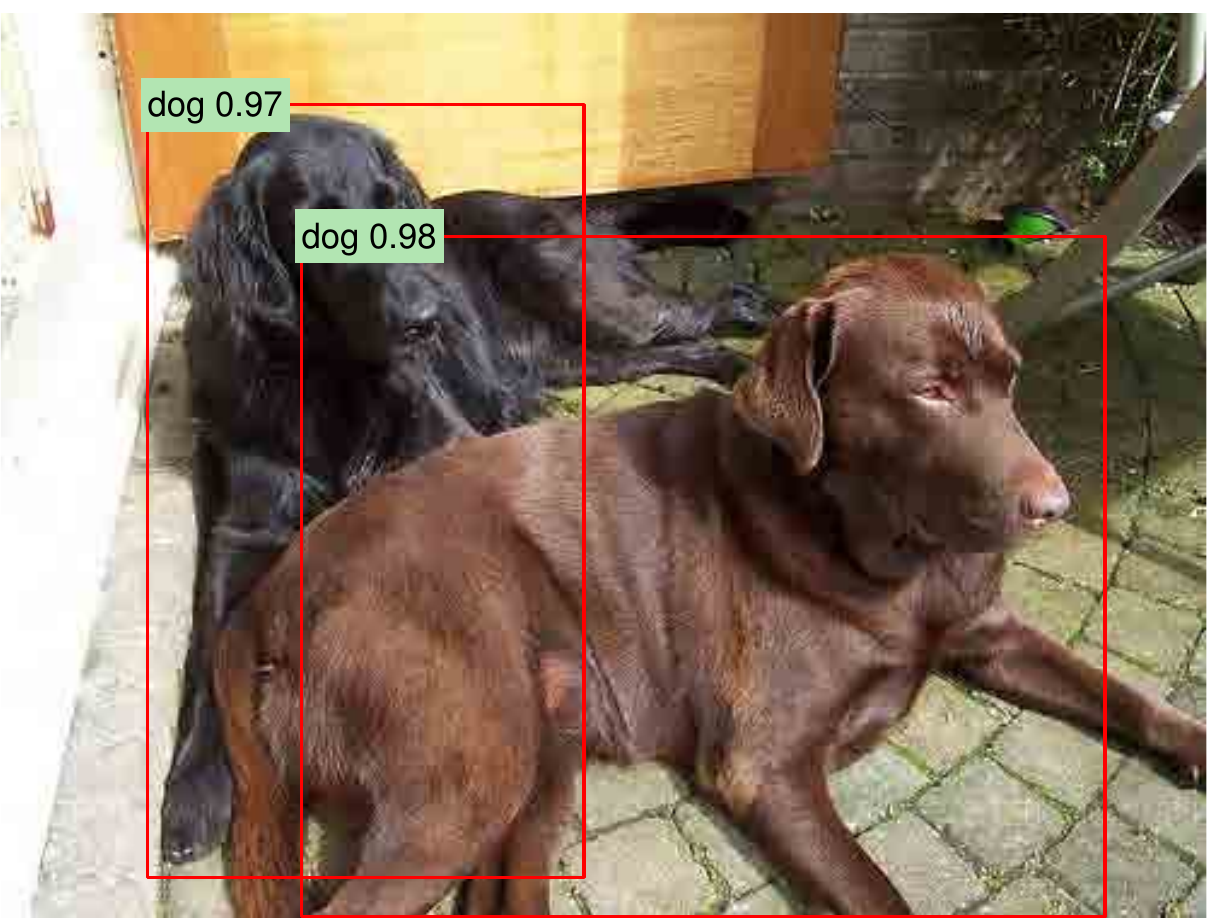}
\includegraphics[height=\sz]{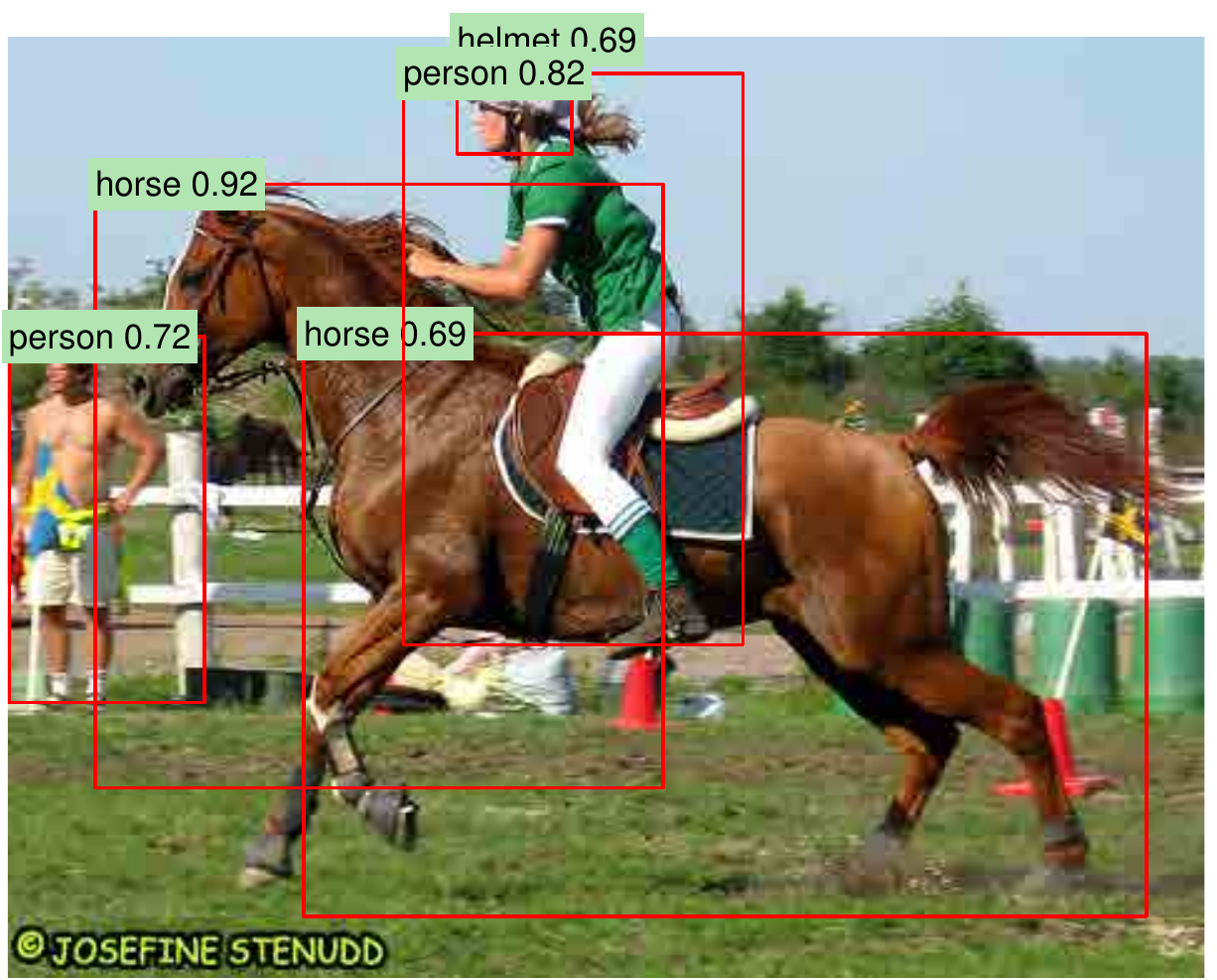}
\includegraphics[height=\sz]{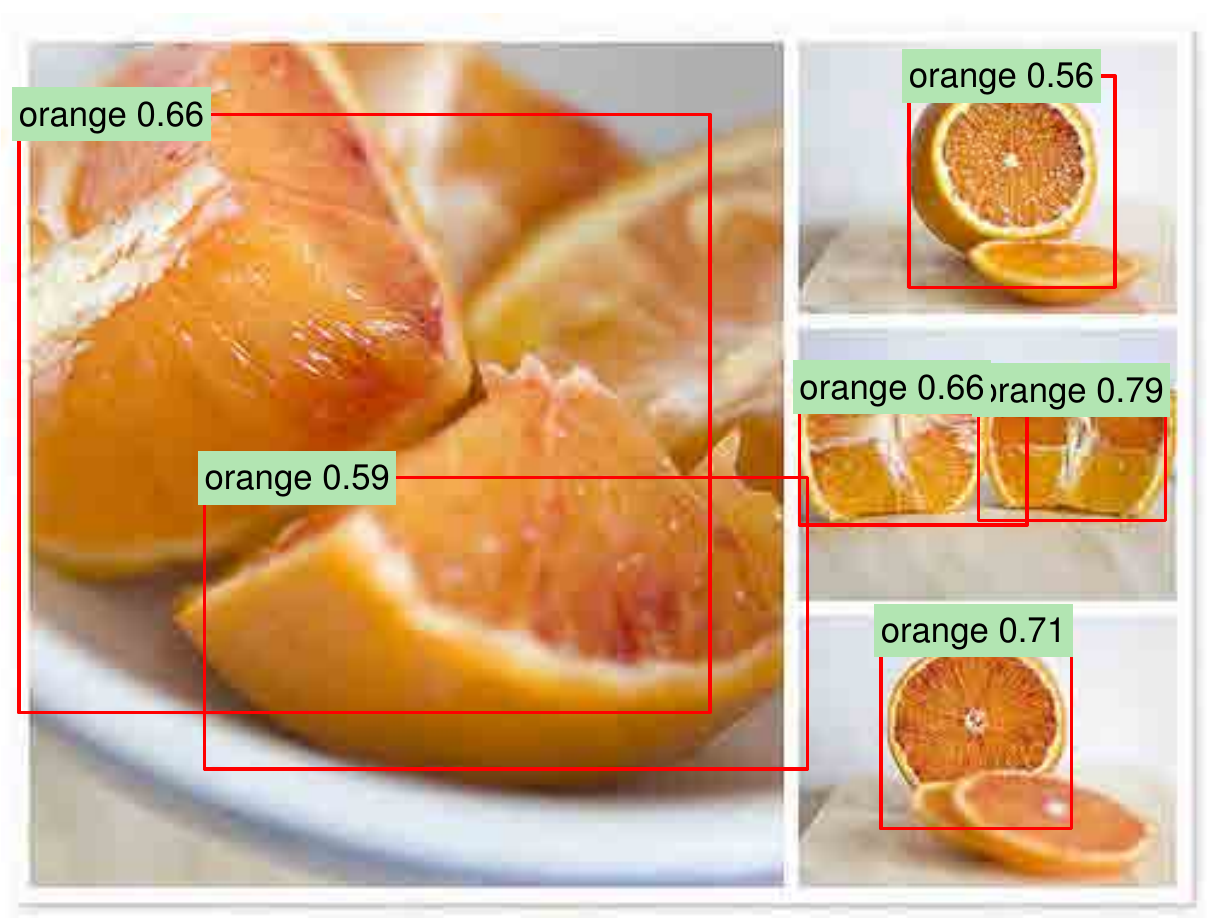}
\includegraphics[height=\sz]{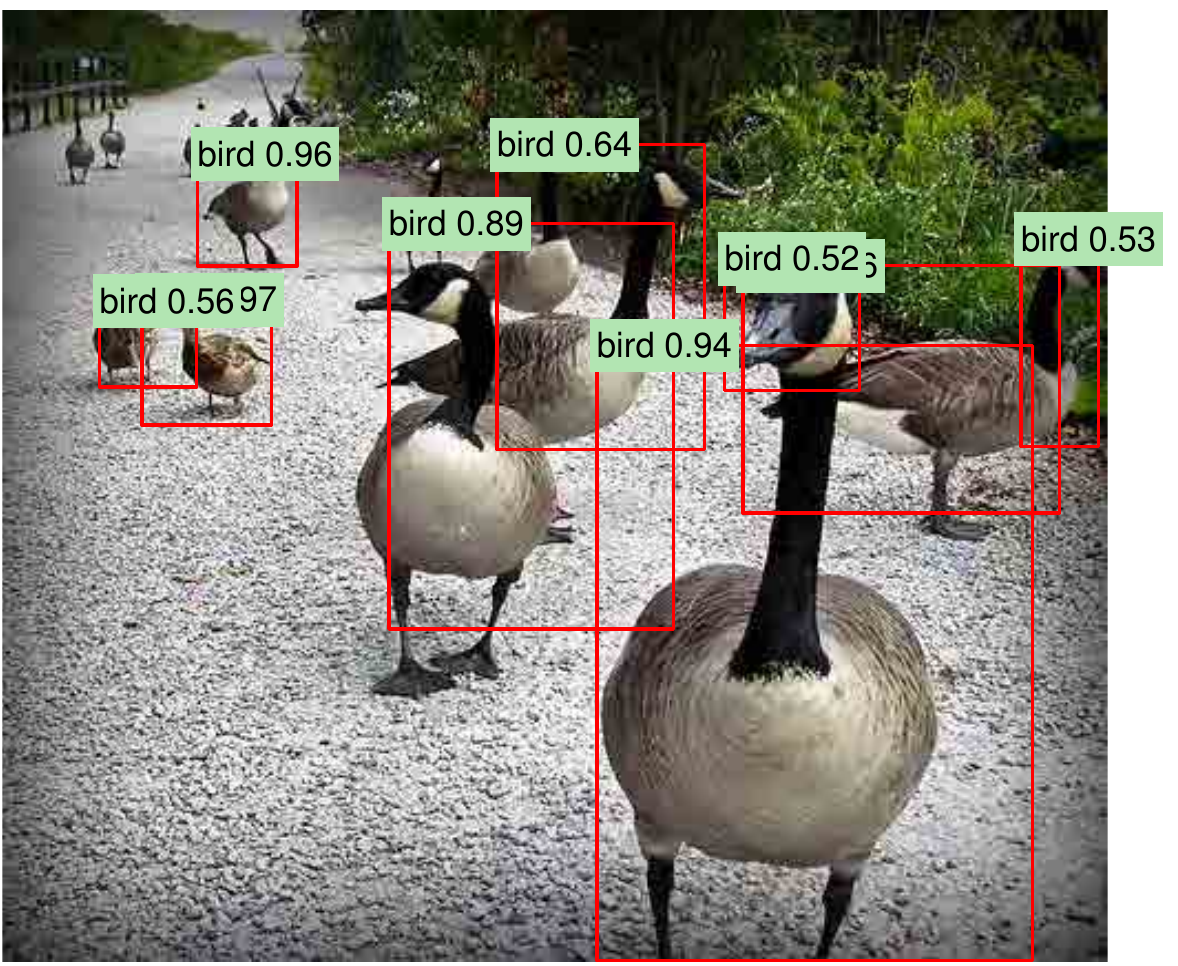}
\includegraphics[height=\sz]{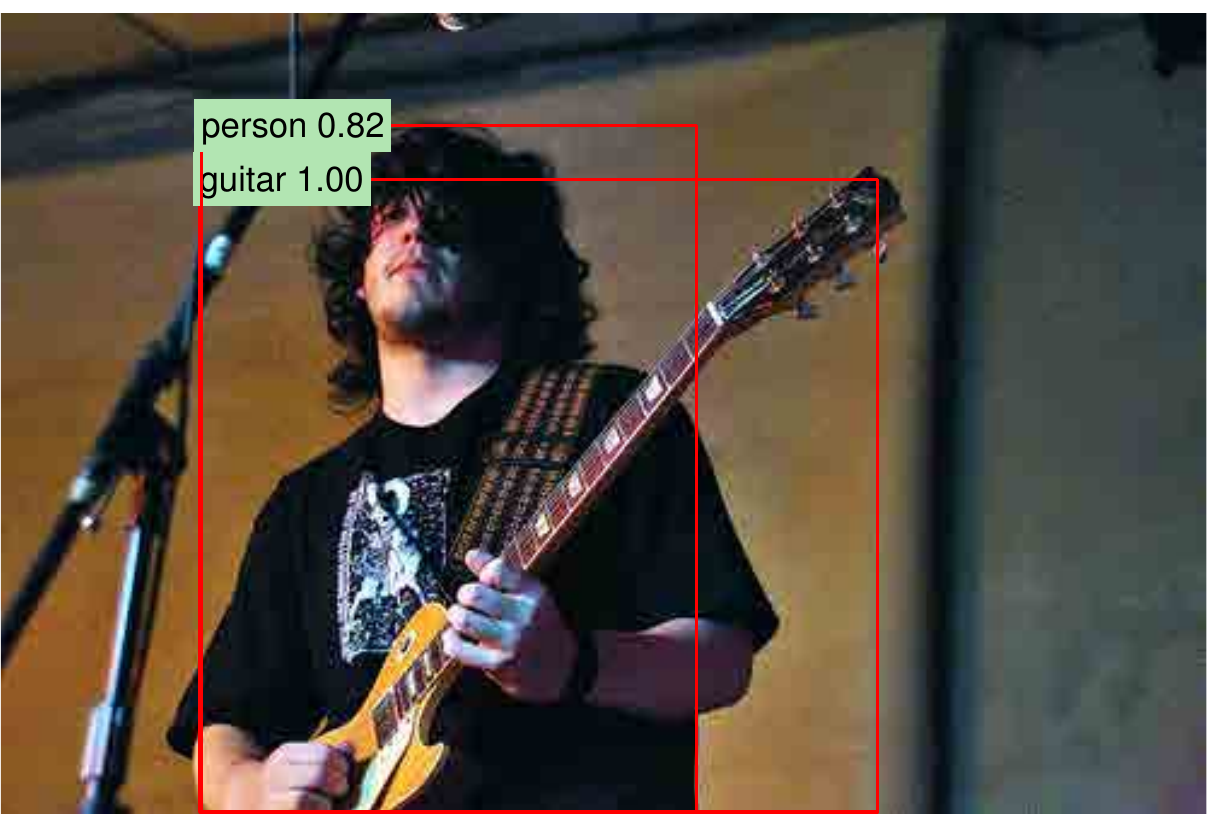}
\includegraphics[height=\sz]{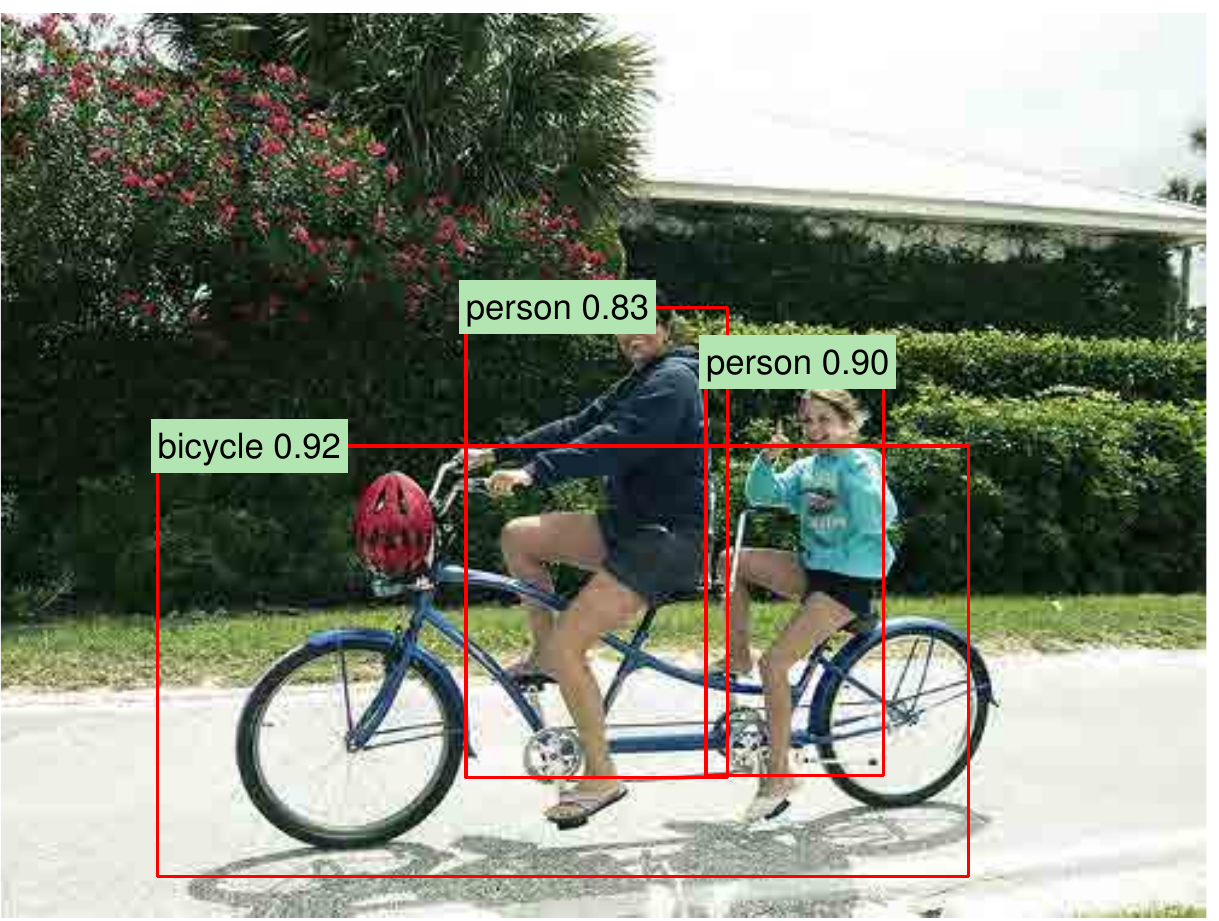}
\includegraphics[height=\sz]{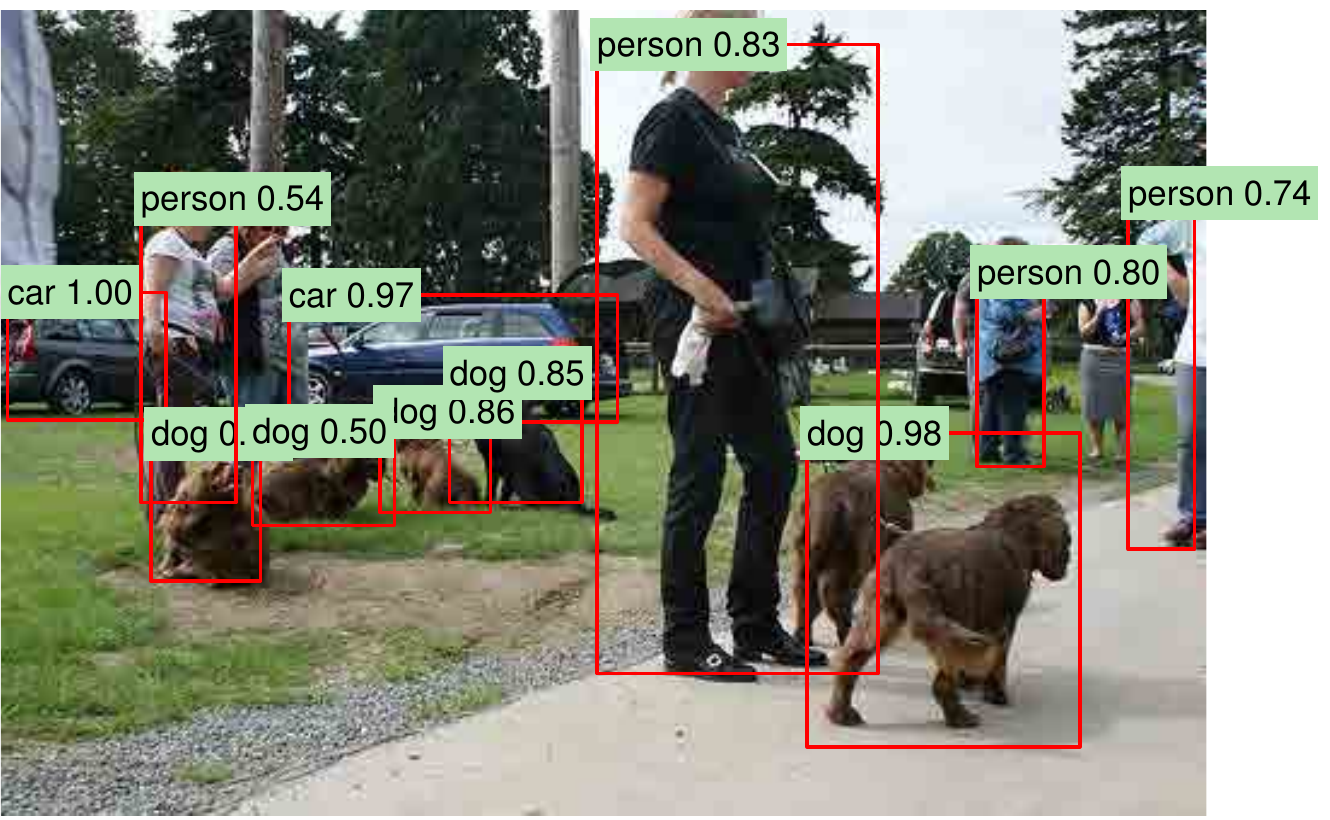}
\includegraphics[height=\sz]{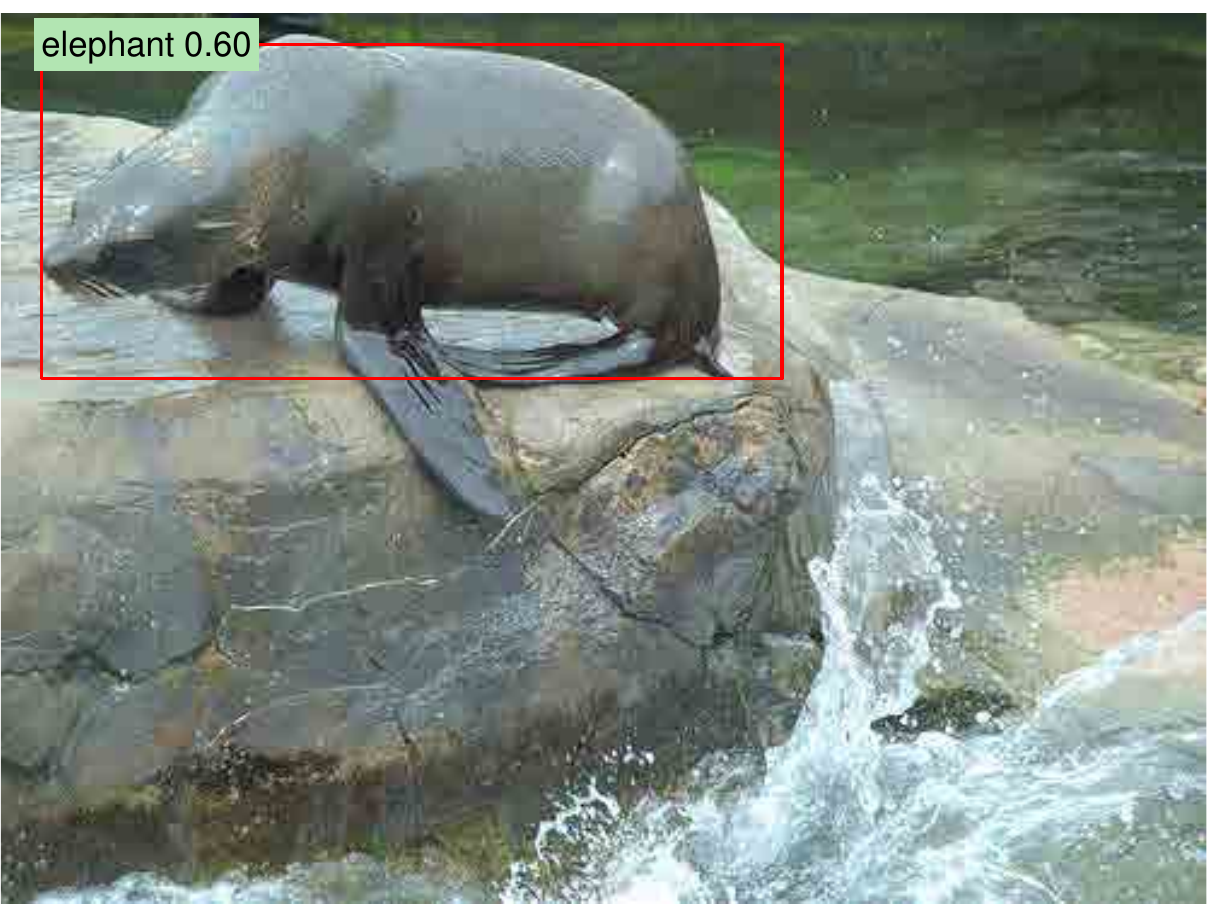}
\end{center}
\caption{More curated examples. See \figref{cexamples1} caption for details.
Viewing digitally with zoom is recommended.
}
\figlabel{cexamples2}
\end{figure*}

\begin{figure*}[t!]
\def \sz {0.46}
\centering
\includegraphics[width=\sz\linewidth]{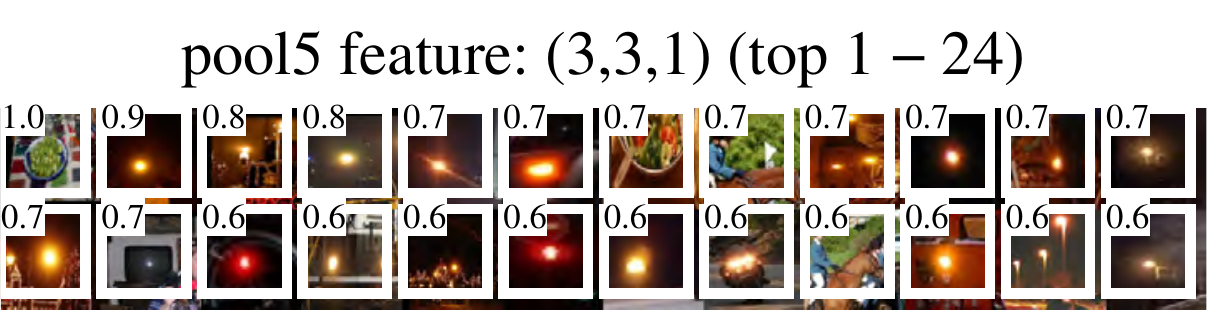}
\hspace{5pt}
\includegraphics[width=\sz\linewidth]{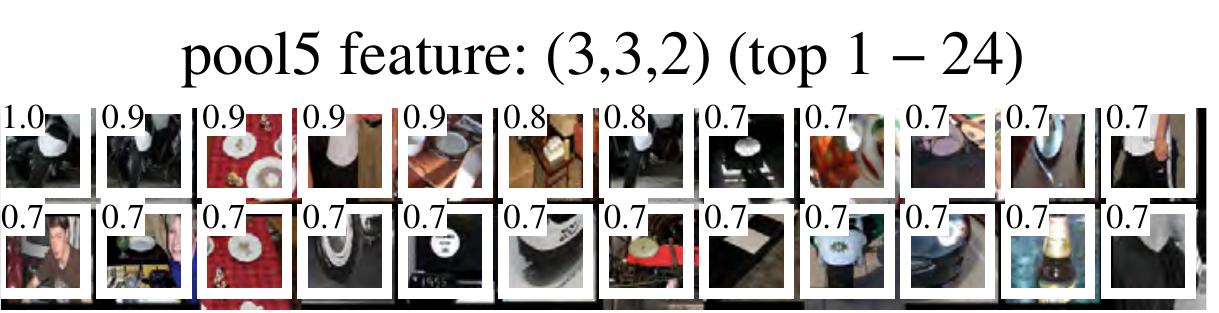}

\includegraphics[width=\sz\linewidth]{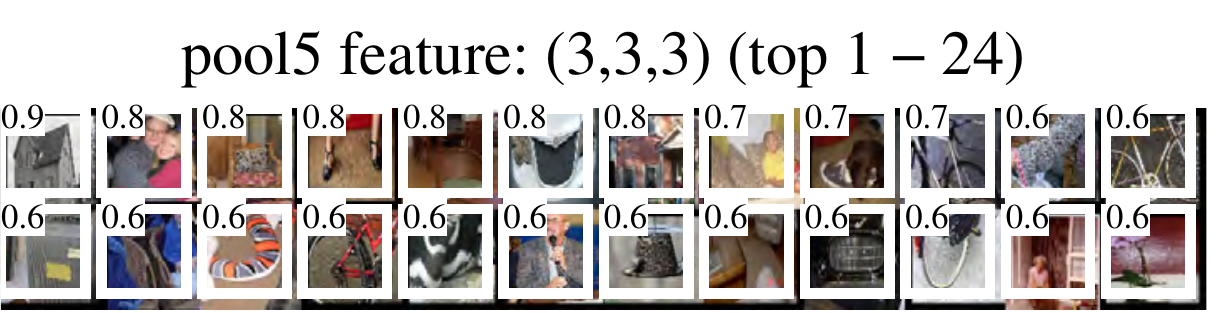}
\hspace{5pt}
\includegraphics[width=\sz\linewidth]{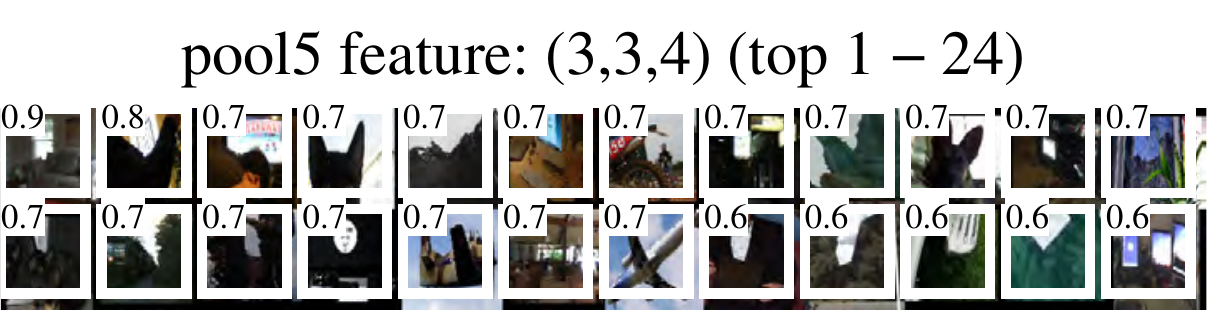}

\includegraphics[width=\sz\linewidth]{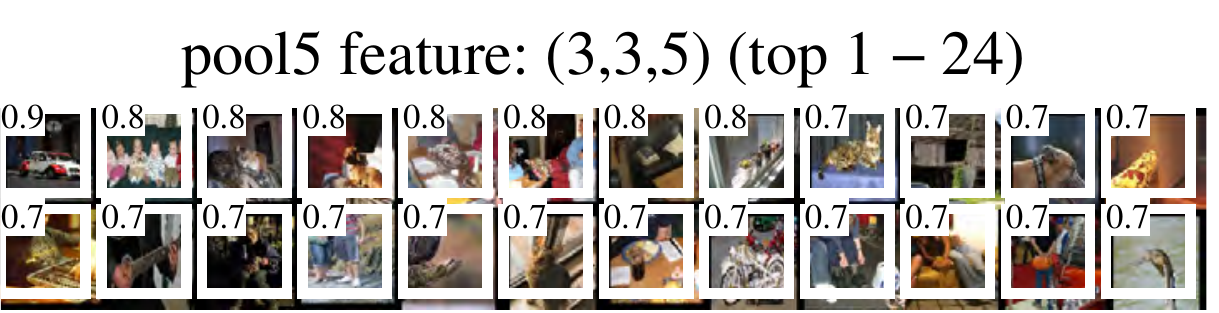}
\hspace{5pt}
\includegraphics[width=\sz\linewidth]{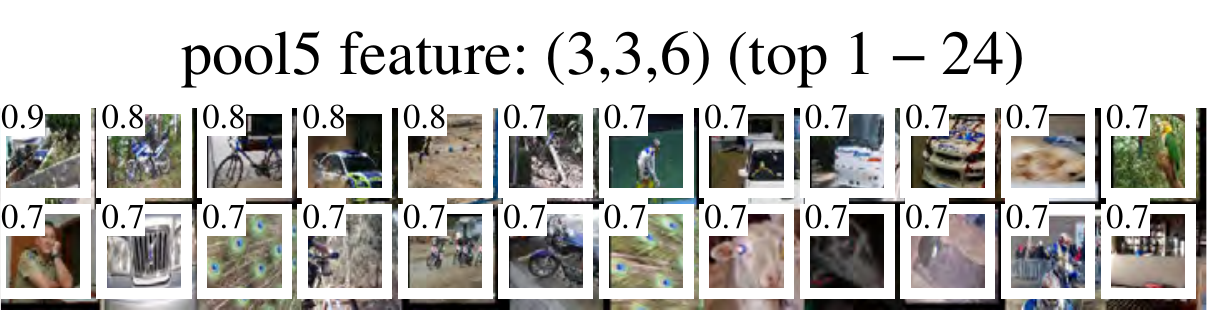}

\includegraphics[width=\sz\linewidth]{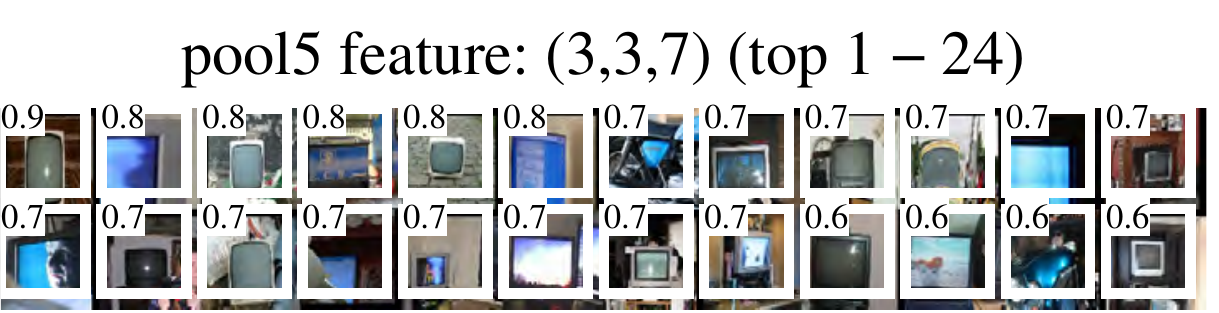}
\hspace{5pt}
\includegraphics[width=\sz\linewidth]{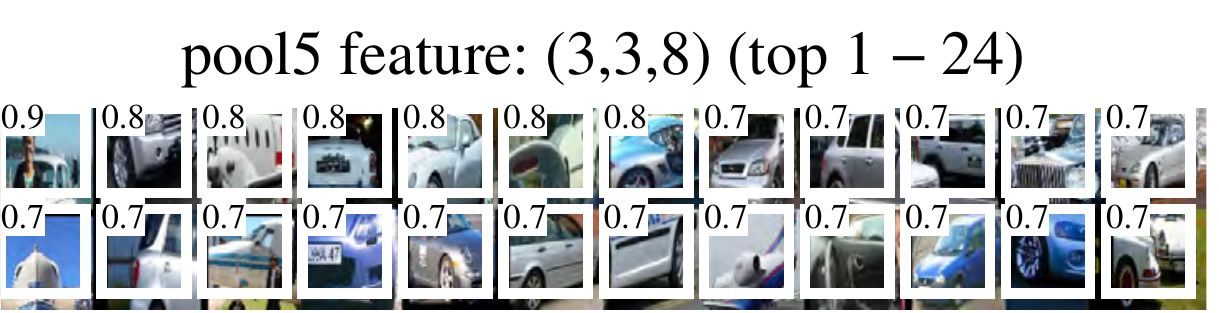}

\includegraphics[width=\sz\linewidth]{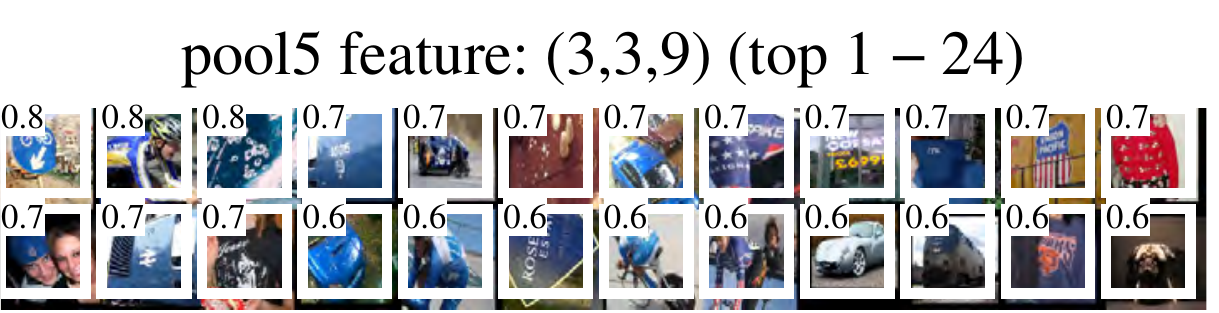}
\hspace{5pt}
\includegraphics[width=\sz\linewidth]{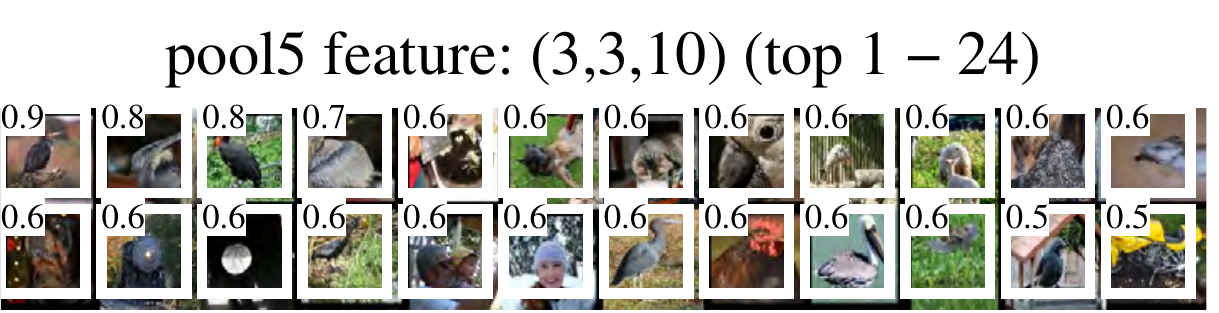}

\includegraphics[width=\sz\linewidth]{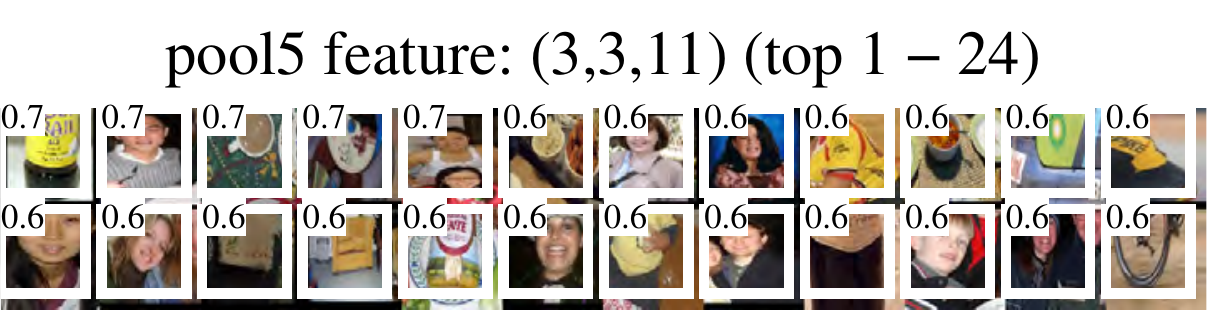}
\hspace{5pt}
\includegraphics[width=\sz\linewidth]{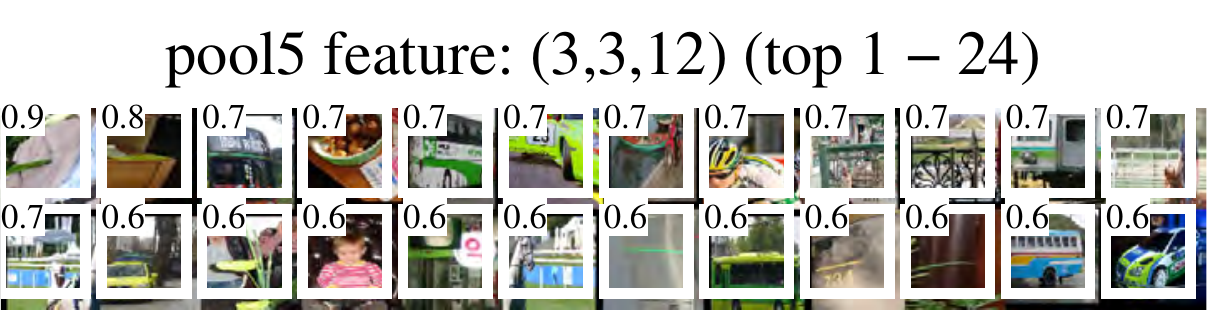}

\includegraphics[width=\sz\linewidth]{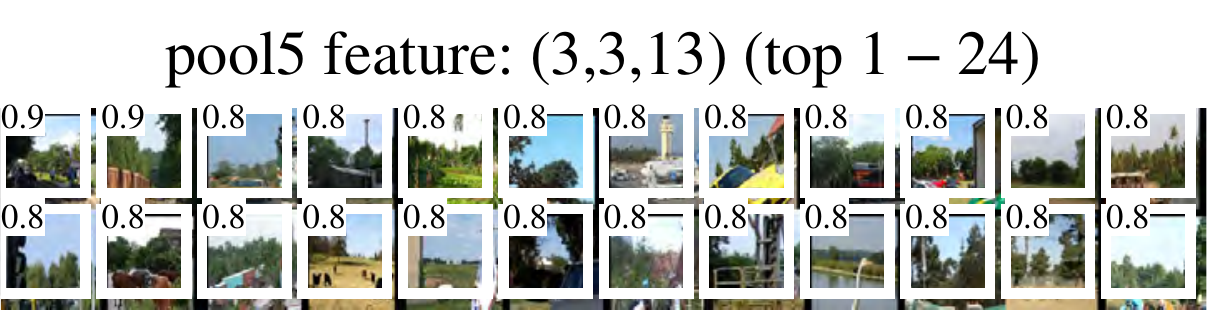}
\hspace{5pt}
\includegraphics[width=\sz\linewidth]{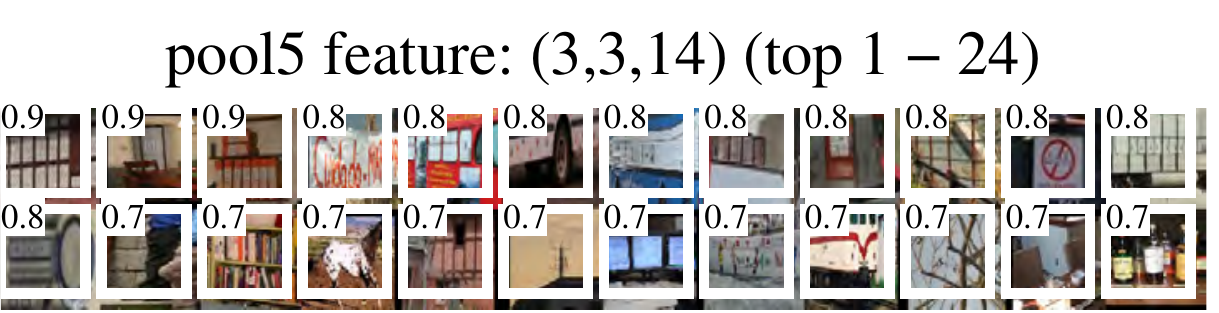}

\includegraphics[width=\sz\linewidth]{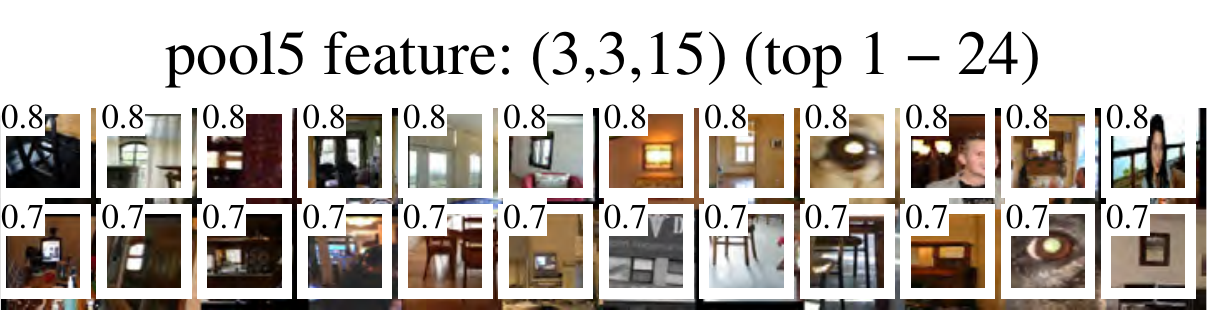}
\hspace{5pt}
\includegraphics[width=\sz\linewidth]{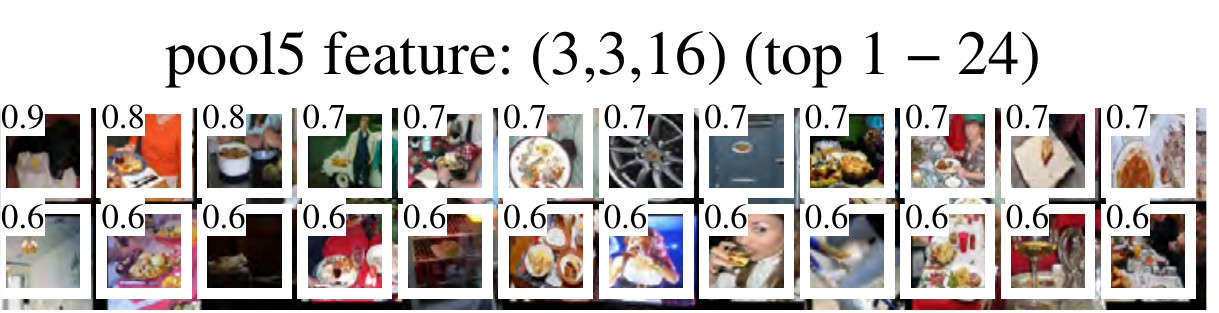}

\includegraphics[width=\sz\linewidth]{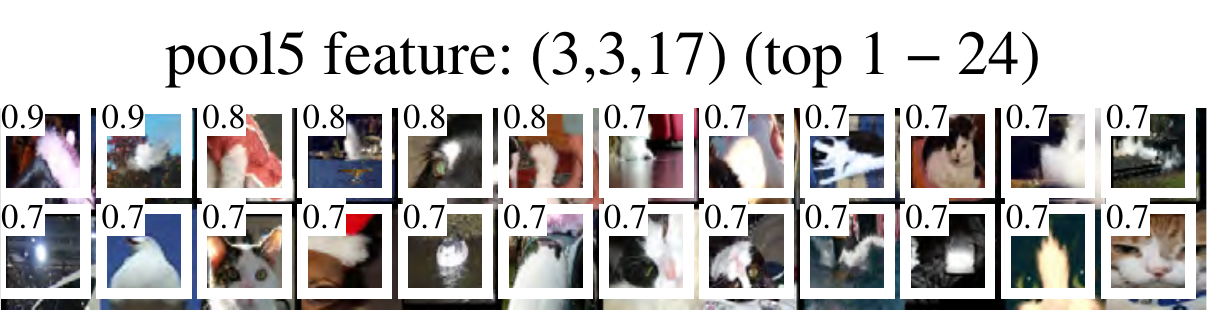}
\hspace{5pt}
\includegraphics[width=\sz\linewidth]{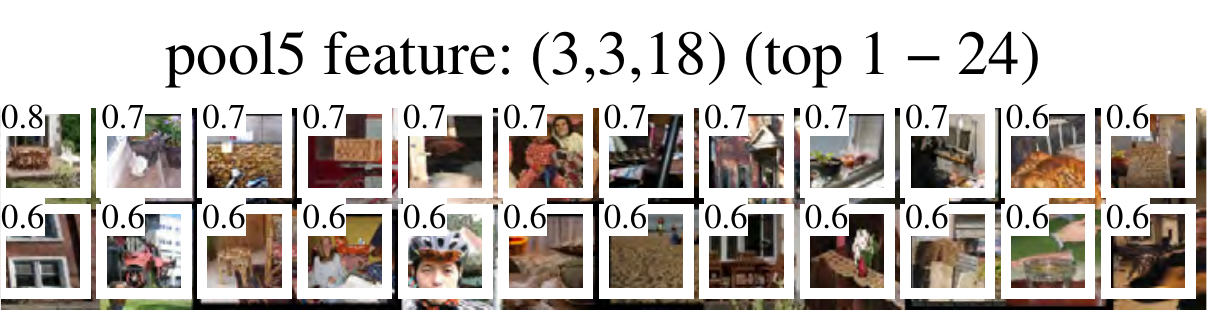}

\includegraphics[width=\sz\linewidth]{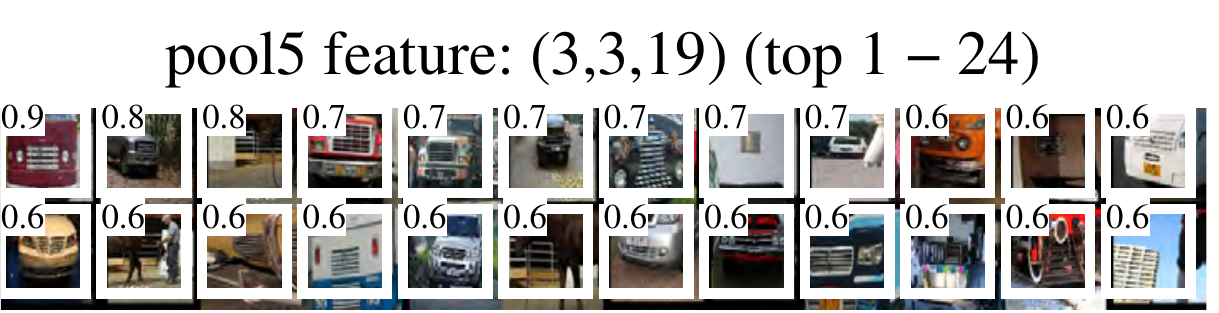}
\hspace{5pt}
\includegraphics[width=\sz\linewidth]{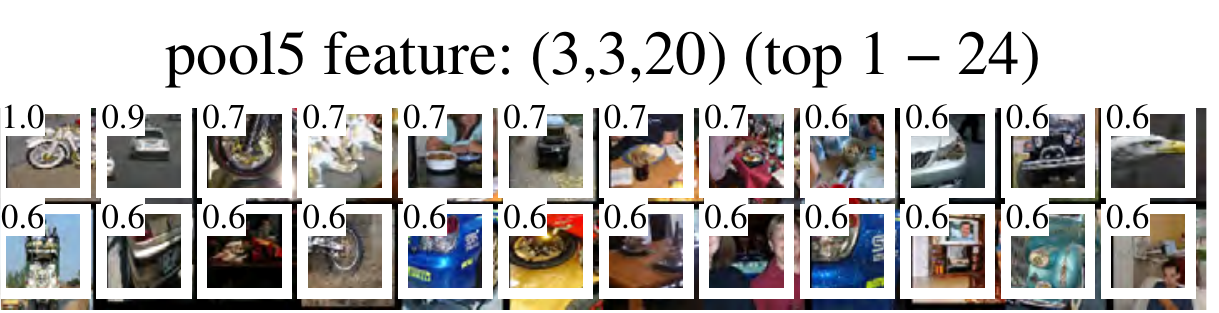}

\caption{\textbf{We show the 24 region proposals, out of the approximately 10 million regions in VOC 2007 test, that most strongly activate each of 20 units.}
Each montage is labeled by the unit's (y, x, channel) position in the $6 \times 6 \times 256$ dimensional \pool{5} feature map.
Each image region is drawn with an overlay of the unit's receptive field in white.
The activation value (which we normalize by dividing by the max activation value over all units in a channel) is shown in the receptive field's upper-left corner.
Best viewed digitally with zoom.}
\figlabel{vis21}
\end{figure*}

\section{Document changelog}

This document tracks the progress of R-CNN.
To help readers understand how it has changed over time, here's a brief changelog describing the revisions.

\paragraph{v1} Initial version.

\paragraph{v2} CVPR 2014 camera-ready revision. Includes substantial improvements in detection performance brought about by (1) starting fine-tuning from a higher learning rate (0.001 instead of 0.0001), (2) using context padding when preparing CNN inputs, and (3) bounding-box regression to fix localization errors. 

\paragraph{v3} Results on the ILSVRC2013 detection dataset and comparison with OverFeat were integrated into several sections (primarily \secref{detection} and \secref{ilsvrc}).

\paragraph{v4} The softmax \vs SVM results in \asecref{posneg} contained an error, which has been fixed. We thank Sergio Guadarrama for helping to identify this issue.

\paragraph{v5} Added results using the new 16-layer network architecture from Simonyan and Zisserman \cite{vggverydeep} to \secref{netarch} and \tableref{voc2007oxfordnet}.

{\small
\bibliographystyle{ieee}
\bibliography{r-cnn-arxiv}

\begin{thebibliography}{10}\itemsep=-1pt

\bibitem{objectness-pami}
B.~Alexe, T.~Deselaers, and V.~Ferrari.
\newblock Measuring the objectness of image windows.
\newblock {\em TPAMI}, 2012.

\bibitem{arbelaez2012semantic}
P.~Arbel{\'a}ez, B.~Hariharan, C.~Gu, S.~Gupta, L.~Bourdev, and J.~Malik.
\newblock Semantic segmentation using regions and parts.
\newblock In {\em CVPR}, 2012.

\bibitem{mcg2014}
P.~Arbel{\'a}ez, J.~Pont-Tuset, J.~Barron, F.~Marques, and J.~Malik.
\newblock Multiscale combinatorial grouping.
\newblock In {\em CVPR}, 2014.

\bibitem{o2p}
J.~Carreira, R.~Caseiro, J.~Batista, and C.~Sminchisescu.
\newblock Semantic segmentation with second-order pooling.
\newblock In {\em ECCV}, 2012.

\bibitem{carreira2012cpmc}
J.~Carreira and C.~Sminchisescu.
\newblock {CPMC}: Automatic object segmentation using constrained parametric
  min-cuts.
\newblock {\em TPAMI}, 2012.

\bibitem{cirecsan2013mitosis}
D.~Cire{\c{s}}an, A.~Giusti, L.~Gambardella, and J.~Schmidhuber.
\newblock Mitosis detection in breast cancer histology images with deep neural
  networks.
\newblock In {\em MICCAI}, 2013.

\bibitem{Dalal05}
N.~Dalal and B.~Triggs.
\newblock Histograms of oriented gradients for human detection.
\newblock In {\em CVPR}, 2005.

\bibitem{dean2013fast}
T.~Dean, M.~A. Ruzon, M.~Segal, J.~Shlens, S.~Vijayanarasimhan, and J.~Yagnik.
\newblock Fast, accurate detection of 100,000 object classes on a single
  machine.
\newblock In {\em CVPR}, 2013.

\bibitem{ILSVRC12}
J.~Deng, A.~Berg, S.~Satheesh, H.~Su, A.~Khosla, and L.~Fei-Fei.
\newblock Image{N}et {L}arge {S}cale {V}isual {R}ecognition {C}ompetition 2012
  ({I}{L}{S}{V}{R}{C}2012).
\newblock \url{http://www.image-net.org/challenges/LSVRC/2012/}.

\bibitem{imagenet_cvpr09}
J.~Deng, W.~Dong, R.~Socher, L.-J. Li, K.~Li, and L.~Fei-Fei.
\newblock Image{N}et: A large-scale hierarchical image database.
\newblock In {\em CVPR}, 2009.

\bibitem{DengCHI14}
J.~Deng, O.~Russakovsky, J.~Krause, M.~Bernstein, A.~C. Berg, and L.~Fei-Fei.
\newblock Scalable multi-label annotation.
\newblock In {\em CHI}, 2014.

\bibitem{decafICML}
J.~{Donahue}, Y.~{Jia}, O.~{Vinyals}, J.~{Hoffman}, N.~{Zhang}, E.~{Tzeng}, and
  T.~{Darrell}.
\newblock {DeCAF: A Deep Convolutional Activation Feature for Generic Visual
  Recognition}.
\newblock In {\em ICML}, 2014.

\bibitem{douze2009evaluation}
M.~Douze, H.~J{\'e}gou, H.~Sandhawalia, L.~Amsaleg, and C.~Schmid.
\newblock Evaluation of gist descriptors for web-scale image search.
\newblock In {\em Proc. of the ACM International Conference on Image and Video
  Retrieval}, 2009.

\bibitem{endres2010category}
I.~Endres and D.~Hoiem.
\newblock Category independent object proposals.
\newblock In {\em ECCV}, 2010.

\bibitem{PASCAL-IJCV}
M.~Everingham, L.~Van~Gool, C.~K.~I. Williams, J.~Winn, and A.~Zisserman.
\newblock The {PASCAL} {V}isual {O}bject {C}lasses ({VOC}) {C}hallenge.
\newblock {\em IJCV}, 2010.

\bibitem{farabet-pami-13}
C.~Farabet, C.~Couprie, L.~Najman, and Y.~LeCun.
\newblock Learning hierarchical features for scene labeling.
\newblock {\em TPAMI}, 2013.

\bibitem{lsvm-pami}
P.~Felzenszwalb, R.~Girshick, D.~McAllester, and D.~Ramanan.
\newblock Object detection with discriminatively trained part based models.
\newblock {\em TPAMI}, 2010.

\bibitem{fidler2013bottom}
S.~Fidler, R.~Mottaghi, A.~Yuille, and R.~Urtasun.
\newblock Bottom-up segmentation for top-down detection.
\newblock In {\em CVPR}, 2013.

\bibitem{fukushima1980neocognitron}
K.~Fukushima.
\newblock Neocognitron: A self-organizing neural network model for a mechanism
  of pattern recognition unaffected by shift in position.
\newblock {\em Biological cybernetics}, 36(4):193--202, 1980.

\bibitem{release5}
R.~Girshick, P.~Felzenszwalb, and D.~McAllester.
\newblock Discriminatively trained deformable part models, release 5.
\newblock \url{http://www.cs.berkeley.edu/~rbg/latent-v5/}.

\bibitem{gu2009recognition}
C.~Gu, J.~J. Lim, P.~Arbel{\'a}ez, and J.~Malik.
\newblock Recognition using regions.
\newblock In {\em CVPR}, 2009.

\bibitem{hariharan2012inverse}
B.~Hariharan, P.~Arbel{\'a}ez, L.~Bourdev, S.~Maji, and J.~Malik.
\newblock Semantic contours from inverse detectors.
\newblock In {\em ICCV}, 2011.

\bibitem{hoiem2012diagnosing}
D.~Hoiem, Y.~Chodpathumwan, and Q.~Dai.
\newblock Diagnosing error in object detectors.
\newblock In {\em ECCV}. 2012.

\bibitem{Jia13caffe}
Y.~Jia.
\newblock {Caffe}: An open source convolutional architecture for fast feature
  embedding.
\newblock \url{http://caffe.berkeleyvision.org/}, 2013.

\bibitem{krizhevsky2012imagenet}
A.~Krizhevsky, I.~Sutskever, and G.~Hinton.
\newblock Image{N}et classification with deep convolutional neural networks.
\newblock In {\em NIPS}, 2012.

\bibitem{lecun-89e}
Y.~LeCun, B.~Boser, J.~Denker, D.~Henderson, R.~Howard, W.~Hubbard, and
  L.~Jackel.
\newblock Backpropagation applied to handwritten zip code recognition.
\newblock {\em Neural Comp.}, 1989.

\bibitem{lecun-98}
Y.~LeCun, L.~Bottou, Y.~Bengio, and P.~Haffner.
\newblock Gradient-based learning applied to document recognition.
\newblock {\em Proc. of the IEEE}, 1998.

\bibitem{lim2013sketch}
J.~J. Lim, C.~L. Zitnick, and P.~Doll{\'a}r.
\newblock Sketch tokens: A learned mid-level representation for contour and
  object detection.
\newblock In {\em CVPR}, 2013.

\bibitem{SIFT}
D.~Lowe.
\newblock Distinctive image features from scale-invariant keypoints.
\newblock {\em IJCV}, 2004.

\bibitem{GIST}
A.~Oliva and A.~Torralba.
\newblock Modeling the shape of the scene: A holistic representation of the
  spatial envelope.
\newblock {\em IJCV}, 2001.

\bibitem{HSC}
X.~Ren and D.~Ramanan.
\newblock Histograms of sparse codes for object detection.
\newblock In {\em CVPR}, 2013.

\bibitem{rowley1998neural}
H.~A. Rowley, S.~Baluja, and T.~Kanade.
\newblock Neural network-based face detection.
\newblock {\em TPAMI}, 1998.

\bibitem{rumelhart86}
D.~E. Rumelhart, G.~E. Hinton, and R.~J. Williams.
\newblock Learning internal representations by error propagation.
\newblock {\em {Parallel Distributed Processing}}, 1:318--362, 1986.

\bibitem{overfeat}
P.~Sermanet, D.~Eigen, X.~Zhang, M.~Mathieu, R.~Fergus, and Y.~LeCun.
\newblock {OverFeat: Integrated Recognition, Localization and Detection using
  Convolutional Networks}.
\newblock In {\em ICLR}, 2014.

\bibitem{sermanetCVPR13}
P.~Sermanet, K.~Kavukcuoglu, S.~Chintala, and Y.~LeCun.
\newblock Pedestrian detection with unsupervised multi-stage feature learning.
\newblock In {\em CVPR}, 2013.

\bibitem{HJL_AAAI12}
H.~Su, J.~Deng, and L.~Fei-Fei.
\newblock Crowdsourcing annotations for visual object detection.
\newblock In {\em AAAI Technical Report, 4th Human Computation Workshop}, 2012.

\bibitem{Sung94}
K.~Sung and T.~Poggio.
\newblock Example-based learning for view-based human face detection.
\newblock Technical Report A.I. Memo No. 1521, Massachussets Institute of
  Technology, 1994.

\bibitem{szegedy2013deep}
C.~Szegedy, A.~Toshev, and D.~Erhan.
\newblock Deep neural networks for object detection.
\newblock In {\em NIPS}, 2013.

\bibitem{UijlingsIJCV2013}
J.~Uijlings, K.~van~de Sande, T.~Gevers, and A.~Smeulders.
\newblock Selective search for object recognition.
\newblock {\em IJCV}, 2013.

\bibitem{lecun94}
R.~Vaillant, C.~Monrocq, and Y.~LeCun.
\newblock Original approach for the localisation of objects in images.
\newblock {\em IEE Proc on Vision, Image, and Signal Processing}, 1994.

\bibitem{regionlets}
X.~Wang, M.~Yang, S.~Zhu, and Y.~Lin.
\newblock Regionlets for generic object detection.
\newblock In {\em ICCV}, 2013.

\bibitem{zeiler2011adaptive}
M.~Zeiler, G.~Taylor, and R.~Fergus.
\newblock Adaptive deconvolutional networks for mid and high level feature
  learning.
\newblock In {\em CVPR}, 2011.

\bibitem{vggverydeep}
K.~Simonyan and A.~Zisserman.
\newblock Very Deep Convolutional Networks for Large-Scale Image Recognition.
\newblock {\em arXiv preprint}, arXiv:1409.1556, 2014.

\end{thebibliography}
}

\end{document}